\documentclass[letterpaper,twocolumn,10pt]{article}
\usepackage{usenix-2020-09}

\usepackage{tikz}
\usepackage{amsmath}

\usepackage{filecontents}

\usepackage{hyperref}
\usepackage{url}
\usepackage{graphicx}
\usepackage{subcaption}
\usepackage{caption}
\usepackage{algorithm, algpseudocode}
\usepackage{amsmath}
\usepackage{amsfonts}
\usepackage{multirow}
\usepackage{color,array, soul}
\usepackage{xcolor}
\usepackage{mdframed}
\usepackage{amssymb}
\usepackage{dsfont}
\usepackage{colortbl}
\usepackage{wrapfig}

\usepackage{booktabs}
\usepackage{placeins}
\usepackage{enumitem}
\usepackage{amsthm}
\newtheorem{theorem}{Theorem}
\usepackage{listings}
\usepackage[utf8]{inputenc}
\definecolor{lavender}{RGB}{230, 230, 250}
\definecolor{gemmagreen}{HTML}{00cc96}
\begin{document}

\date{}

\title{\Large \bf Revisiting Privacy, Utility, and Efficiency Trade-offs\\when Fine-Tuning Large Language Models}

\author{
{\rm Soumi Das}\\
MPI-SWS\\
Germany
\and
{\rm Camila Kolling}\\
MPI-SWS\\
Germany
\and
{\rm Mohammad Aflah Khan}\\
MPI-SWS\\
Germany
\and
{\rm Mahsa Amani}\\
MPI-SWS\\
Germany
\and
{\rm Bishwamittra Ghosh}\\
MPI-SWS\\
Germany
\and
{\rm Qinyuan Wu}\\
MPI-SWS\\
Germany
\and
{\rm Till Speicher}\\
Aleph Alpha\\
Germany
\and
{\rm Krishna P. Gummadi}\\
MPI-SWS\\
Germany
} 

\maketitle

\begin{abstract}
We study the inherent trade-offs in \emph{minimizing privacy risks and maximizing utility, while maintaining high computational efficiency}, when fine-tuning large language models (LLMs).
A number of recent works in privacy research have attempted to mitigate privacy risks posed by memorizing fine-tuning data by using differentially private training methods (e.g., DP-SGD), albeit at a significantly higher computational cost (inefficiency). 
In parallel, several works in systems research have focussed on developing (parameter) efficient fine-tuning methods (e.g., LoRA), but few works, if any, investigated whether such efficient methods enhance or diminish privacy risks.
In this paper, we investigate this gap and arrive at a surprising conclusion: efficient fine-tuning methods like LoRA mitigate privacy-risks similar to private fine-tuning methods like DP-SGD.  
Our empirical finding directly contradicts prevailing wisdom that privacy and efficiency objectives are at odds during fine-tuning.
Our finding is established by (a) carefully defining measures of privacy and utility that distinguish between memorizing \emph{sensitive and non-sensitive} tokens in training and test datasets used in fine-tuning and (b) extensive evaluations using multiple open-source language models from Pythia, Gemma, Llama, and Qwen families and different domain-specific datasets.

\end{abstract}

\section{Introduction}

Large language models (LLMs) have shown  proficiency across diverse natural language tasks~\cite{naveed2023comprehensive}, finding applications in education~\cite{wang2024large}, medical chatbots~\cite{thirunavukarasu2023large}, and AI assistants~\cite{dong2023towards}. Their capabilities stem from a two-phase process: (a) pre-training on extensive web data~\cite{kaplan2020scaling} to develop general language understanding~\cite{brown2020language}, (b) fine-tuning on domain-specific data for specific tasks~\cite{zhang2023instruction}. In both phases, the key challenges involve enhancing \emph{privacy} and \emph{efficiency} while maintaining the models' \emph{utility}. 
Privacy is related to reducing the risk of LLMs leaking sensitive user information contained in the training data, efficiency is related to reducing the computational cost of training, while utility is related to generative performance over test data.
%

A long line of recent research in the privacy community has focussed on methods to mitigate privacy risks when training LLMs~\cite{shi-etal-2022-selective, zhao-etal-2022-provably, 10.1145/2976749.2978318,yu2022differentiallyprivatefinetuninglanguage}. A notable example of such methods is differential privacy based stochastic gradient descent (DP-SGD)\cite{10.1145/2976749.2978318}.
Simultaneously, a flurry of recent research in the systems community has explored parameter-efficient fine-tuning in  LLMs~\cite{han2024parameterefficientfinetuninglargemodels}. A notable example of this class of methods is low-rank adaptation (LoRA)~\cite{hu2022lora}.
However, no prior works, to the best of our knowledge, have investigated the privacy risks associated with efficient training methods.
The central question driving our research here is: \emph{do efficient fine-tuning methods enhance or mitigate privacy risks during training?}
As DP-SGD incurs significant additional computational overhead~\cite{dupuy2022efficient}, while LoRA significantly reduces the computational costs, the answer to the above question can have significant consequences for achieving good privacy-efficiency-utility tradeoffs when fine-tuning. 
For instance, if LoRA mitigates privacy risks of training that would suggest that it can simultaneously achieve both privacy and efficiency objectives, contradicting the conventional wisdom drawn from DP literature that privacy comes at a computational cost.
A key (surprising) finding of our work lies in establishing that \textit{LoRA does indeed mitigate privacy risks}.
A conjecture that might explain our finding is rooted in the following high-level observations about DP-SGD and LoRA: methodologically, both DP-SGD and LoRA restrict the impact that training examples can have on model parameters -- DP-SGD deliberately through its noisy gradient update, and LoRA through low-rank adaptation. We formalize this intuition in this work.  
%
%
%
%

When attempting to answer the above question, we encountered a more foundational question: \emph{how should one quantify such privacy risks associated with a fine-tuning method, so that it allows for a performance comparison across different methods?} 
Numerous studies have highlighted privacy risks in LLMs due to their tendency to memorize and regurgitate training data containing sensitive personally identifiable information (PII) such as names, emails, and credentials ~\cite{mattern2023membership, fu2023practical, kaneko2024sampling, carlini2021extracting, mireshghallah-etal-2022-empirical, panda2024teach}. 
A natural way to account for privacy risks from memorization might be to measure loss on recollecting tokens in training data sequences.

\begin{figure}
    \centering
    \begin{subfigure}[b]{.49\linewidth}
    \centering
    \includegraphics[width=\linewidth]{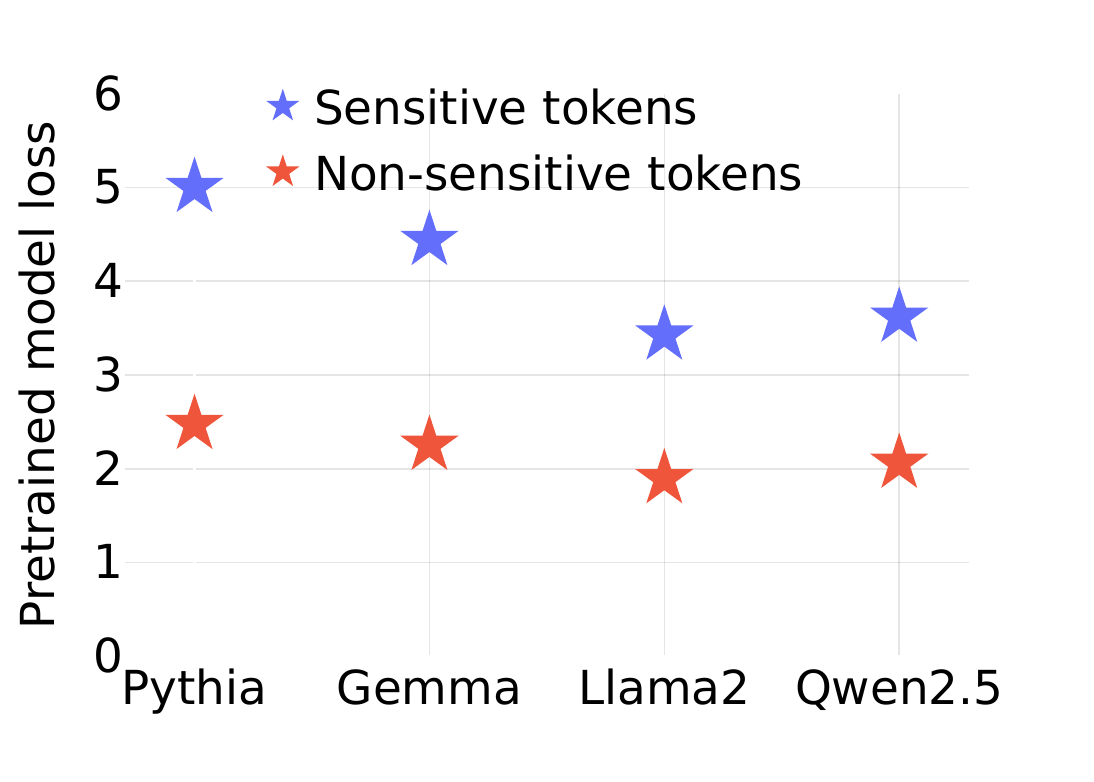}
    \caption{Pretrained}
    \label{fig:a}
    \end{subfigure}
    \begin{subfigure}[b]{.49\linewidth}
    \centering
    \includegraphics[width=\linewidth]{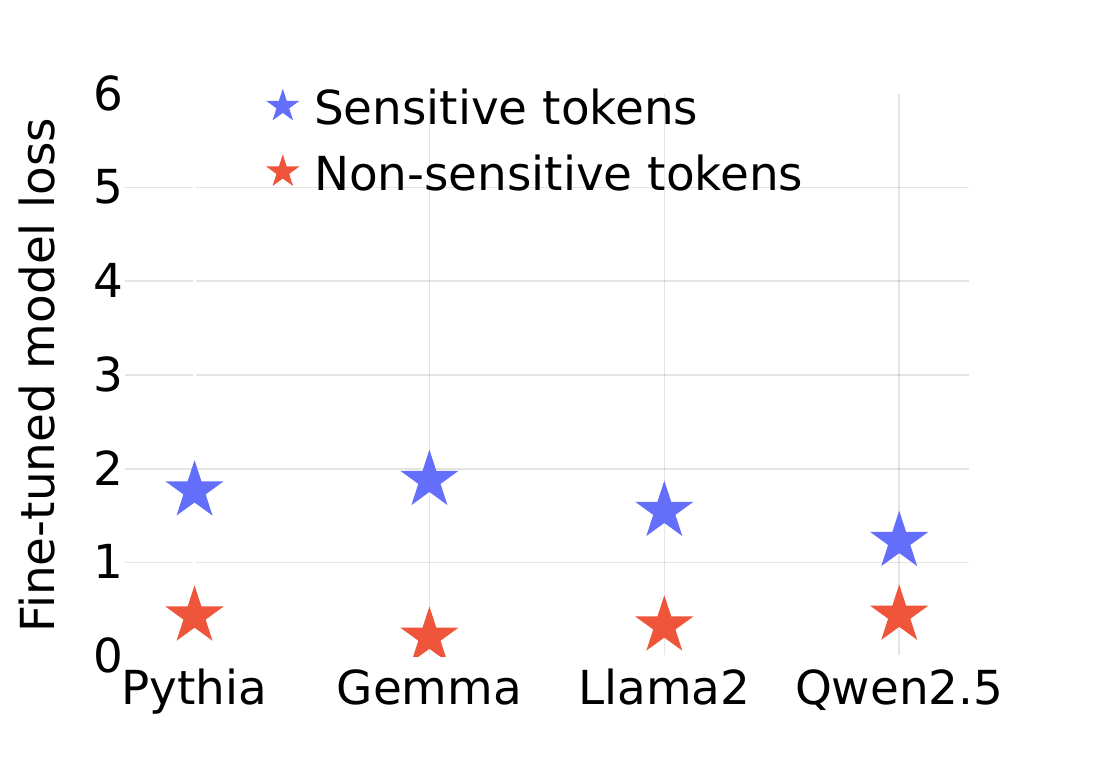}
    \caption{Fine-tuned}
    \label{fig:b}
    \end{subfigure}
    \caption{
    Sensitive and non-sensitive tokens have different predictability, measured as the recollection loss by pre-trained  models (Figure~\ref{fig:a}) and fine-tuned models (Figure~\ref{fig:b}). This distinction motivates to quantify privacy using sensitive tokens from the training data.
    }
    \label{fig:illustrative_example}
\end{figure}

However, we find that LLMs exhibit very different losses in recollecting sensitive vs. non-sensitive tokens in training data (see Figure~\ref{fig:illustrative_example}).
This difference is due to inherent randomness and unpredictability of sensitive data (e.g., phone numbers, SSNs) compared to non-sensitive data (e.g., “The dog chases the \_”), which is often more structured and predictable. 
Consequently, we propose a new privacy measure that explicitly account for this difference: aiming for high loss on sensitive tokens from training data.

\noindent \textbf{Our contributions and findings.} The primary contribution of this work is to explore if efficient fine-tuning methods inherently mitigate privacy risks during training. To the best of our knowledge, no prior works have investigated these privacy risks associated with efficient training methods. We summarize our main contributions and findings as follows:

\noindent \emph{1. Quantifying privacy and utility, when training LLMs:} We conceptually argue and empirically demonstrate that LLM's ability to recollect (predict) sensitive and non-sensitive data in training (test) datasets are so starkly different that we need to account for them when quantifying privacy and utility. Privacy is best captured by a model's ability to \textit{recollect sensitive tokens in training data}, while utility is best captured by the model's ability to \textit{predict non-sensitive tokens in test data}.


\noindent \emph{2. Comparing privacy-utility-efficiency tradeoffs for three different fine-tuning methods:} Our measures allow us to conduct a systematic and extensive empirical study of three different fine-tuning methods: full fine-tuning (FFT), DP-SGD, and LoRA, using models from four different LLM families, Pythia~\cite{biderman2023pythia}, Gemma~\cite{team2024gemma}, Llama~\cite{touvron2023llama} and Qwen \cite{yang2024qwen2} over two datasets. 
Our comparative study yields several interesting insights:  we find that -- FFT results in poor utility-privacy tradeoffs; DP offers reasonable utility-privacy tradeoffs, but is computationally very expensive; LoRA is almost on par with DP in terms of privacy with good privacy-utility tradeoffs, but is more computationally efficient. To re-confirm our results, we evaluate the three methods using existing privacy measures—\textit{privacy loss} \cite{10.1145/2976749.2978318} and \textit{canary exposure} \cite{carlini2019secret}.  We also show formally that the effect of fine-tuning a model using LoRA and DP on a datapoint is analogous, leading to privacy benefits.

\noindent \emph{3. Feasibility of achieving all the three privacy, utility and efficiency objectives simultaneously:} Our findings about LoRA performance challenge prevailing wisdom that enhancing privacy during training is more computationally expensive. This calls for investigating privacy benefits of existing and new parameter efficient fine-tuning methods.

\section{Related Work}
\label{sec:related_work}
\textbf{Privacy Attacks and Quantification.}
Privacy concerns in LLMs have gained attention in recent years \cite{KIBRIYA2024109698, YAO2024100211}, particularly the possibility of exposing data via membership inference attacks (MIA) \cite{mireshghallah-etal-2022-quantifying, mattern2023membership, fu2023practical, kaneko2024sampling} and training data extraction attacks \cite{carlini2021extracting, mireshghallah-etal-2022-empirical, panda2024teach}. While membership inference attacks have the goal of identifying if a certain datapoint was in the training dataset using model confidence scores, training data extraction attacks try to extract specific parts of data from the training dataset using different prompting strategies. A recent study \cite{zhang2024membershipinferenceattacksprove} underscores that membership inference attacks may not be reliable due to their practical limitations of requiring knowledge of the entire training data, thus proving data extraction attacks to be more viable.

Existing works have quantified privacy attacks in large language models especially through the lens of memorization. For example, the authors in \cite{carlini2023quantifying} assess memorization by quantifying how closely LLM generation matches the exact training data phrases when prompted with tailored prefixes. The metric of `exposure' is introduced in \cite{10.5555/3361338.3361358} to evaluate a model’s vulnerability when exposed to data that was artificially introduced into the dataset (also known as "canaries") several times during training. 
This metric has been widely adopted in subsequent research \cite{shi-etal-2022-selective, zhao-etal-2022-provably} as a measure of privacy. However, the metric’s formulation is limited by its reliance on assumptions about the surrounding knowledge of other canaries, which do not hold in practical scenarios. 

Intuitively, privacy evaluation should be more related to sensitive data seen by the model during its training. Some of the existing studies ~\cite{10.5555/3666122.3667033, 10179300, zhao-etal-2022-provably, shi-etal-2022-selective} focusing on privacy attacks in LLMs may consider sensitive and non-sensitive counterparts in the data explicitly during training, but they do not retain this distinction during privacy quantification. We believe (and later show) that this lack of distinction can lead to an inaccurate (sometimes, overestimated) assessment of privacy threats. In our paper, we carefully make the distinction between sensitive and non-sensitive data and propose a revised quantification.

\noindent
\textbf{Privacy-Utility Tradeoffs.}
To mitigate privacy leakage, differential privacy (DP-SGD-SGD) measures have been proposed, which add theoretical privacy guarantees to the training process \cite{10.1145/2976749.2978318}. The key idea of DP-SGD-SGD is to clip the gradients of each datapoint and add noise to the clipped gradients in every iteration.  Its primary goal is to reduce the influence of individual datapoints in the training procedure thus preventing its leakage during inference. 

Existing work utilising DP-SGD~\cite{yu2022differentiallyprivatefinetuninglanguage} has shown a clear trade-off between privacy and utility where privacy is measured in terms of theoretical guarantee and utility in terms of overall performance on the end task. The authors in \cite{10.1145/3531146.3534642} also question the notion of privacy considered in these privacy preserving techniques. Another line of work from \cite{ shi-etal-2022-selective} and \cite{zhao-etal-2022-provably} distinguishes between sensitive and non-sensitive data using techniques like regex and redaction and proposes a customised training strategy using DP-SGD. However, their evaluation measure also relies on theoretical guarantees for privacy and overall performance for utility. The question that arises here is whether this measure of quantification of privacy and utility is sufficient in the context of large language models, or whether it needs to be more nuanced.

Additionally, to ensure privacy guarantees, differential privacy requires higher computational resources (time and memory) during training. For example, compared to vanilla SGD, DP-SGD-SGD may incur up to $20$x training time, which is often a bottleneck in resource-intensive tasks \cite{dupuy2022efficient}. Therefore, the conventional wisdom is that privacy comes at the cost of efficiency. This leads us to probe ways in which the above limitation can be prevented.

\noindent
\textbf{Utility-Efficiency Tradeoffs.} Fine-tuning a pre-trained LLM on a task specific context has been used in several applications like medical chatbots, AI assistants, etc. However, full fine-tuning of large language models is expensive as it involves updating all the parameters. Recent work has focused on reducing the training cost while maintaining utility, with the introduction of parameter efficient fine-tuning (PEFT) techniques~\cite{han2024parameterefficientfinetuninglargemodels}. Well-known PEFT methods include adapter-based fine-tuning \cite{houlsby2019parameter, lei2023conditional, zhu2021counter, he2021towards}, soft prompt-based fine-tuning \cite{li2021prefix, li2023prefix, liu2021p, liu2024gpt, lester2021power}, and parameterized fine-tuning \cite{hu2022lora,liu2024dora}. Among different PEFT methods, the Low-Rank Adaptation method, called LoRA \cite{hu2022lora} has emerged as one of the most widely used methods. LoRA updates fewer parameters in the model via low-rank approximation, providing computational efficiency with a relatively low cost to utility.

In recent work on private fine-tuning \cite{liu2023differentially, yu2021differentially,ma2024efficient}, the authors combine LoRA with DP-SGD to reduce the additional computational overhead induced by differential privacy. In such a context, we ask whether DP-SGD is the only method towards ensuring privacy. LoRA's training procedure of updating fewer parameters can be thought of as analogous to the noisy gradient update in DP-SGD. This leads us to the question of whether \textit{LoRA has any privacy benefits} besides having control over utility-efficiency tradeoffs.

\noindent
\textbf{Privacy-Utility-Efficiency Tradeoffs.}
To the best of our knowledge, our work is the first to \textit{investigate the privacy benefits of LoRA and systematically examine the privacy, utility, and efficiency tradeoffs among different fine-tuning methods}. Besides, instead of relying on the existing measures of privacy and utility, we distinguish between sensitive and non-sensitive data to \textit{propose a nuanced quantification of privacy and utility}.  

\section{Quantifying Privacy and Utility}
\label{sec:quantifying_privacy_utility}
In this section, we introduce the distinction between sensitive and non-sensitive entities when quantifying privacy and utility of an LLM. We conduct case studies to compare our quantification with existing measures. Finally, we demonstrate how the privacy threat is unintentionally exaggerated in existing methods due to the lack of distinction between sensitive and non-sensitive entities.

\subsection{Rethinking Privacy and Utility}

Existing studies at the intersection of privacy and natural language processing \cite{zhao-etal-2022-provably,shi-etal-2022-selective,li2022large,yu2022differentially} seek to enhance privacy while maintaining model utility. Utility is generally assessed based on model performance, such as loss, accuracy, or perplexity \textit{across the entire test dataset}. Privacy is evaluated in terms of performance measures on the \textit{entire training dataset} or theoretical guarantees in differential privacy (DP-SGD).

Natural language text may contain both sensitive and non-sensitive words, referred to as entities. For example, sensitive entities include names, addresses, phone numbers, order IDs, and other personally identifiable information. In contrast, non-sensitive entities generally involve semantic and/or syntactic completions following predictable patterns in language generation tasks. Informally, sensitive entities are drawn from a large search space (e.g., \textit{a random sequence of digits}), resulting in high entropy and low predictability. In contrast, non-sensitive entities are more restricted in their occurrences (e.g., \textit{a subject is typically followed by a verb}), leading to low entropy and high predictability. 
Several studies \cite{biderman2024emergent, shi-etal-2022-selective, zhao-etal-2022-provably} distinguish between sensitive and non-sensitive entities in their proposed privacy leakage mitigation methods. However, the distinction is not leveraged in the \textit{quantification of privacy and utility}, which is essential for a granular evaluation as discussed next.

\noindent
\textbf{Quantification of privacy and utility.} In this work, we quantify privacy and utility by accounting for sensitive and non-sensitive entities. Considering a training dataset and a test dataset in a general LLM training pipeline, we quantify \textbf{privacy} as the \textit{recollection of sensitive entities in the training data} and \textbf{utility} as the \textit{prediction of non-sensitive entities in the test data}. Our motivation for the quantification is two-fold: (1) privacy of a model is generally related to training data, while utility is the model's performance on the test data. (2) when quantifying privacy, we deliberately disregard non-sensitive entities, since they are more predictable and not sensitive to a specific person or entity. Similarly, in quantifying utility, we ignore sensitive entities in the test data, since the sensitive entities are rare (and possibly unseen during training), whereas predicting non-sensitive entities indicates the general language understanding ability of LLMs. Next, we provide two pieces of evidence supporting why the distinction is important.

\subsection{Why do we distinguish between sensitive and non-sensitive entities?}
\label{sec:sens_non_sens_distinction}

In this section, we present evidence supporting the importance of distinguishing between sensitive and non-sensitive entities in natural language text while quantifying privacy and utility of an LLM. 

\begin{figure}[t!]
    \centering
        \begin{subfigure}{.48\linewidth}
       \includegraphics[scale=0.25]{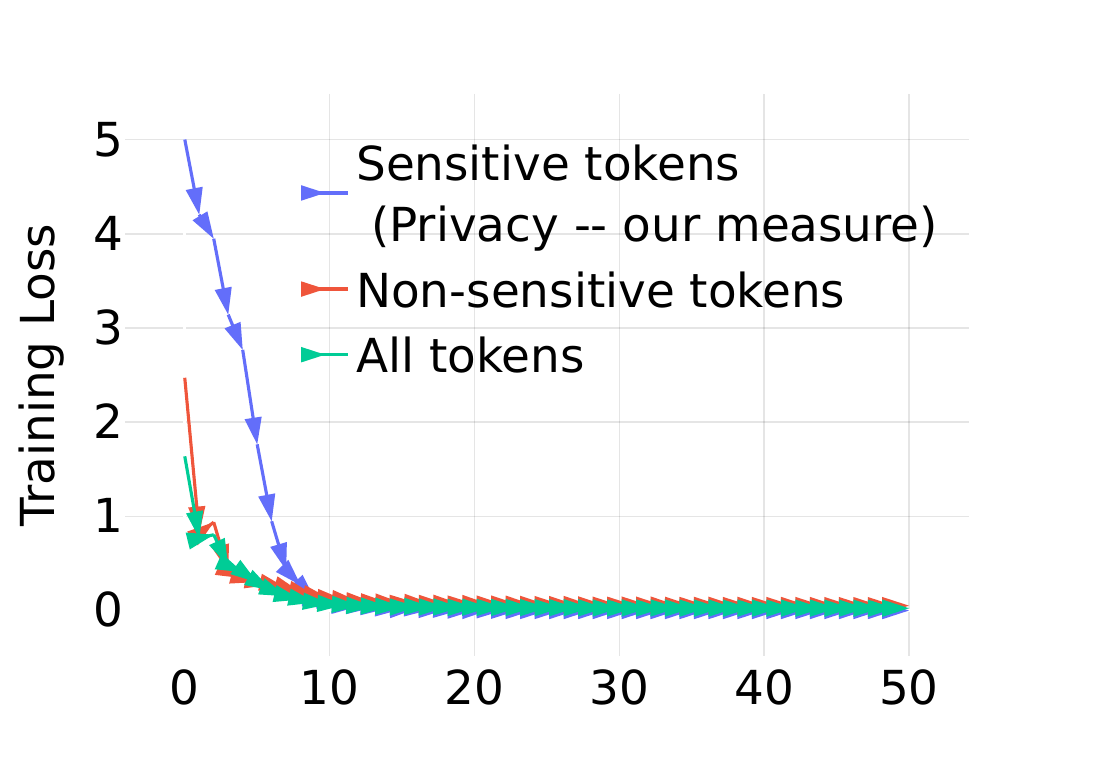}
        \caption{Privacy measure in Pythia}
        \label{fig:pyt_priv}
    \end{subfigure}
    \hfil
    \begin{subfigure}{.48\linewidth}
        \includegraphics[scale=0.25]{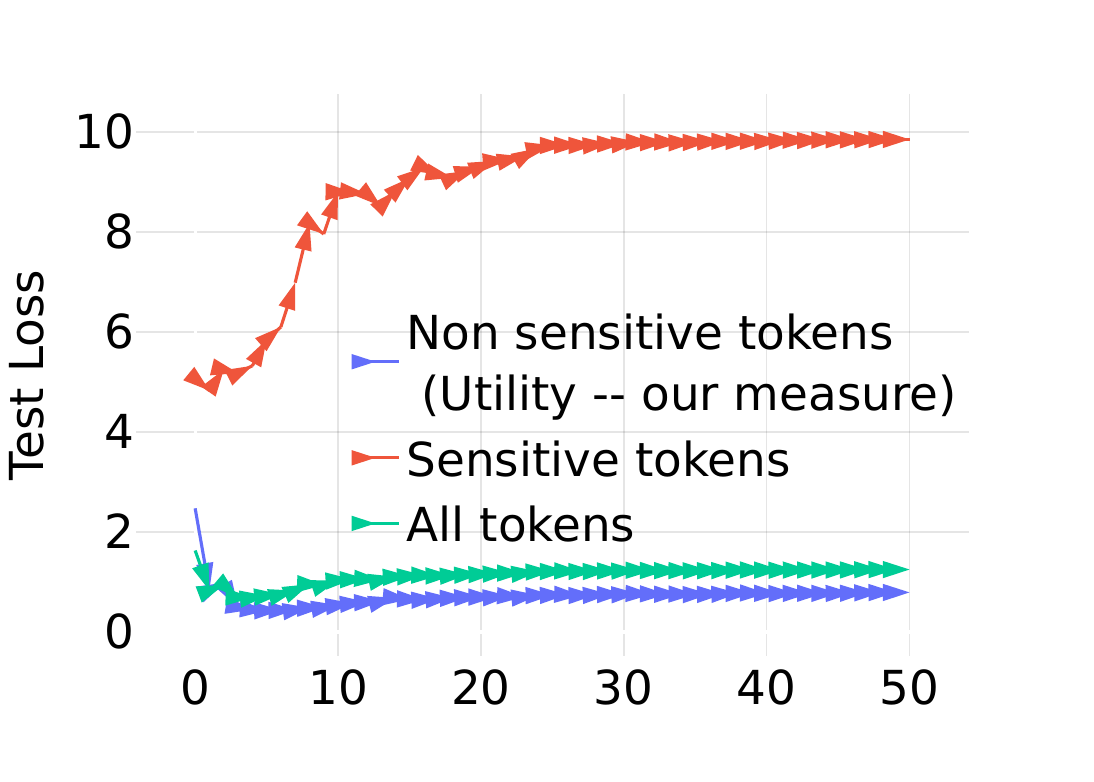}
        \caption{Utility measure in Pythia}
        \label{fig:pyt_util}
    \end{subfigure}
    
    \begin{subfigure}{.48\linewidth}
       \includegraphics[scale=0.25]{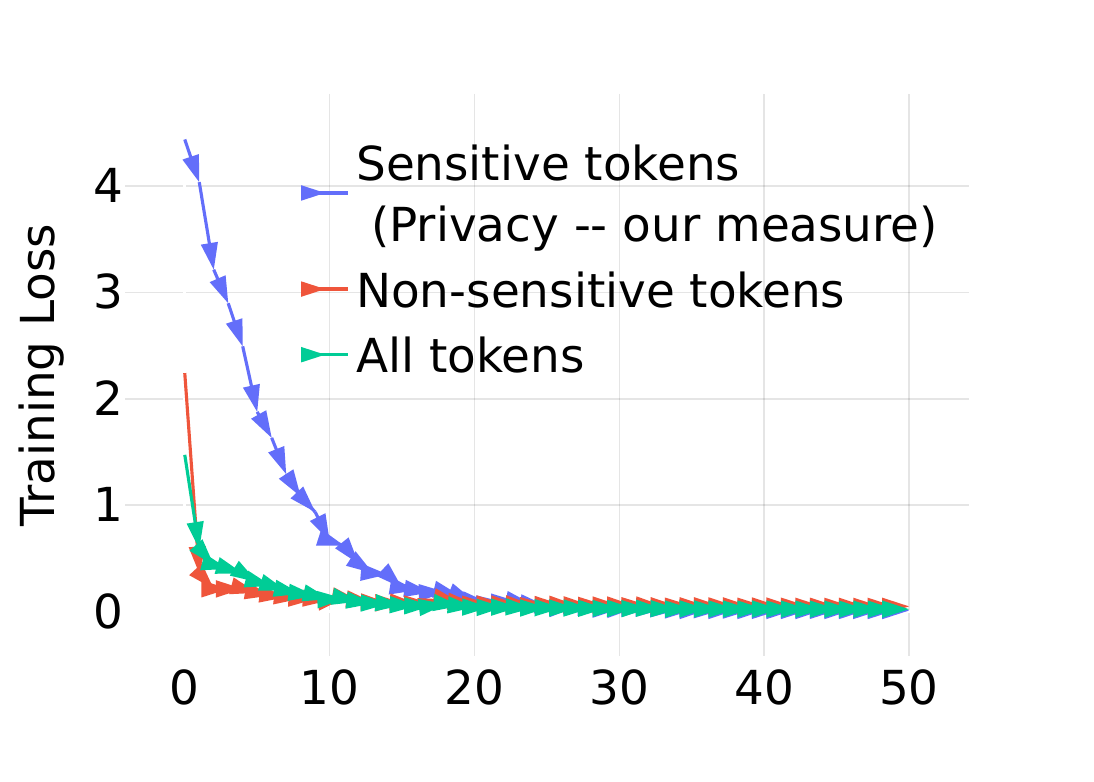}
        \caption{Privacy measure in Gemma}
        \label{fig:gemma_priv}
    \end{subfigure}
    \hfil
    \begin{subfigure}{.48\linewidth}
        \includegraphics[scale=0.25]{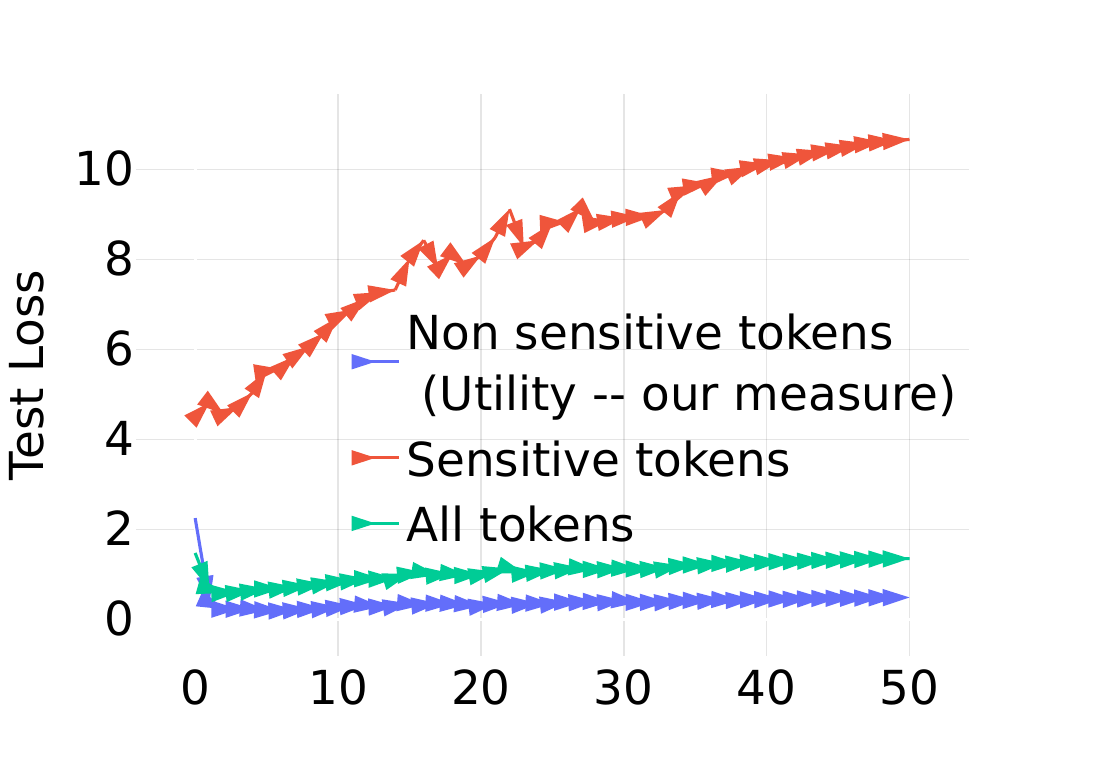}
        \caption{Utility measure in Gemma}
        \label{fig:gemma_util}
    \end{subfigure}

    \begin{subfigure}{.48\linewidth}
       \includegraphics[scale=0.25]{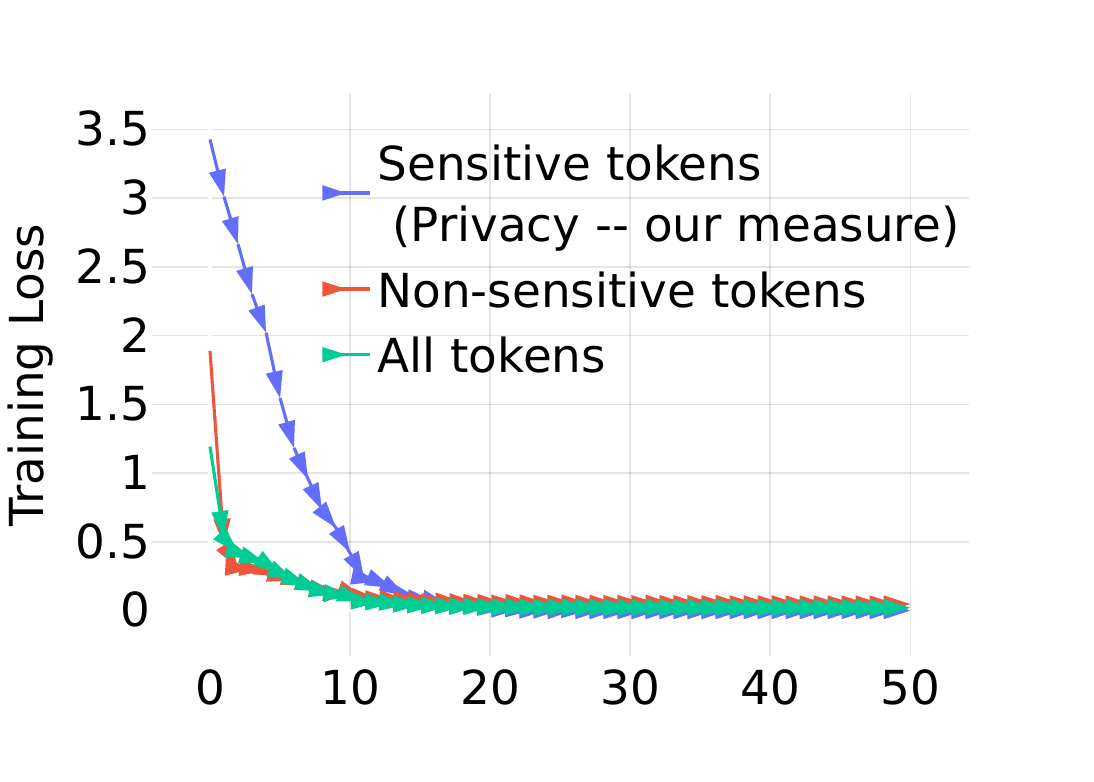}
        \caption{Privacy measure in Llama2}
        \label{fig:llama_priv}
    \end{subfigure}
    \hfil
    \begin{subfigure}{.48\linewidth}
        \includegraphics[scale=0.25]{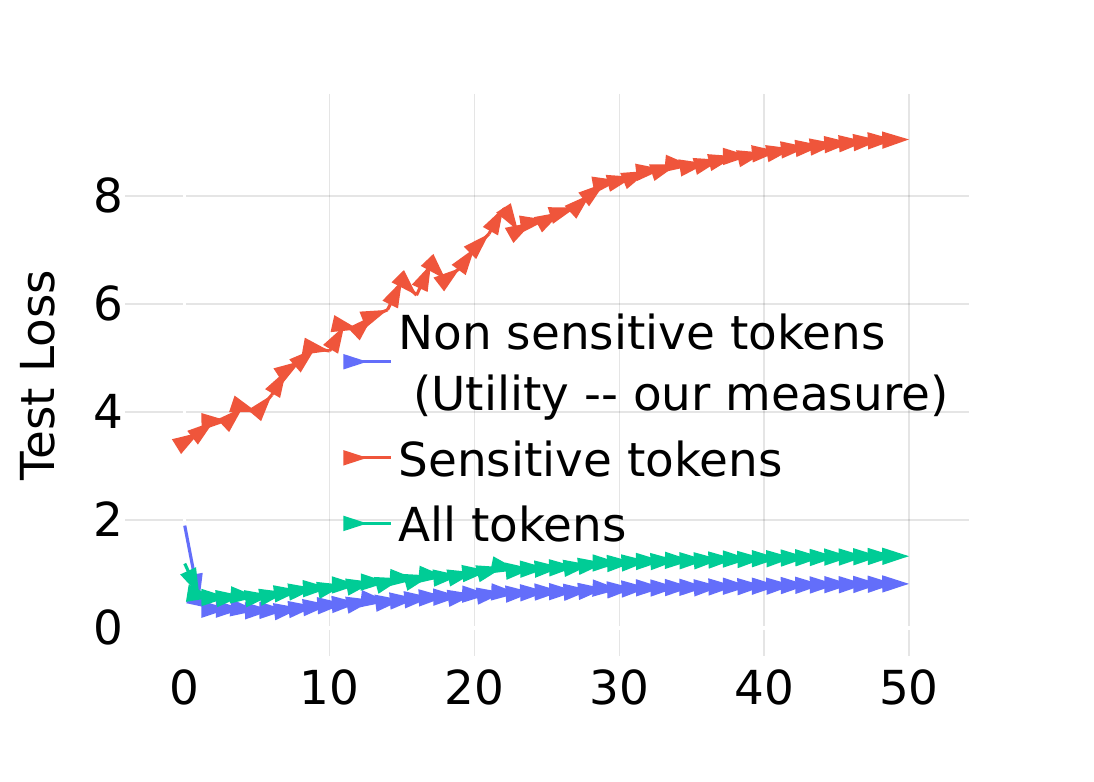}
        \caption{Utility measure in Llama2}
        \label{fig:llama_util}
    \end{subfigure}

    \begin{subfigure}{.48\linewidth}
       \includegraphics[scale=0.25]{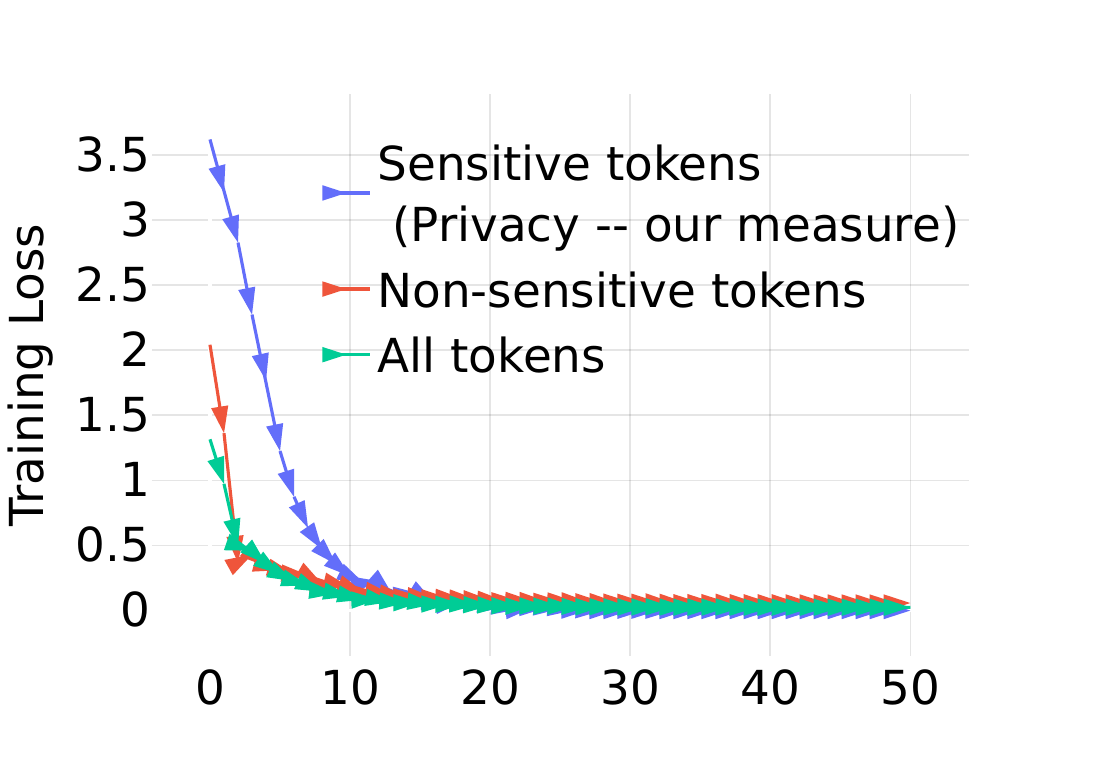}
            \caption{Privacy measure in Qwen2.5}
        \label{fig:qwen_priv}
    \end{subfigure}
    \hfil
    \begin{subfigure}{.48\linewidth}
        \includegraphics[scale=0.25]{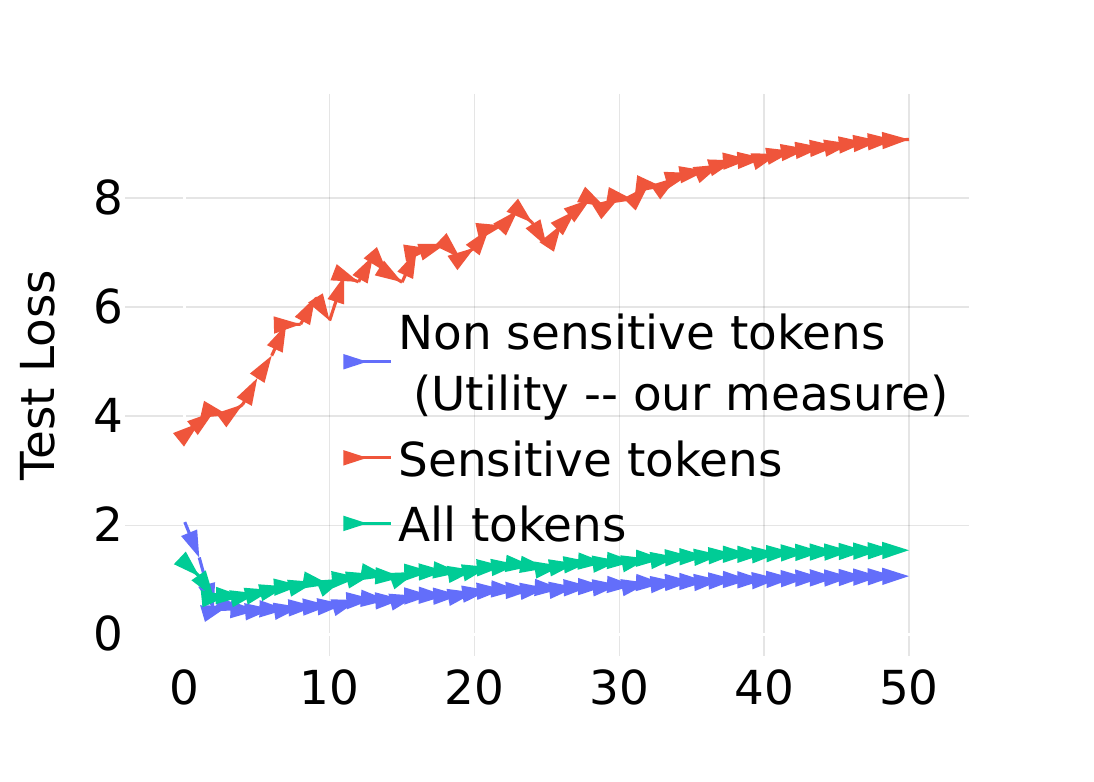}
        \caption{Utility measure in Qwen2.5}
        \label{fig:qwen_util}
    \end{subfigure}

    \caption{
    Our measures offer a more precise assessment of privacy and utility when fine-tuning LLMs by distinguishing between sensitive and non-sensitive tokens, revealing higher privacy (higher loss) for sensitive tokens and better utility (lower loss) for non-sensitive tokens compared to traditional measures that overlook this sensitivity-based distinction.
    }
    \label{fig:FFT_All}
\end{figure}

\paragraph{\textbf{I. Analyzing privacy and utility while fine-tuning an LLM
}}
In order to align with LLM terminology, hereafter, we use tokens to denote entities. Fine-tuning involves iterating an LLM on a specific dataset containing both sensitive and non-sensitive tokens. 
We illustrate how our measure of privacy and utility compares to existing measure in a typical fine-tuning scenario, highlighting a key difference: our approach distinguishes between sensitive and non-sensitive tokens, whereas the existing measure does not.

\textbf{Results.} In Figure~\ref{fig:FFT_All}, we demonstrate measures of privacy (left column) and utility (right column) while fine-tuning three LLM models on Customersim dataset \cite{shi-etal-2022-selective} (experimental details
are provided at the end of this section). In particular, we show training loss on the left column and test loss on the right column. Importantly, we separately compute the loss for both sensitive and non-sensitive tokens in both training and test datasets. Intuitively, a higher loss denotes more privacy and less utility.

\textbf{Privacy is overestimated in the existing measure.} In Figure~\ref{fig:pyt_priv}, we compute privacy using our measure, as well as the existing one. The existing measure of privacy considers \textit{all tokens in the training data}, where low training loss denotes less privacy, while our measure considers \textit{only the sensitive tokens in the training data}. Using our measure, a notable disparity emerges: \textit{sensitive tokens exhibit significantly higher loss than non-sensitive ones}, particularly in the initial training epochs, as sensitive tokens are less predictable. This eventually indicates that the loss over all tokens (existing measure) would be much lower initially than the loss over only sensitive tokens (our measure), thus overestimating privacy threats much earlier. 
Similar trends are observed for other models in Figures \ref{fig:gemma_priv}, \ref{fig:llama_priv}, and \ref{fig:qwen_priv}.

\textbf{Utility is underestimated in the existing measure.}
In Figure \ref{fig:pyt_util}, we compute utility using our measure and the prevailing one. The existing utility measure is related to the test loss of all tokens, where lower test loss indicates better utility. We can observe that our measure that considers the test loss on only non-sensitive tokens provides better utility than the existing measure. Similar trends are observed for other models in Figures \ref{fig:gemma_util}, \ref{fig:llama_util}, and \ref{fig:qwen_util}. 
Existing measure would lead to better utility at earlier epochs compared to our measure
i.e. one would select checkpoints at epochs $4$, $2$, $2$, and $2$ using existing measures for Pythia, Gemma, Llama2, and Qwen2.5 respectively and at epochs $5$, $6$, $5$, and $6$ using our measure. Thus, utility is underestimated in the existing measure. 
For the same fine-tuning scenario, our measure certifies better utility than the existing measure, which unintentionally considers the loss of sensitive tokens that are rare. The key to our findings is how we annotate sensitive and non-sensitive tokens in real-world datasets. Next, we expand our experimental details leading to our results.

\begin{table}[t!]
\scriptsize
\centering
\caption{Example of samples from datasets with sensitive tokens highlighted.}
\label{tab:snippets}
\begin{tabular}{|ll|}
\hline
\multicolumn{2}{|c|}{\textbf{CustomerSim}}                                                                                                                                                                                                                                                                                                                                                                                                                                                                                                                                                                                                                                                                                                                                                                                                                                                                                                                                                                                                                                                                                                                                                                                                                                                                                                                                                                                                                                                                                                                                                                                                                                                                                                     \\ \hline
\multicolumn{1}{|l|}{\begin{tabular}[c]{@{}l@{}}SYS: Hello, I am the customer \\ support bot. What can \\ I do for you?\\ USR: Hello robot. I ordered a pot \\ several days ago but I can't track it.\\ SYS: Could you verify your full name?\\ USR: \hl{Patrick Schug}\\ SYS: Verify your order number please.\\ USR: It's \hl{843-58572-7002}.\\ SYS: You can track your package \\ with your tracking number \\, which is \hl{AGZIM5T6KL}. \\ Are you happy about my answer?\\ USR: All good. See you.\\ SYS: Have a nice day! Bye.\end{tabular}}                                                                                                                                                                                                                                                                              & \begin{tabular}[c]{@{}l@{}}SYS: Hello, I am the customer \\ support bot. What can \\ I do for you?\\ USR: Hi. Where is my package?\\ SYS: Could you verify \\ your full name?\\ USR: I am \hl{Catherine Pena}.\\ SYS: Verify your phone number.\\ USR: You can reach me at \\ \hl{547.302.3744}.\\ SYS: The tracking number is \\ \hl{VVTPHDB6VK}. \\ Anything else?\\ USR: All good.\end{tabular}                                                                                                                                                                                                                                                                                                                                                                                                                                                                                                      \\ \hline
\multicolumn{2}{|c|}{\textbf{SynBio}}                                                                                                                                                                                                                                                                                                                                                                                                                                                                                                                                                                                                                                                                                                                                                                                                                                                                                                                                                                                                                                                                                                                                                                                                                                                                                                                                                                                                                                                                                                                                                                                                                                                                                                            \\ \hline
\multicolumn{1}{|l|}{\begin{tabular}[c]{@{}l@{}}My name is \hl{Alexander Tanaka}, and I'm a \\ saleswoman with a year of experience. I \\ recently completed a project that \\ involved developing and implementing \\ a new sales strategy  for \\ my company. I started by analyzing our \\ current sales data to identify areas where \\ we could improve...
\end{tabular}} & \begin{tabular}[c]{@{}l@{}}My name is \hl{Phillip Martinez}, and \\ I would like to share some aspects \\ of my life's journey with \\ you. I have had the pleasure of  \\ living in various places throughout \\ my life,  but I currently reside \\ at \hl{4537 Tanglewood Trail}\\
... you can reach me via email at \\ \hl{phillip-martinez@outlook.com}\\ or by phone at \hl{+86 19144 1648}.\end{tabular} \\ \hline
\end{tabular}
\end{table}

\noindent
\textbf{Experimental setup and methodology.} We perform our analysis on two datasets: CustomerSim~\cite{shi-etal-2022-selective}, a simulated dialog dataset for conversation generation and SynBio (originally called PII)~\cite{pii}, an LLM generated dataset representing student biographies containing personal identifiable information. Table \ref{tab:snippets} shows some excerpts from the datasets. We use four open-source models during evaluation:  \texttt{Pythia-1B}~\cite{biderman2023pythia}, \texttt{Gemma-2B}~\cite{team2024gemma}, \texttt{Llama2-7B}~\cite{touvron2023llama}, and \texttt{Qwen2.5-7B}~\cite{yang2024qwen2}

We leverage two tools for annotating sensitive information in a given text: Presidio~\cite{MsPresidio}, which helps in identification of private entities in text, and GPT-4 \cite{achiam2023gpt}, which is provided with a particular prompt for returning the annotated portions. An example of such a prompt for annotating samples is provided in Appendix~\ref{appendix:example-priv-annotatation}.

\begin{figure}[!h]
    \centering
     \includegraphics[scale=0.28]{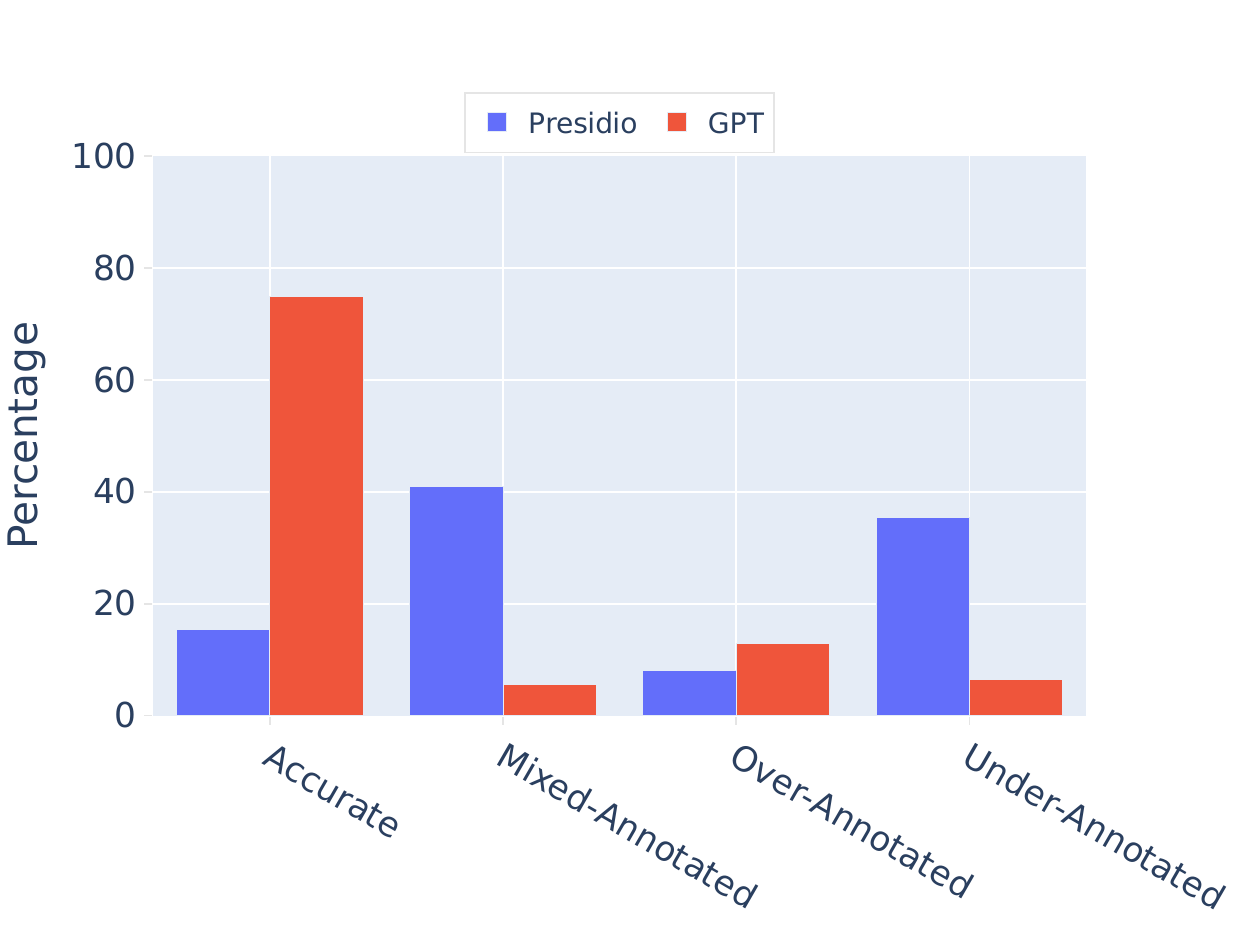}
    \caption{
    GPT-4 shows higher annotation accuracy, with 75\% of participants rating its annotations to be accurate while Presidio annotations were mostly mixed or under-annotated.
    }
    \label{fig:survey_results}
\end{figure}

We run two surveys, each among 40 Prolific\footnote{\url{https://www.prolific.com}} users, to gauge the usefulness of the two tools.
We provide the details of the survey in Appendix~\ref{appendix:human-survey-priv-annotatations}.
Figure~\ref{fig:survey_results} shows the results on the CustomerSim dataset which depicts that 75\% participants found GPT4's annotations to be accurate while Presidio annotations were mostly mixed or under-annotated.
\textit{Hence, throughout the rest of the paper we show our results using GPT-4 annotations}, and those using Presidio annotations are shown in Appendix~\ref{appendix:exploring-tradeoffs}.

To summarize, the degree of difference in computing privacy and utility using our measure and the existing one depends on the ratio of sensitive to non-sensitive tokens. A higher ratio would result in a higher difference in the measure, and vice versa. Considering the distinction between sensitive and non-sensitive tokens, we show that the existing measure can both exaggerate privacy threats and underestimate the utility in LLMs. 
In this context, we re-examine a prior study~\cite{biderman2024emergent} to better support our claim that reported privacy threats are exaggerated.

\begin{table}[]
\caption{
Examples of memorized sequences from \cite{biderman2024emergent}, often containing predictable and non-sensitive patterns, like mathematical series and licensing text.
}
\label{tab:memseq}
\scriptsize
\centering
\scalebox{0.9}{
\begin{tabular}{p{0.5\columnwidth}|p{0.5\columnwidth}}
    Prompt  & Generation \\
    \toprule
264. 
  265. 
  266. 
  267. 
  268. 
  269. 
  270. 
  271. & 272. 
  273. 
  274. 
  275. 
  276. 
  277. 
  278. 
  279. \\
  \midrule
active.disabled:focus,
.datepicker table tr td.active.disabled:hover:focus,
.datepicker table tr td.active:active, & .datepicker table tr td.active:hover:active,
.datepicker table tr td.active.disabled:active,
.datepicker table tr td \\
\midrule
$\langle$rel=``Chapter" href=``Char.html"$\rangle$$ \langle$link title=``Clflags" rel=``Chapter" href=``Clflags.html"$\rangle$ $\langle$ & link title=``Complex" rel=``Chapter" href=``Complex.html"$\rangle$ $ \langle$link title=``Condition" rel=``Chapter" href=``Condition.html"$\rangle$ \\

\midrule
    amp amp amp amp amp amp amp amp  amp amp amp amp amp amp amp amp & amp amp amp amp amp amp amp amp amp  amp amp amp amp amp amp amp  \\

\midrule
	.word 0
	.word 0
	.word 0
	.word 0
	.word 0
	.word 0 & .word 0
	.word 0
	.word 0
	.word 0
	.word 0
	.word 0
	.word \\
  \bottomrule
\end{tabular}}
\end{table}

\paragraph{\textbf{II. Examining memorized sequences from \cite{biderman2024emergent}:}}
We consider a case study to analyze the reported memorized strings by~\cite{biderman2024emergent}, 
Our goal is to examine whether the memorized strings contain sensitive information or mere syntactic and semantic patterns.

\textbf{Experimental setup.} The authors in \cite{biderman2024emergent} considered the task of predicting whether a model memorizes specific training data points from the Pile dataset~\cite{gao2020pile}, which is used to train base LLM models. Among published memorized strings, we randomly choose $5,000$ strings from the \textit{pythia-1b-dup} split~\cite{biderman2024emergent}. A representative list of memorized strings is in Table~\ref{tab:memseq}, where the strings often follow syntactic and semantic patterns, such as completion of mathematical series, code snippets, licensing agreements, etc. Therefore, \textit{our hypothesis is that most of the memorized strings contain a great amount of non-sensitive and highly predictable tokens.} To validate our hypothesis, we query for the source of memorized strings with respect to the training dataset, Pile, which aggregates data from multiple sources such as Pile-Cc, OpenWebText, ArXiv, etc. We leverage GPT-4 model to accomplish our task -- given a memorized string, we ask for the source of the string from the list of Pile sections. The prompt template for GPT-4 is the following.

\noindent
\begin{mdframed}[backgroundcolor=lavender, linewidth=0pt]
\small
\textit{You are provided with the following text: \{\textcolor{blue}{memorized--sequence}\}.\newline
Which section of the Pile dataset does the text belong to? Choose from the list below. You can select 1 or 2 options separated by a comma. Please respond with only the option number.\newline
a. Pile-CC
b. PubMed Central
c. Books3 \\
d. OpenWebText2
e. ArXiv
f. GitHub
g. FreeLaw \\
h. Stack Exchange
i. USPTO Backgrounds \\
j. PubMed Abstracts
k. Gutenberg (PG-19) \\
l. OpenSubtitles
m. Wikipedia (en) \\
n. DM Mathematics
o. Ubuntu IRC \\
p. BookCorpus2
q. EuroParl 
r. HackerNews \\
s. YoutubeSubtitles 
t. PhilPapers \\
u. NIH ExPorter
v. Enron Emails}
\end{mdframed}

\begin{figure}[!h]
    \centering
        \subfloat{
       \includegraphics[scale=0.22]{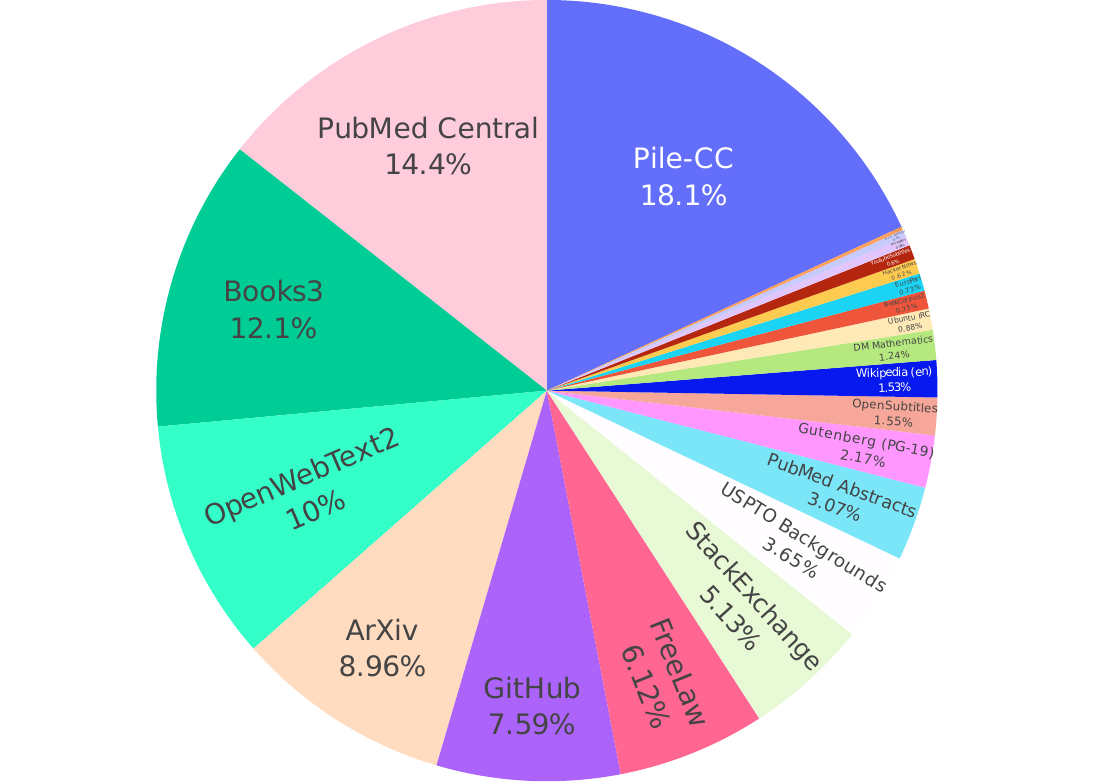}
        }
        \subfloat{
       \includegraphics[scale=0.22]{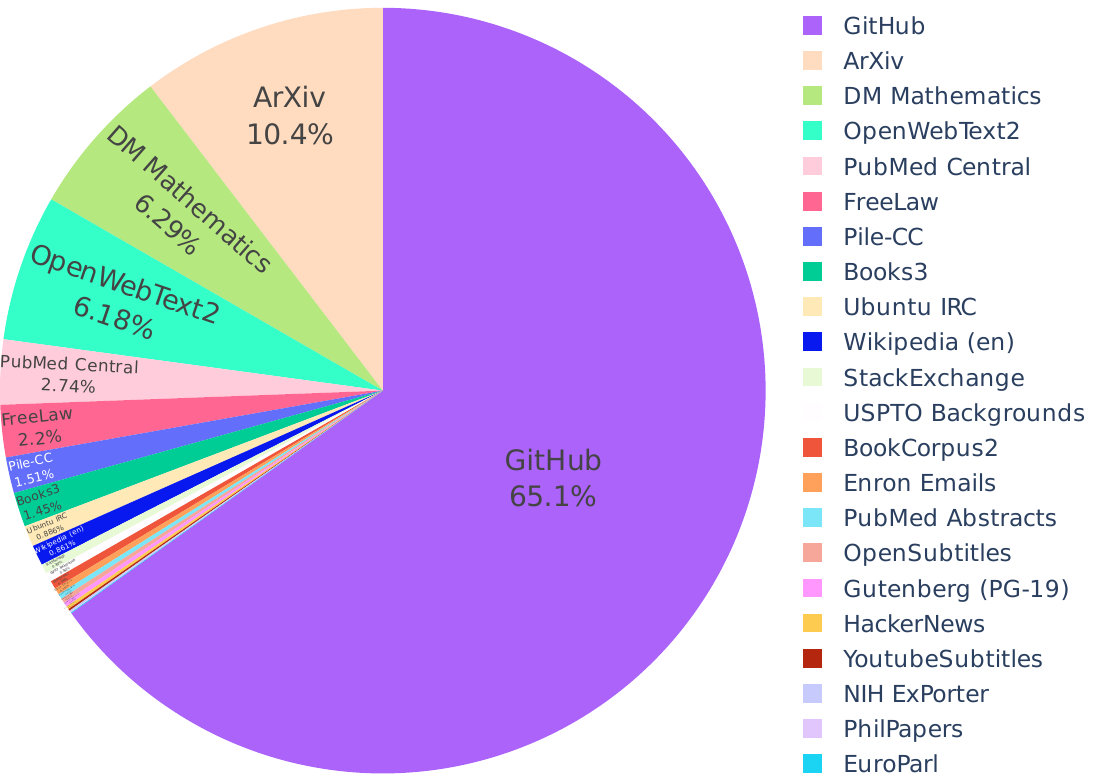}
        }
    \caption{
    Memorized sequences are predominantly sourced from GitHub and ArXiv, despite these sections being mid-range in the original Pile dataset, suggesting that memorized content is largely non-sensitive and may pose a lower privacy risk than previously assumed.
    }
    \label{fig:pile-data}
\end{figure}

\textbf{Results.} In Figure~\ref{fig:pile-data}, we present two pie charts illustrating the distributions across $22$ distinct sections or data sources within the Pile dataset. The left chart represents the original content distribution of sections within the Pile dataset, while the right chart depicts the distribution of sources of memorized sequences as predicted by GPT-4. 

In the right chart in Figure~\ref{fig:pile-data}, the memorized strings are predicted mostly from GitHub, followed by ArXiv, while the rest of the sources are largely under-represented. Herein, both GitHub and ArXiv are relatively in the middle range in terms of contents in the original dataset in the top chart.  However, analyzing the typical data in these sections, GitHub appears as a source of structured format code with repeated predictable patterns, which is commonly tagged as non-sensitive data. Similarly, the Pile dataset includes {\LaTeX} files uploaded to ArXiv, since {\LaTeX} is a common typesetting language for scientific research papers~\cite{gao2020pile}. As such, highly memorized strings in the Pile dataset are non-sensitive in nature.

\begin{figure}[!t]
    \centering
        \includegraphics[scale=0.35]{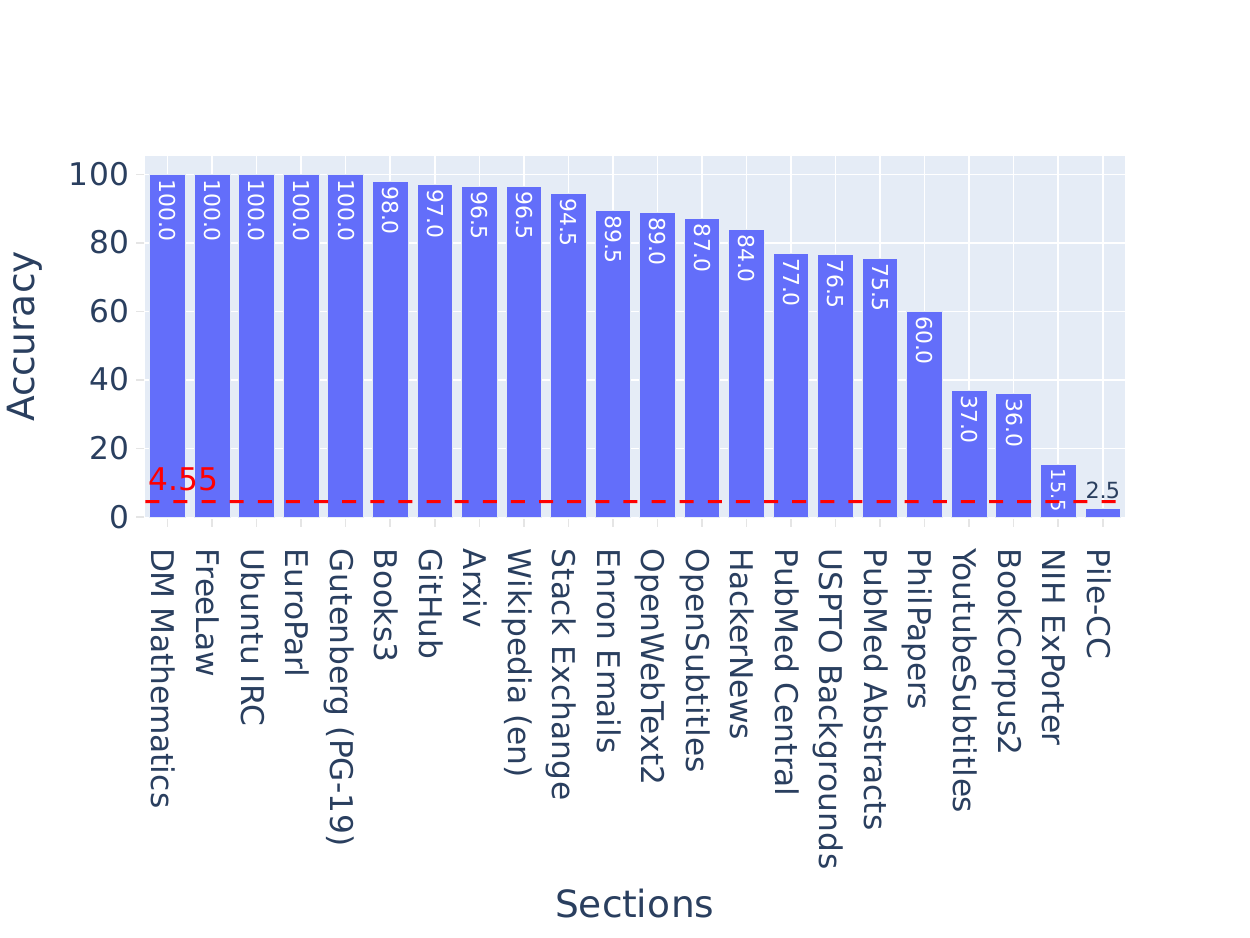}
    \caption{
    GPT-4 achieves an average accuracy of 78\% in predicting the source of memorized strings across Pile dataset sections,
    reinforcing the reliability of GPT-4 and supporting our position that privacy concerns in prior work are overestimated without distinguishing token sensitivity.
    }
    \label{fig:pile-data-verify}
\end{figure}

\textbf{Validating GPT-4 predictions.} GPT-4 predictions may be erroneous. Hence, we conduct a verification test to evaluate the accuracy of GPT-4's predictions. For this assessment, we sample $200$ random strings from each of the 22 sections of the Pile dataset \cite{gao2020pile}, and prompt GPT-4 to predict the source of the strings. Unlike the previous experiment, \textit{the ground-truth of string source is known in this validation experiment}. Figure~\ref{fig:pile-data-verify} illustrates the accuracy for each section, indicating that $50\%$ of the sections exhibit an accuracy rate of at least $90\%$ with $4.5\%$ being the base accuracy of a random predictor.
Furthermore, GPT-4 predicts the correct source on an average of $78\%$ strings across all $22$ sections of the Pile dataset. In addition, misclassified strings are often assigned to sections of a similar category, e.g., \textit{NIH Explorer misclassified as PubMed Central} (more details in Appendix~\ref{appendix:validating-gpt4-preds}). \textit{Therefore, the GPT-4 predictions can be considered as reliable.}

\begin{figure}[!t]
    \centering
       \includegraphics[scale=0.3]{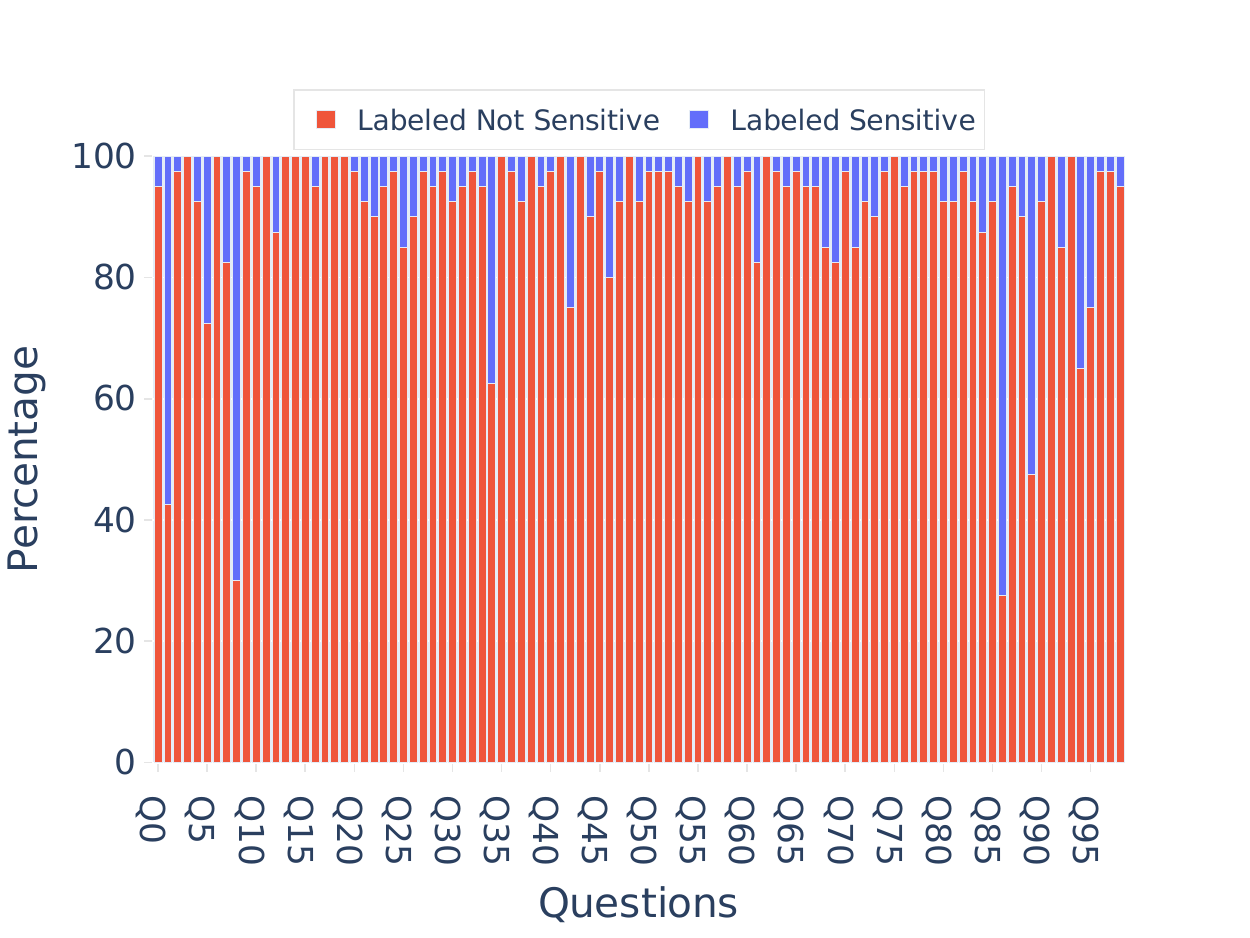}
    
    \caption{
    Most participants classified the memorized sequences detected by~\cite{biderman2024emergent} as non-sensitive, with fewer than 10\% marking them as privacy-sensitive, indicating that the perceived privacy risk of these strings is generally low.
    }
    \label{fig:human-survey-mem-txt}
\end{figure}

Finally, we conduct a human survey on Prolific to evaluate the extent of sensitive information present in  randomly chosen $100$ memorized strings from \cite{biderman2024emergent}. 
The survey results are summarized in Figure~\ref{fig:human-survey-mem-txt}. The majority of participants classified memorized strings as non-sensitive, while  $< 10\%$ participants disagree and mark the strings as containing privacy-sensitive information.
\textit{Therefore, most crowdsourced participants do not perceive the sampled stings as containing privacy-sensitive content.} 
Further details on the setup are provided in Appendix~\ref{appendix:human-survey-priv-annotatations}. Thus, by distinguishing between sensitive and non-sensitive entities, we demonstrate a deeper understanding of actual privacy threat.
\section{Privacy-Utility-Efficiency Interplay}
\label{sec:trade_off}

\begin{figure}[t]
    \centering
    \begin{subfigure}{0.49\linewidth}
    \centering
    \includegraphics[scale=0.22]{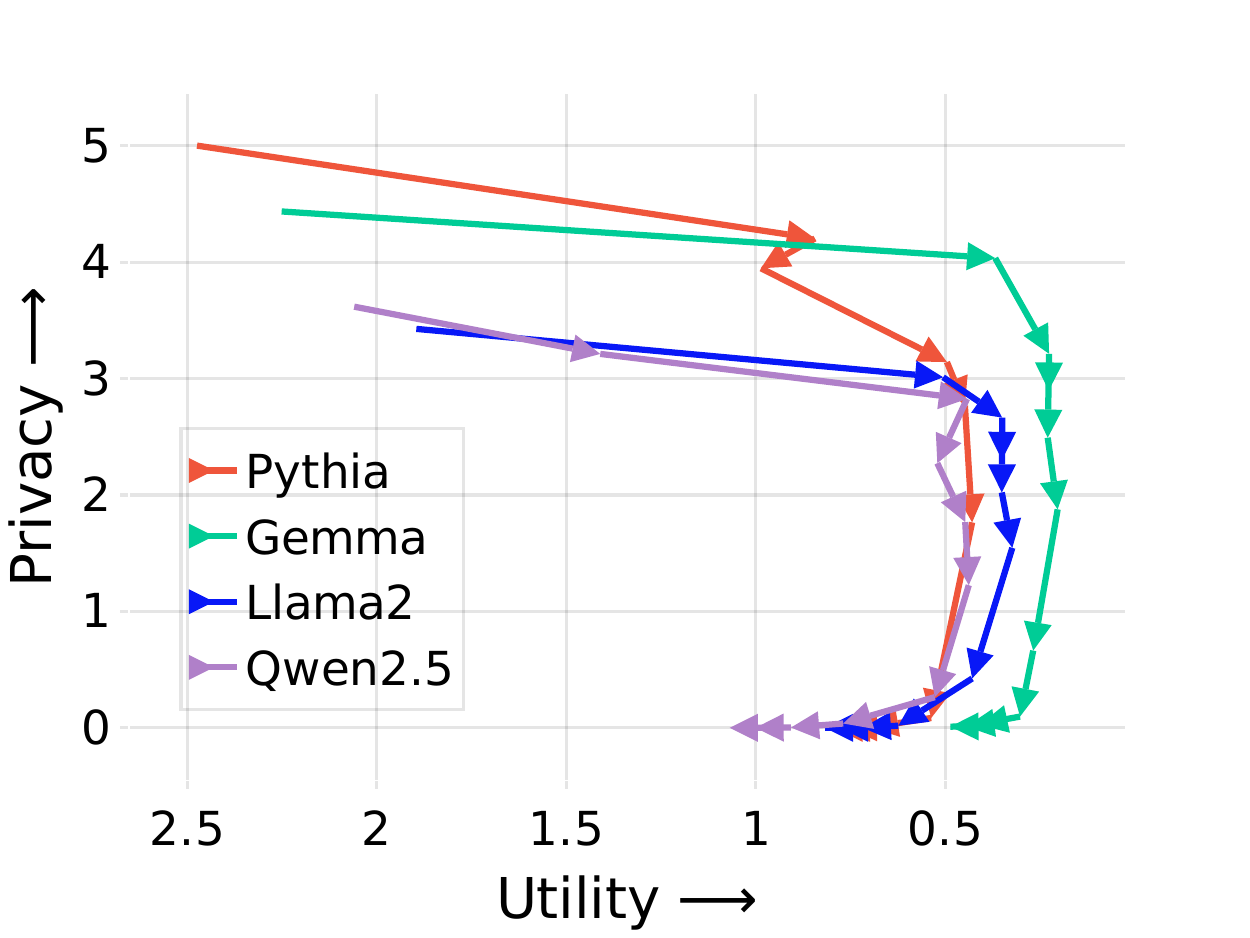}
    \caption{CustomerSim}
    \label{fig:ffta}
    \end{subfigure}
    \begin{subfigure}{0.49\linewidth}
    \centering
    \includegraphics[scale=0.22]{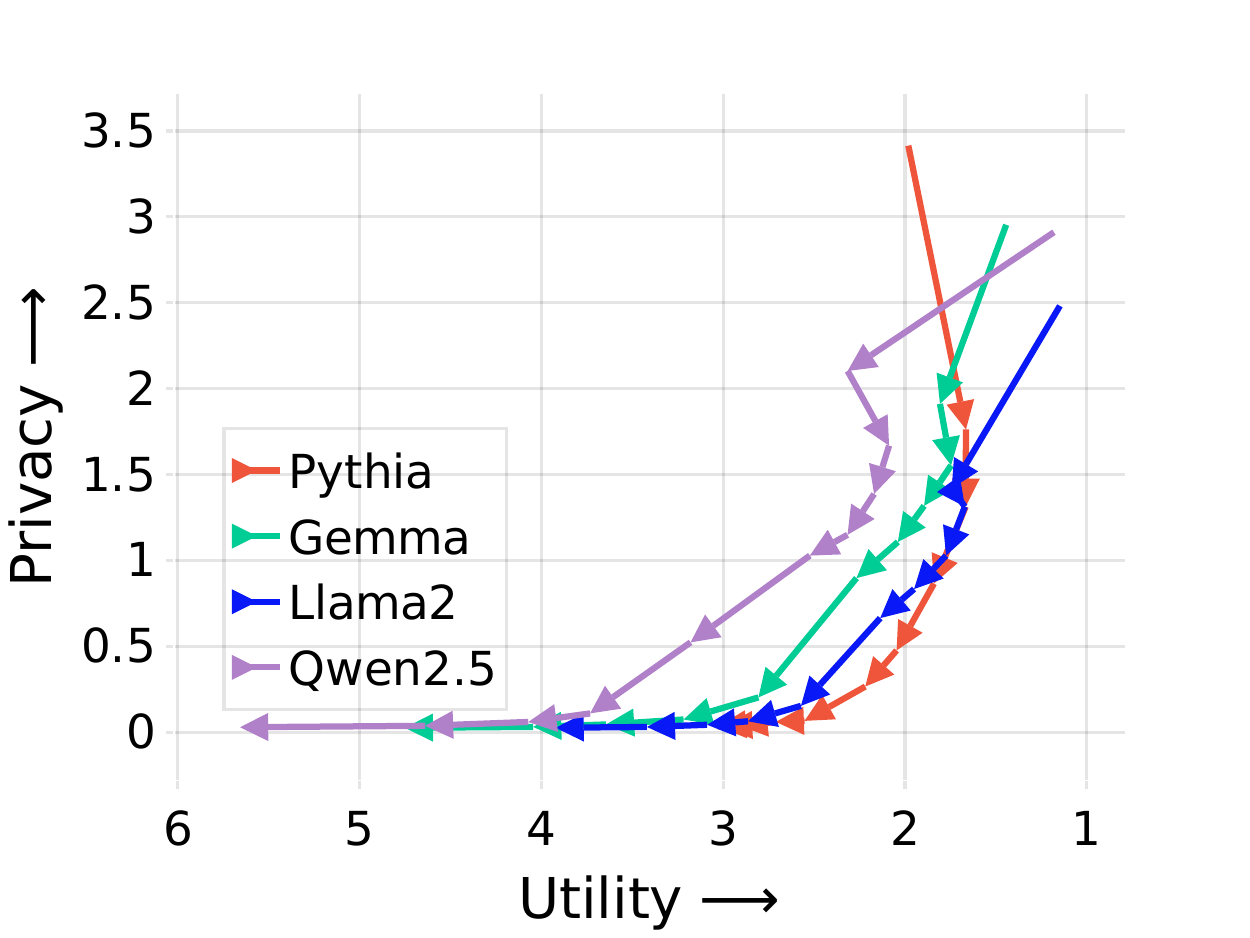}
    \caption{SynBio}
    \label{fig:fftb}    
    \end{subfigure}
    \caption{
    Privacy-utility trade-off shows that privacy increases with higher training loss on sensitive tokens, while utility improves with lower test loss on non-sensitive tokens, enabling desired checkpoint selection to balance both objectives.
    }
    \label{fig:fft}
\end{figure}

\begin{figure*}[h!]
    \centering
        \begin{subfigure}{.32\linewidth}
       \includegraphics[scale=0.25,height=3.5cm]{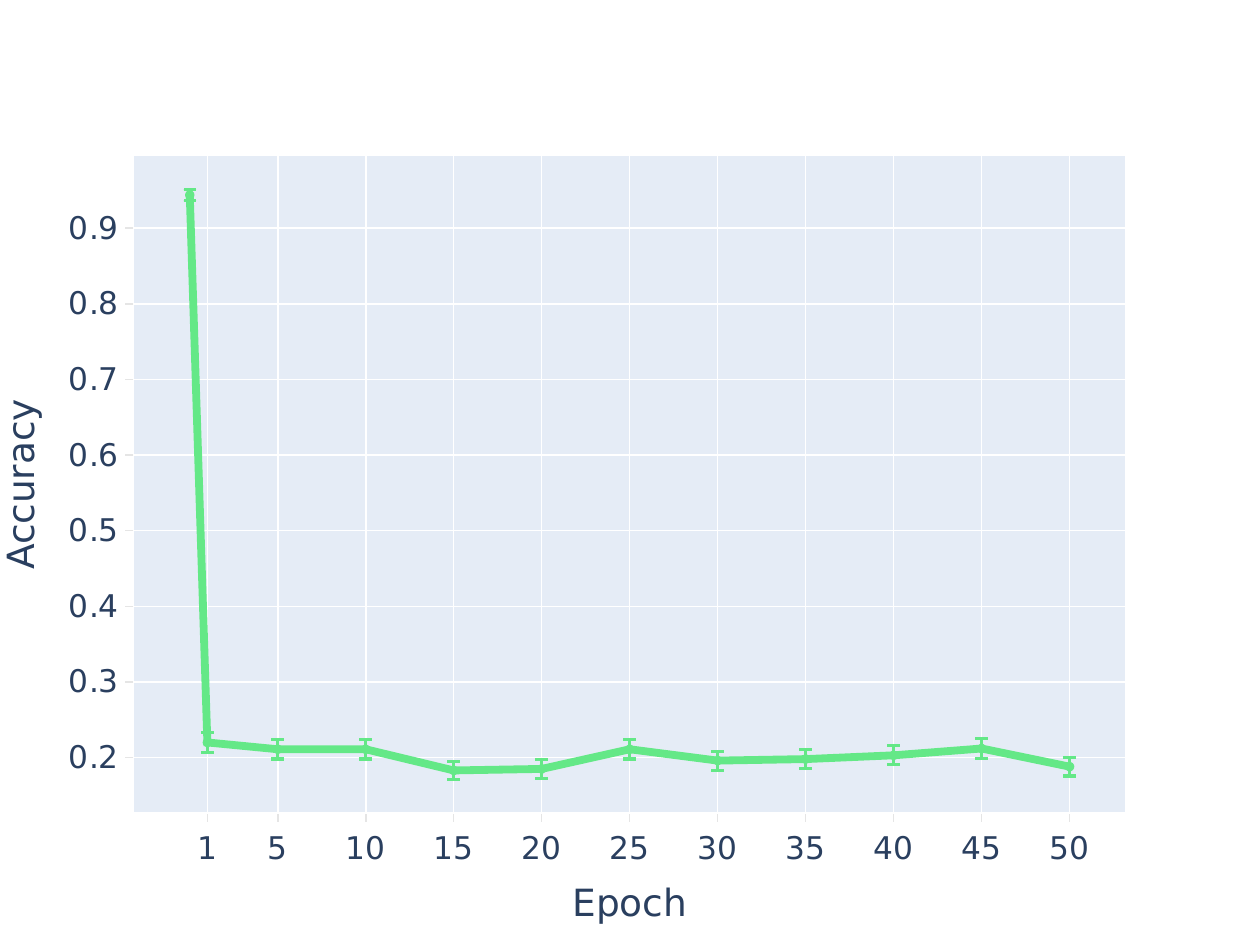}
        \caption{SCIQ Benchmark}
        \label{fig:sciq_fft}
    \end{subfigure}
    \begin{subfigure}{.32\linewidth}
       \includegraphics[scale=0.25,height=3.5cm]{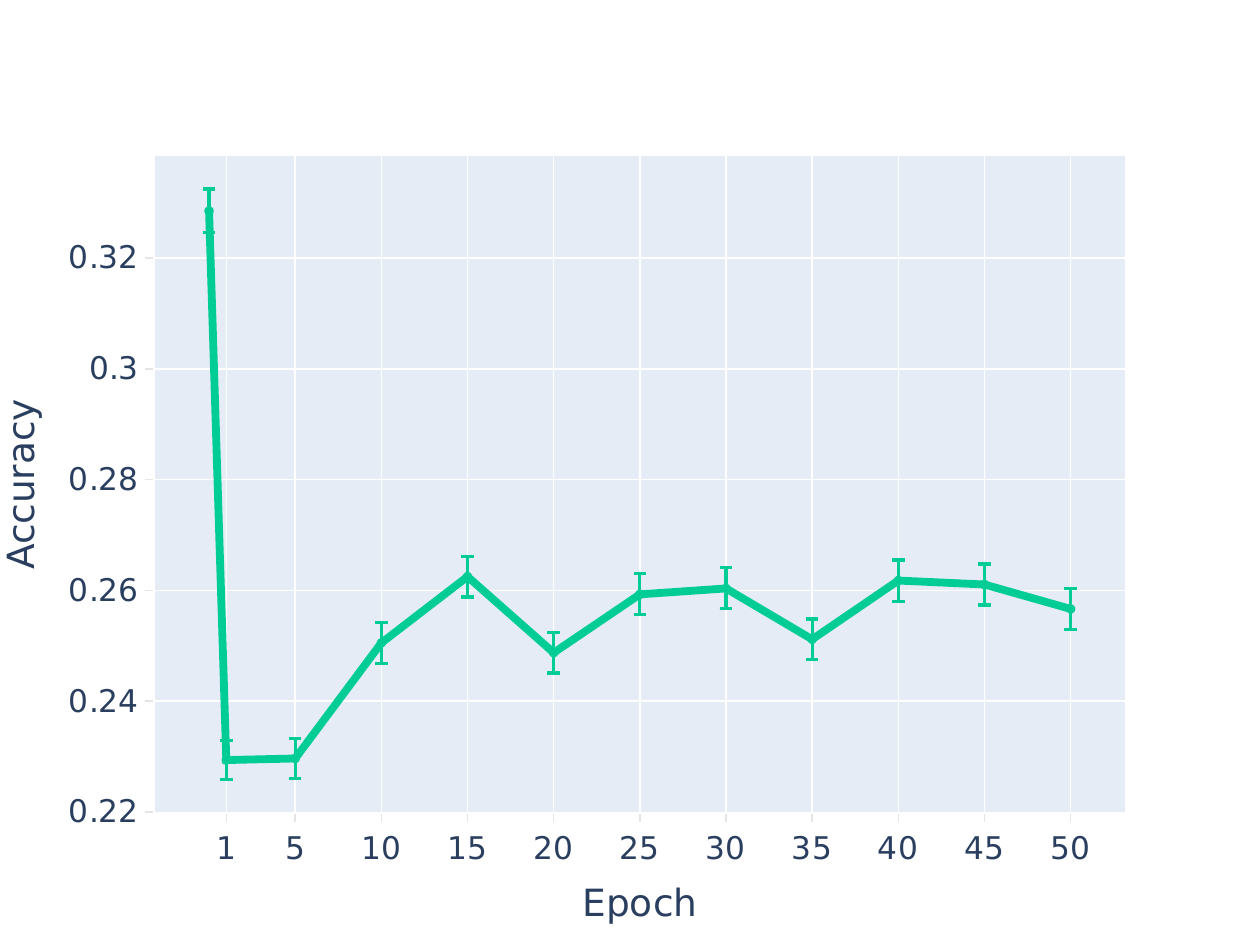}
        \caption{MMLU Benchmark}
        \label{fig:mmlu_fft}
    \end{subfigure}
     \begin{subfigure}{.32\linewidth}
       \includegraphics[scale=0.25,height=3.5cm]{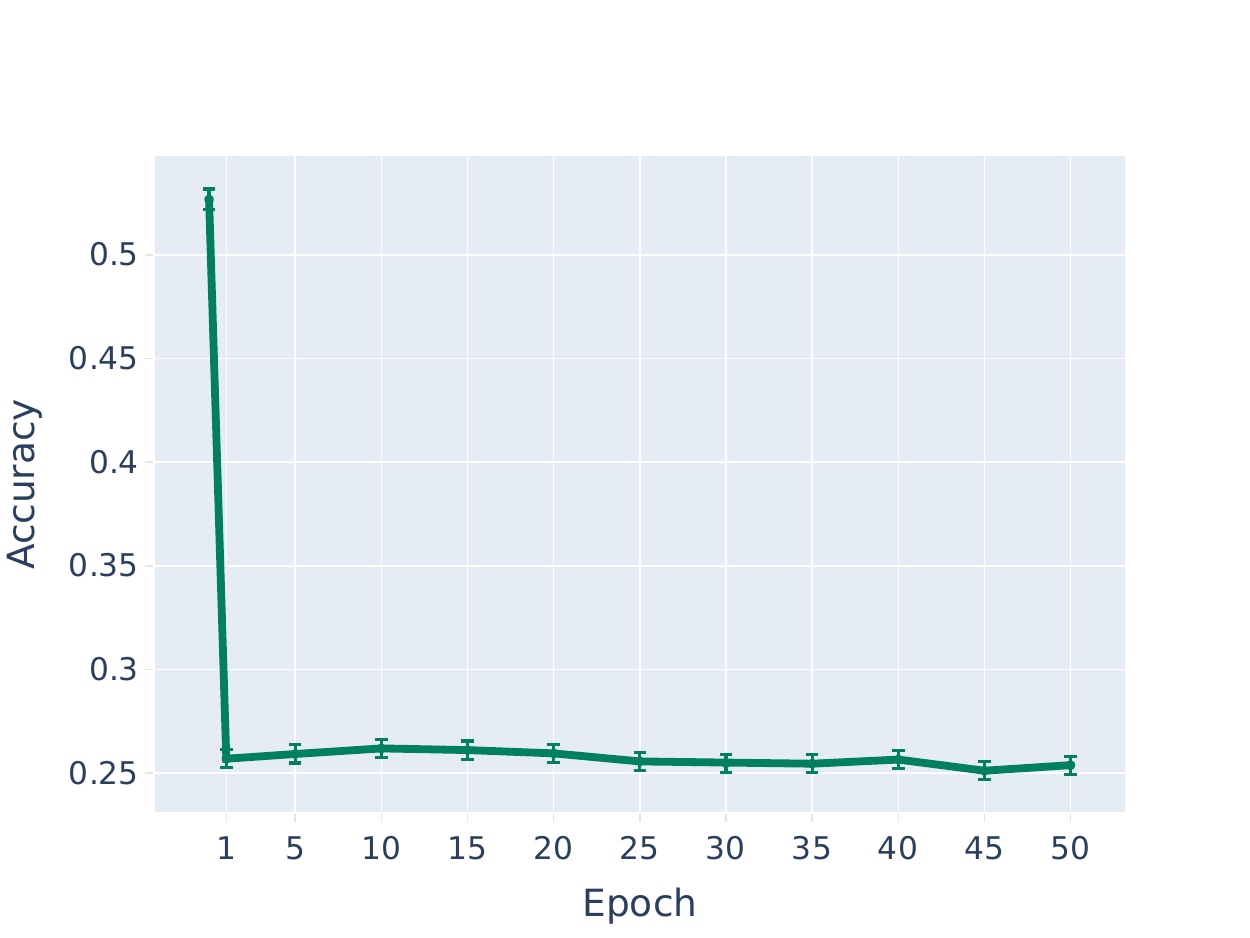}
        \caption{HellaSwag Benchmark}
        \label{fig:hs_fft}
    \end{subfigure}
    \caption{
    Full fine-tuning of the Gemma model leads to a significant drop in accuracy on benchmark datasets, with declines of approximately 75\%, 9\%, and 30\% for SCIQ, MMLU, and HellaSwag, respectively.
    }
    \label{fig:bench-fft}
\end{figure*}

We use the distinction between sensitive and non-sensitive tokens to study the privacy and utility impact of training models with three different fine-tuning methods: full fine-tuning (FFT), Differentially Privacy (DP-SGD), and Low-Rank Adaptation (LoRA).
We also investigate the computational efficiency of each method.
Our goal is to answer the following questions: ``\textit{How prone is each method to recollecting the sensitive parts of the training data? (Privacy)'' , ``How effective is each method at predicting non-sensitive parts of test data? (Utility)'' , ``What is the computational cost associated with each method? (Efficiency})''.

To answer these questions, we use each fine-tuning method to train the four models -- \texttt{Pythia-1B}~\cite{biderman2023pythia}, \texttt{Gemma-2B}~\cite{team2024gemma}, \texttt{Llama2-7B}~\cite{touvron2023llama}, and \texttt{Qwen2.5-7B}~\cite{yang2024qwen2} on two datasets, CustomerSim and SynBio.
More information about the datasets and our methodology for distinguishing between sensitive and non-sensitive tokens can be found in Section~\ref{sec:sens_non_sens_distinction}.
We train the models for 50 epochs on each dataset.
Details on hyperparameters can be found in Appendix~\ref{appendix:hp-models}.

For each fine-tuning method, model and dataset we report three metrics: 1) privacy as the loss on sensitive tokens (annotated by GPT-4) in the training data, 2) utility as the loss on non-sensitive tokens (the remaining tokens) on a held-out test set, and 3) efficiency based on the relative amount of computation and memory usage of each method.
Additionally, to assess how fine-tuning affects the underlying abilities and knowledge of the base model, we measure the performance of the fine-tuned Gemma model (trained with CustomerSim data) on general language understanding benchmarks: SCIQ \cite{SciQ}, a dataset of over 13,000 crowdsourced questions on Physics, Chemistry, and Biology; MMLU \cite{hendryckstest2021, hendrycks2021ethics}, a large multi-task dataset covering various domains of knowledge; and Hellaswag \cite{zellers-etal-2019-hellaswag}, a dataset for commonsense inference.

\noindent
\textbf{Update rules:}
For each fine-tuning method, we describe how updated weights $W_{t + 1}$ are computed from the previous weights $W_t$ in each step, where $W_0$ are the weights of the pre-trained base model before fine-tuning.
We use $X$ to refer to a batch of $|X|$ datapoints and $x_i$ to refer to individual datapoint, $\mathcal{M}_{W}$ to refer to the model parameterized by weights $W$, and $\mathcal{L}(\mathcal{M}_{W}(X), X)$ to refer to the autoregressive cross-entropy loss of the model on data $X$.
We denote by $\nabla_{W} \mathcal{L}(...)$ as the gradient of the loss wrt. weights $W$ and $\eta$ is the learning rate.

\noindent
\textbf{Efficiency:}
The efficiency of each method is determined by the amount of computation it requires, and also other factors such as memory requirements, which can affect the usable batch-size and thus the overall training throughput.
Following~\cite{kaplan2020scaling}, we estimate the amount of training compute ($C$) in floating point operations (FLOPs) for full fine-tuning as $C_{\text{FFT}} = 6 D N$, where $D$ is the number of training tokens and $N$, the number of model parameters.
For each method, we report its compute requirements relative to the FFT-baseline based on measurements using the PyTorch profiler\footnote{\url{https://pytorch.org/docs/stable/profiler}}.
We also comment on other factors that affect training throughput.

\subsection{Full fine-tuning for utility costs privacy}

\textbf{Update rules:}
Full fine-tuning (FFT) updates all model parameters at each step:
\begin{equation}
    W_{t + 1} = W_t - \eta \nabla_{W_t} \mathcal{L}(\mathcal{M}_{W_{t}}(X), X)
\end{equation}

\noindent
\textbf{Privacy-Utility trade-off:}
Figures~\ref{fig:ffta} and~\ref{fig:fftb} show the \textit{privacy-utility trade-off} for the CustomerSim and SynBio datasets, respectively.
In these figures, privacy increases with the training loss on sensitive tokens (\textit{up $\Uparrow$ on the y-axis}), while utility increases when the test loss on non-sensitive tokens decreases (\textit{right $\Longrightarrow$ on the x-axis}).
Each curve starts with the baseline performance of the pre-trained model.
For CustomerSim (Figure~\ref{fig:ffta}), as training advances (\textit{denoted by an arrow $\rightarrow$} on the lines), privacy progressively decreases (\textit{lower on the y-axis}), while utility improves (\textit{rightward on the x-axis}) for approximately the first 5 epochs across all models before stabilizing and eventually declining (\textit{leftward on the x-axis}).
However, for SynBio (Figure \ref{fig:fftb}), the privacy-utility trade-off primarily worsens for Gemma and Llama models. On examining these curves, one can select a desired checkpoint that aligns with specified privacy and utility thresholds.

\noindent
\textbf{Impact on benchmark datasets:}
Figures \ref{fig:sciq_fft}, \ref{fig:mmlu_fft}, and \ref{fig:hs_fft} show the fully fine-tuned Gemma model's accuracy at each epoch for the three benchmarks: SCIQ, MMLU, and Hellaswag, respectively. Note that the accuracy corresponding to the first point represents the performance of the pre-trained model. We observe that full fine-tuning shows a substantial decline in accuracy (around $0.75$, $0.09$, and $0.3$ decrease in accuracy in SCIQ, MMLU, and HellaSwag,  respectively).

\noindent
\textbf{Efficiency:}
FFT serves as our efficiency baseline.
It has moderate compute requirements (discussed above), and relatively high memory requirements, since in addition to the input-dependent activations, we need to keep four numbers per model parameter in GPU memory: the parameter value, its gradient, and two optimizer states (first and second moments of the gradient for Adam \cite{kingma2014adam}).

\noindent
\textbf{Takeaway:}
FFT offers poor privacy-utility trade-offs, since gains in utility in most cases come at the cost of a significant loss in privacy.
During FFT, models learn to both predict the training distribution better, but also quickly learn to recollect sensitive tokens.
In addition, FFT deteriorates the base performance of the model, as can be seen by the rapid decline of the benchmark scores.
FFT is moderately efficient and has relatively high memory requirements. The degree of measures along the \textit{trade-off, knowledge retention, and efficiency} are: 

\indent Utility-privacy trade-offs: \textit{poor} \\
\indent Retention of base performance: \textit{poor} \\
\indent Efficiency: \textit{moderate}

\begin{figure*}[ht]
    \centering
    \begin{subfigure}[b]{0.24\linewidth}
        \includegraphics[width=\linewidth]{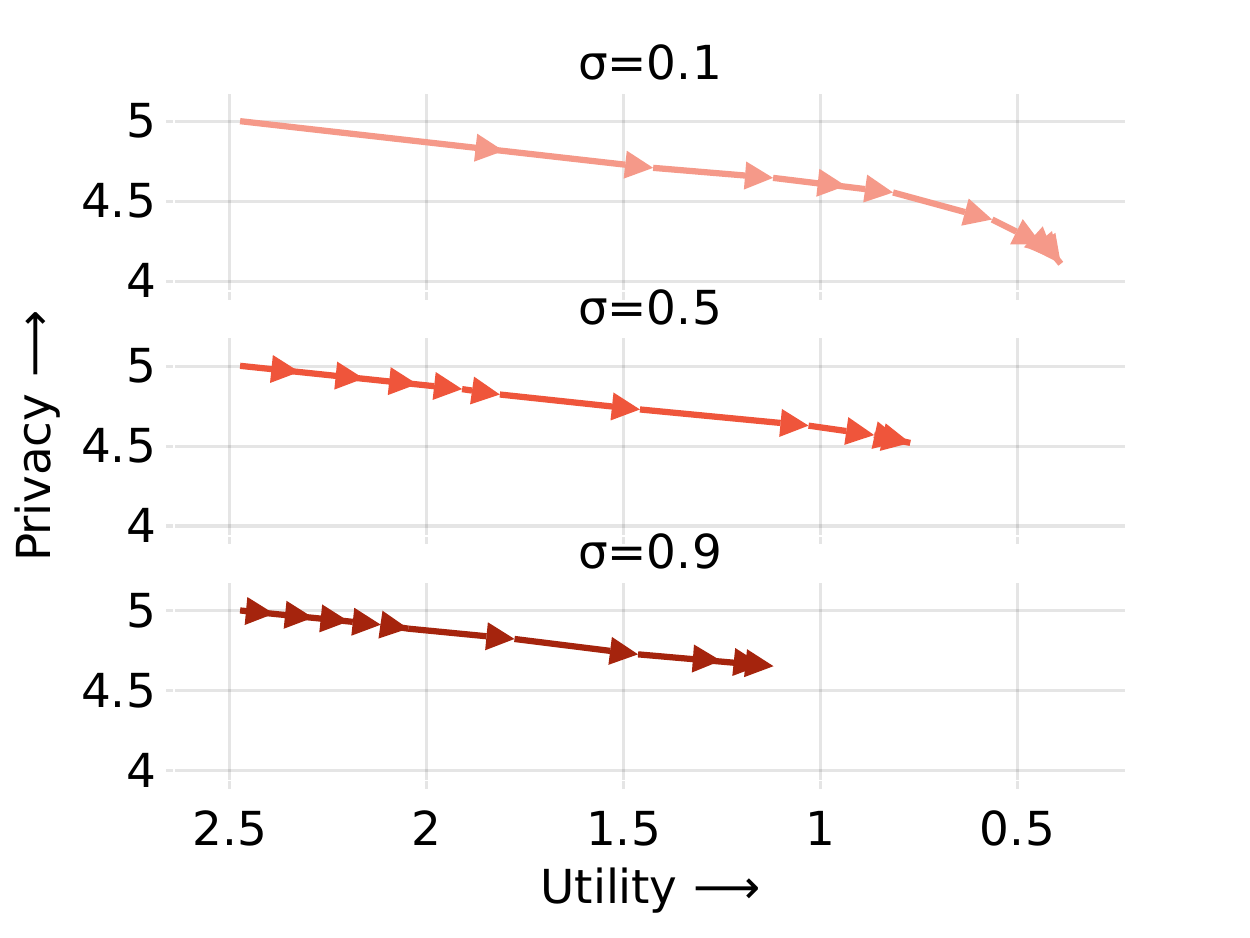}
        \caption{Pythia (CustSim)}
        \label{fig:dp_csima}
    \end{subfigure}
    \begin{subfigure}[b]{0.24\linewidth}
        \includegraphics[width=\linewidth]{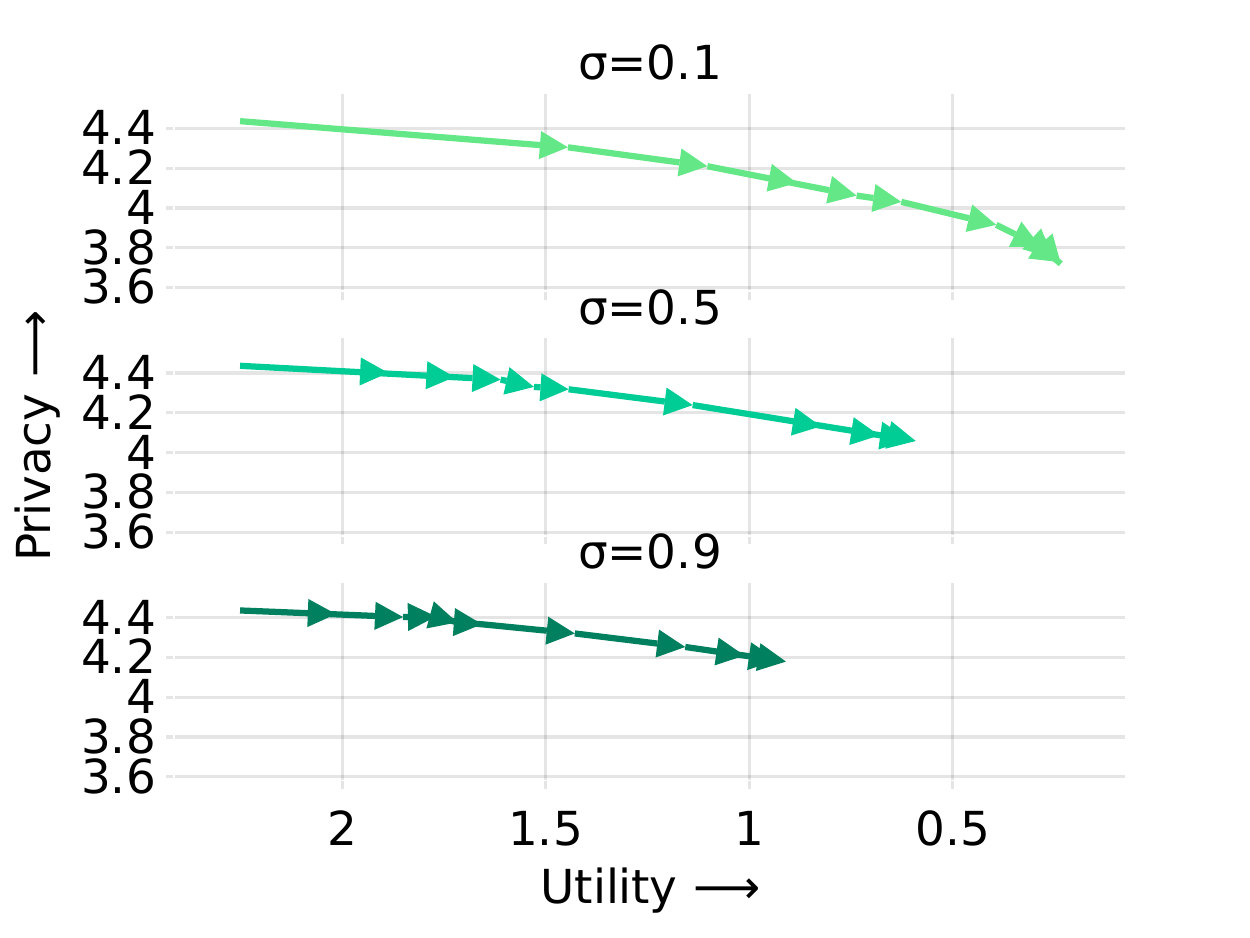}
        \caption{Gemma (CustSim)}
        \label{fig:dp_csimb}
    \end{subfigure}
    \begin{subfigure}[b]{0.24\linewidth}
        \includegraphics[width=\linewidth]{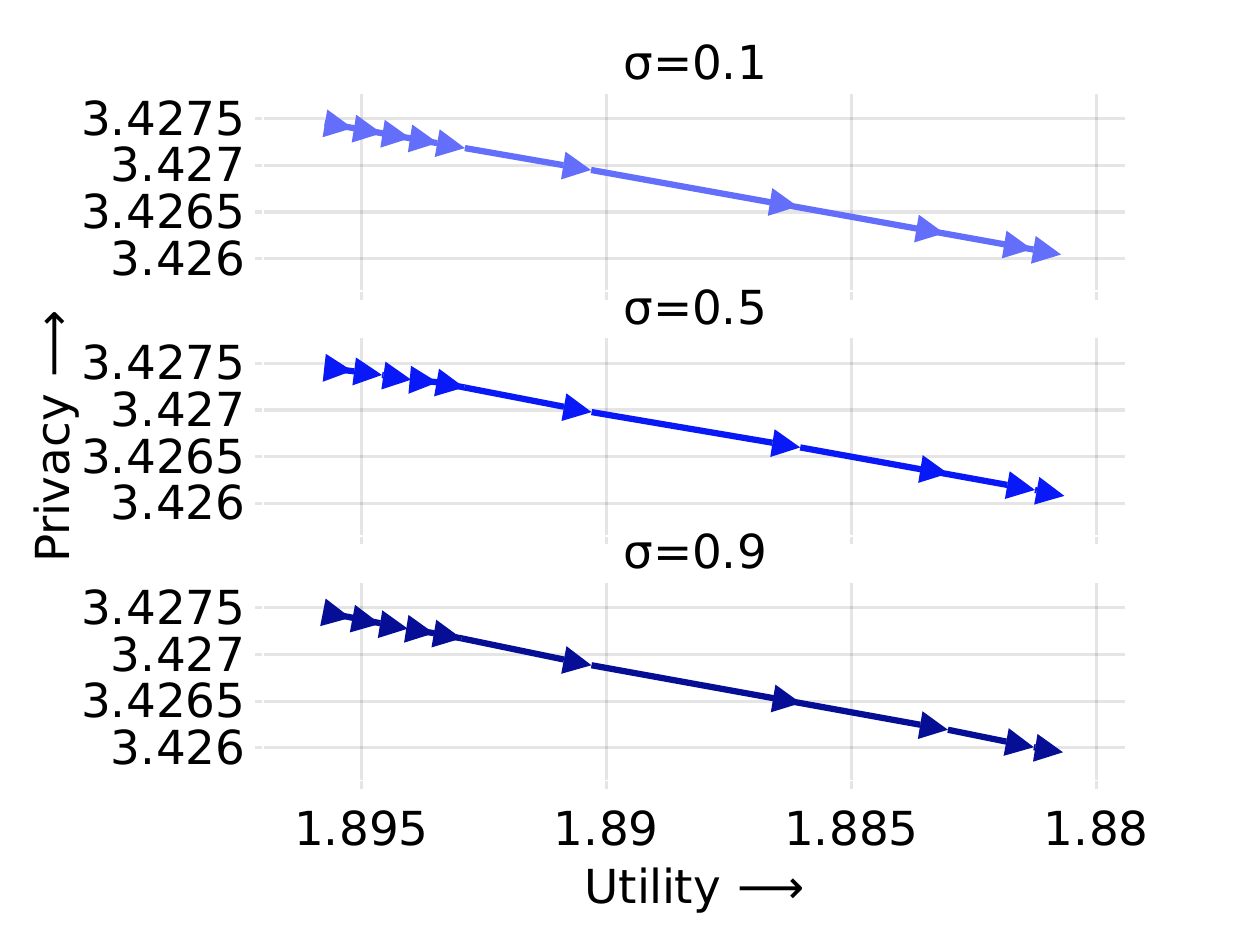}
        \caption{Llama2 (CustSim)}
        \label{fig:dp_csimc}
    \end{subfigure}
    \begin{subfigure}[b]{0.24\linewidth}
        \includegraphics[width=\linewidth]{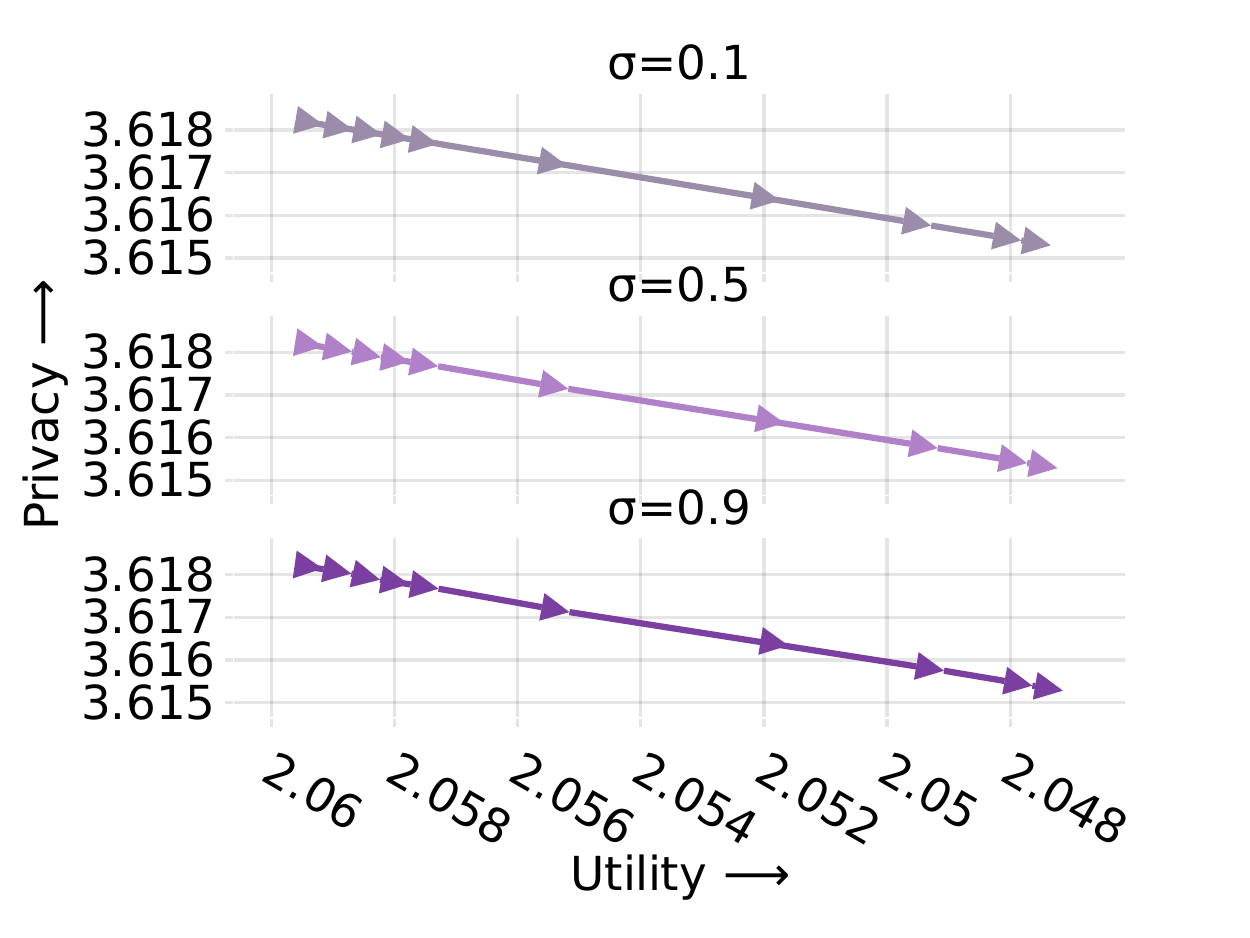}
        \caption{Qwen2.5 (CustSim)}
        \label{fig:dp_csimd}
    \end{subfigure}

    \begin{subfigure}[b]{0.24\linewidth}
        \includegraphics[width=\linewidth]{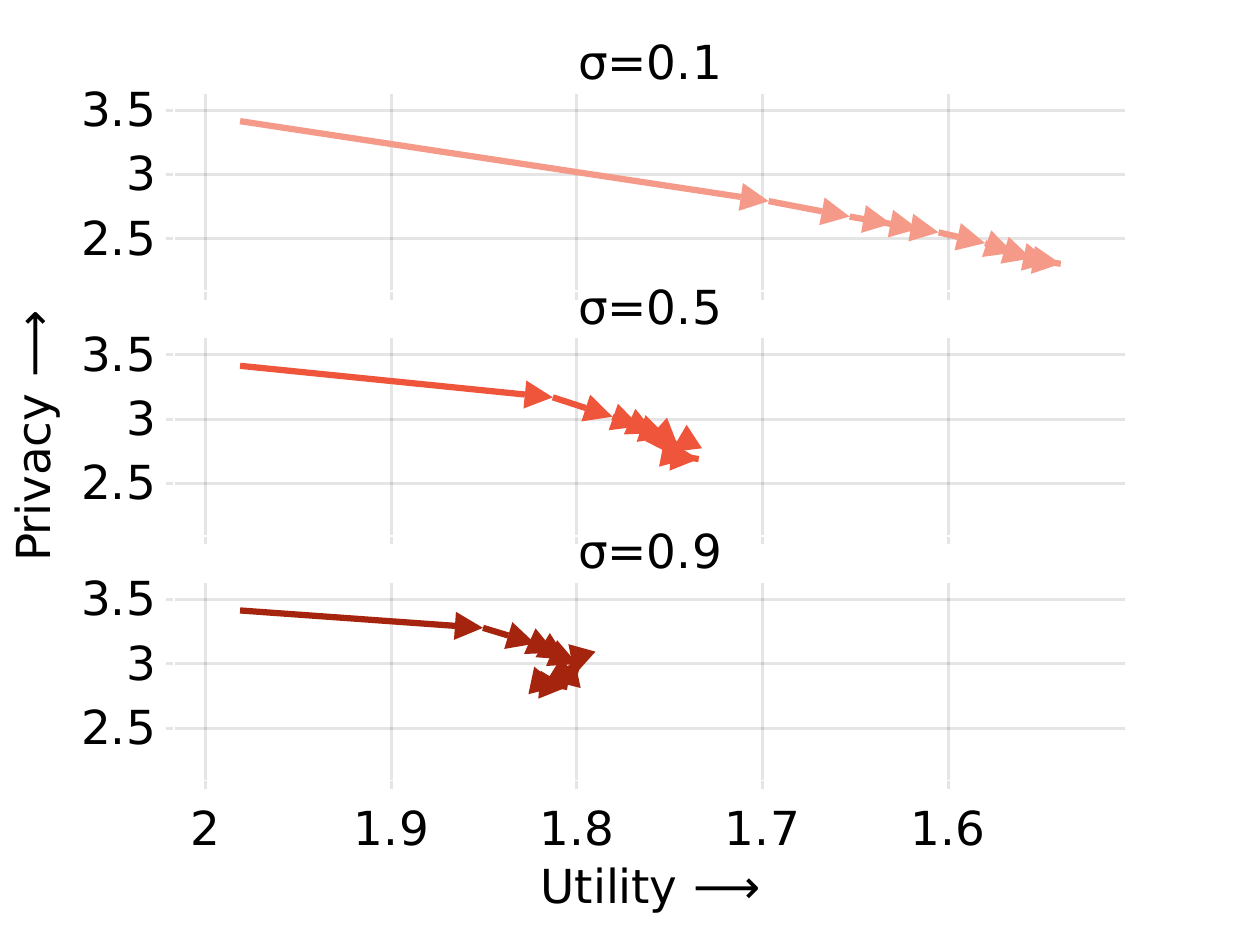}
        \caption{Pythia (SynBio)}
        \label{fig:dp_piia}
    \end{subfigure}
    \begin{subfigure}[b]{0.24\linewidth}
        \includegraphics[width=\linewidth]{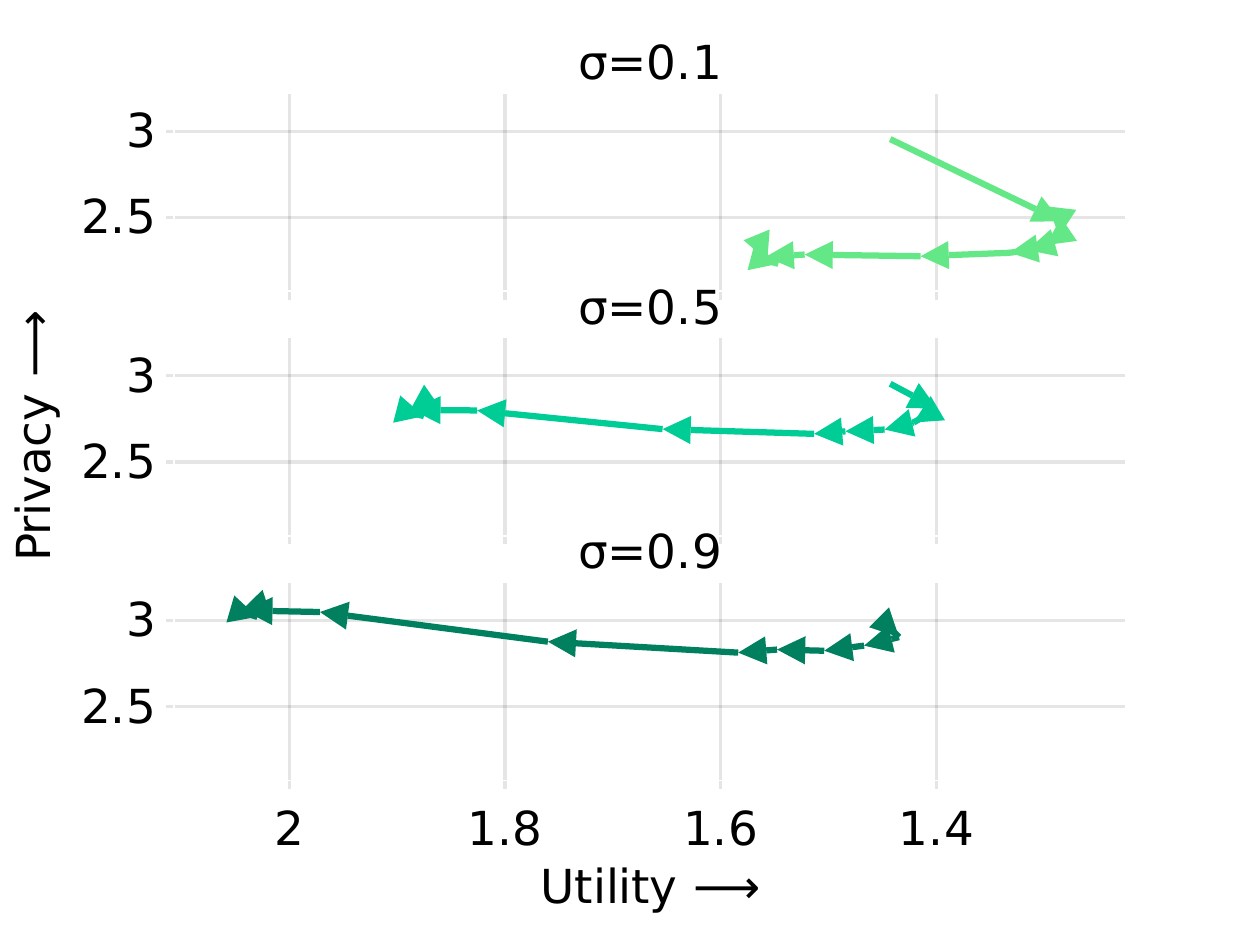}
        \caption{Gemma (SynBio)}
        \label{fig:dp_piib}
    \end{subfigure}
    \begin{subfigure}[b]{0.24\linewidth}
        \includegraphics[width=\linewidth]{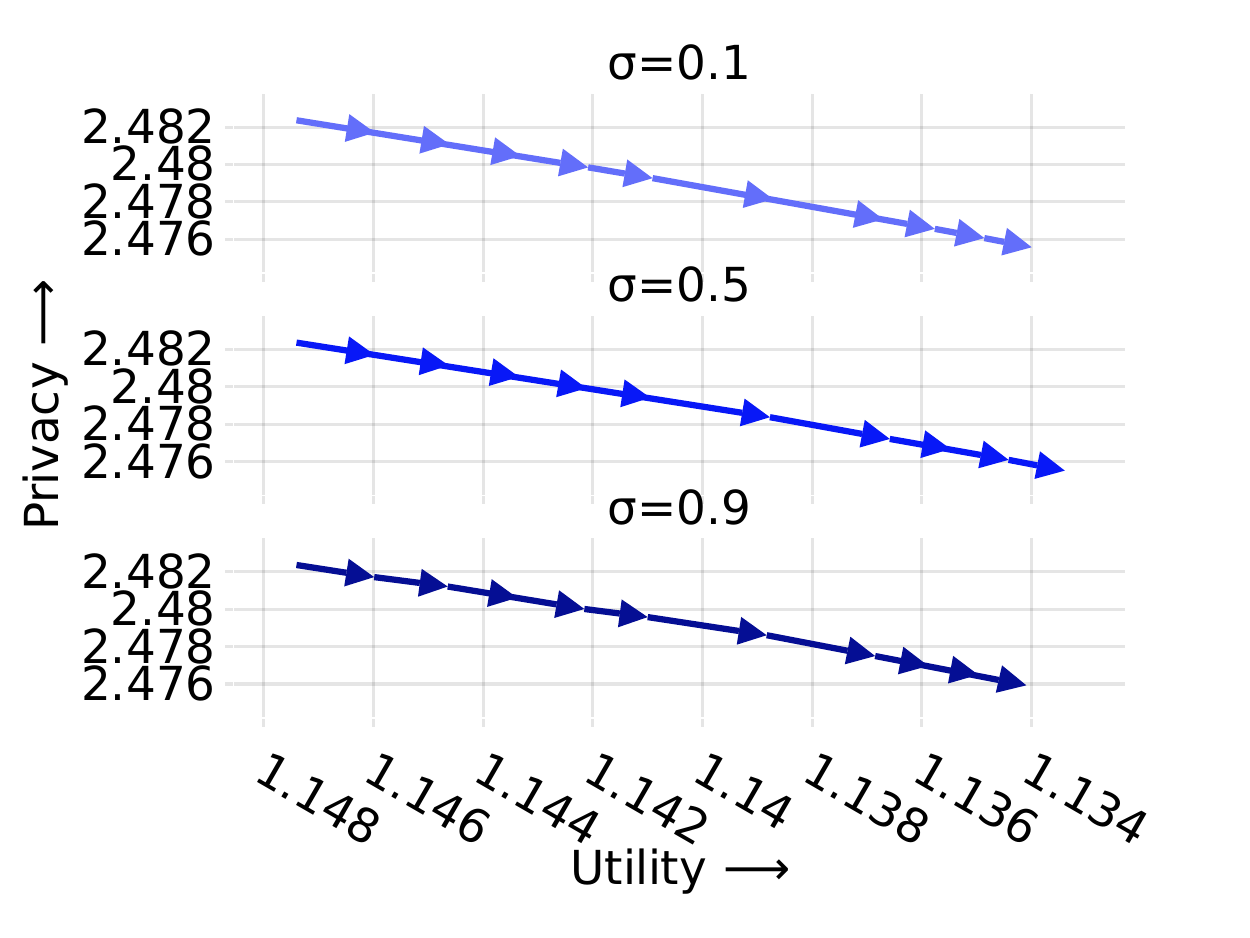}
        \caption{Llama2 (SynBio)}
        \label{fig:dp_piic}
    \end{subfigure}
    \begin{subfigure}[b]{0.24\linewidth}
        \includegraphics[width=\linewidth]{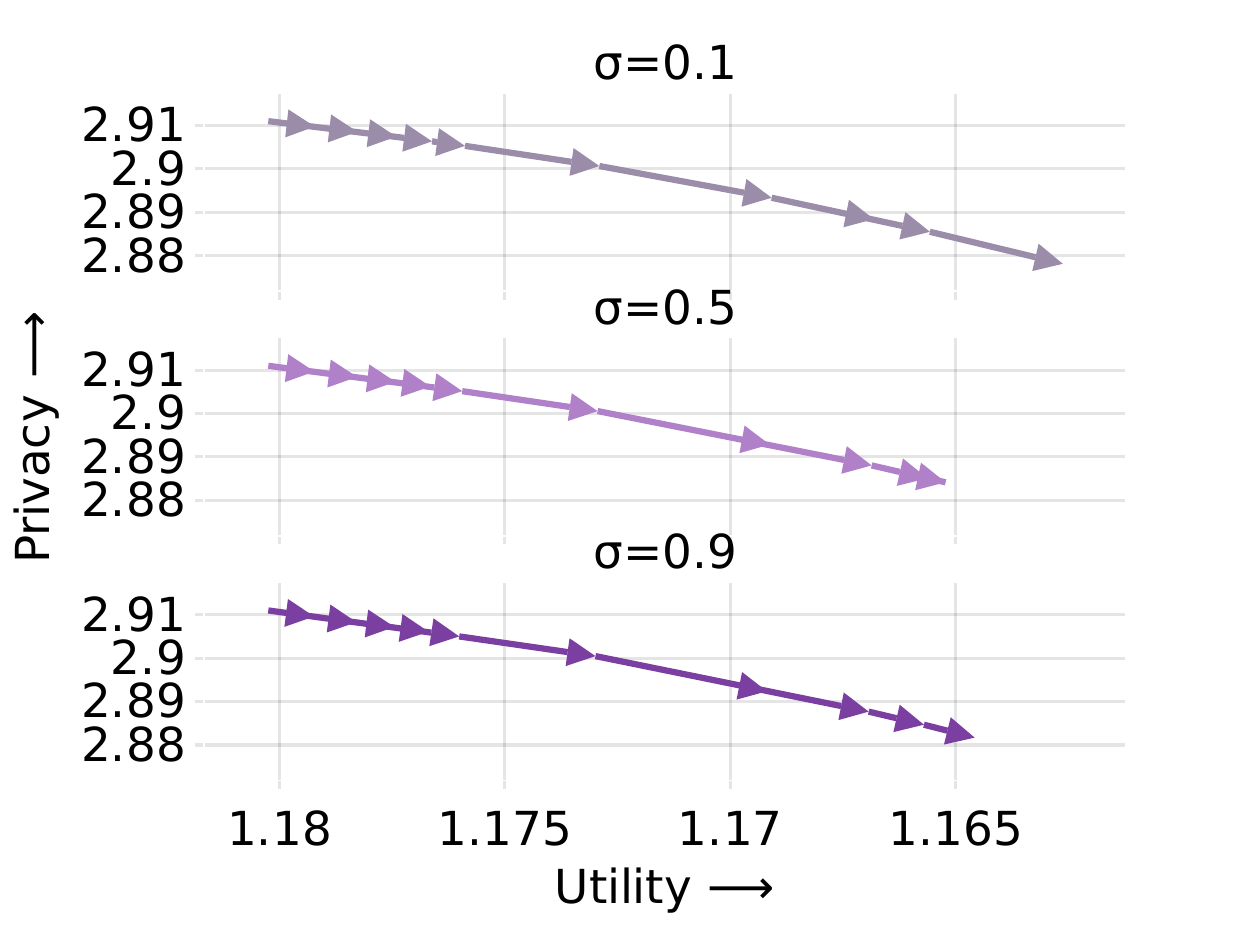}
        \caption{Qwen2.5 (SynBio)}
        \label{fig:dp_piid}
    \end{subfigure}
    \caption{
        Privacy–utility trade-offs for differential privacy (DP-SGD) across models and datasets with varying noise levels ($\sigma$). Larger $\sigma$ increases privacy but reduces utility, with the decline more pronounced in larger models. On SynBio, DP-SGD exhibits a similar pattern. Gemma shows a unique utility drop after two epochs, highlighting the complexity of the privacy-utility trade-off.
    }
    \label{fig:combined_dp}
\end{figure*}

\subsection{DP-SGD fine-tuning for privacy and utility costs efficiency}

Differential Privacy (DP-SGD) algorithms ~\cite{10.1145/2976749.2978318} aim to safeguard the privacy of individual training data points by limiting their influence on the gradient updates during training.

\noindent
\textbf{Update rules:}
DP-SGD clips the \textit{$l_2$} norm of each data point’s gradient at a threshold $T$, followed by adding noise with magnitude $\sigma$ to each clipped gradient.
The purpose of clipping is to reduce data sensitivity by ensuring that the impact of data points with high gradient magnitudes on the model parameters is limited.
Adding noise further obscures the contribution of individual data points, making it difficult to infer specific data points from the model.

\vspace{-10pt}
\small{
\begin{equation}
\begin{split}
    W_{t + 1} &= W_t - \eta \, \text{Noise} \left( \frac{1}{B} \sum_i \text{Clip}\left( \nabla_{W_t} \mathcal{L}(\mathcal{M}_{W_t}(x_i), x_i) \right) \right) \\
    \text{Clip}(y) &= y / \max \left( 1, \frac{\lVert y \rVert_2}{T} \right) \\
    \text{Noise}(y) &= y + \mathcal{N}(0, \sigma^2 T^2 \mathds{1})
\end{split}
\end{equation}
}
\normalsize
\vspace{-10pt}

We vary the noise hyperparameter $\sigma$ for the experiments.
The clipping gradient norm $T$ is fixed at $10^{-2}$, as in \cite{shi-etal-2022-selective}.

\begin{figure*}[h!]
    \centering
        \begin{subfigure}{.3\linewidth}
       \includegraphics[scale=0.25,height=3.5cm]{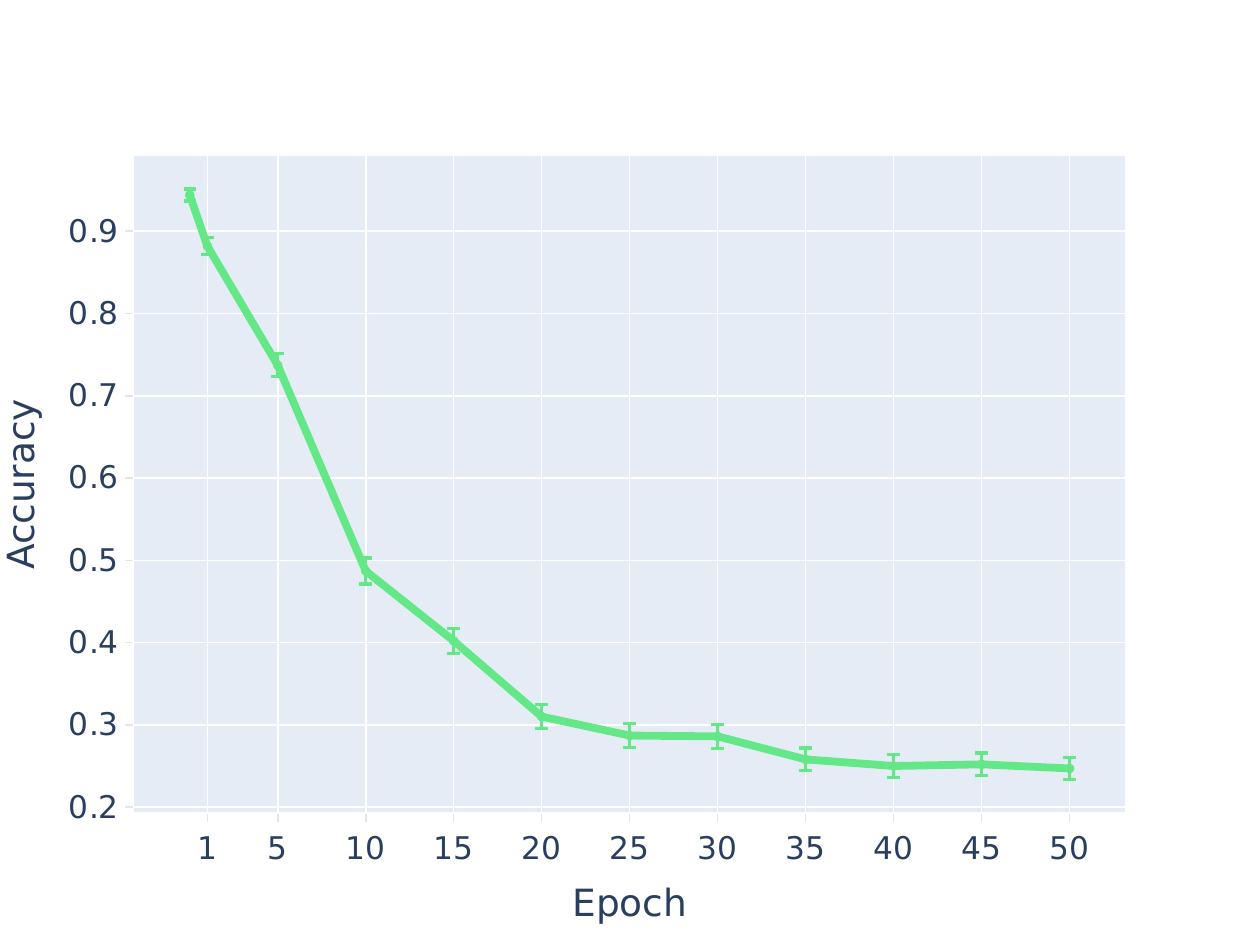}
        \caption{SCIQ Benchmark}
        \label{fig:sciq_dpsgd}
    \end{subfigure}
    \begin{subfigure}{.3\linewidth}
       \includegraphics[scale=0.25,height=3.5cm]{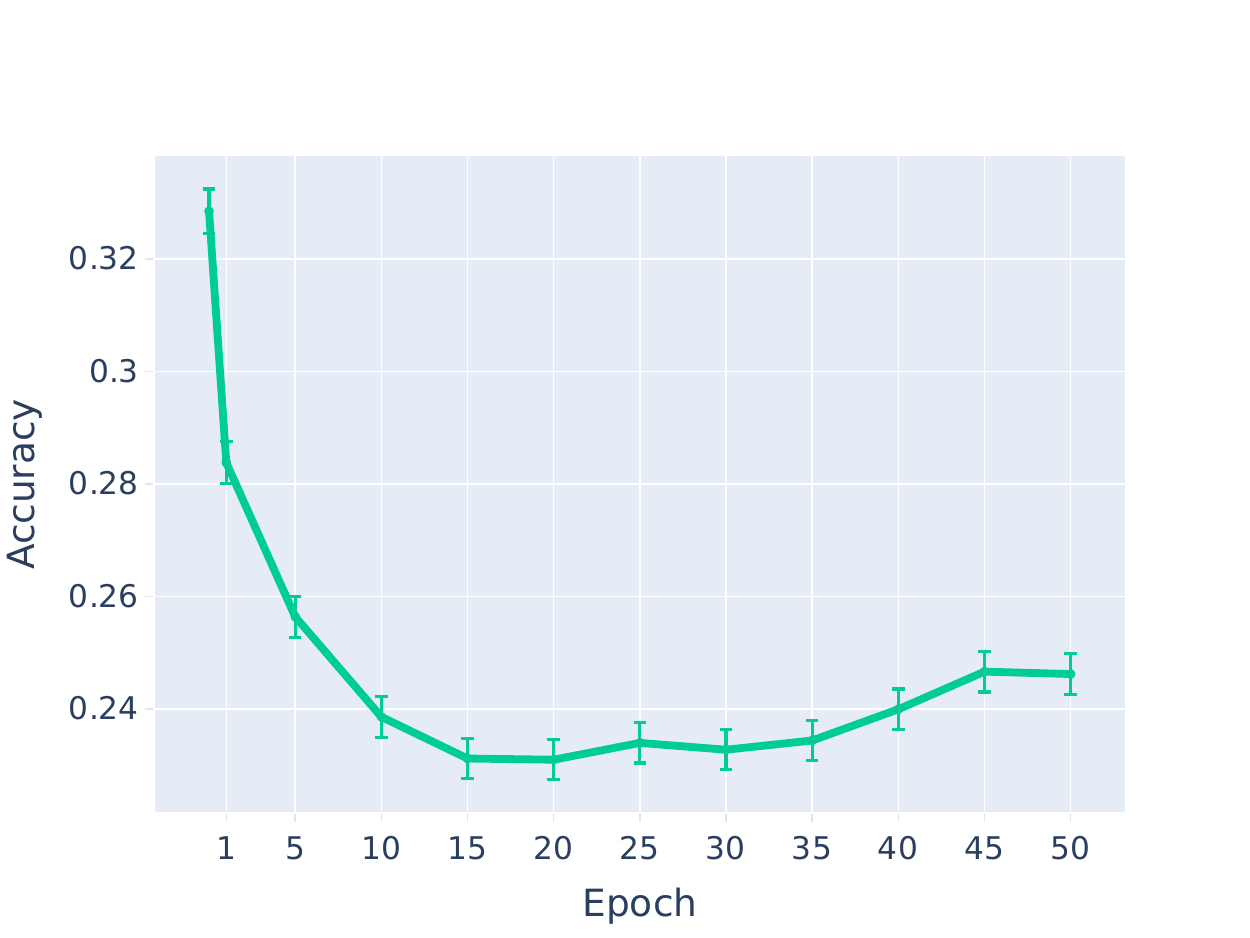}
        \caption{MMLU Benchmark}
        \label{fig:mmlu_dpsgd}
    \end{subfigure}
     \begin{subfigure}{.3\linewidth}
       \includegraphics[scale=0.25,height=3.5cm]{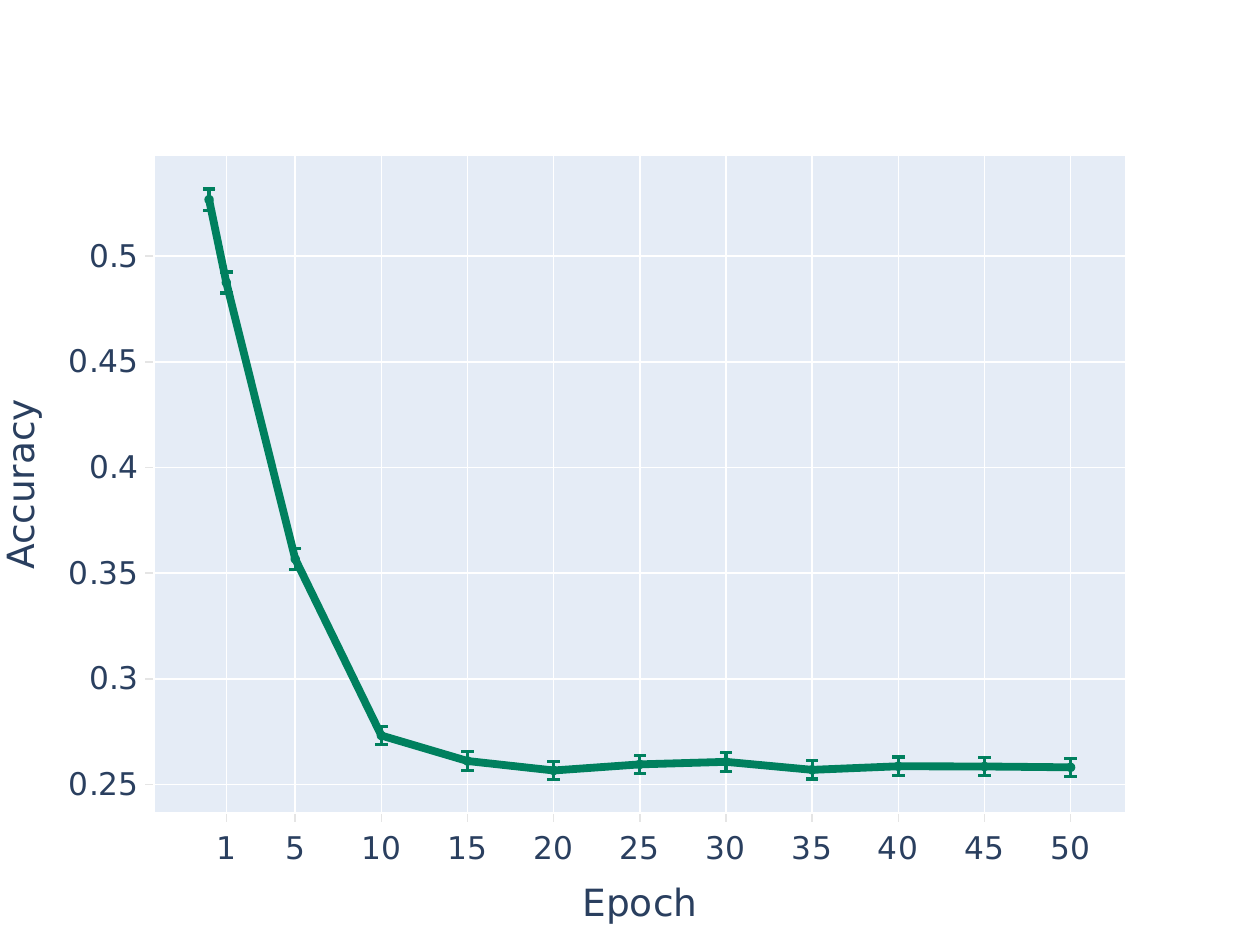}
        \caption{HellaSwag Benchmark}
        \label{fig:hs_dpsgd}
    \end{subfigure}
    \caption{
    Fine-tuning Gemma with DP-SGD results in a substantial accuracy declines of 70\%, 10\%, and 30\% across the benchmarks.
    }
    \label{fig:bench-dpsgd}
\end{figure*}

\noindent
\textbf{Privacy-Utility trade-off:}
Figures \ref{fig:dp_csima}, \ref{fig:dp_csimb}, and \ref{fig:dp_csimc} illustrate the privacy-utility trade-off when varying the noise $\sigma$ on the CustomerSim dataset across the Pythia, Gemma, and Llama2 models.
As in previous plots, each curve begins with the performance of the pre-trained model.
It is evident that DP-SGD maintains privacy effectively with minimal degradation.
But there is a trade-off between privacy and utility.
For all models, lower noise values such as $\sigma = 0.1$ are able to achieve better utility than higher ones such as $\sigma = 0.5$ and $\sigma = 0.9$, but they also decrease privacy more.
While the total amount of utility achievable with DP-SGD is limited, especially for the larger Llama2-7B model, overall, it provides good privacy-utility trade-offs.
Similar trends can be observed for the SynBio dataset in Figures~\ref{fig:dp_piia} and \ref{fig:dp_piic}, but for Figure~\ref{fig:dp_piib} (with Gemma) utility declines after approximately two epochs.  

\noindent
\textbf{Impact on benchmark datasets:} Figures~\ref{fig:sciq_dpsgd}, \ref{fig:mmlu_dpsgd} and \ref{fig:hs_dpsgd} show the benchmark accuracy over epochs for the DP-SGD fine-tuned Gemma model with $\sigma=0.1$ on SCIQ, MMLU, and HellaSwag datasets, respectively. Accuracy drops gradually with increasing epochs for all benchmarks, stabilizing at lower levels, indicating that fine-tuning with DP-SGD significantly reduces the knowledge retention capacity.

\noindent
\textbf{Efficiency:}
Differential privacy comes with a high computational cost, since the gradients of each datapoint need norm and clipping computations, and additional noise values are added.
Empirically, we observe a relative FLOPs requirement of $C_{\text{DP-SGD}} / C_{\text{FFT}} = 1.33$.
Additionally, the per-sample operations required for clipping mean that we need to keep copies of the gradient for each datapoint in memory, which requires substantially more GPU memory, which decreases the feasible batch size and the overall training throughput even further.

\noindent
\textbf{Takeaway:} DP-SGD offers a reasonable privacy-utility trade-off and is less susceptible to learning sensitive data. Increasing noise rates lead to poor utility across all models.
Convergence with DP-SGD
is not guaranteed, particularly in larger models such as Llama2.
The privacy gains come at the cost of efficiency.
Additionally, fine-tuning with DP-SGD leads to quick decline in the benchmark performance. The degree of measures along the \textit{tradeoff, knowledge retention, and efficiency} are: 

\indent 
Utility-privacy trade-offs: \textit{moderate} \\
\indent Retention of base performance: \textit{poor} \\
\indent Efficiency: \textit{poor}

\begin{figure*}[t]
    \centering
    \begin{subfigure}{0.245\linewidth}
        \centering
        \includegraphics[width=\linewidth]{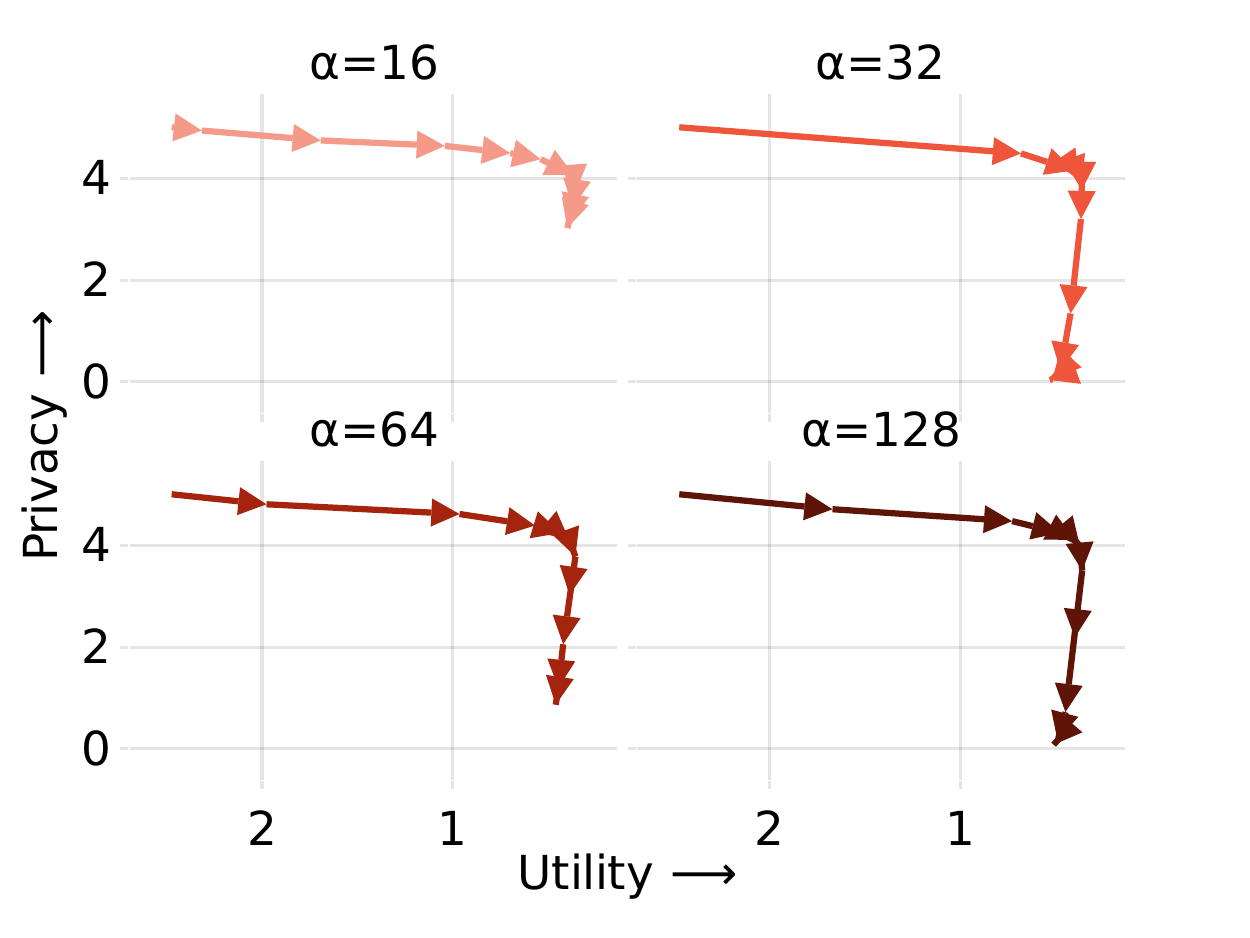}
        \caption{Pythia (CustSim)}
        \label{fig:lora_16_csima}
    \end{subfigure}
    \begin{subfigure}{0.245\linewidth}
        \centering
        \includegraphics[width=\linewidth]{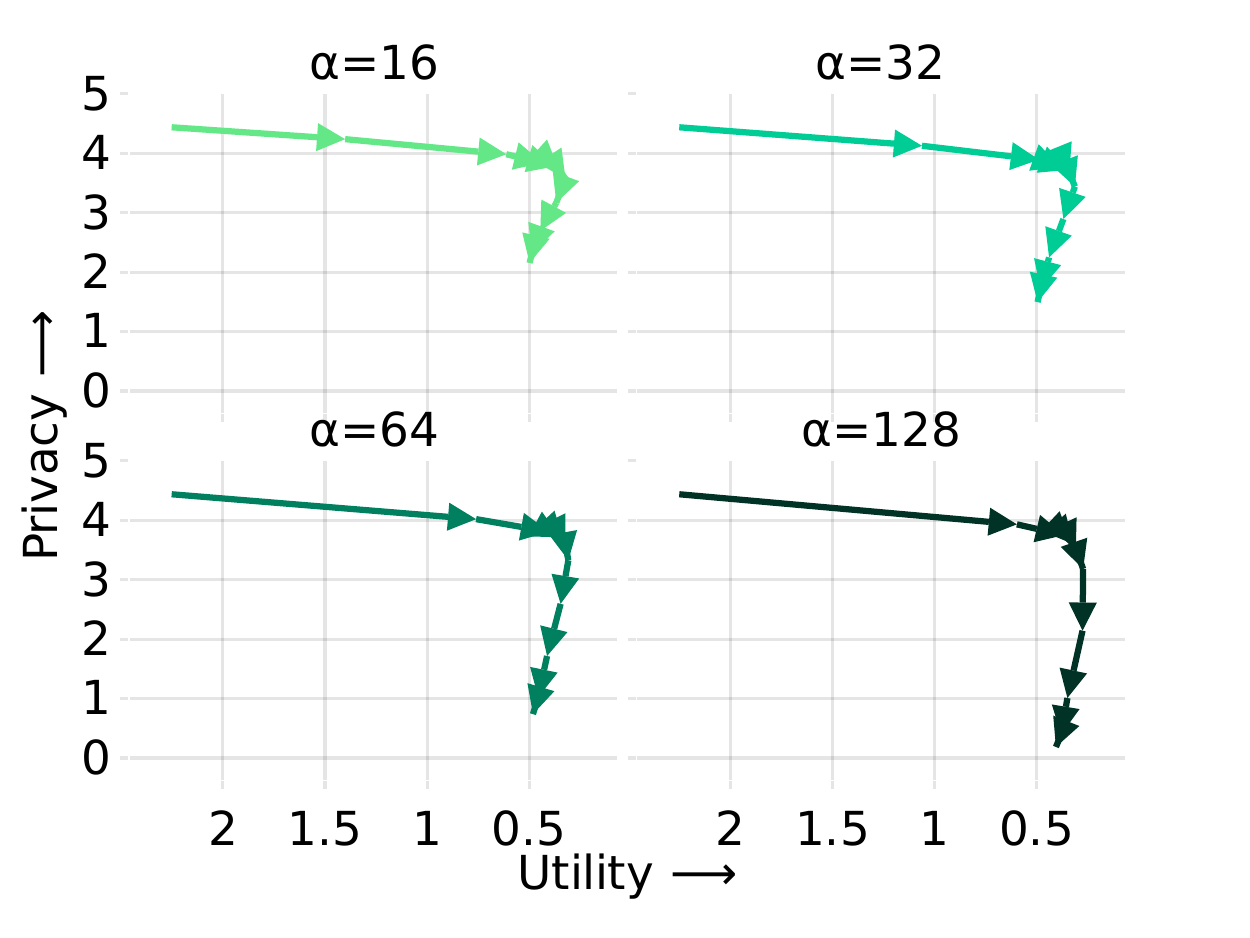}
        \caption{Gemma (CustSim)}
        \label{fig:lora_16_csimb}
    \end{subfigure}
    \begin{subfigure}{0.245\linewidth}
        \centering
        \includegraphics[width=\linewidth]{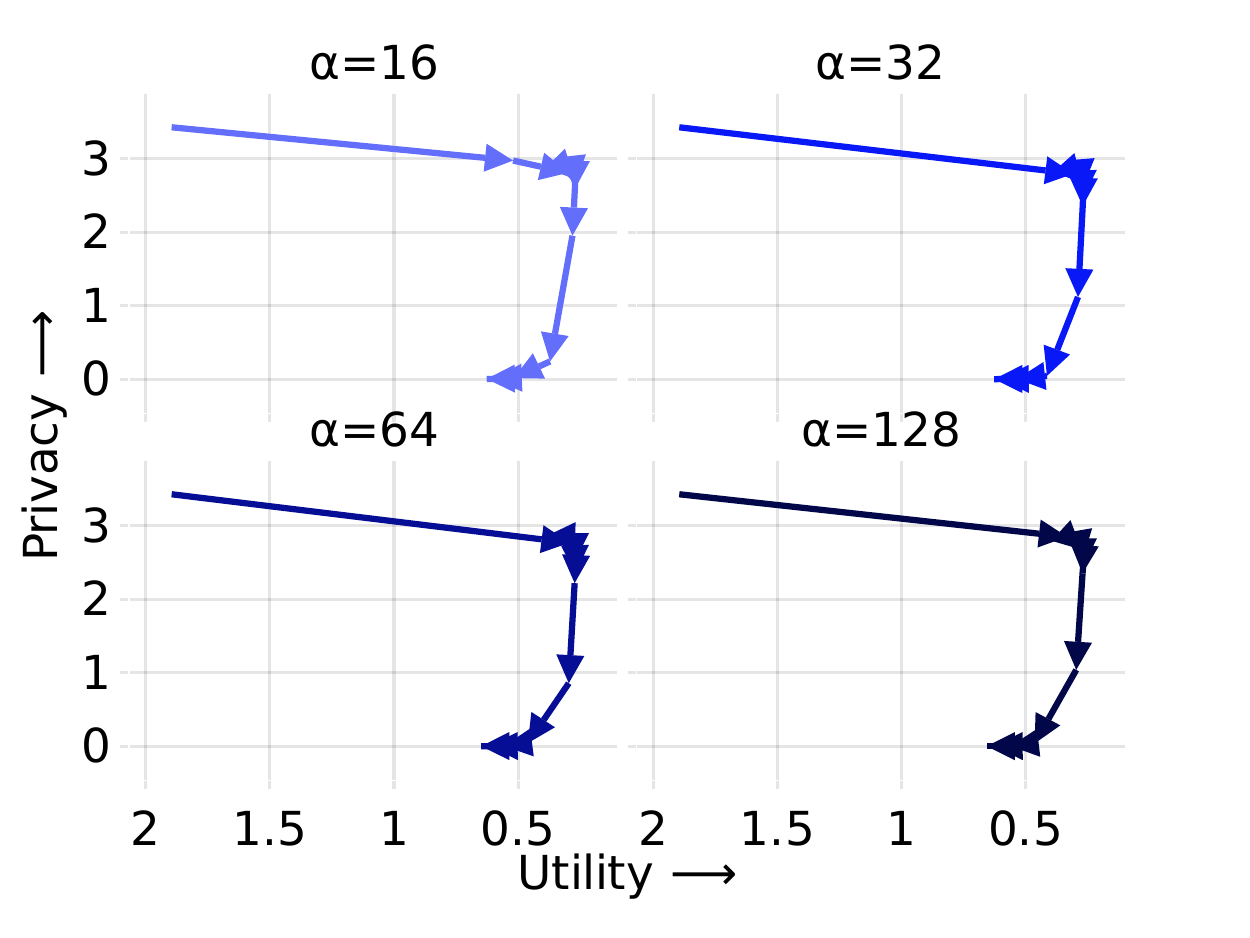}
        \caption{Llama2 (CustSim)}
        \label{fig:lora_16_csimc}
    \end{subfigure}
    \begin{subfigure}{0.245\linewidth}
        \centering
        \includegraphics[width=\linewidth]{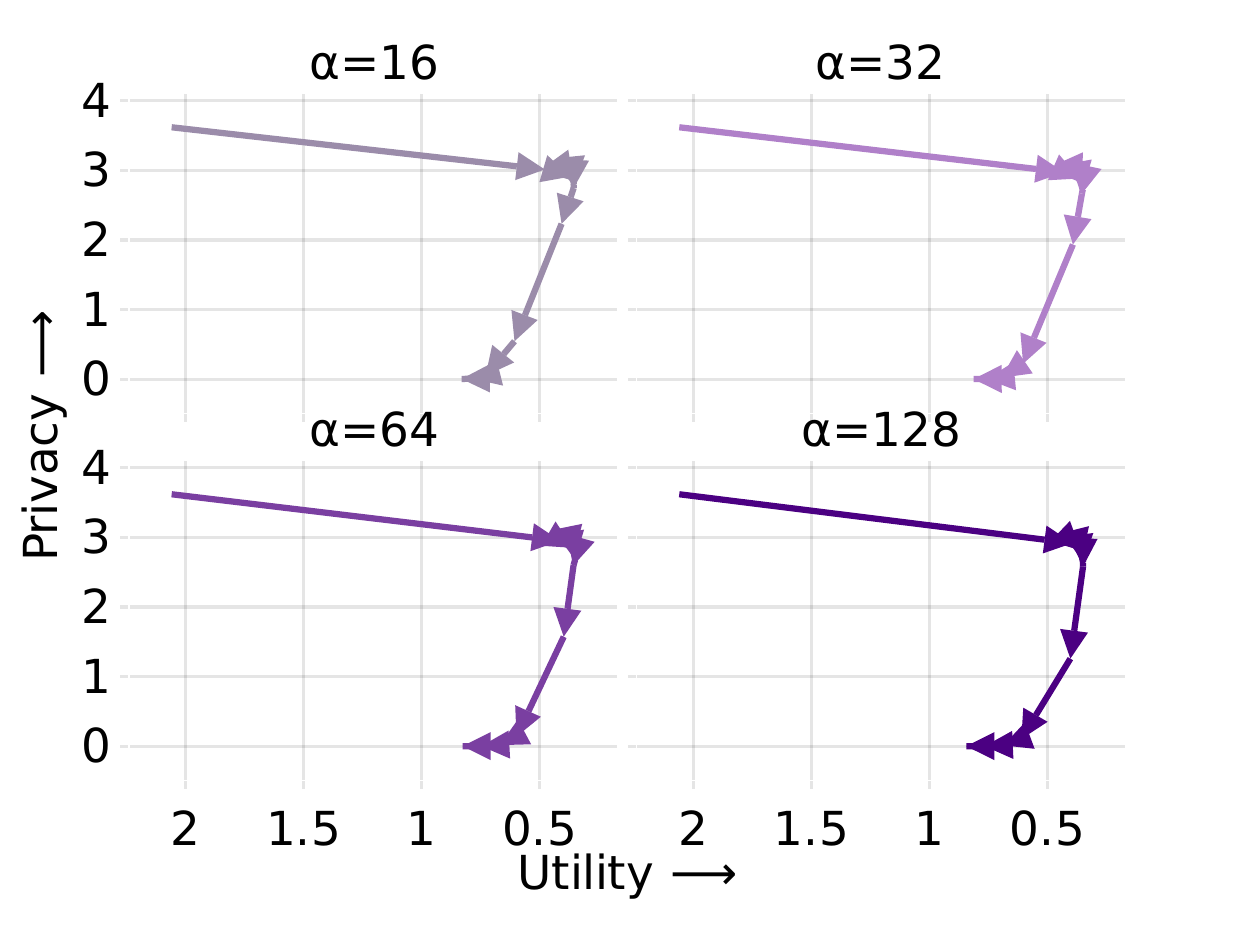}
        \caption{Qwen2.5 (CustSim)}
        \label{fig:lora_16_csimd}
    \end{subfigure}
    
    \begin{subfigure}{0.245\linewidth}
        \centering
        \includegraphics[width=\linewidth]{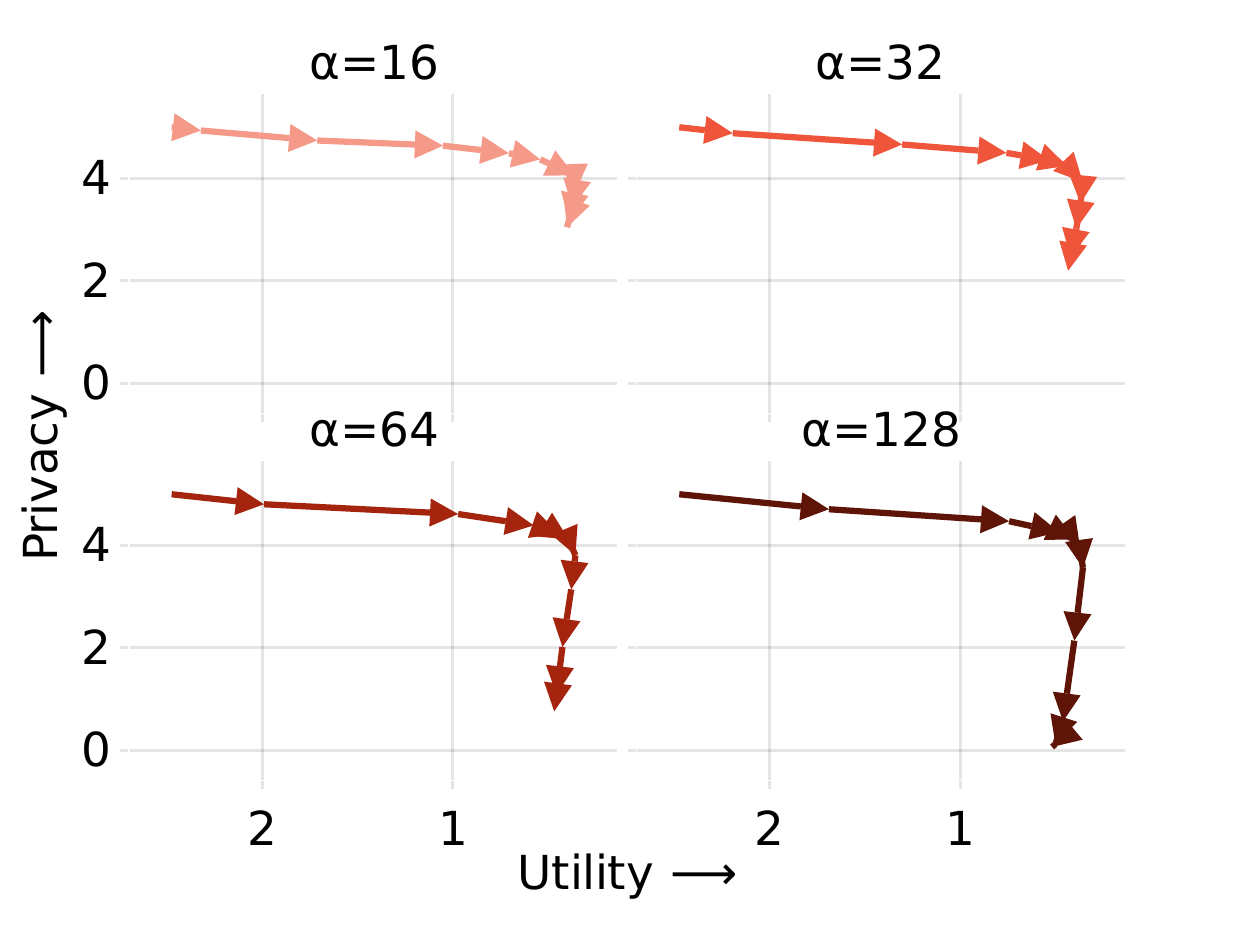}
        \caption{Pythia (Rank 32)}
        \label{fig:lora_32_csima}
    \end{subfigure}
    \begin{subfigure}{0.245\linewidth}
        \centering
        \includegraphics[width=\linewidth]{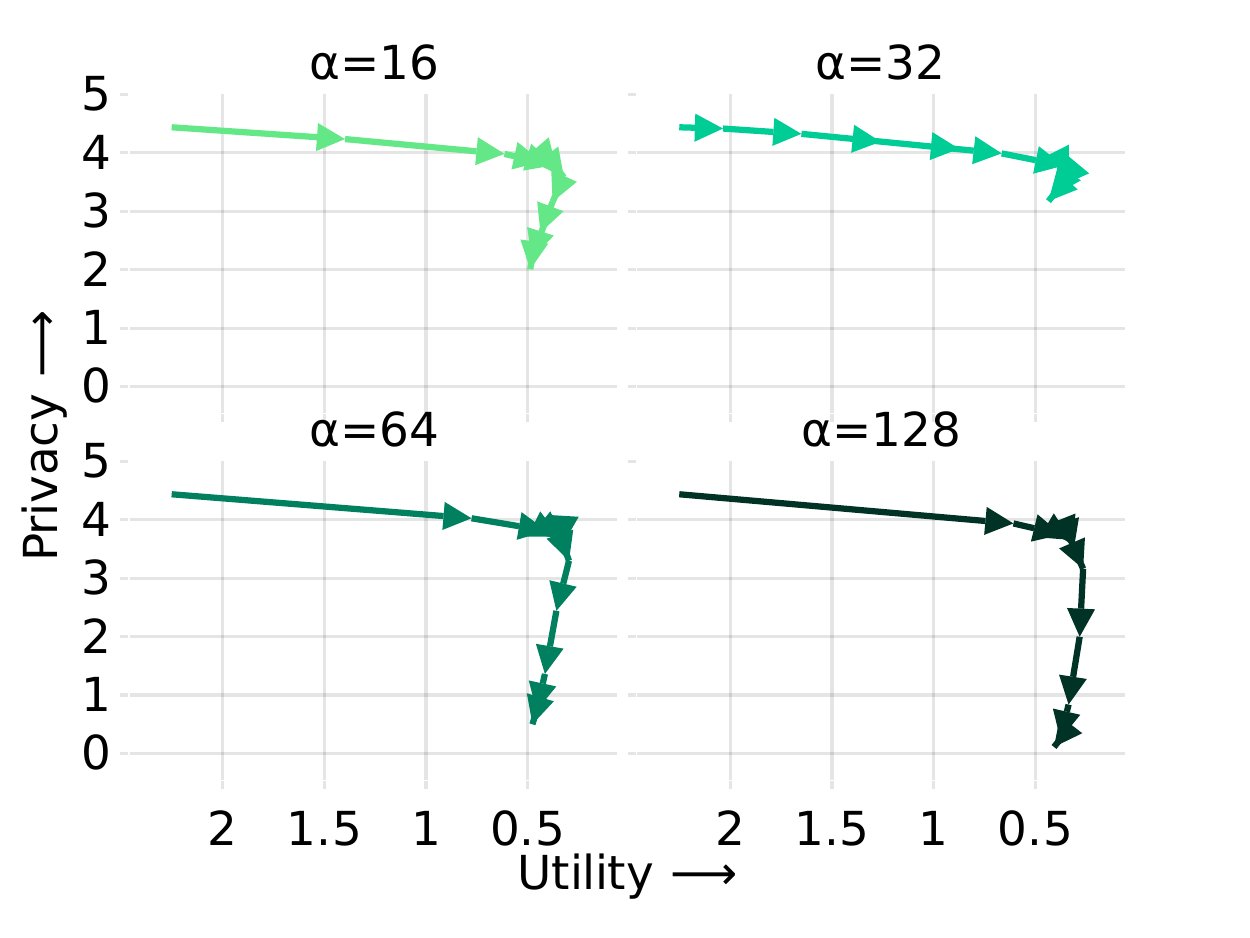}
        \caption{Gemma (Rank 32)}
        \label{fig:lora_32_csimb}
    \end{subfigure}
    \begin{subfigure}{0.245\linewidth}
        \centering
        \includegraphics[width=\linewidth]{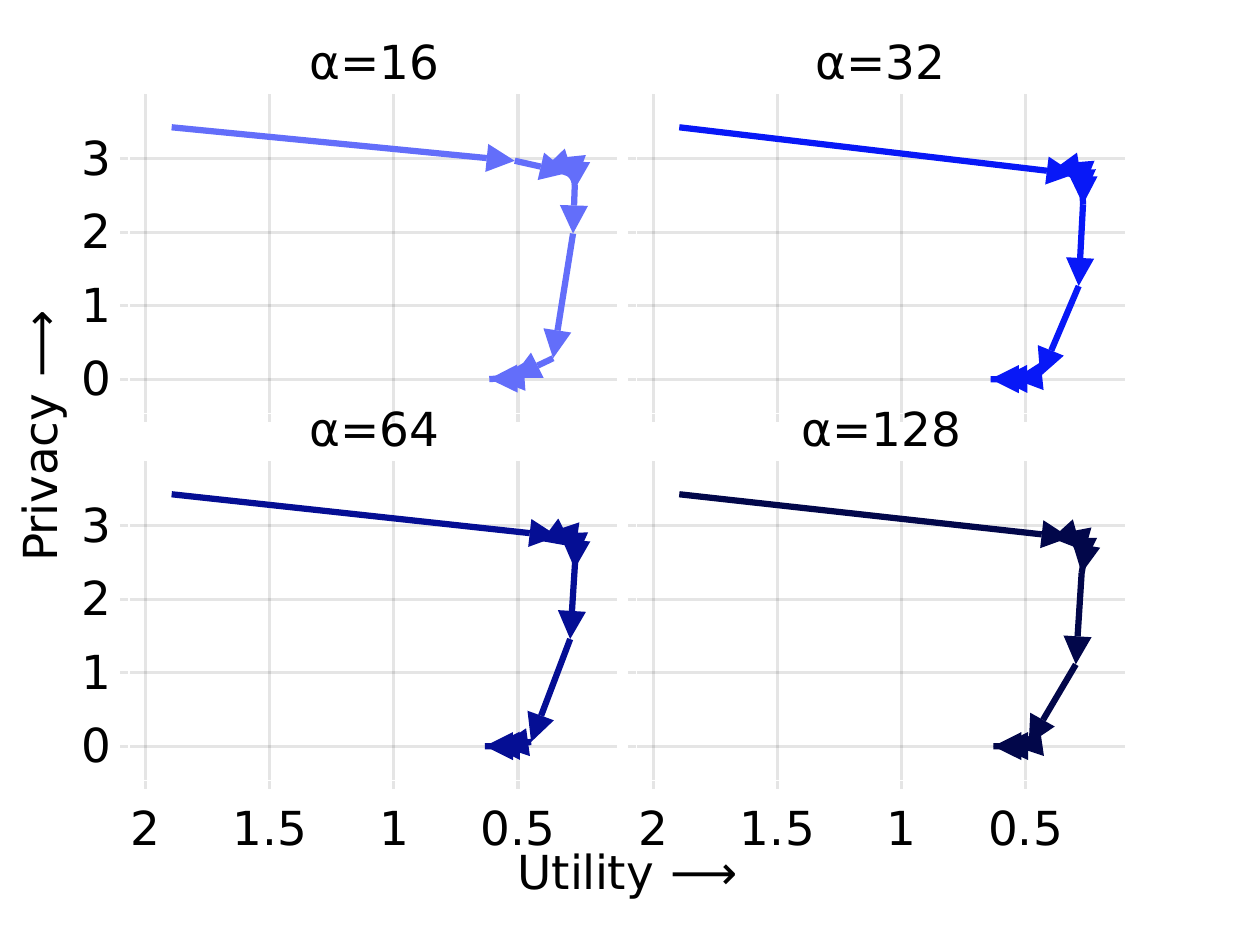}
        \caption{Llama2 (Rank 32)}
        \label{fig:lora_32_csimc}
    \end{subfigure}
    \begin{subfigure}{0.245\linewidth}
        \centering
        \includegraphics[width=\linewidth]{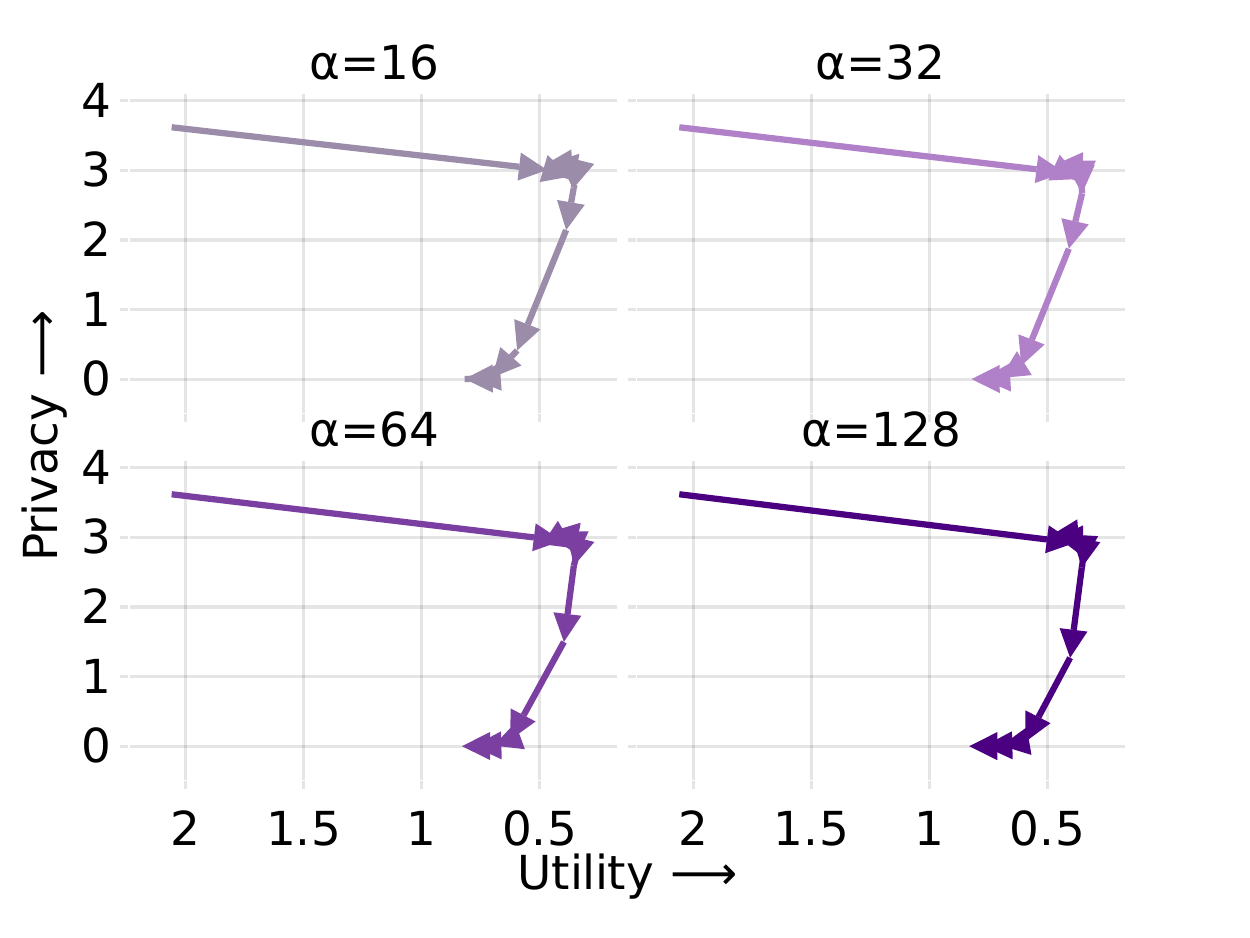}
        \caption{Qwen2.5 (Rank 32)}
        \label{fig:lora_32_csimd}
    \end{subfigure}

    \caption{
       Privacy–utility trade-offs for LoRA fine-tuning with ranks 16 and 32 on CustomerSim and varying scaling factor $\alpha$. Increasing $\alpha$ generally improves or maintains utility but reduces privacy. Smaller models achieve more favorable privacy–utility trade-offs, while larger models retain utility at the cost of reduced privacy during extended training.
    }
    \label{fig:combined_lora}
\end{figure*}

\begin{figure*}[t]
    \centering
   \begin{subfigure}{0.245\linewidth}
        \centering
        \includegraphics[width=\linewidth]{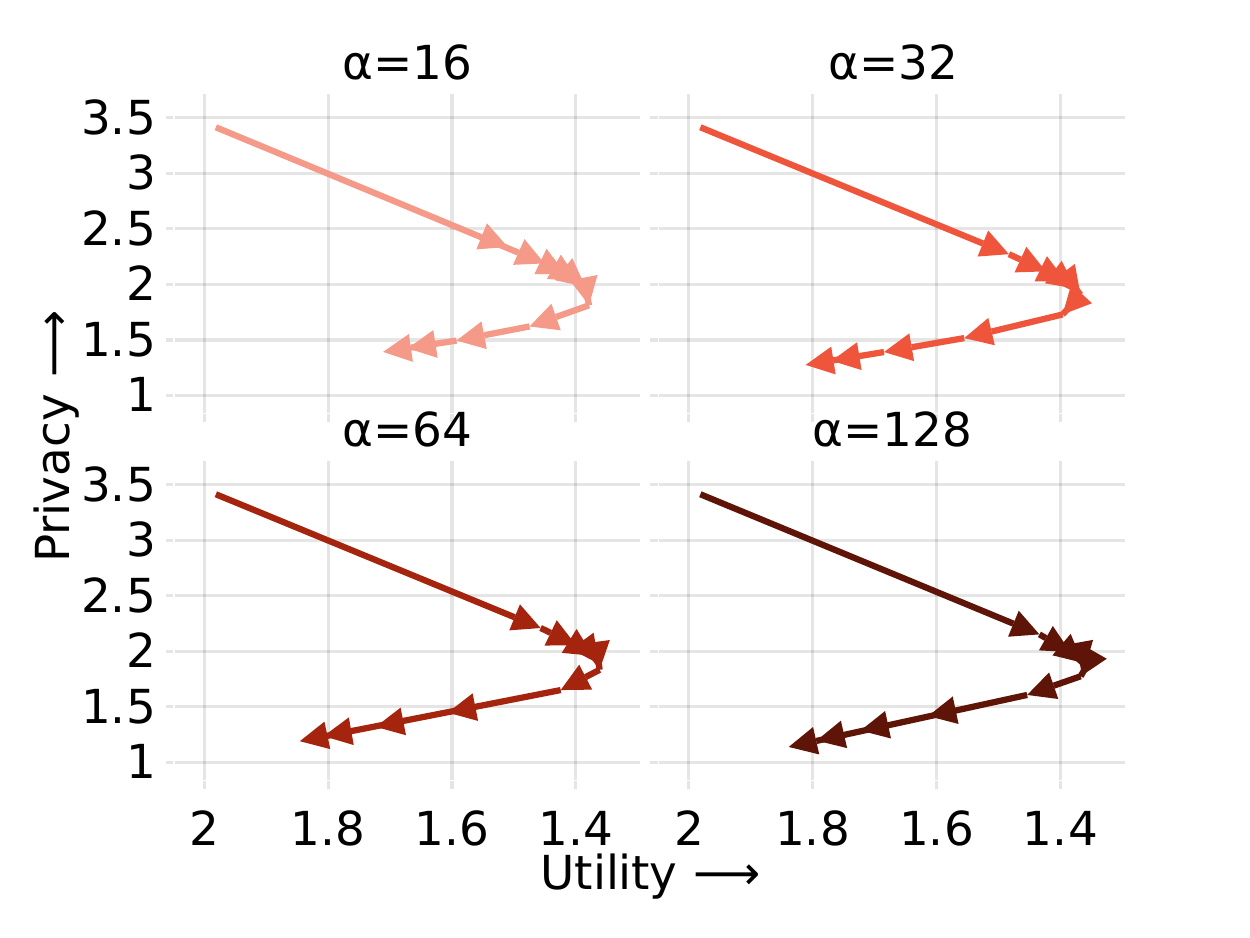}
        \caption{Pythia (Rank 16)}
        \label{fig:lora_16_piia}
    \end{subfigure}
   \begin{subfigure}{0.245\linewidth}
        \centering
        \includegraphics[width=\linewidth]{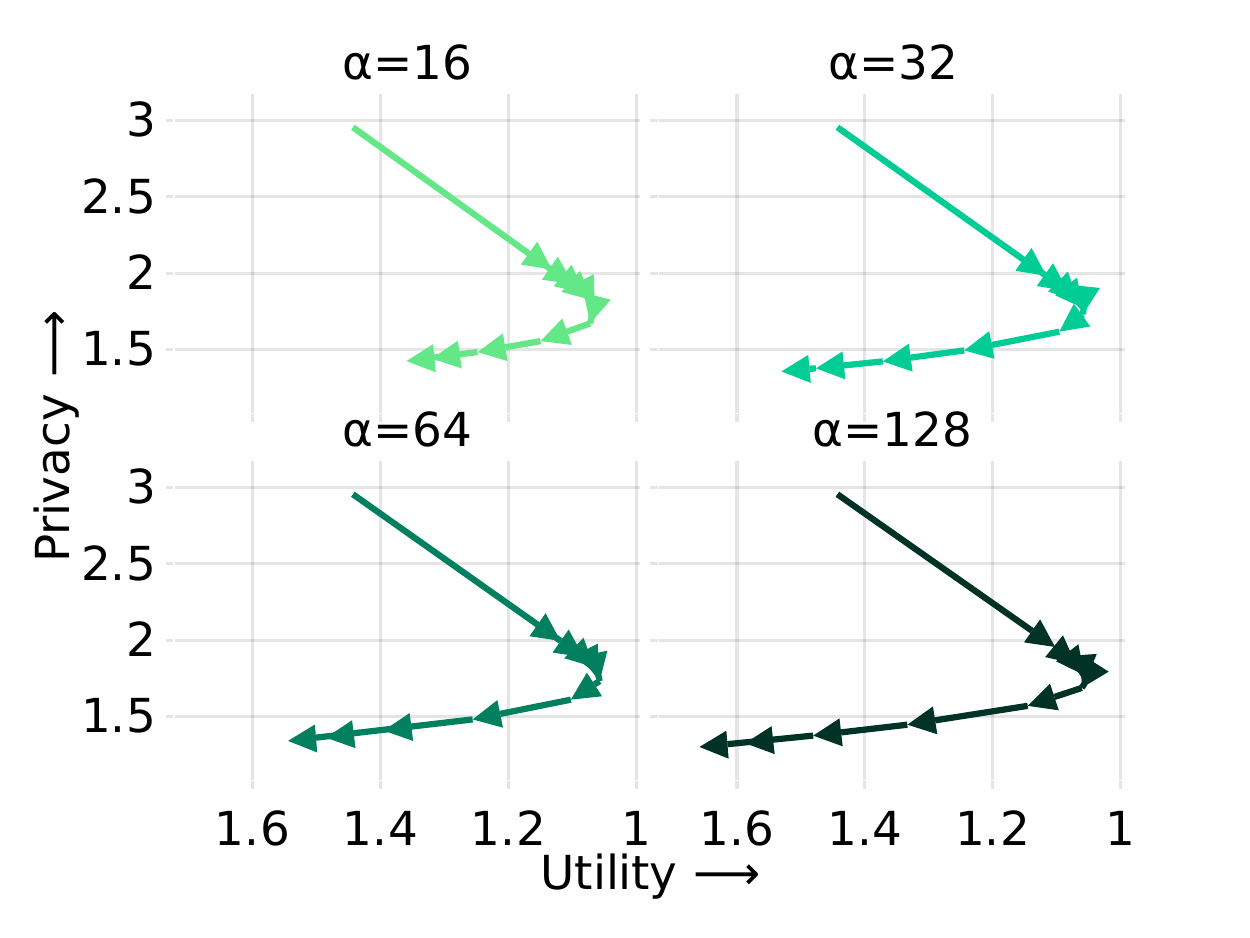}
        \caption{Gemma (Rank 16)}
        \label{fig:lora_16_piib}
    \end{subfigure}
   \begin{subfigure}{0.245\linewidth}
        \centering
        \includegraphics[width=\linewidth]{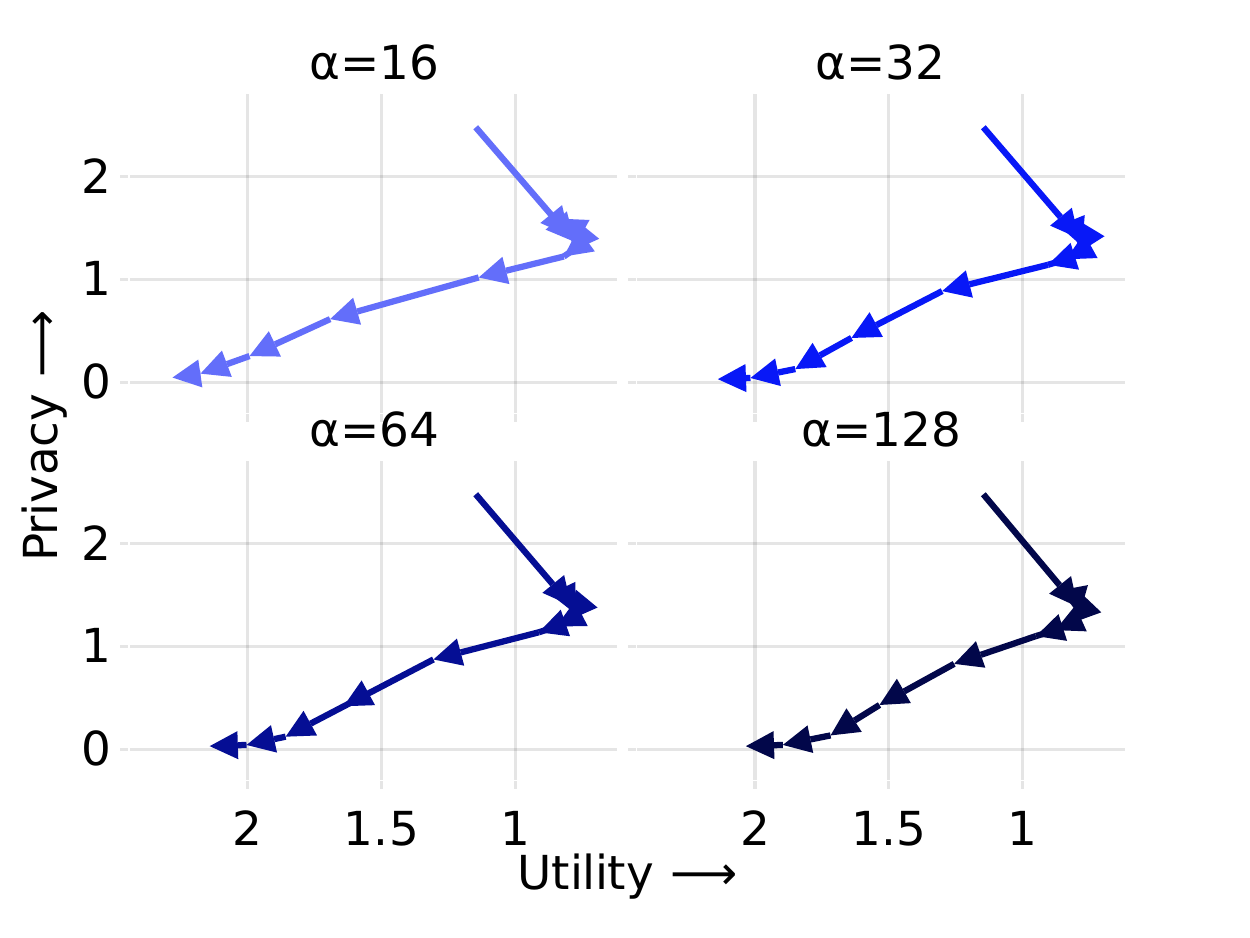}
        \caption{Llama2 (Rank 16)}
        \label{fig:lora_16_piic}
    \end{subfigure}
   \begin{subfigure}{0.245\linewidth}
        \centering
        \includegraphics[width=\linewidth]{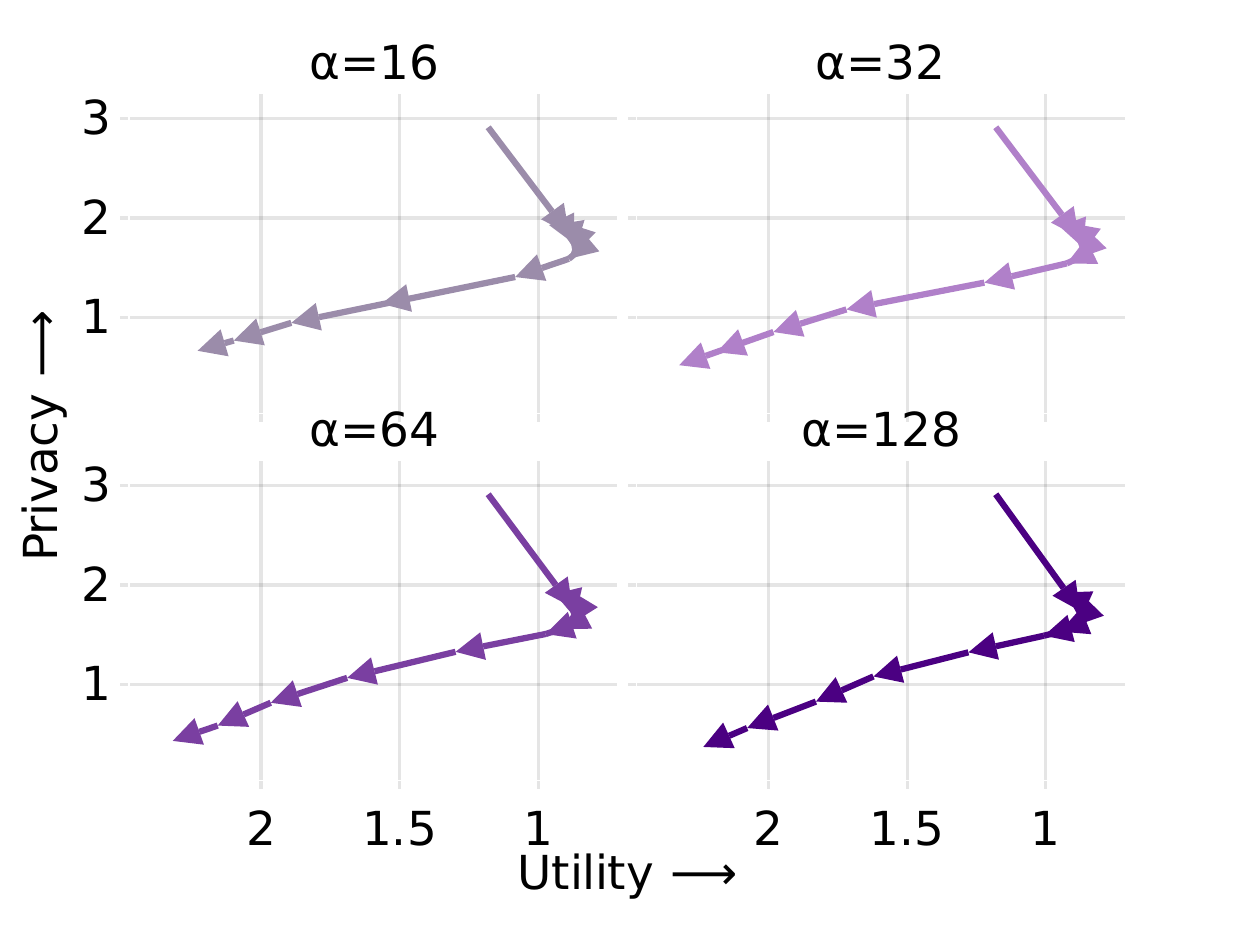}
        \caption{Qwen2.5 (Rank 16)}
        \label{fig:lora_16_piic}
    \end{subfigure}
   \begin{subfigure}{0.245\linewidth}
        \centering
        \includegraphics[width=\linewidth]{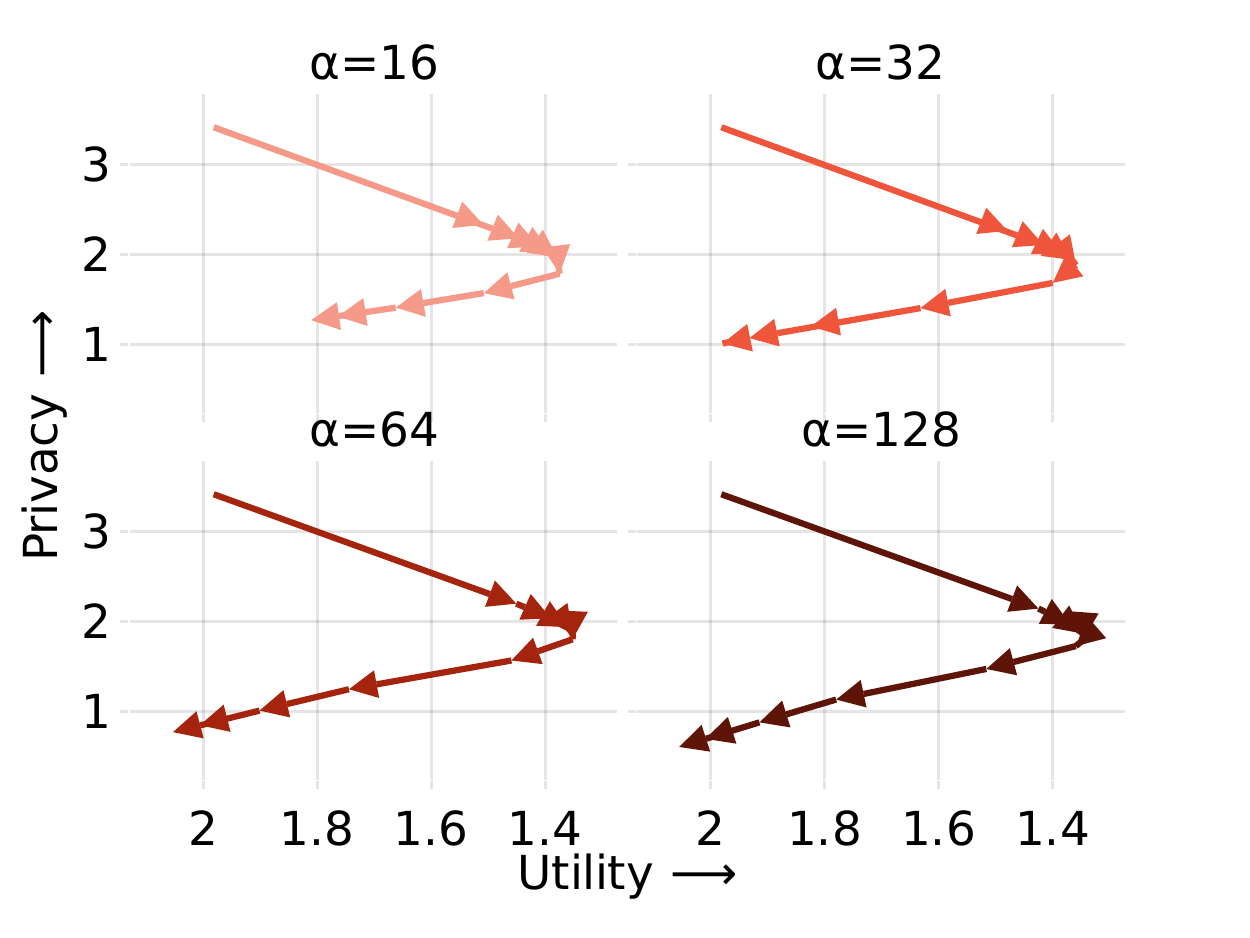}
        \caption{Pythia (Rank 32)}
        \label{fig:lora_32_piia}
    \end{subfigure}
   \begin{subfigure}{0.245\linewidth}
        \centering
        \includegraphics[width=\linewidth]{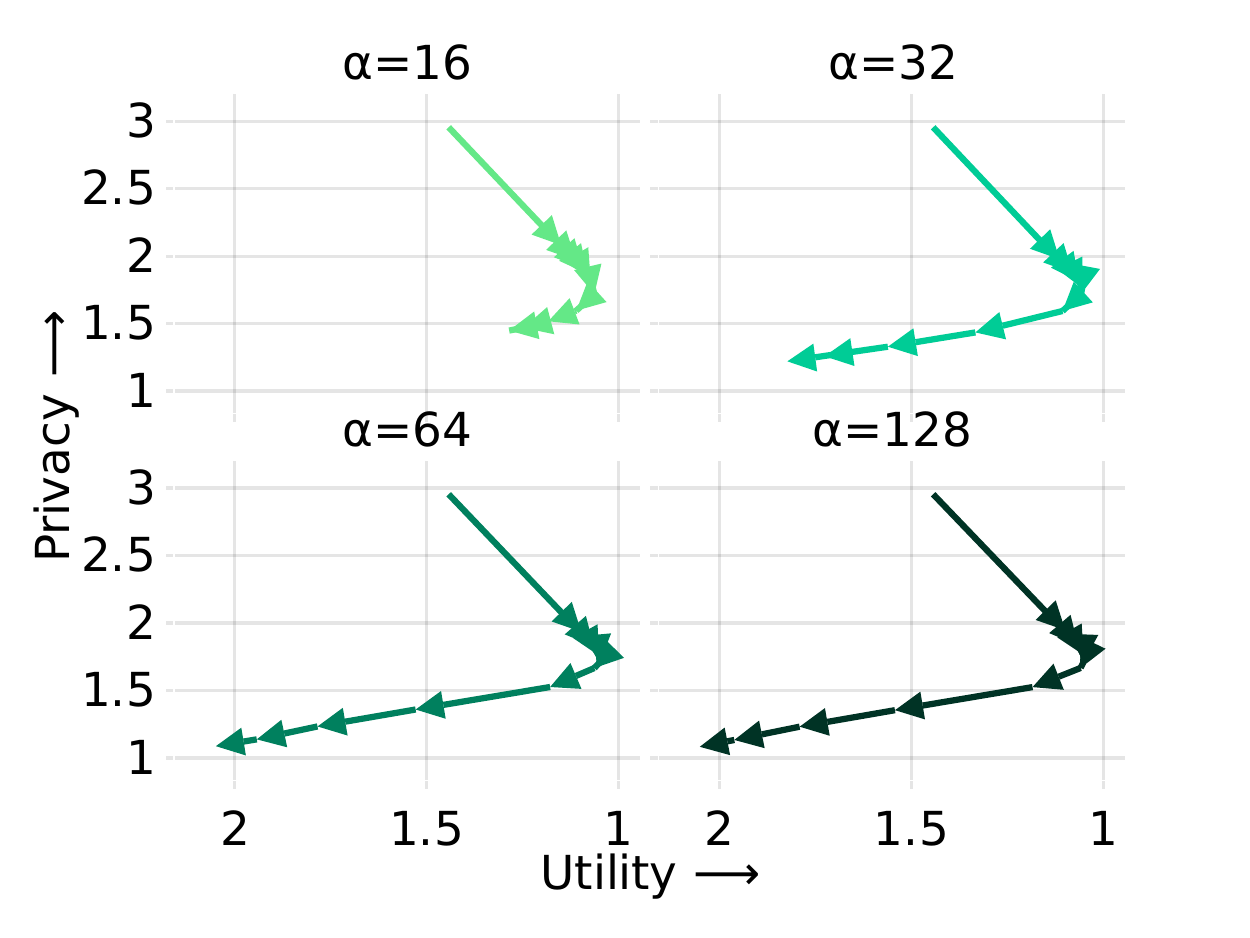}
        \caption{Gemma (Rank 32)}
        \label{fig:lora_32_piib}
    \end{subfigure}    
   \begin{subfigure}{0.245\linewidth}
        \centering
        \includegraphics[width=\linewidth]{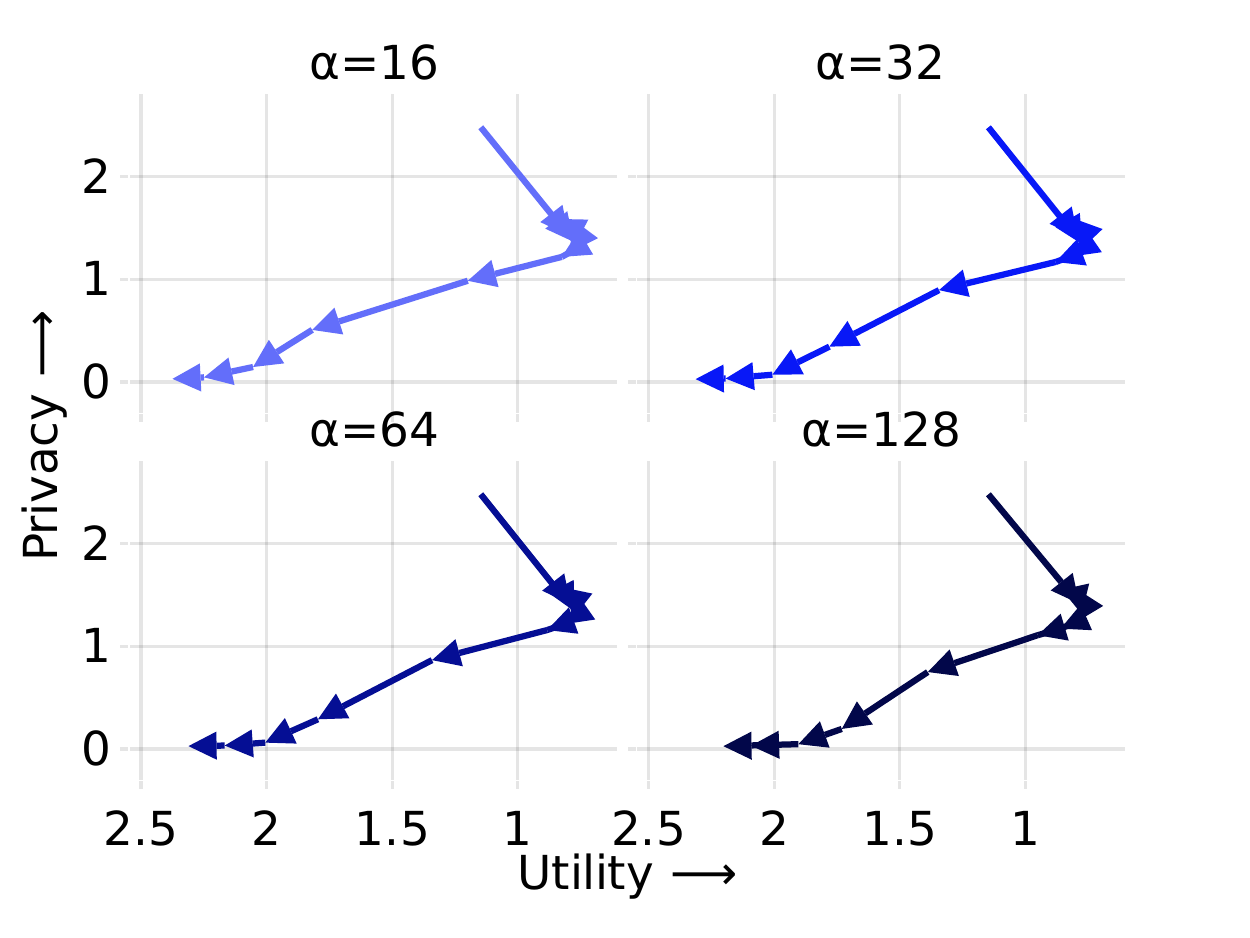}
        \caption{Llama2 (Rank 32)}
        \label{fig:lora_32_piic}
    \end{subfigure}
   \begin{subfigure}{0.245\linewidth}
        \centering
        \includegraphics[width=\linewidth]{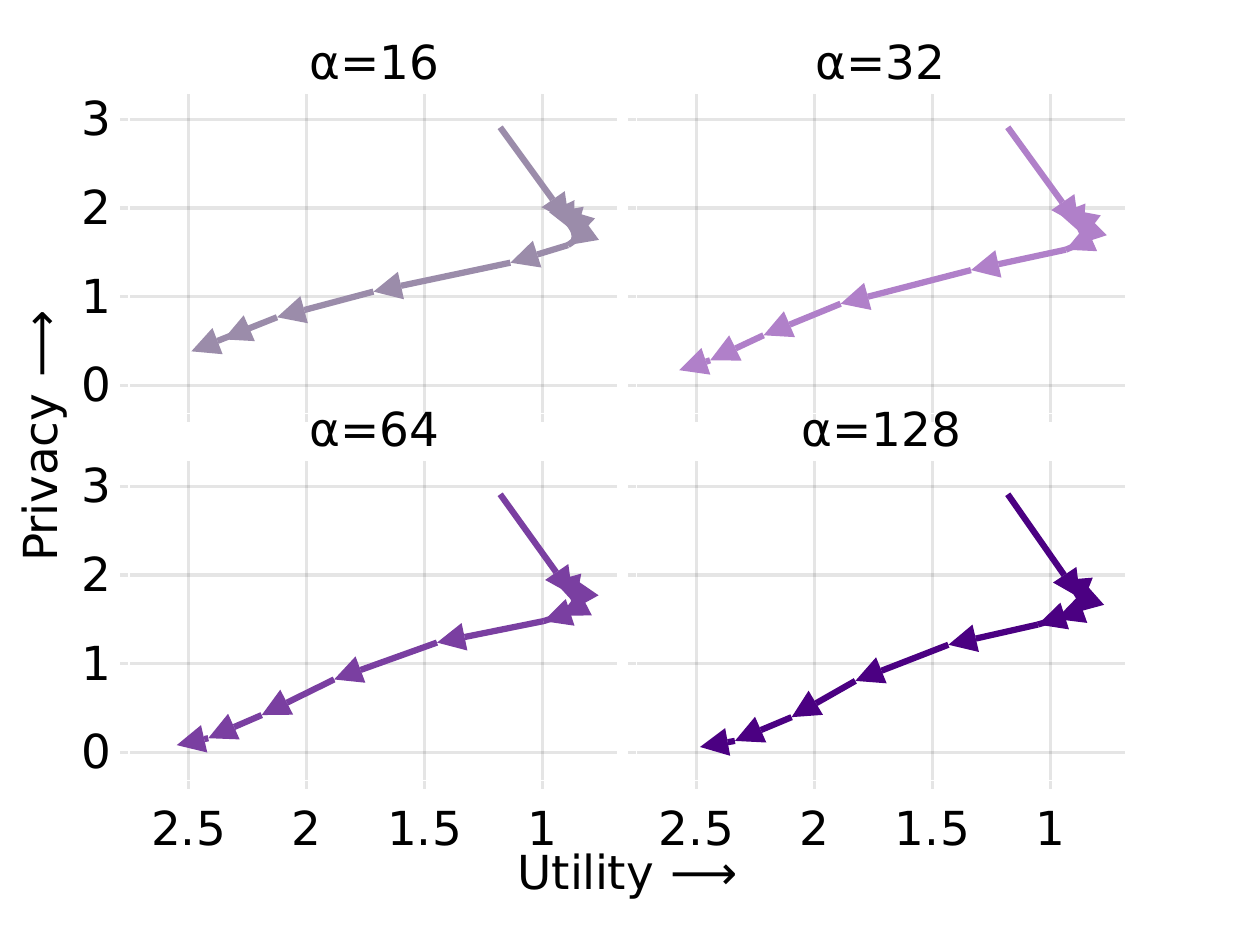}
        \caption{Qwen2.5 (Rank 32)}
        \label{fig:lora_32_piid}
    \end{subfigure}
    \caption{
       Privacy–utility trade-offs for LoRA fine-tuning with ranks 16 and 32 on SynBio and varying scaling factor $\alpha$. Increasing $\alpha$ generally improves or maintains utility but reduces privacy. Smaller models achieve more favorable privacy–utility trade-offs, while larger models retain utility at the cost of reduced privacy during extended training.
    }
    \label{fig:combined_lora_synbio}
\end{figure*}

\begin{figure*}[ht!]
    \centering
        \begin{subfigure}{.3\linewidth}
       \includegraphics[scale=0.3,height=3.5cm]{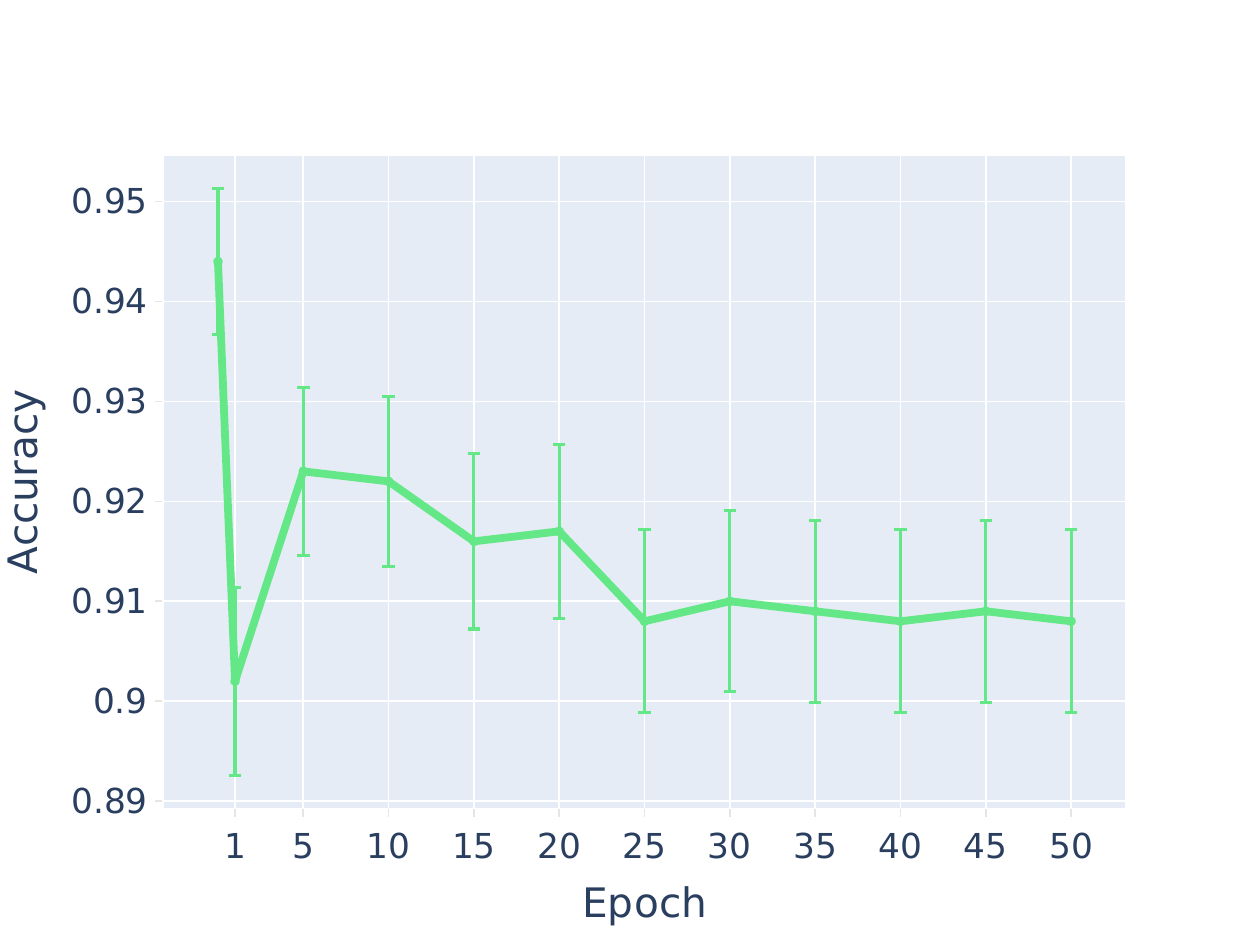}
        \caption{SCIQ Benchmark}
        \label{fig:sciq_lora}
    \end{subfigure}
    \begin{subfigure}{.3\linewidth}
       \includegraphics[scale=0.3,height=3.5cm]{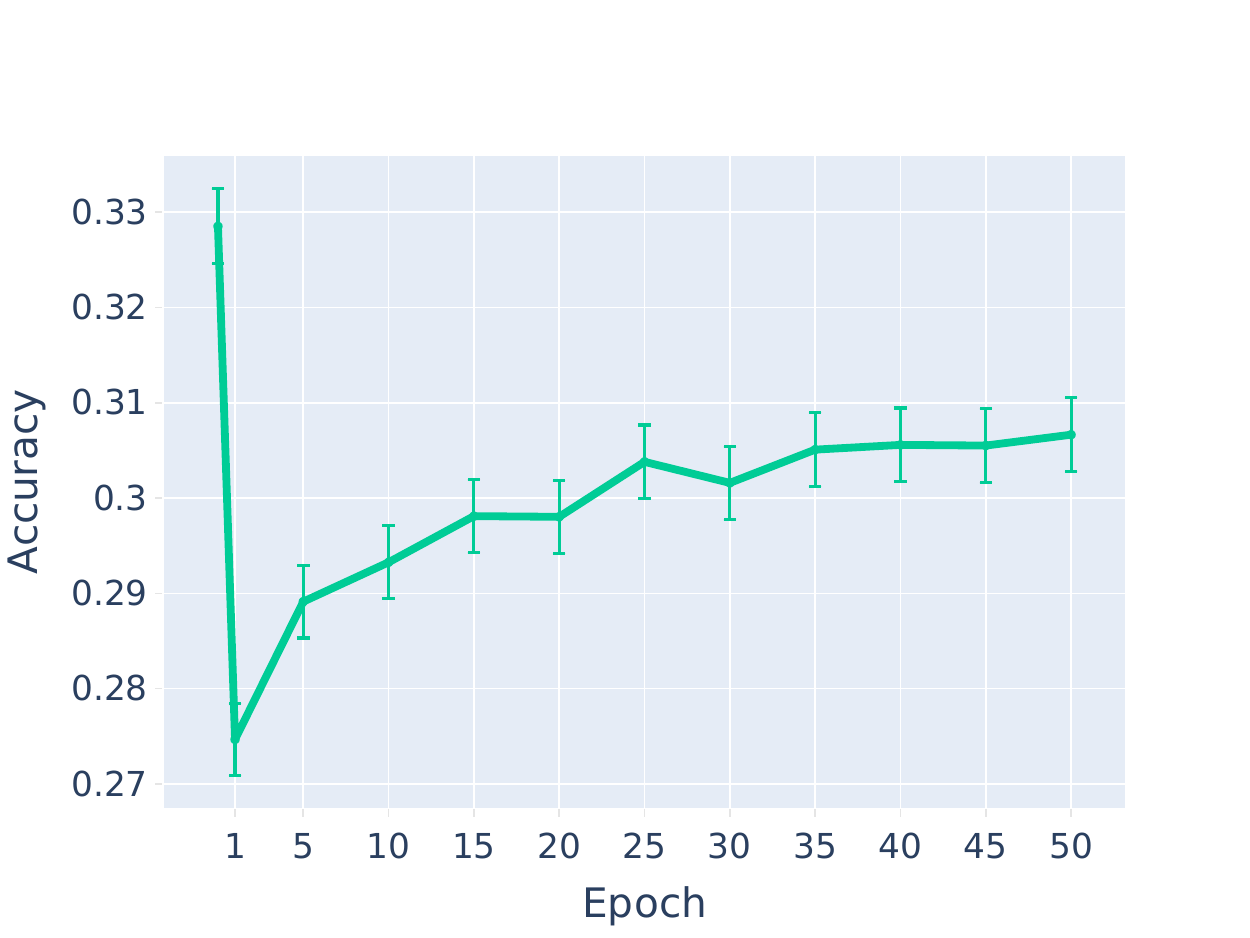}
        \caption{MMLU Benchmark}
        \label{fig:mmlu_lora}
    \end{subfigure}
     \begin{subfigure}{.3\linewidth}
       \includegraphics[scale=0.3,height=3.5cm]{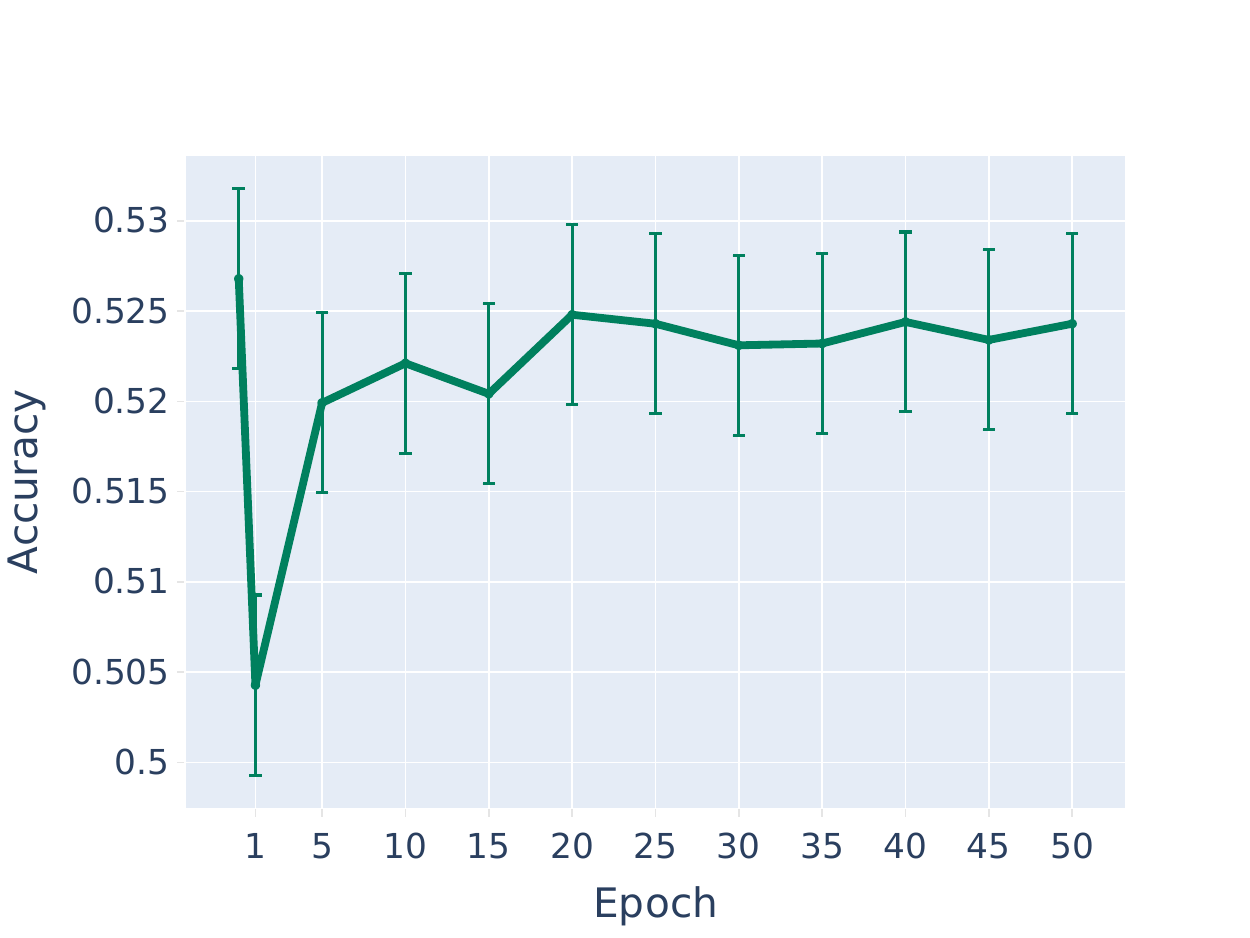}
        \caption{HellaSwag Benchmark}
        \label{fig:hs_lora}
    \end{subfigure}
    \caption{
    LoRA fine-tuned model maintains accuracy levels close to the pre-trained model with declines of 5\%, 3\% and 3\% across SCIQ, MMLU, and Hellaswag benchmarks, highlighting its effectiveness in knowledge retention.
    }
    \label{fig:bench-lora}
\end{figure*}

\subsection{LoRA fine-tuning for efficiency and utility improves privacy}

LoRA~\cite{hu2022lora} is a parameter-efficient fine-tuning method developed to reduce the compute and memory requirements of fine-tuning LLMs, and to reduce the size of storing fine-tuned checkpoints.
It enjoys large popularity for LLM fine-tuning.

\textbf{Update rules:}

\begin{equation}
\begin{split}
    W_{t + 1} &= W_0 + \frac{\alpha}{r} \Delta W_{t + 1}
 \\
    \Delta W_{t + 1} &= \Delta W_t - \eta \nabla_{\Delta W_t} \mathcal{L}(\mathcal{M}_{W_t}(X), X)
\end{split}
\label{eq:compute_fft}
\end{equation}

LoRA freezes the weights of the pretrained base model $W_0$ and only fine-tunes an adapter matrix $\Delta W_t$.
During training, $\Delta W_t$ is stored separately from $W_0$ as two low-rank matrices with rank $r$: $\Delta W_t = B_t A_t$ with $B_t \in \mathbb{R}^{d \times r}, A_t \in \mathbb{R}^{r \times k}, W_0 \in \mathbb{R}^{d \times k}$.
$r$ is typically very small, often between $4$ and $32$.
$\alpha$ is a constant that controls how much the LoRA adapters affect the behavior of the base weights $W_0$.

We hypothesize that LoRA's low-rank updates restrict the model's capacity to memorize precise details, which could have an effect similar to the noisy updates in DP-SGD.
Therefore, LoRA has the potential to provide better privacy-utility trade-offs than FFT, similar to DP-SGD, while also being computationally more efficient.

\noindent
\textbf{Privacy-Utility trade-off:}
We investigate the effects of varying both the rank $r$ and scaling parameter $\alpha$ of LoRA.
We use common rank values of $16$ and $32$ and $\alpha \in \{16,32,64,128\}$.
While LoRA has been explored for privacy in conjunction with DP-SGD~\cite{yu2022differentiallyprivatefinetuninglanguage}, there has been no prior work that specifically examines the privacy benefits of LoRA alone. 

Figures~\ref{fig:lora_16_csima}-~\ref{fig:lora_16_csimc} show the trade-off with rank $16$ and varying parameters of $\alpha$ for the CustomerSim dataset across Pythia, Gemma and Llama2 models. As seen in Figures \ref{fig:lora_16_csima} and \ref{fig:lora_16_csimb}, smaller-scale models (Pythia and Gemma) exhibit a better privacy-utility trade-off when the rank and $\alpha$ values are equal, compared to the other configurations. For the larger model, the privacy declines at later epochs, while utility is mostly retained.
This trend is similar to the one observed for rank $32$ in Figures~\ref{fig:lora_32_csima}-~\ref{fig:lora_32_csimc}. 
We also analyze the trade-off for the SynBio dataset in Figures~\ref{fig:lora_16_piia}-~\ref{fig:lora_16_piic} for rank 16, and Figures~\ref{fig:lora_32_piia}-~\ref{fig:lora_32_piic} for rank 32, which make similar observations. 
However, due to the more unstructured nature of the SynBio dataset, there is a larger reduction in utility after certain epochs compared to the CustomerSim dataset.

\noindent
\textbf{Impact on benchmark datasets:} We use the LoRA model with configuration $r=16, \alpha=16$ for this experiment. Figures \ref{fig:sciq_lora}-~\ref{fig:hs_lora} illustrate the LoRA fine-tuned model's accuracy at each epoch for the three benchmarks. 
The LoRA-based fine-tuned Gemma model retains performance levels close to those of the pre-trained model.

\noindent
\textbf{Efficiency:}
During training, LoRA has slightly larger compute requirements for the forward pass than full fine-tuning, since additional FLOPs are required for the adapter matrices $\Delta W$, though they are much smaller than the full base weight $W_0$.
However, during the backward pass, LoRA requires less compute, since no gradients need to be computed for the base weights $W_0$.
We observe a relative FLOPs requirement of $C_{\text{LoRA}} / C_{\text{FFT}} \approx 0.65$.
The original paper~\cite{hu2022lora} reports a 25\% speedup during training.
LoRA has needs of less GPU memory than FFT, since no optimizer states and gradients need to be stored for the base weights $W_0$, which makes it possible to run it with larger batch-sizes and thus an overall increased training throughput.

\noindent
\textbf{Takeaway:}
We are one of the first works to explore the privacy benefits of parameter-efficient fine-tuning methods, particularly LoRA.
We vary LoRA's $r$ and $\alpha$ hyperparameters and observe that all configurations are able to achieve high utility, while especially lower $r$ and $\alpha$ values also preserve a high degree of privacy.
For each model, the optimal privacy-utility trade-off value is achieved with $r = \alpha$.
In addition, we observe that LoRA, after being fine-tuned on the CustomerSim dataset, did not lose much of its abilities on the benchmark datasets, maintaining results comparable to those of pre-trained model.
Finally, LoRA is much more computationally and memory efficient than FFT and especially DP-SGD.
Overall it provides the best trade-offs in terms of utility, privacy and efficiency and shows that privacy can be achieved without additional computational costs. The degree of measures along the \textit{trade-off, knowledge retention, and efficiency} are: 

\indent Utility-privacy trade-offs: \textit{good} \\
\indent Retention of base performance: \textit{moderate} \\
\indent Efficiency: \textit{good}
\section{Comparison of Fine-tuning Methods}
\label{sec:comparison}

In this section, we compare all fine-tuning methods along three key dimensions—privacy, utility, and efficiency—using the best hyperparameter configurations identified in Section \ref{sec:trade_off}. We present Pareto-optimal curves under a fair setting, where each method is evaluated at its best configuration. By examining these curves, one can select the checkpoint most appropriate for the task at hand.

Recall that \textit{utility is measured as test loss on non-sensitive tokens}. Besides measuring privacy as the \textit{training loss on sensitive tokens}, we define two more metrics for privacy : (a) Privacy Loss over sensitive tokens \cite{10.1145/2976749.2978318} and (b) Canary exposure \cite{carlini2019secret}.

\textbf{Privacy Loss}: Let $\mathcal{D}$ be the dataset with the sensitive token $d$ under consideration, and $\mathcal{D'}$ be the dataset without it. Considering $\mathcal{M}(\mathcal{D})$ as the fine-tuned model that has seen the datapoint $d$, $\mathcal{M}(\mathcal{D'})$ as the  model that has not seen $d$, and leveraging the definition from \cite{10.1145/2976749.2978318}, we can define the privacy loss (PL) as : 
\begin{equation}
    \begin{split}
         PL &= log \frac{P(\mathcal{M}(\mathcal{D})=d)}{P(\mathcal{M}(\mathcal{D'})=d)} = log \frac{P_{\mathcal{D}}(d)}{P_{\mathcal{D'}}(d)} 
    \end{split}
\end{equation}
A natural approximation of such a model with the data unseen is the pretrained model. Recall that negative log-likelihood for a token $d$ is $NLL_{\mathcal(D)}(d) = -log P_{\mathcal{D}}(d)$. We may rewrite the privacy loss as the difference between the negative log-likelihood of the pretrained model and the fine-tuned model over the token $d$ : 
\begin{equation}
    \begin{split}
    PL &= log \frac{P_{\mathcal{D}}(d)}{P_{\mathcal{D'}}(d)} = log P_{\mathcal{D}}(d) - log P_{\mathcal{D'}}(d) \\ &= NLL_{\mathcal{D'}}(d) - NLL_{\mathcal{D}}(d)         
    \end{split}
\end{equation}

Unlike our measure where we observe the absolute loss on sensitive tokens from the training data, this metric observes the relative change in loss w.r.t a model that has never seen the datapoint.

\textbf{Canary exposure}: Originally proposed by \cite{carlini2019secret} to measure unintended memorization, canaries are random sequences (e.g., \textit{$s$=``My ID is $\circ\circ\circ\circ\circ$"}) inserted into training data. Exposure is computed by enumerating all possible sequences from the randomness space $\mathcal{R}$ and calculating the negative log-rank. Following the definition of exposure in \cite{carlini2019secret}, the exposure of a canary $s[r]$ in a model $\mathcal{M}$ over randomness space $\mathcal{R}$ is defined as:
\begin{equation}
    exposure_{\mathcal{M}} = log_2|\mathcal{R}| - log_2 rank_{\mathcal{M}}(s[r])
\end{equation}
where the rank of a canary is its index in the list of all possibly-instantiated canaries, ordered by the
model perplexity of all those sequences. In this setting, we inserted the canary ``My ID is 34175" into the training data for $10$ times and measured its exposure as a metric for privacy.

For efficiency, we measure floating point operations (FLOPs) based on the number of operations incurred (e.g., matrix multiplication, addition, etc.) during training.

\begin{figure*}[t]
    \centering
    \begin{subfigure}{0.245\linewidth}
        \centering
        \includegraphics[width=\linewidth]{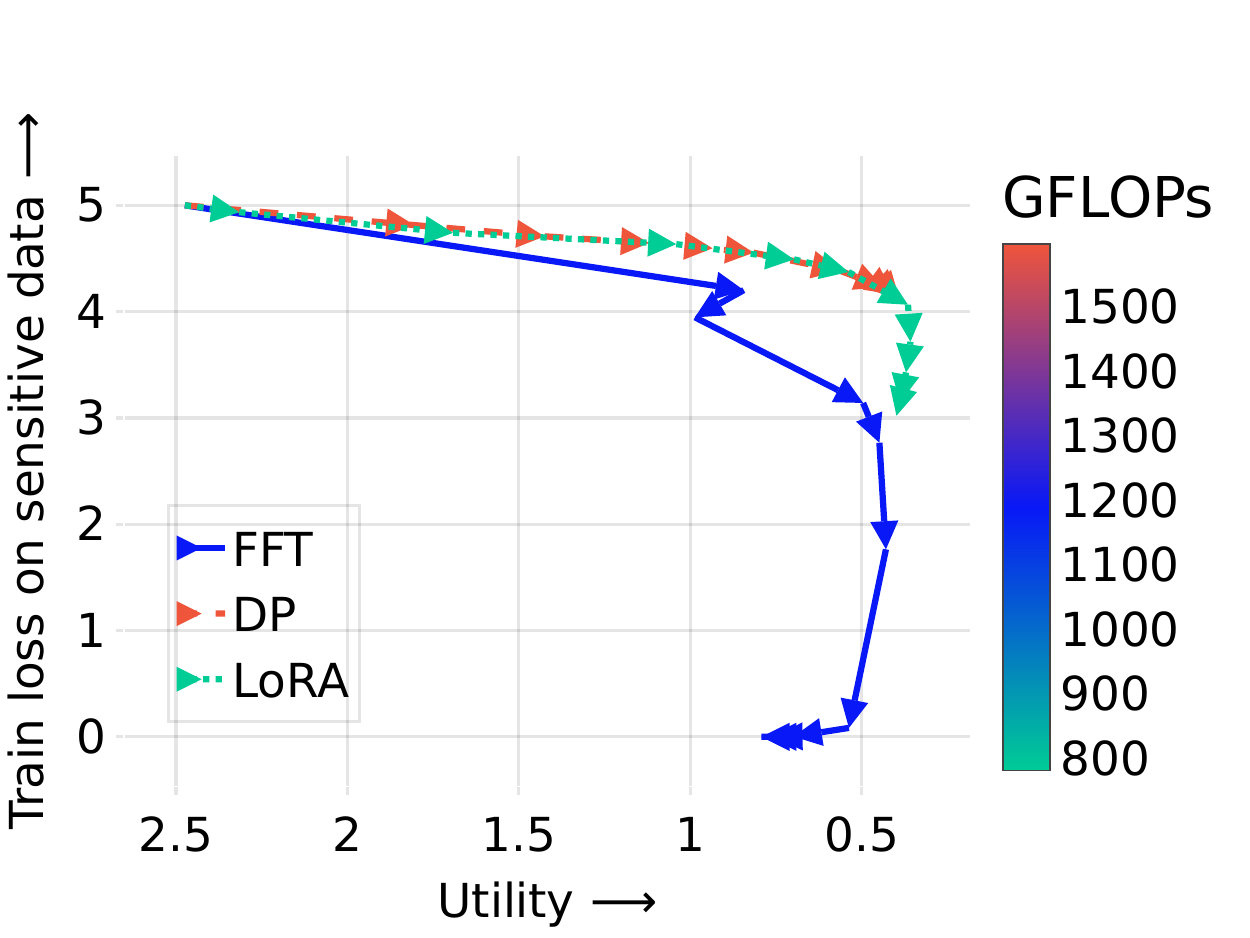}
    \end{subfigure}
    \begin{subfigure}{0.245\linewidth}
        \centering
        \includegraphics[width=\linewidth]{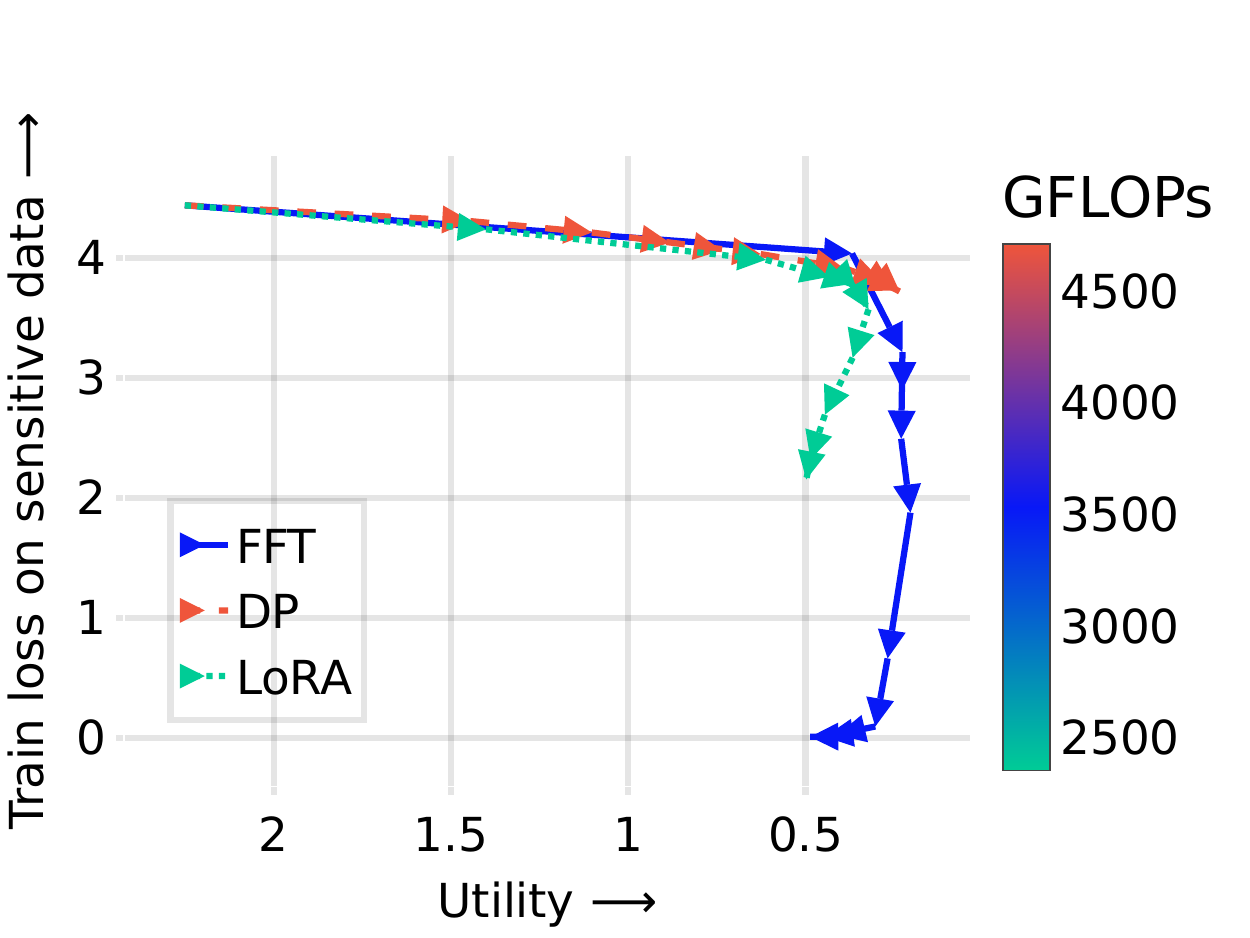}
    \end{subfigure}
    \begin{subfigure}{0.245\linewidth}
        \centering
        \includegraphics[width=\linewidth]{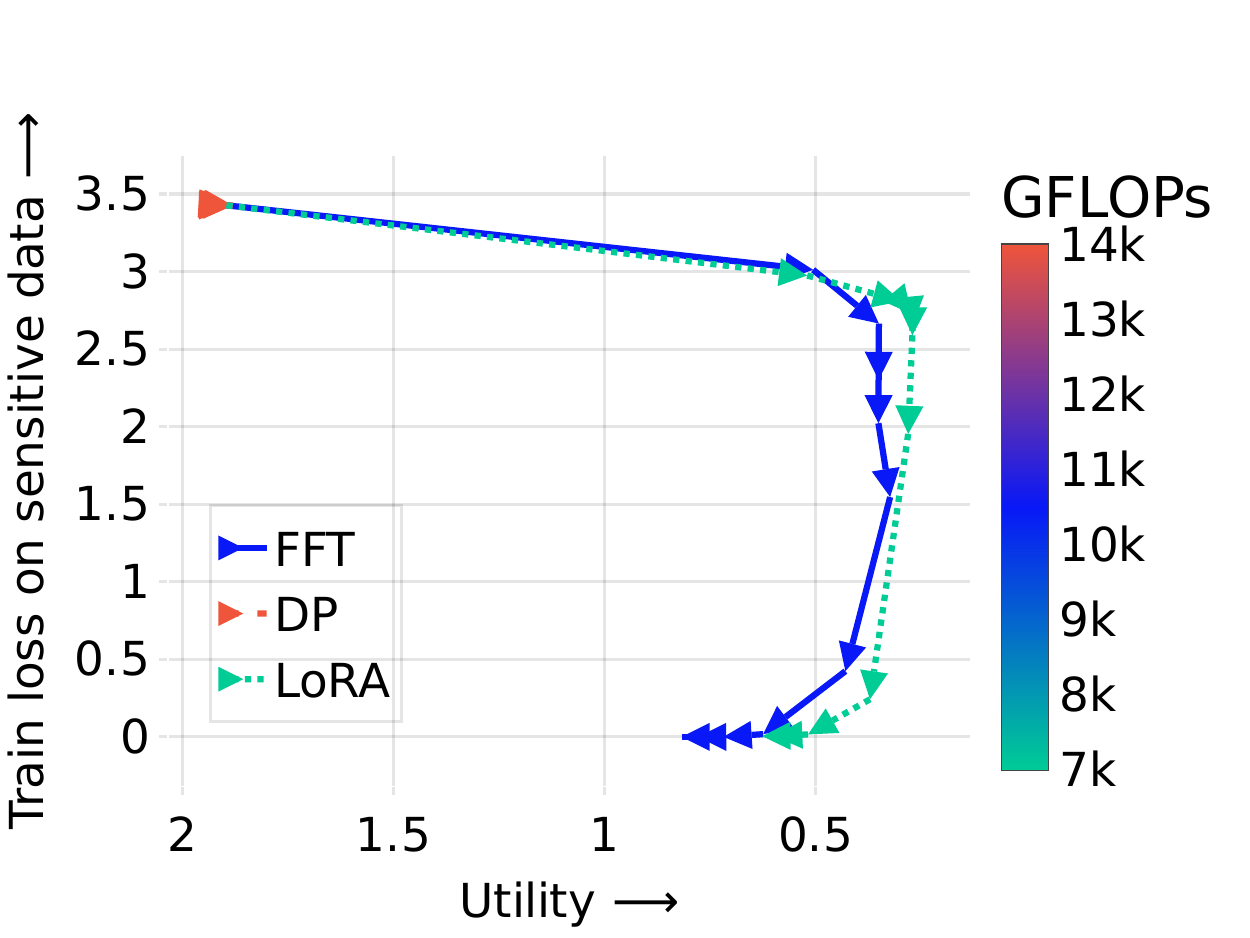}
    \end{subfigure}
    \begin{subfigure}{0.245\linewidth}
        \centering
        \includegraphics[width=\linewidth]{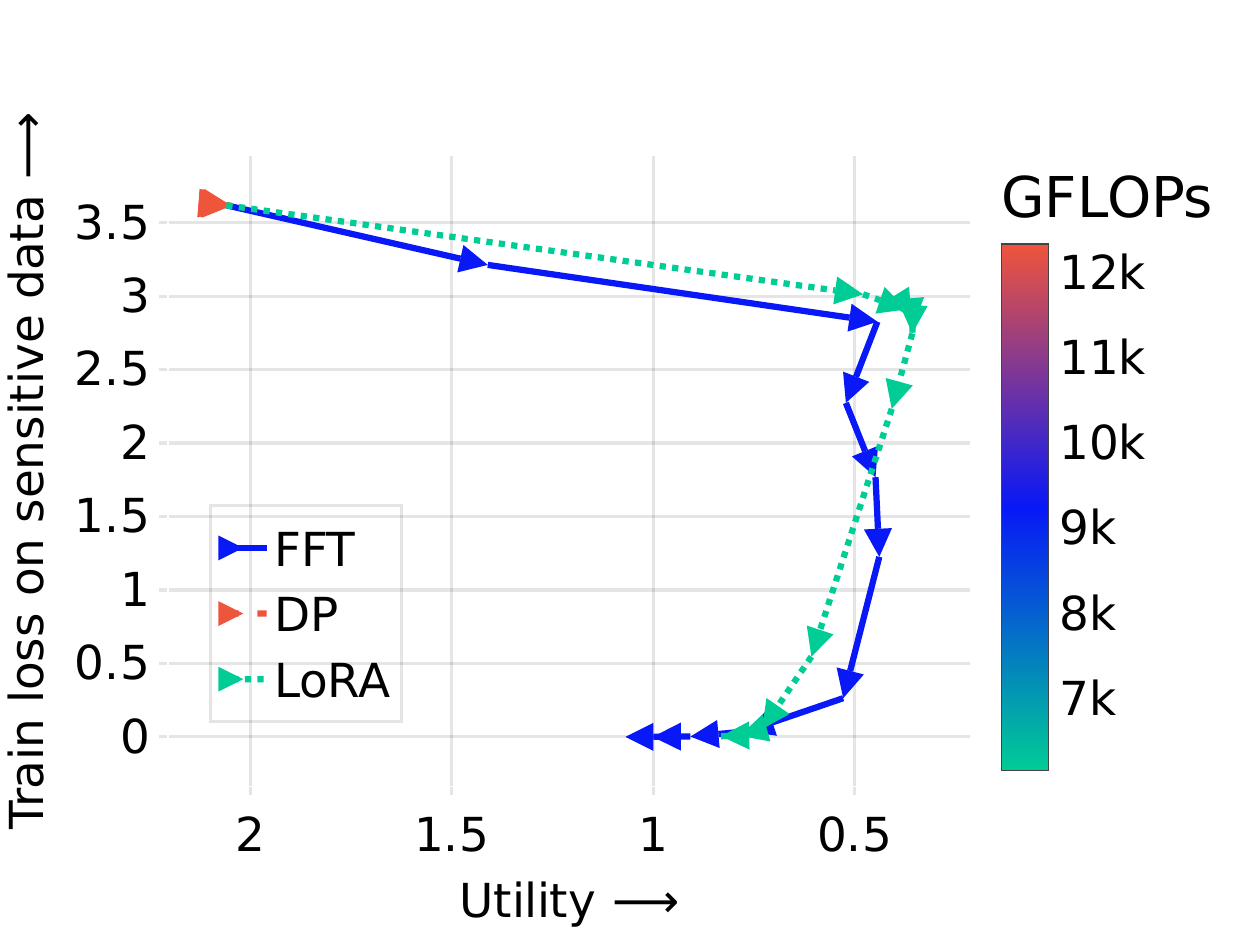}
    \end{subfigure}

    \begin{subfigure}{0.245\linewidth}
        \centering
        \includegraphics[width=\linewidth]{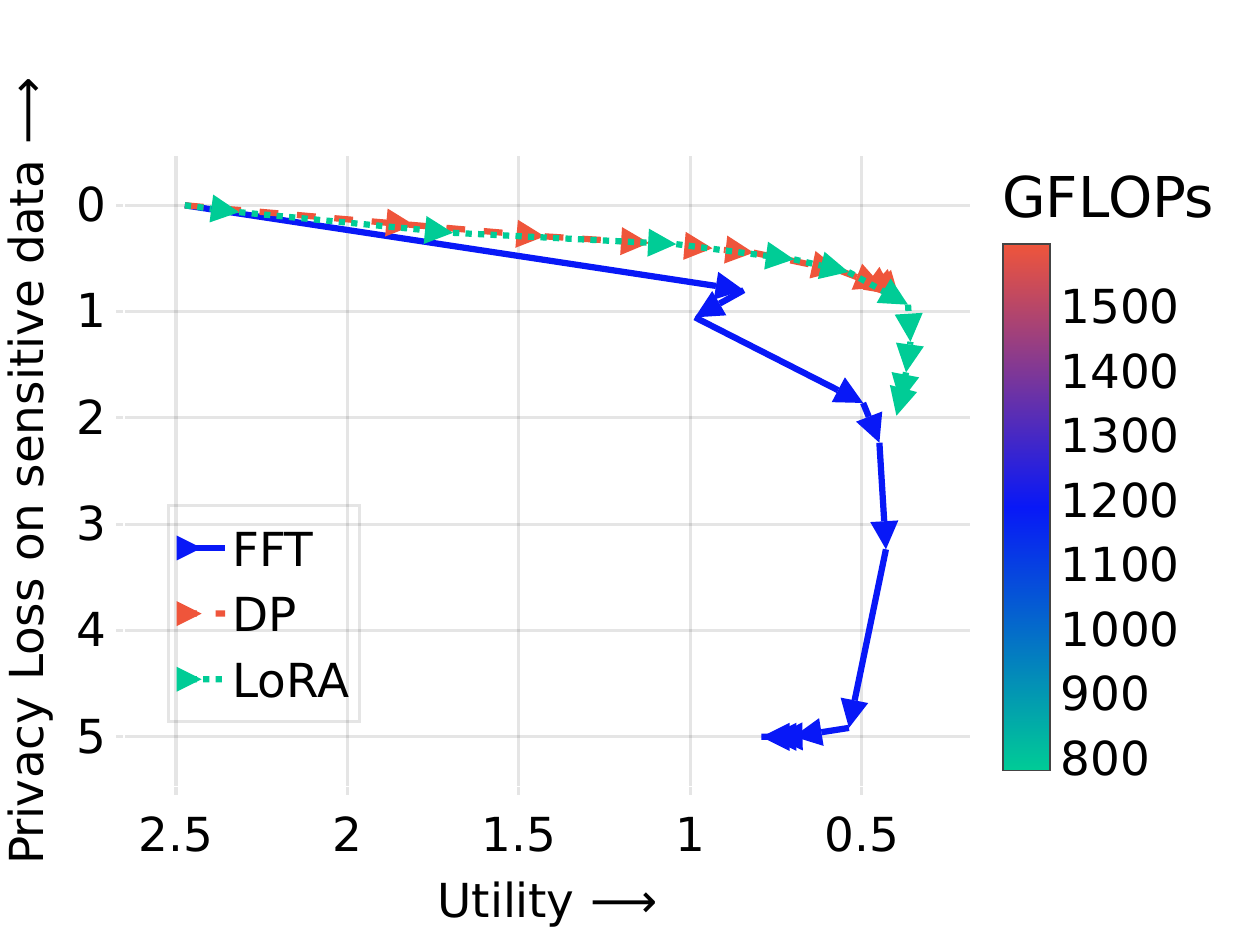}
    \end{subfigure}
    \begin{subfigure}{0.245\linewidth}
        \centering
        \includegraphics[width=\linewidth]{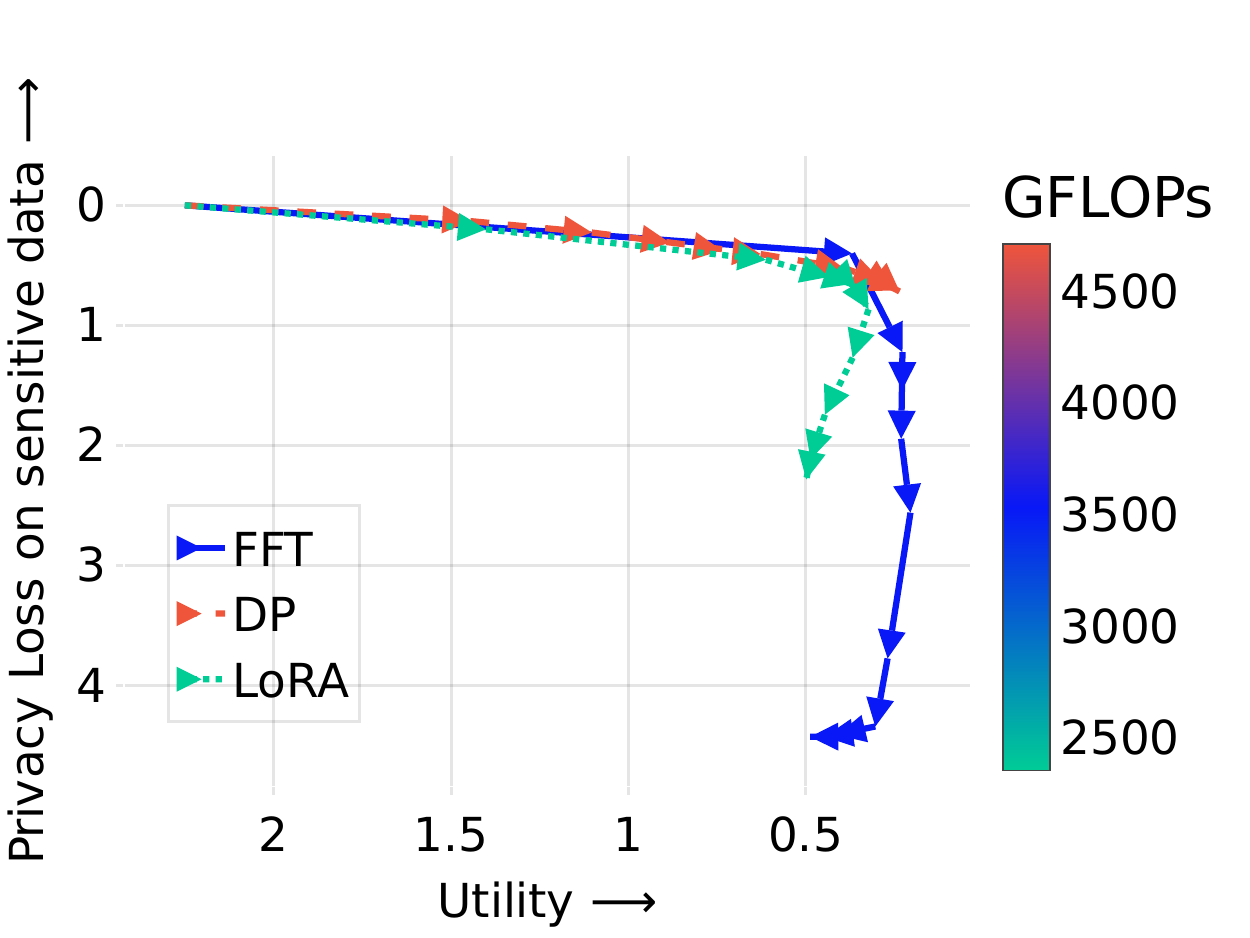}
    \end{subfigure}
    \begin{subfigure}{0.245\linewidth}
        \centering
        \includegraphics[width=\linewidth]{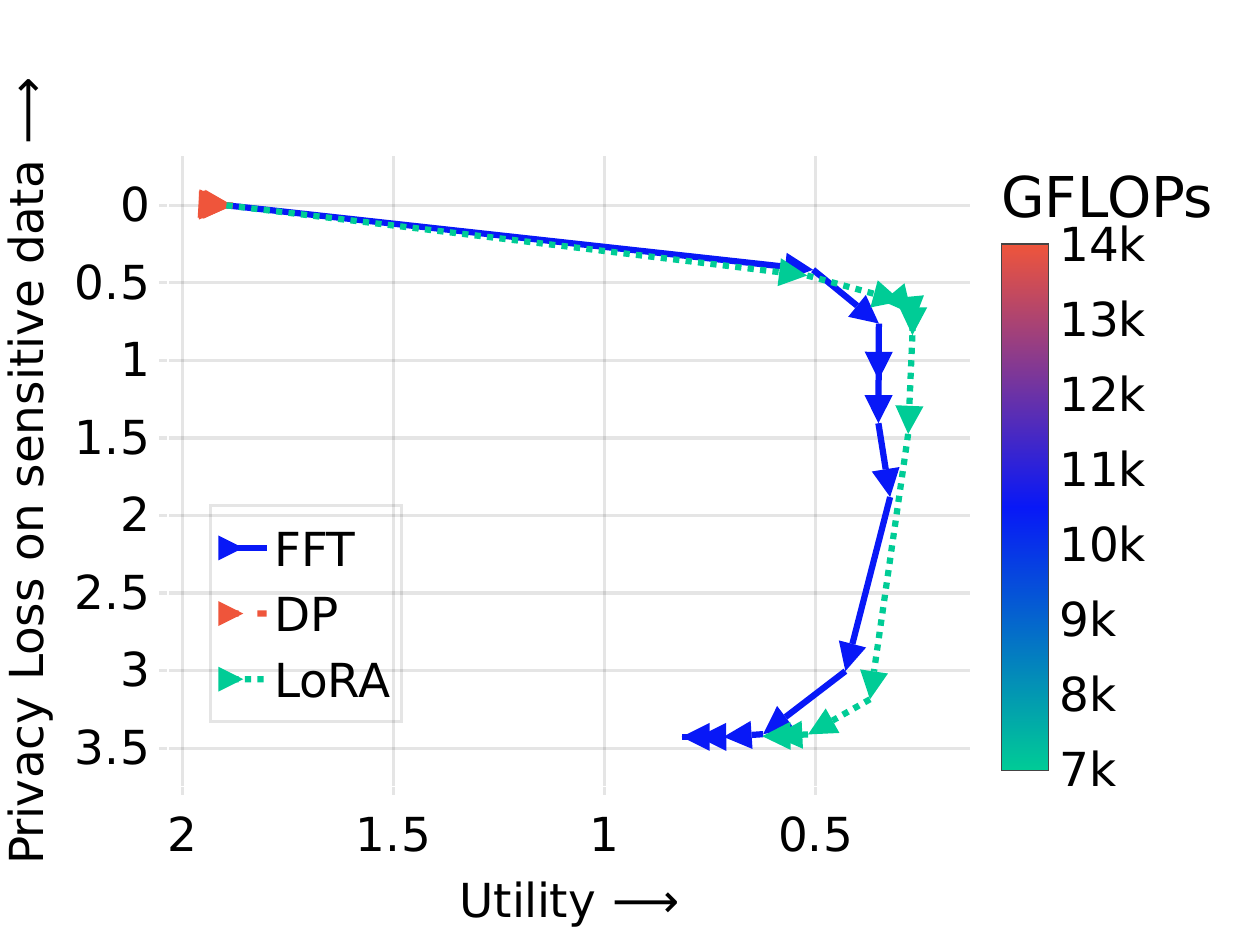}
    \end{subfigure}
    \begin{subfigure}{0.245\linewidth}
        \centering
        \includegraphics[width=\linewidth]{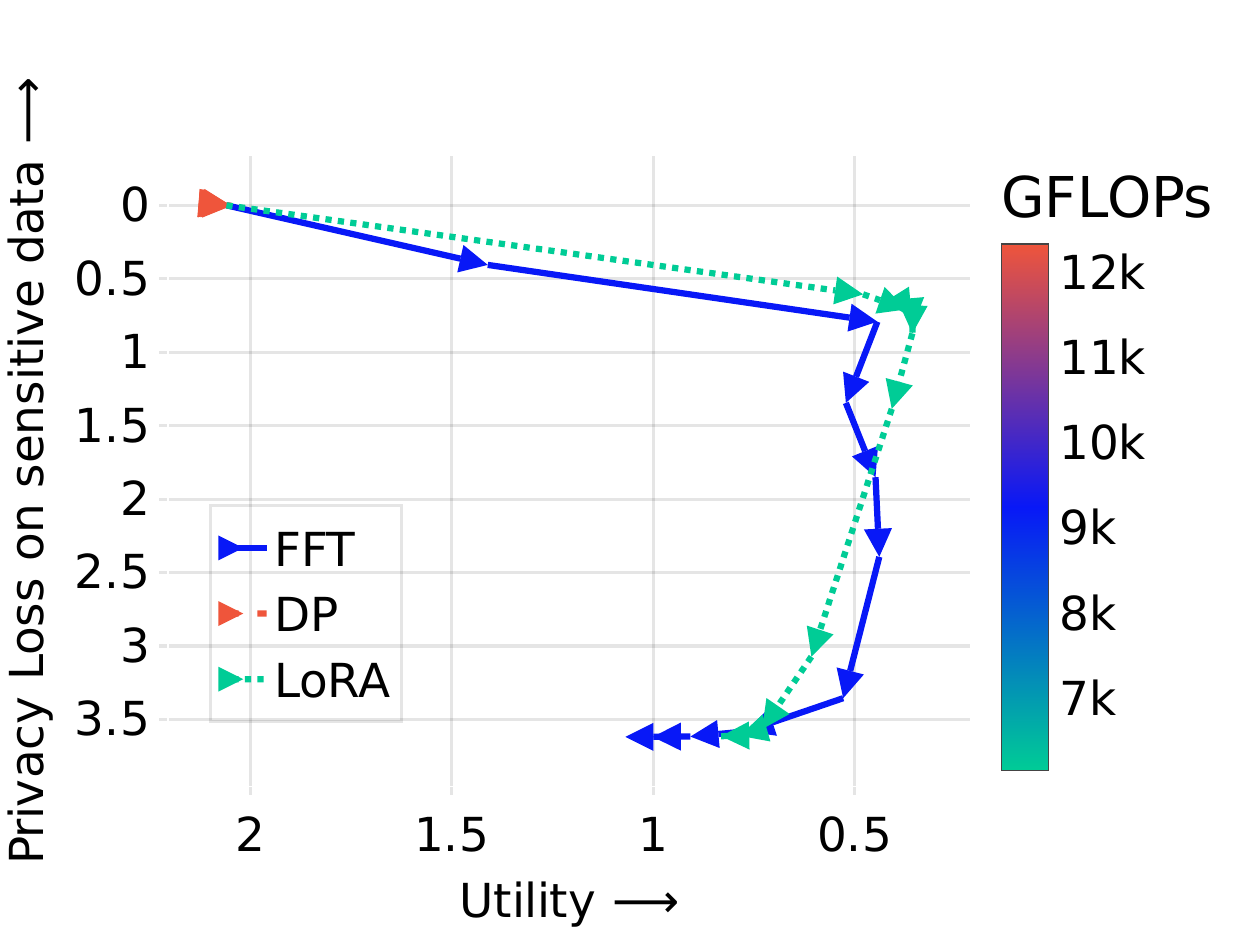}
    \end{subfigure}

    \begin{subfigure}{0.245\linewidth}
        \centering
        \includegraphics[width=\linewidth]{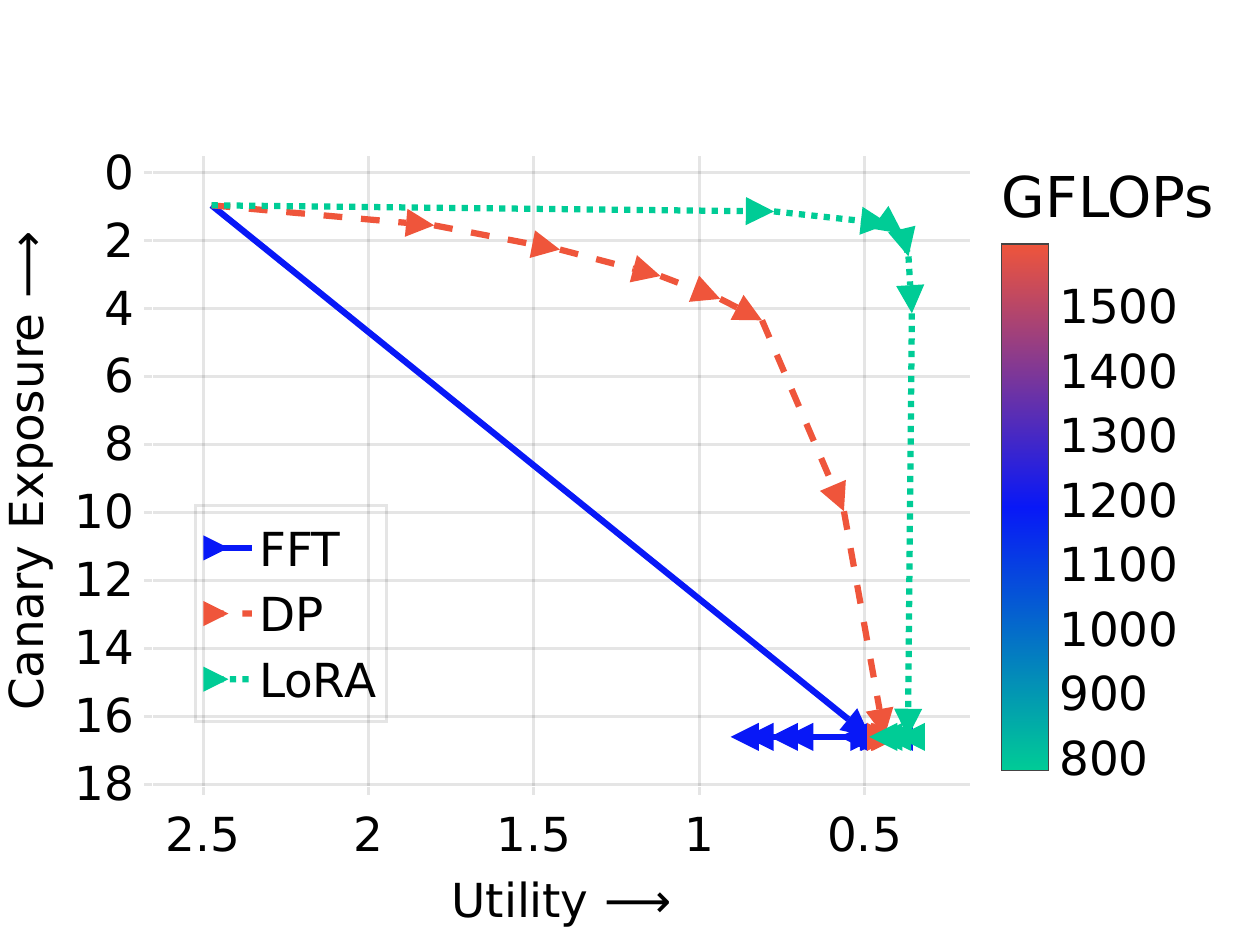}
        \caption{Pythia}
        \label{fig:fdl_csim_pl}
    \end{subfigure}
    \begin{subfigure}{0.245\linewidth}
        \centering
        \includegraphics[width=\linewidth]{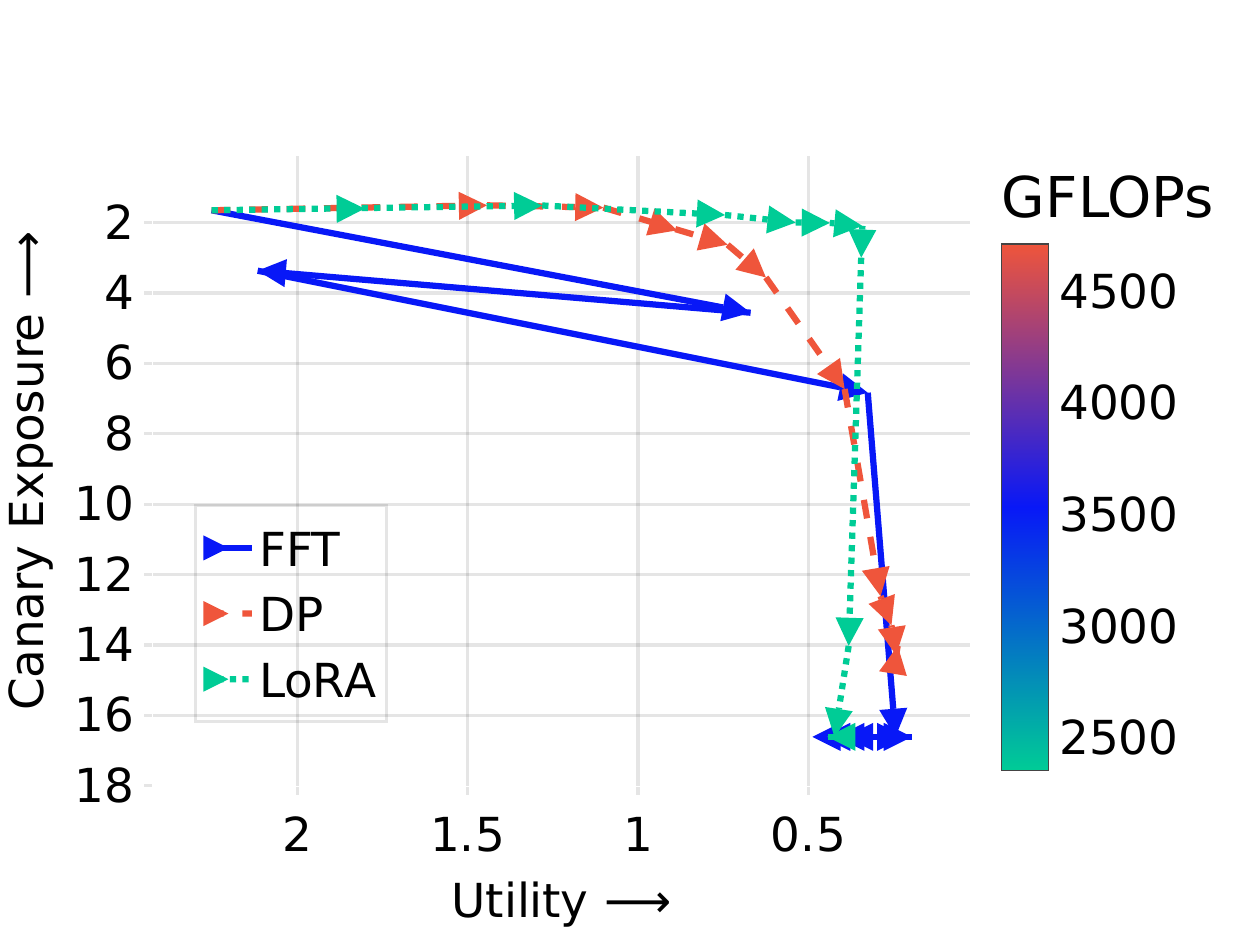}
        \caption{Gemma}
        \label{fig:fdl_csim_pl}
    \end{subfigure}
    \begin{subfigure}{0.245\linewidth}
        \centering
        \includegraphics[width=\linewidth]{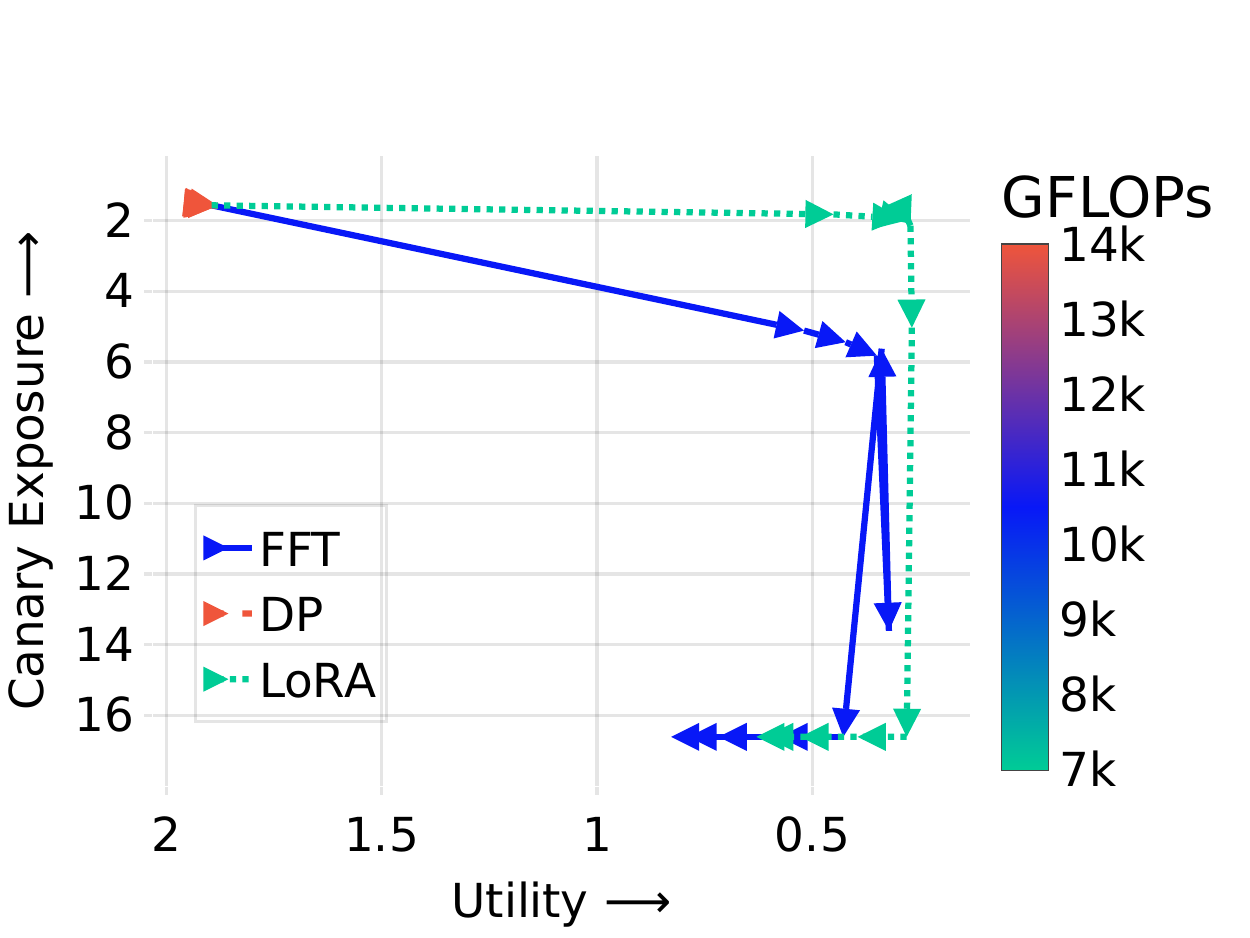}
        \caption{Llama2}
        \label{fig:fdl_csim_pl}
    \end{subfigure}
    \begin{subfigure}{0.245\linewidth}
        \centering
        \includegraphics[width=\linewidth]{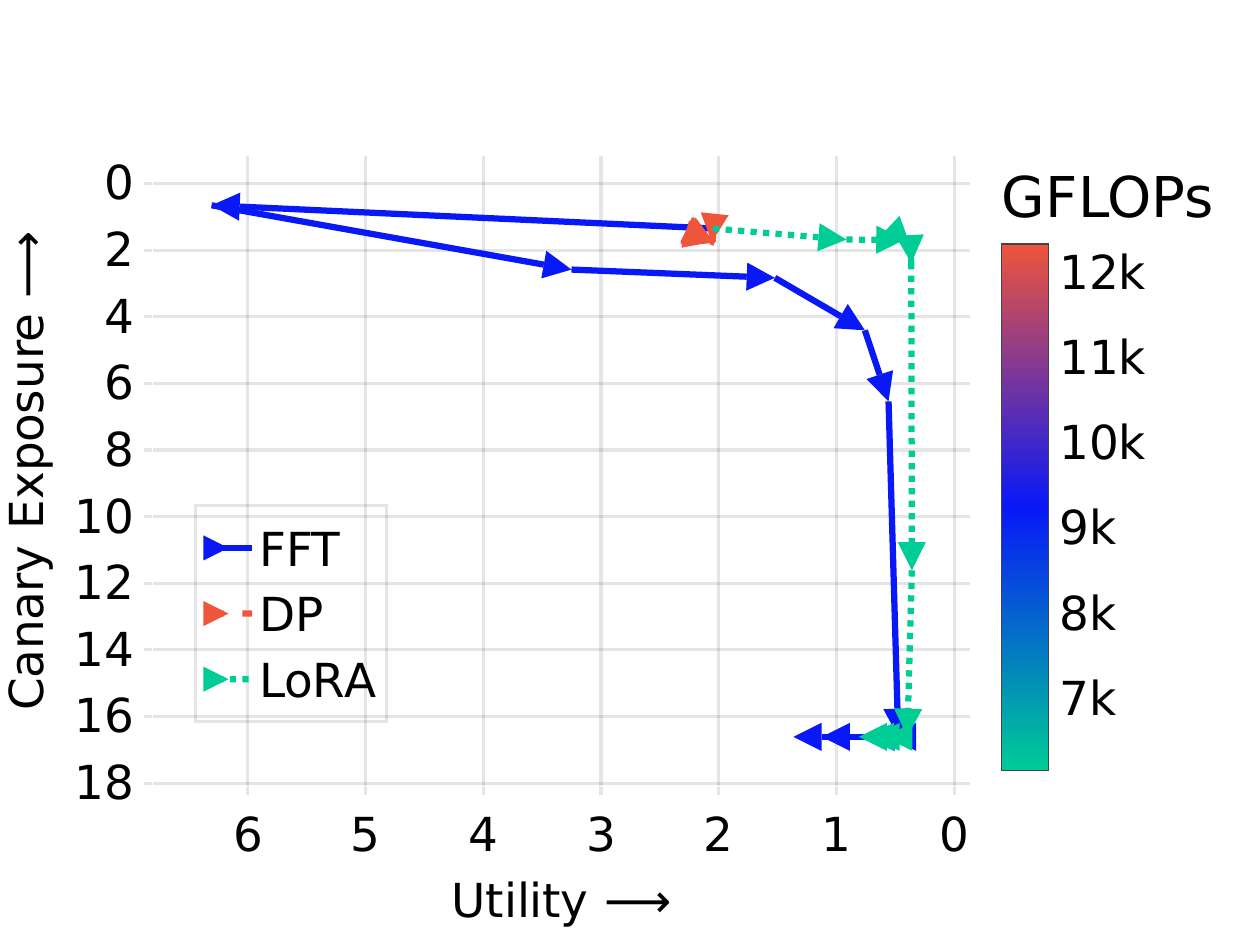}
        \caption{Qwen2.5}
        \label{fig:fdl_csim_pl}
    \end{subfigure}

    \caption{
        Pareto-optimal tradeoffs (\textit{top-row} : Training loss on sensitive tokens ; \textit{middle row} : Privacy Loss on sensitive tokens ; \textit{bottom-row} : Canary exposure). DP-SGD achieves best privacy-utility tradeoffs on small models with reduced utility on large models, albeit with a significantly high computational cost; FFT offers reasonable efficiency-utility tradeoffs with the worst privacy; and LoRA provides a balanced trade-off among all objectives. 
    }
    \label{fig:combined_fdl}
\end{figure*}

We obtain the best configuration for DP-SGD with noise ratio of $\sigma = 0.1$, and for LoRA with $r= \alpha=16 , \sigma=16 $. Figure \ref{fig:combined_fdl} presents the pareto-optimal curves for CustomerSim, comparing the fine-tuning strategies over all the metrics.




\noindent
\textbf{Privacy}: Regarding \textit{privacy}, \textit{FFT} shows \textit{poor privacy} over extended training over all 3 metrics, while \textit{DP} achieves the \textit{highest privacy levels}.
\textit{LoRA }provides \textit{similar privacy as DP} throughout most epochs but declines gradually with extended training, especially on the large-scale model. 
\noindent
\textbf{Utility:}  \textit{FFT} maintains \textit{relatively strong utility} on CustomerSim.
\textit{DP-SGD} while \textit{yielding good utility in smaller models} (Pythia,Gemma), \textit{performs poorly in the larger models} (Llama2,Qwen2.5).
In contrast, \textit{LoRA consistently preserves higher utility} across the entire training period.
\newline
\noindent
\textbf{Efficiency}: The color bar in Figure~\ref{fig:combined_fdl} highlights the FLOPs intensity associated with each fine-tuning strategy.
\textit{DP-SGD} requires the \textit{highest number of FLOPs} (in \textcolor{red}{red}) due to the need for per-sample gradient computation, where each sample corresponds to a token within each training sequence.
\textit{FFT} demands a \textit{moderate number of FLOPs} (in \textcolor{blue}{blue}), proportional to the total number of parameters.
Finally, \textit{LoRA} requires the \textit{fewest FLOPs} (in \textcolor{green}{green}), as especially during backpropagation, most operations only operate on the low-rank matrices.

\begin{figure*}[!t]
    \centering
        \begin{subfigure}{.32\linewidth}
       \includegraphics[scale=0.25]{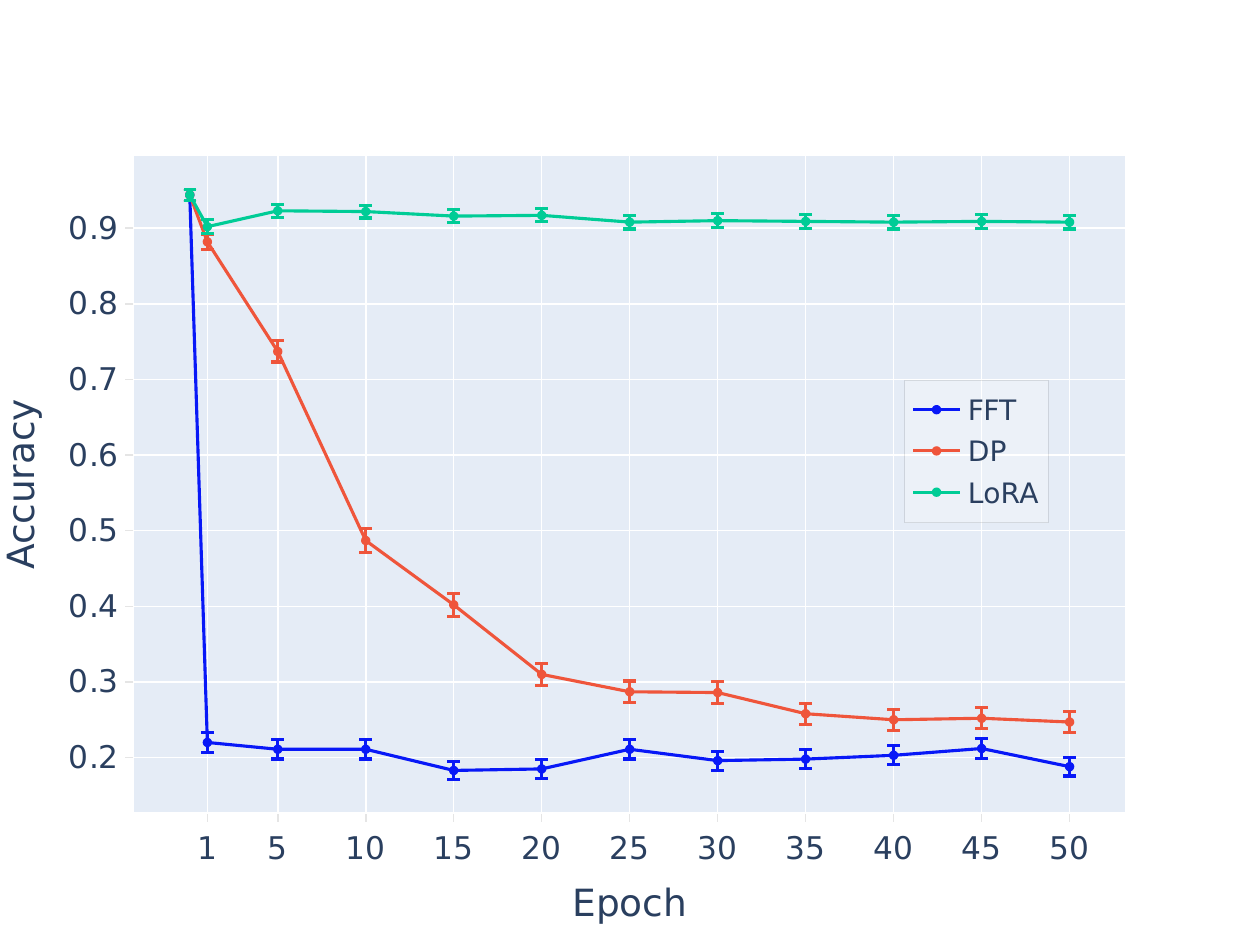}
        \caption{Benchmark SCIQ}
        \label{fig:sciq_all}
    \end{subfigure}
    \begin{subfigure}{.32\linewidth}
       \includegraphics[scale=0.25]{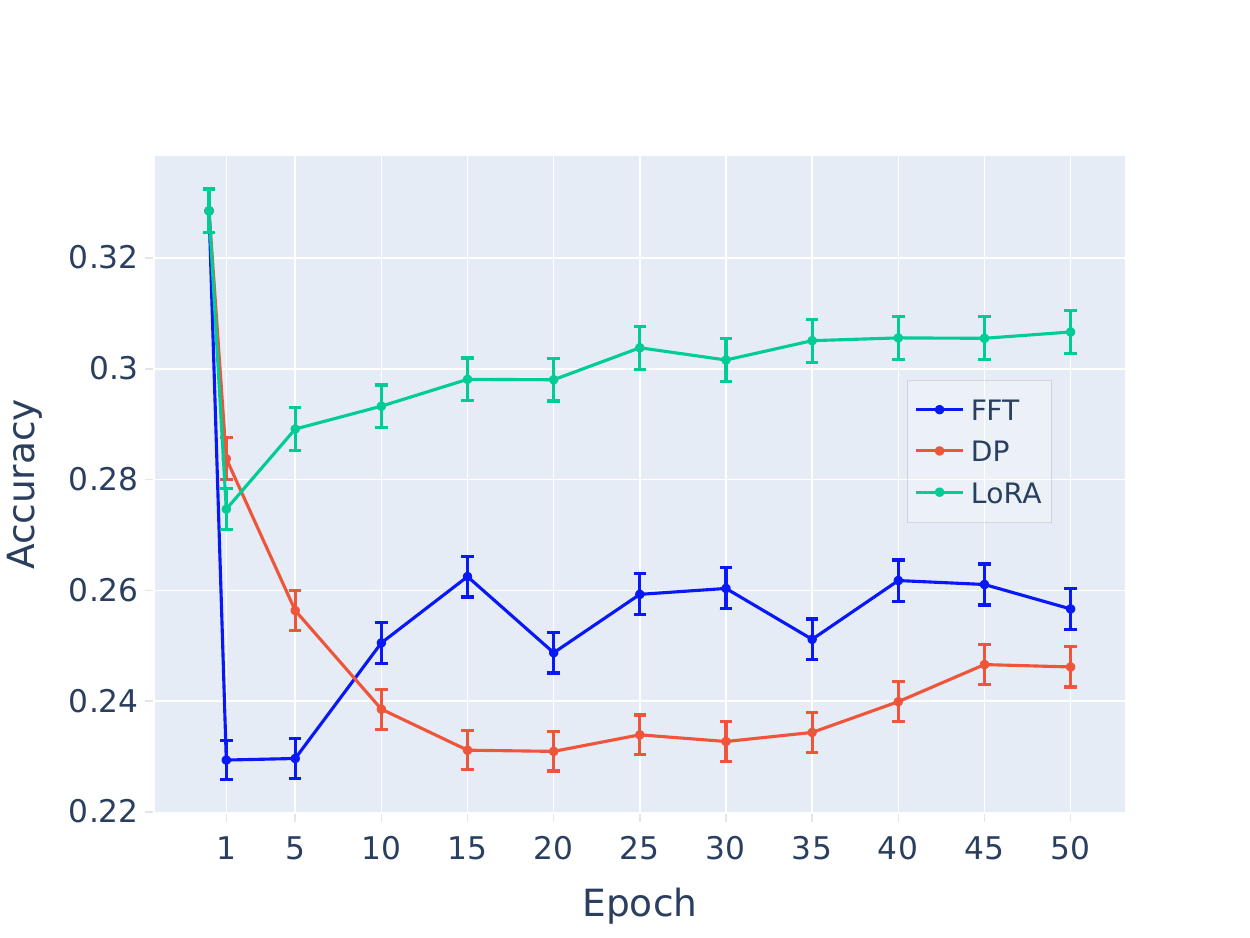}
        \caption{Benchmark MMLU}
        \label{fig:mmlu_all}
    \end{subfigure}
     \begin{subfigure}{.32\linewidth}
       \includegraphics[scale=0.25]{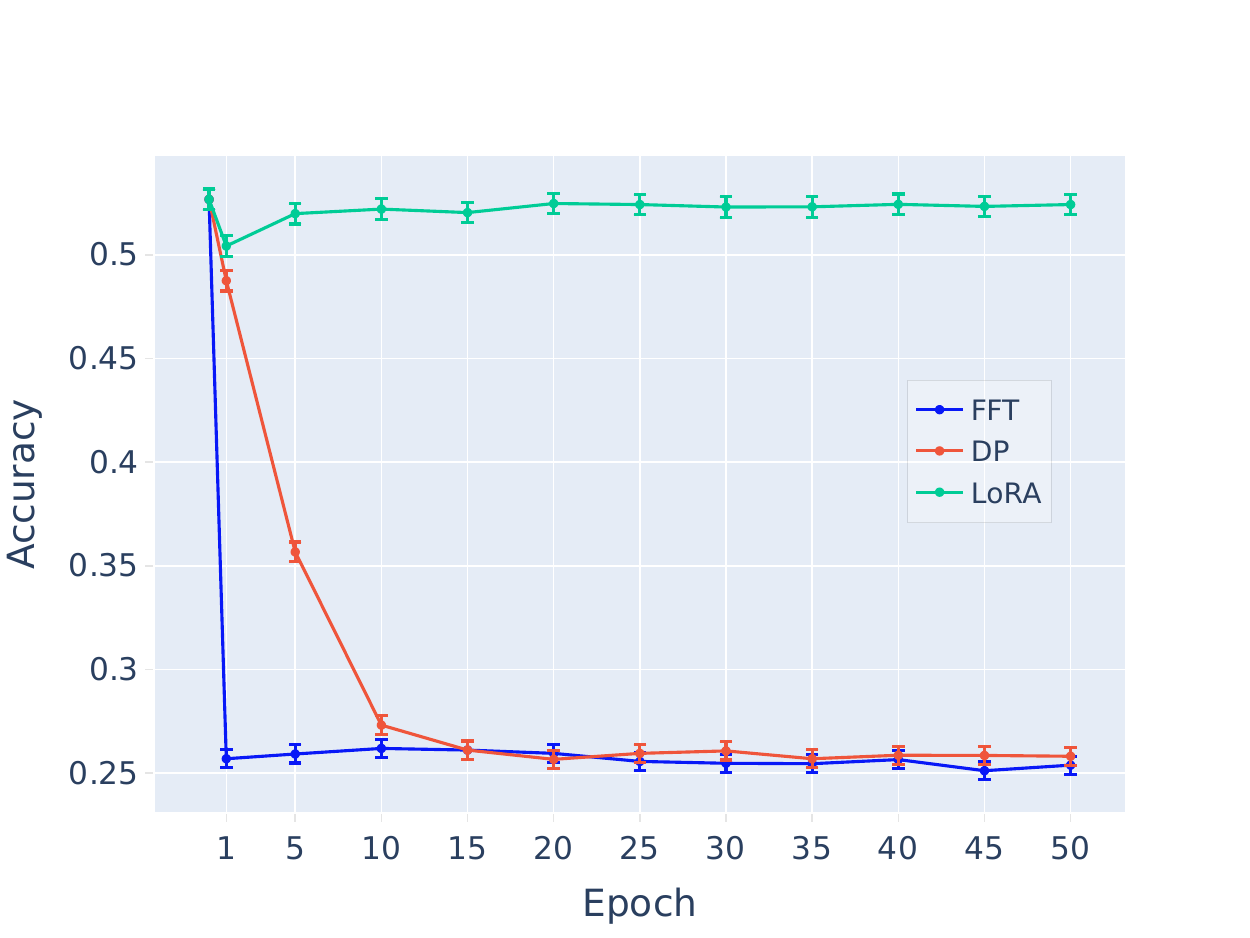}
        \caption{Benchmark HellaSwag}
        \label{fig:hs_all}
    \end{subfigure}
    \caption{
    LoRA demonstrates superior knowledge retention on benchmark datasets throughout the training regime on the CustomerSim dataset, while FFT and DP show fragility with sharp and gradual performance declines, respectively.
    }
    \label{fig:bench-all}
\end{figure*}

\noindent
\textbf{Benchmark performance:} Figure~\ref{fig:bench-all} shows strong knowledge retention capabilities of LoRA on the three benchmarks after being fine-tuned on the CustomerSim dataset for 50 epochs.
FFT and DP, on the other hand, decline sharply and gradually, respectively from the pretrained base model performance.

\noindent
Assessing all the above aspects, we can observe the following: 

1. \textit{Full fine-tuning} achieves \textit{high utility initially, but starts diminishing} after a few epochs, and also witnesses a significant \textit{drop in privacy}.
It has a relatively \textit{high computational cost}.
Additionally, the fully fine-tuned model's performance \textit{diminishes significantly on benchmark datasets}.

2. \textit{DP} offers the \textit{strongest privacy protection} and achieves a \textit{reasonable privacy-utility tradeoff in smaller models}.
However, this tradeoff deteriorates in larger models.
DP incurs the \textit{highest computational cost} as its per-sample noisy gradient updates significantly increase FLOPs and also memory requirements.
Additionally, models fine-tuned with DP exhibit a \textit{gradual decline in their benchmark performance} over the course of training.

3. \textit{LoRA}, a parameter-efficient fine-tuning method, maintains \textit{high utility} and achieves \textit{privacy levels comparable to DP} in smaller models, though this advantage reduces in larger ones. 
Figure~\ref{fig:combined_fdl} shows that while LoRA preserves less privacy as training progresses, it is possible to select checkpoints that balance strong privacy with utility.
This finding challenges the prevailing notion that privacy must come at the cost of high efficiency, demonstrating that \textit{\textbf{LoRA can offer privacy benefits}}.
Moreover, LoRA-tuned models \textit{retain performance on benchmark datasets} close to that of pre-trained models throughout training. Figure \ref{tab:overall} in Appendix \ref{appendix:comparison} shows a comprehensive comparison across all the fine-tuning methods. 
\section{Conjecture: LoRA’s Privacy Benefits Relative to DP-SGD}
\label{sec:conjecture}

In this section, we present a formal conjecture on LoRA’s privacy benefits, drawing an analogy to differential privacy (DP-SGD).

\begin{figure}
    \centering
    \includegraphics[scale=0.25]{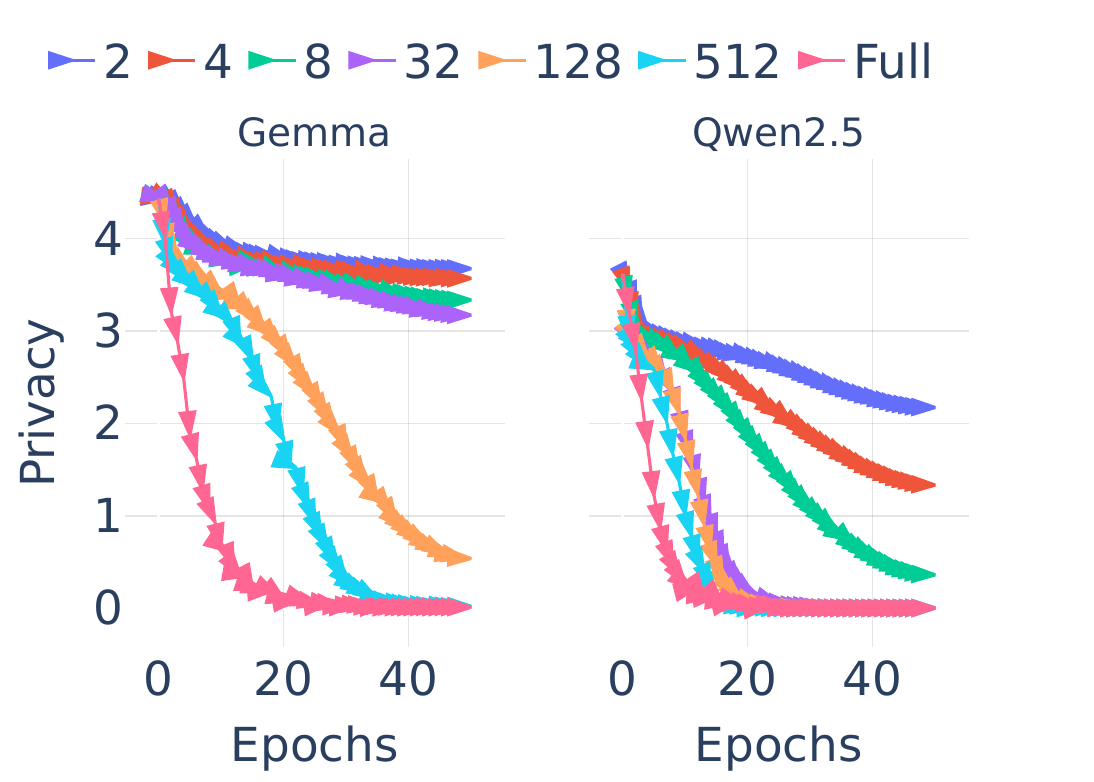}
    \caption{
    Privacy improves with low-rank matrix.
    }
    \label{fig:rank-privacy}
\end{figure}


\textbf{Analogy between LoRA and DP-SGD:} We hypothesize that LoRA's low-rank constraint helps in preserving privacy of the sensitive data. By forcing all the parameter updates into a low-rank subspace, LoRA can reduce the influence a training point can have on the model. Essentially, this is the same quantity that DP-SGD also aims to reduce by clipping and adding noise to the gradients. DP-SGD follows the route of adding randomness, and LoRA follows the route of compressing updates. The effect becomes more on the sensitive data as they are rarely occurring in the training set, which may possibly lead to it being absent in the subspace learnt by LoRA. Empirically, Figure \ref{fig:rank-privacy} shows that low-rank LoRA model results in better privacy than its higher rank counterparts or full fine-tuning. We formally show the analogy in the theorem below:

\begin{theorem}
    Let $D$ and $D'$ be two neighbouring datasets differing by a datapoint $i$. Both LoRA and DP-SGD reduce the influence of $i$ in comparison to a fully trained model, i.e. $\Delta_{DP}^i \leq \Delta_{FFT}^i$ and $\Delta_{LoRA}^i \leq \Delta_{FFT}^i$. This indicates privacy preservation of datapoint $i$ as less the influence, less will be its distinguishability. 
\end{theorem}

 \begin{proof}
     Consider we have $D$ and $D'$ as two neighbouring datasets differing by a datapoint $i$. Let $W(D)$ and $W(D')$ be the respective model parameters where $W(.) \in \mathbb{R}^{d \times k}$. Similarly, let $\Delta W(D)$ and $\Delta W(D')$ be the change in respective aggregate model gradients. Since $D$ and $D'$ differ by a point $i$, we can attribute the difference between $\Delta W(D)$ and $\Delta W(D')$ to the contribution or influence of datapoint $i$ on the aggregate update.

     $\therefore \Delta_{FFT}^i$ = $|| \Delta W(D) - \Delta W(D')||_F = ||g_i||_F$

     For DP-SGD, it is well established that noise (of scale $\sigma$) gets added to the clipped gradients (gradients clipped by a threshold of $T$) that masks the impact of the datapoint \cite{10.1145/2976749.2978318}. We can thereby frame the influence of the datapoint $i$ w.r.t DP-SGD model as:


     $\Delta_{DP}^i$ = $\frac{||g_i||_F}{max(1,\frac{||g_i||_F}{T})} + \mathcal{N}(0,\sigma^2 \mathcal{I}) \approx  \frac{||g_i||_F/T}{\sigma} = \frac{||g_i||_F}{T\sigma} \leq \Delta_{FFT}^i $

     $\therefore \Delta_{DP}^i \leq \Delta_{FFT}^i $

     When LoRA with low rank $r$ is applied, the weight updates get projected to a subspace $\mathcal{S}$ of rank $r$. Let the projection be $P_r : \mathbb{R}^{d \times k} \rightarrow \mathcal{S}_r$. We can frame the influence of the datapoint $i$ w.r.t LoRA model as:

     $\Delta_{LoRA}^i$ = $|| \mathcal{P}_r(\Delta W(D)) - \mathcal{P}_r(\Delta W(D'))||_F = ||\mathcal{P}_r(g_i)||_F$ ; $\mathcal{P}_r$ being a linear projection operator.

     Now, $||\mathcal{P}_r(g_i)||_F \leq ||g_i||_F$ as projection $\mathcal{P}_r$ will drop any point from $g_i$ lying outside its subspace.
     
     $\therefore \Delta_{LoRA}^i \leq \Delta_{FFT}^i $  
 \end{proof}
We can see that both LoRA and DP-SGD reduce the influence of the datapoint ; while DP-SGD does it probabilistically using noise $\sigma$, LoRA does it deterministically through low-rank projection $\mathcal{P}_r$. This aligns with prior work \cite{chaudhuri2013near,kapralov2013differentially} linking principal component analysis (PCA) or low-rank factorization to DP, where PCA limits leakage by discarding components (similar as in LoRA) and DP-SGD by adding noise.
\section{Concluding Discussion}
\label{sec:conclusion}
In this paper, we examine the  trade-offs among privacy, utility, and efficiency while fine-tuning an LLM. The traditional wisdom of achieving privacy comes at the cost of computational inefficiency using dedicated methods like DP. In contrast, we demonstrate that parameter efficient fine-tuning methods like LoRA, initially designed for efficiency, achieves privacy of sensitive data without any computational overhead. Simultaneously, LoRA retains the utility of general language understanding compared to DP, or even full-fine-tuning, realizing the superiority of LoRA in optimizing all three aspects. Towards our investigation, we  establish the significance of redefining privacy and utility using a careful distinction between sensitive and non-sensitive counterparts of the fine-tuned data. Through case studies, we demonstrate how existing measures exaggerate privacy threats and undermine the utility of an LLM. Our paper calls for a joint venture of   privacy and systems communities in achieving privacy-aware efficient fine-tuning of LLMs while retaining utility.
\section{Ethics considerations}
\label{sec:ethics}

This study utilized publicly available datasets~\cite{shi-etal-2022-selective, pii}, some of which included identifiable information such as personal details. However, third-party organizations pre-processed and validated the datasets to ensure that no real individuals’ data were present, thus mitigating potential privacy concerns.

This project received ethical clearance from the Ethical Review Board of the affiliated institution on October 21, 2024 (Approval No. 24-09-4), with no ethical concerns raised.

\section{Open science}
\label{sec:openscience}
This work promotes transparency and reproducibility in research on privacy and utility in large language models (LLMs). To enable further investigation, we will release:
\begin{itemize}
    \item[1.] Code and Framework: The implementation of our proposed privacy measurement framework, which distinguishes between sensitive and non-sensitive tokens, along with scripts for privacy leakage analysis and utility-efficiency evaluation.
    \item[2.] Datasets and Preprocessing information: Links to publicly available datasets used in this study, along with preprocessing scripts to ensure reproducibility. Sensitive data were excluded or anonymized to comply with ethical standards.
    \item [3.] Evaluation Pipeline: An open-source pipeline for assessing privacy leakage and the trade-offs between privacy, utility, and efficiency in LLMs. 
\end{itemize}

These resources aim to support reproducibility and further research into privacy-aware, efficient LLM development.

\bibliographystyle{plain}
\bibliography{sections/references}

\section*{Appendix}
\noindent{\textbf{Overview of Appendices}}
\begin{itemize}
    \item Appendix~\ref{app:limitations}: Limitations
    \item Appendix~\ref{app:llm-usage}: Use of LLMs
    \item Appendix~\ref{appendix:hp-models}: Hyperparameters
    \item Appendix~\ref{appendix:validating-gpt4-preds}: Validating GPT-4 predictions
    \item Appendix~\ref{appendix:example-priv-annotatation}: Prompt for annotating data.
    \item Appendix~\ref{appendix:human-survey-priv-annotatations}: Assessing quality of annotations through human survey
    \item Appendix~\ref{appendix:exploring-tradeoffs}: Privacy Utility Efficiency tradeoffs
    \item Appendix~\ref{appendix:comparison}: Overall comparison
\end{itemize}

\section{Limitations}
\label{app:limitations}
One of the limitations of our work lies in fine-tuning the models for unsupervised setup and not extending it to other supervised tasks like question-answering and so on. However we believe that our findings would hold in any task setup, given that the nature of these fine-tuning techniques would not change. We experimented with LoRA as one the most generic PEFT methods -- however, testing out other PEFT methods would be an interesting extension of our work to explore privacy benefits extensively in the systems community. 

We consider the sensitive data to be from PIIs defined in GDPR Article 4(1) \cite{EU2016GDPR}. However, sensitive data can also be context-dependent. While one piece of information may be sensitive for one, it may not be for the other,. However, we also acknowledge that while contextual integrity is an important task, it is still hard to formalize or implement as seen in \cite{shvartzshnaider2025position}.
Lastly, while this work provides an empirical measure of privacy across different fine-tuning methods, one can definitely use such a measure for optimisation during training and establish a theoretical bound on the privacy benefits that would then also be empirically validated. We intend to explore these directions in the future. 

\section{Use of LLMs for Paper-writing}
\label{app:llm-usage}
In this paper, we use LLMs for the following purposes:

\begin{enumerate}
\item \textbf{Polishing the writing}: We check for grammatical mistakes, and make minor para-phrasing to improve the flow and coherence of the paper.
\item \textbf{Related Work}: Besides traditional search, we use the OpenAI Deep Research to identify relevant literature.
\end{enumerate}

\section{Hyperparameters}
\label{appendix:hp-models}

The following hyperparameters were used for fine-tuning our models:
\begin{table}[H]
\centering
\scriptsize
\caption{Hyperparameters used for fine-tuning methods}
\begin{tabular}{l|c|c|c}
\textbf{Hyperparameter} & \textbf{Full Fine-Tuning (FFT)} & \textbf{LoRA} & \textbf{DP} \\
\toprule
Learning Rate           & 0.00025                         & 0.00025       & 0.00005         \\
Scheduler               & Linear                          & Linear        & Linear          \\
Warmup Steps            & 10                              & 10            & 10              \\
Clipping gradient          & -                               & -             &  (\(1 \times 10^{-2}\)) \\
\bottomrule
\end{tabular}
\end{table}

\begin{table}[H]
\centering
\scriptsize
\caption{Batch sizes for CustomerSim and SynBio}
\begin{tabular}{l|l|c|c|c}
\textbf{Dataset} & \textbf{Model} & \textbf{Full Fine-Tuning (FFT)} & \textbf{LoRA} & \textbf{DP} \\
\toprule
\multirow{3}{*}{CustomerSim} & Pythia & 16 & 32 & 8 \\
                             & Gemma  & 8  & 16 & 4 \\
                             & Llama2 & 4  & 8  & 2 \\
\midrule
\multirow{3}{*}{SynBio}         & Pythia & 8  & 16 & 4 \\
                             & Gemma  & 4  & 8  & 2 \\
                             & Llama2 & 2  & 4  & 2 \\
\bottomrule
\end{tabular}
\end{table}

\section{Validating GPT-4 Predictions}
\label{appendix:validating-gpt4-preds}

\begin{figure}[!h]
    \centering
    \includegraphics[scale=0.2]{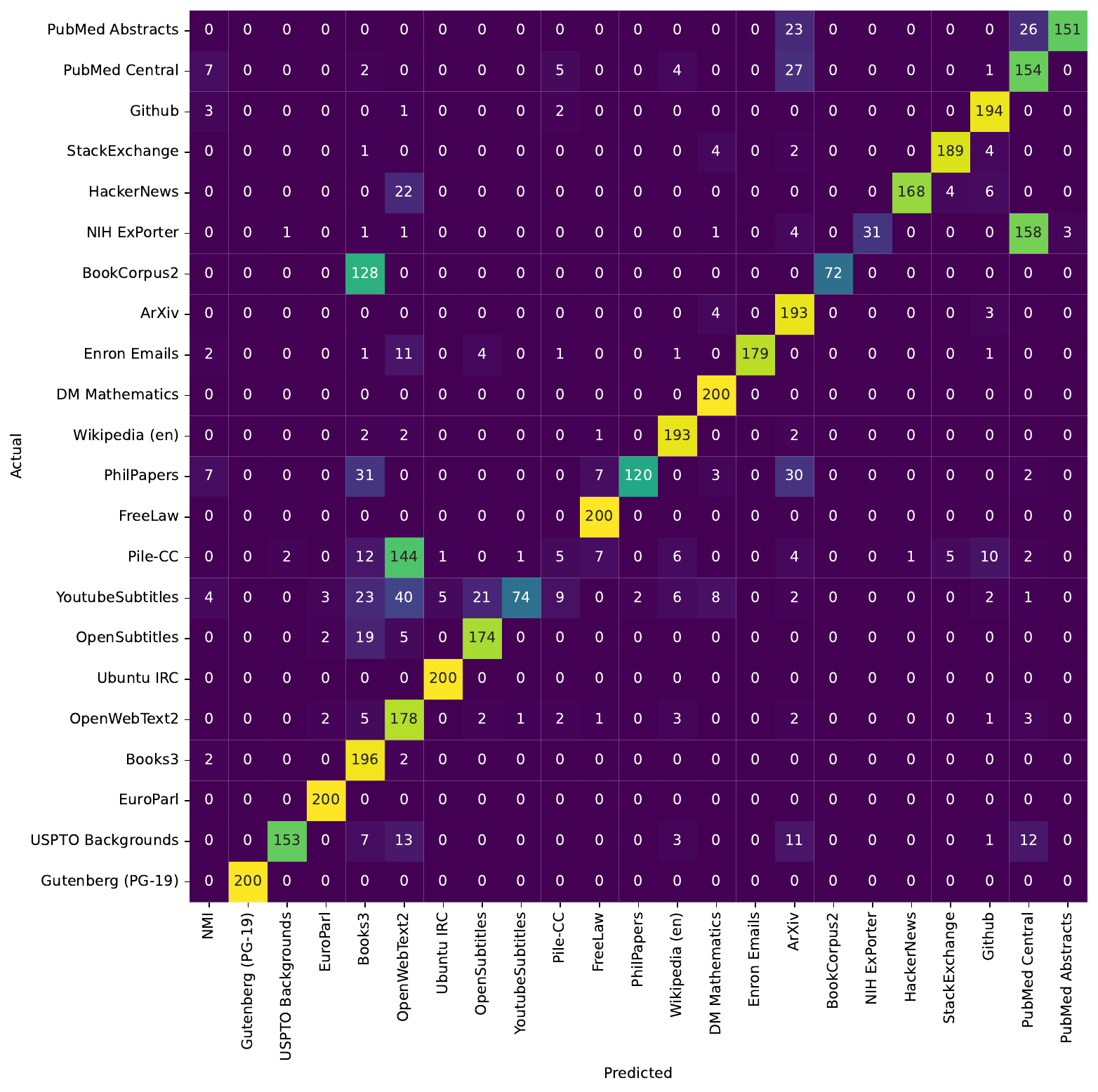}
    \caption{
    Misclassified GPT-4 instances were often placed in similar sections (e.g., \textit{NIH Explorer misclassified as PubMed Central}), suggesting these misclassifications may not reflect clear inaccuracies, complicating definitive accuracy assessment.
    }
    \label{fig:pile-heatmap}
\end{figure}

We analyzed the predictions shown in the heatmap in Figure~\ref{fig:pile-heatmap} and observed that misclassified instances were frequently assigned to sections of a similar nature (e.g., \textit{NIH Explorer misclassified as PubMed Central}).
This overlap suggests that the nature of the misclassifications may not always indicate clear inaccuracies, making it difficult to definitively assess the accuracy of certain misclassified instances.

\section{Prompt for Annotating Privacy-Sensitive Information}
\label{appendix:example-priv-annotatation}

We used the following prompt to obtain the privacy-sensitive information from GPT-4o for CustomerSim:
\begin{mdframed}[backgroundcolor=lavender, linewidth=0pt]
\small
\textit{``You are a private data identifying bot. You will be provided a conversation between a user and a customer service bot. You need to identify ALL instances of private data in the conversation. Private data includes any information that can be used to identify an individual, such as names, phone numbers, email addresses, and locations. It can also include any other sensitive information, such as credit card numbers, social security numbers, tracking data, health information etc.. Identify all of these use your discretion to identify any other information that can be used to identify an individual.''} 
\end{mdframed}
\normalsize



\section{Human Survey: Assessing the Quality of Privacy-Sensitive Annotations from Presidio and GPT-4}
\label{appendix:human-survey-priv-annotatations}

\begin{figure*}[!h]
    \centering
     \begin{subfigure}{.49\linewidth}
        \includegraphics[width=\textwidth]{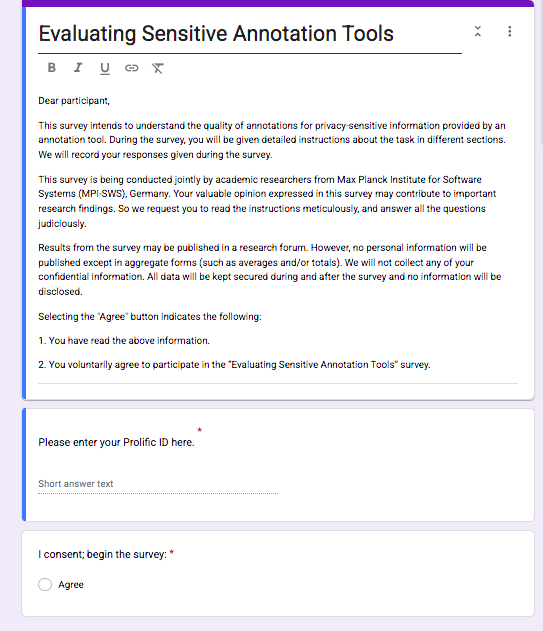}
        \caption{Instructions for annotators}
        \label{fig:survey-1}
    \end{subfigure}
     \begin{subfigure}{.49\linewidth}
        \includegraphics[width=\textwidth]{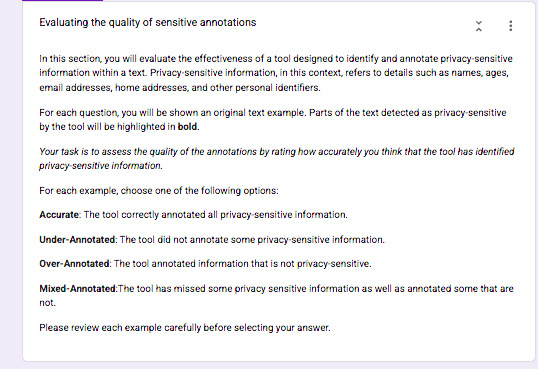}
        \caption{Guidelines for annotators}
        \label{fig:survey-2}
    \end{subfigure}
    \begin{subfigure}{.49\linewidth}
        \includegraphics[width=\textwidth]{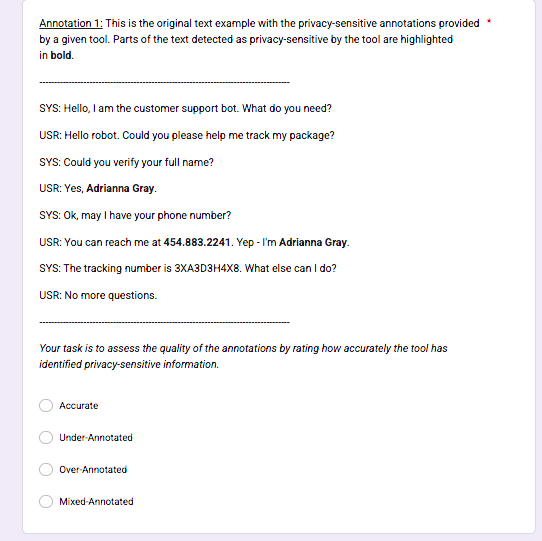}
        \caption{Sample Question 1}
        \label{fig:survey-3}
    \end{subfigure}
     \begin{subfigure}{.49\linewidth}
        \includegraphics[width=\textwidth]{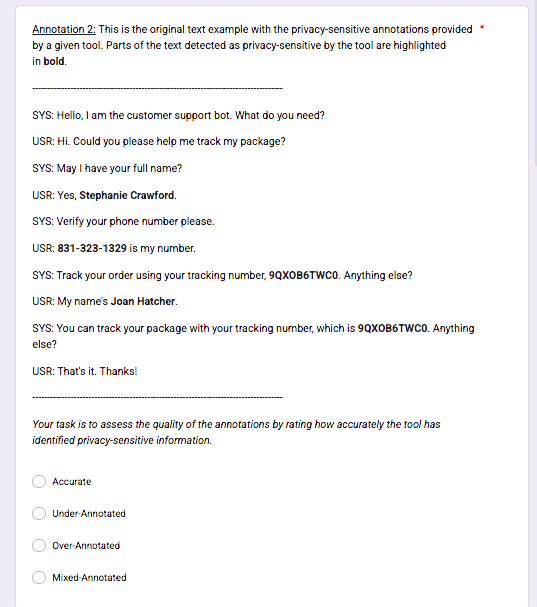}
        \caption{Sample Question 2}
        \label{fig:survey-4}
    \end{subfigure}
    \caption{
    Annotators were provided with these instructions and guidelines to mark the samples as accurately annotated, under-annotated, over-annoted or mixed-annotated.
    }
    \label{fig:survey-images}
\end{figure*}

\begin{figure}[!h]
    \centering
    \begin{subfigure}{.49\linewidth}
        \includegraphics[scale=0.22]{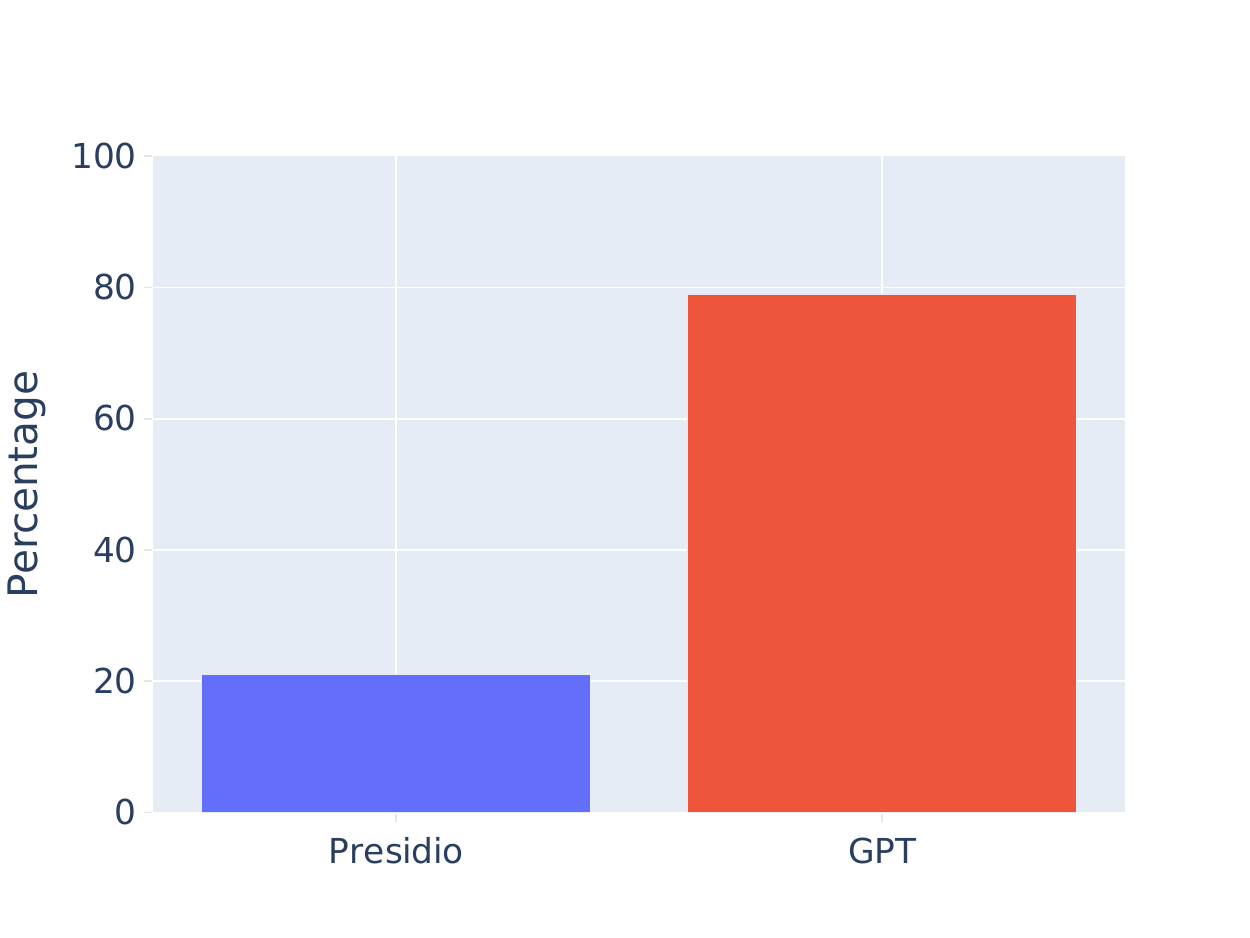}
        \caption{Binary survey}
        \label{fig:binary-pii}
    \end{subfigure}
    \begin{subfigure}{.49\linewidth}
        \includegraphics[scale=0.22]{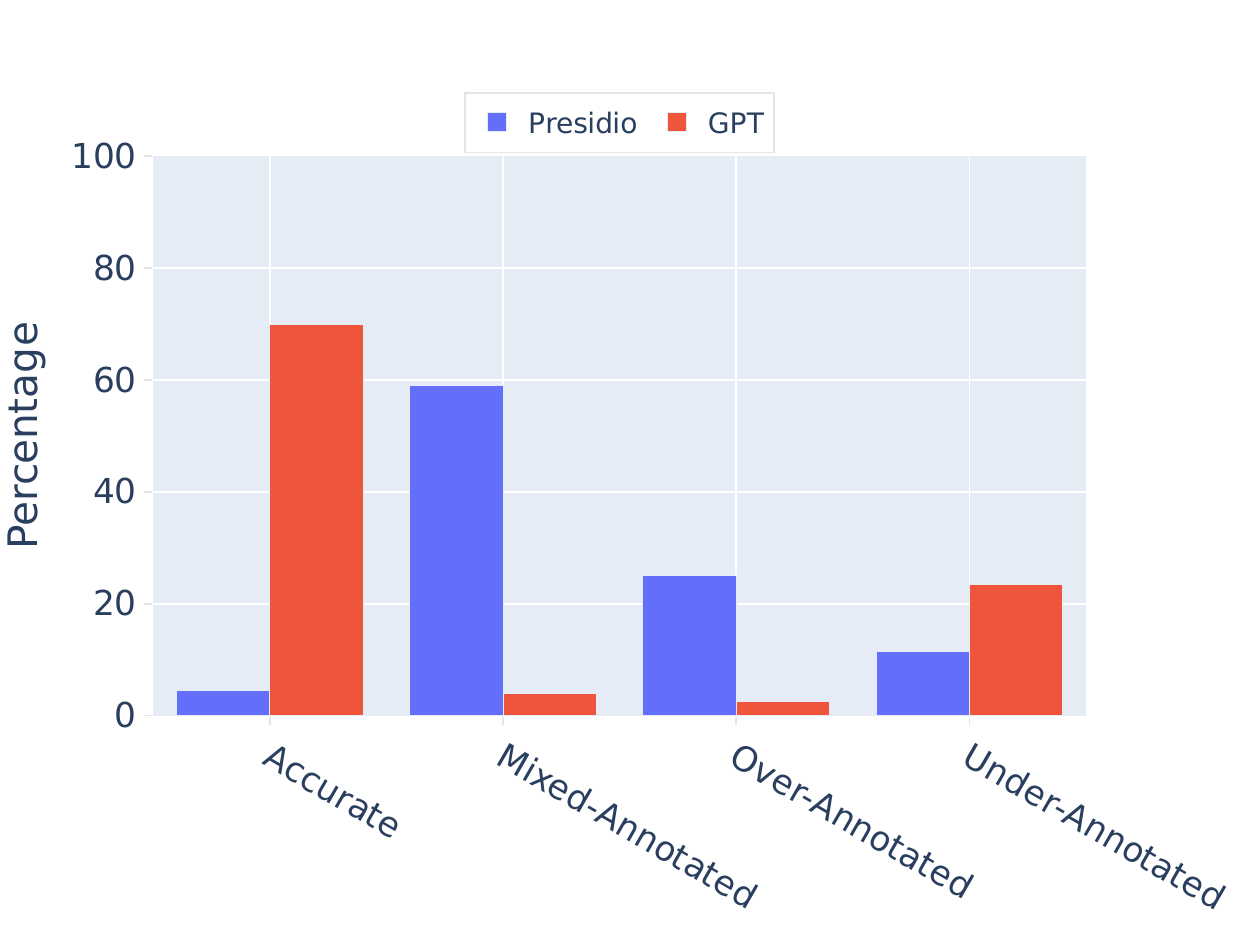}
        \caption{Multiple-choice survey}
        \label{fig:mult-pii}
    \end{subfigure}
    \caption{
    (a) GPT-4 was also preferred in SynBio by human annotators for identifying privacy-sensitive information.
    (b) GPT-4 also demonstrates in SynBio higher accuracy than Presidio for identifying privacy-sensitive information.
    }
    \label{fig:human-survey-quest-tools}
\end{figure}

We evaluated privacy-sensitive annotations from Presidio and GPT-4 through two Prolific surveys, each with 10 samples for 40 participants.  Figure \ref{fig:survey-images} shows sample screenshots of the conducted survey. \textbf{Binary Survey:} For each sample, participants viewed side-by-side annotations from both tools (order randomized) and chose the more accurate one. Figure~\ref{fig:binary-pii} shows GPT-4 was consistently preferred over Presidio for identifying privacy-sensitive information. \textbf{Multiple-Choice Survey:} Each of the 10 instances included 5 samples from each tool, displayed randomly. Participants had four options: (1) \emph{Accurate} (all sensitive info correctly annotated), (2) \emph{Under-Annotated} (some sensitive info missed), (3) \emph{Over-Annotated} (non-sensitive info included), and (4) \emph{Mixed-Annotated} (both missed and non-sensitive info annotated).  Each was given to 40 Prolific workers. Figure~\ref{fig:mult-pii} shows GPT-4 rated as ``Accurate'' across datasets.

\section{Exploring the Tradeoffs between Privacy and Utility}
\label{appendix:exploring-tradeoffs}

\begin{figure}[H]
    \centering
    \begin{subfigure}{0.49\linewidth}
    \centering
    \includegraphics[scale=0.22]{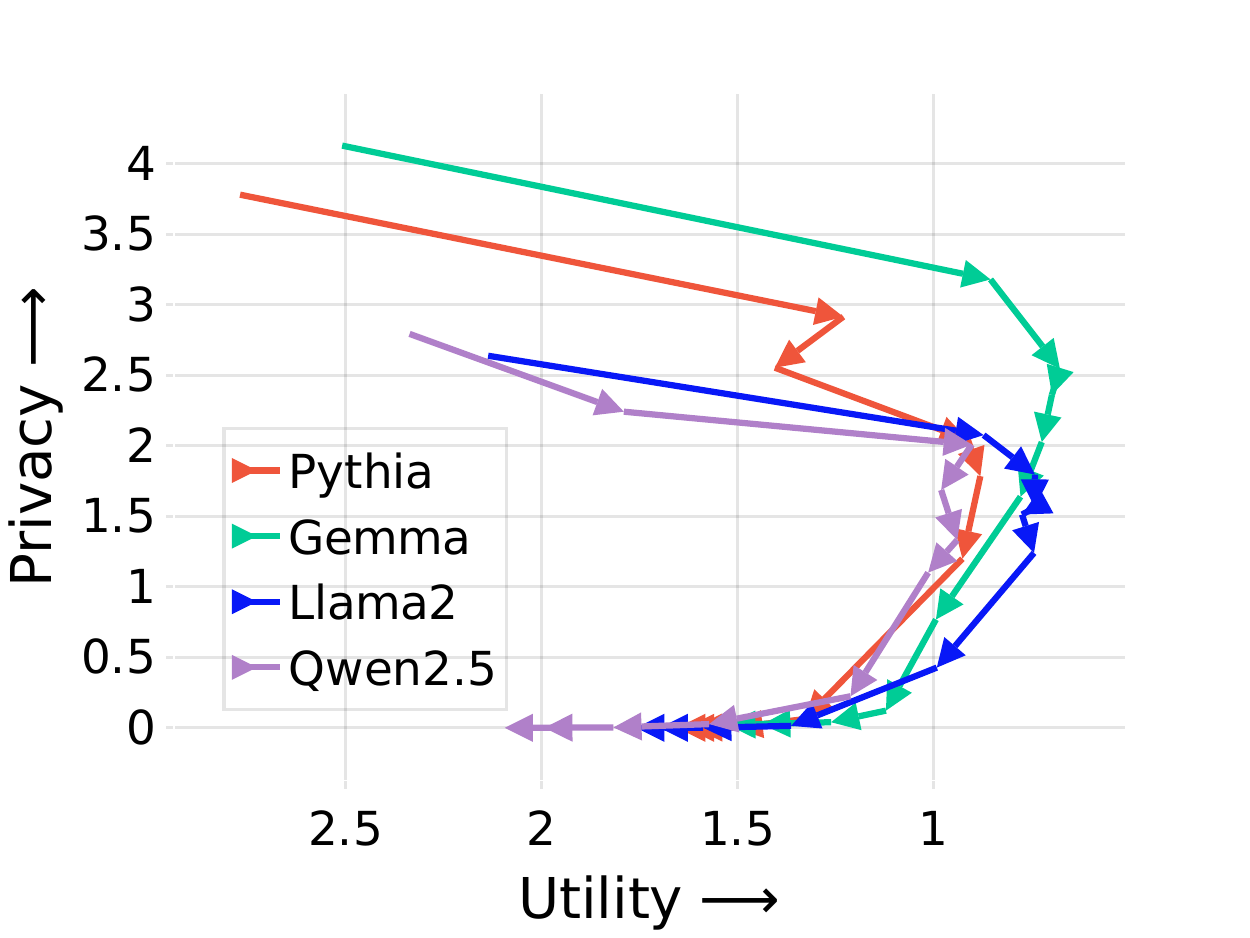}
    \caption{\textbf{CustomerSim}}
    \label{fig:appendix-ffta}
    \end{subfigure}
    \begin{subfigure}{0.49\linewidth}
    \centering
    \includegraphics[scale=0.22]{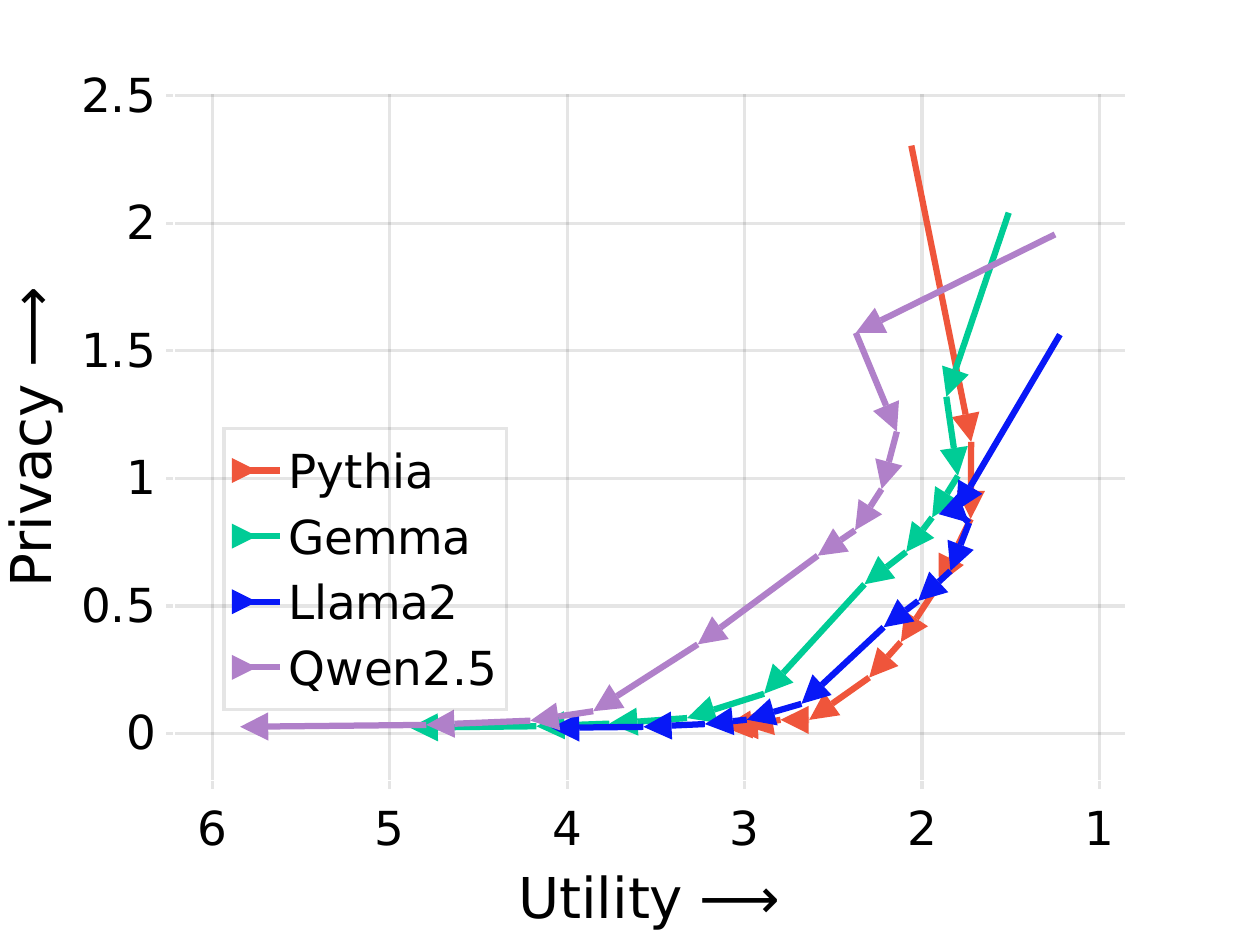}
    \caption{\textbf{SynBio}}
    \label{fig:appendifftb}    
    \end{subfigure}
    \caption{\textbf{Full fine-tuning} on (a) CustomerSim and  (b) SynBio using Presidio to annotate the privacy-sensitive information.}
    \label{fig:appendix-fft}
\end{figure}

\textbf{Full Fine-tuning}: We see a similar trade-off between utility and privacy when using Presidio annotations in Figure ~\ref{fig:appendix-fft}, as observed in Figure~\ref{fig:fft}. For the CustomerSim dataset, we notice that privacy decreases at the beginning of training, while utility improves. However, as training progresses, both metrics start to worsen. In contrast, the SynBio dataset shows a decline in both utility and privacy, except for the Pythia model, which initially experiences an improvement in utility during the first few epochs.

\begin{figure}[!h]
    \centering
    \begin{subfigure}{.48\linewidth}
        \includegraphics[scale=0.2,height=3cm]{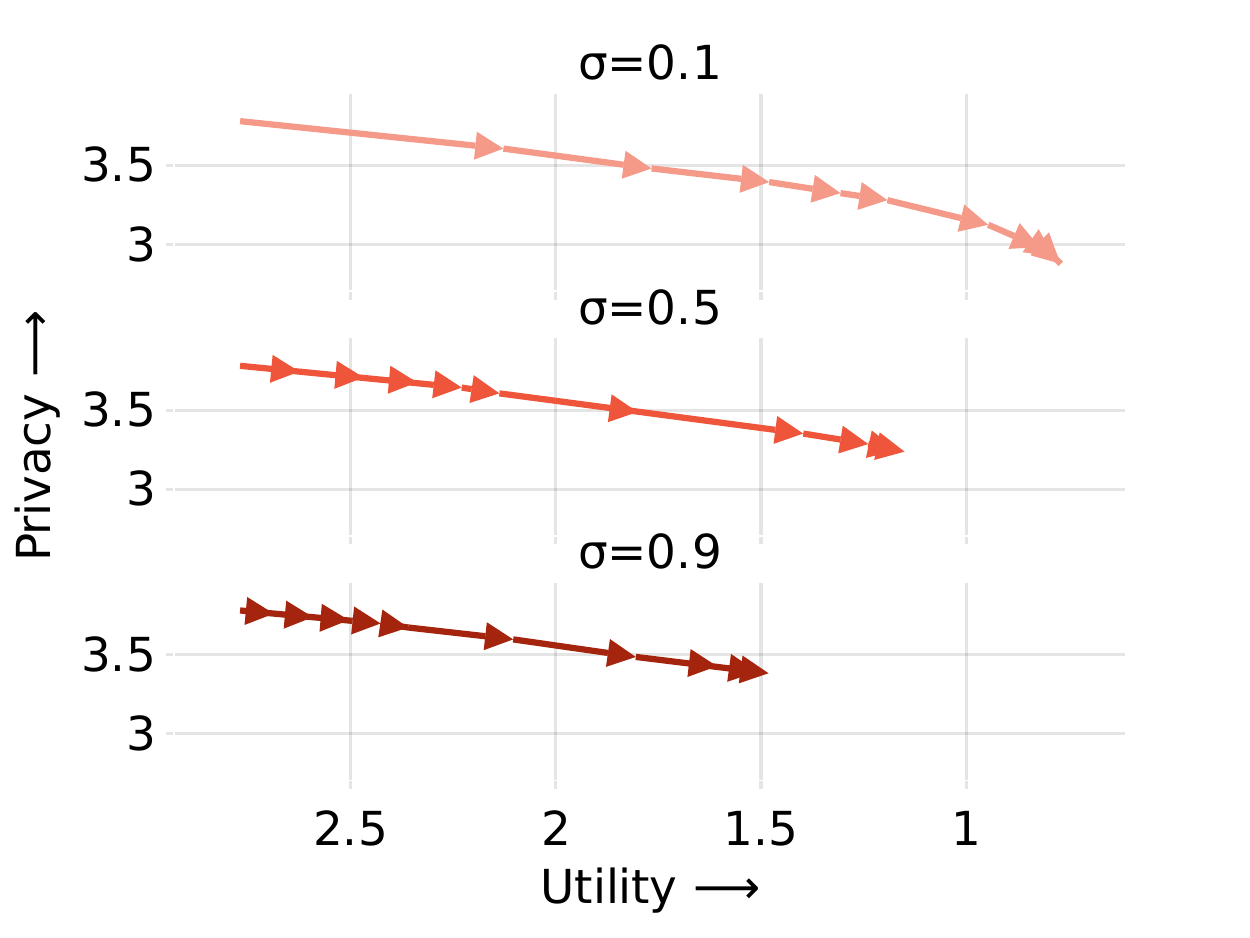}
        \caption{\textbf{Pythia}}
        \label{fig:appendix-dp_csima}
    \end{subfigure}
    \begin{subfigure}{.48\linewidth}
        \includegraphics[scale=0.2,height=3cm]{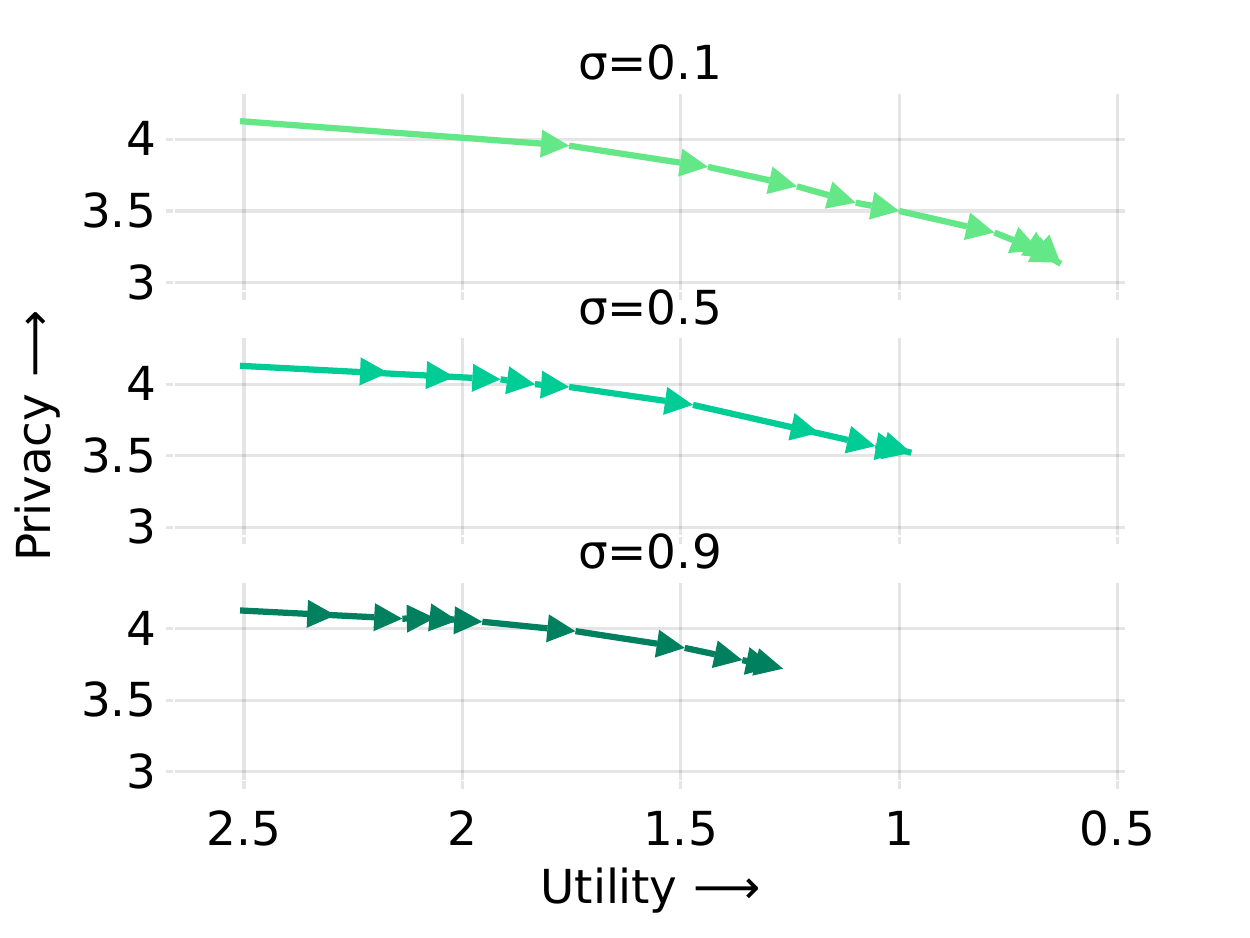}
        \caption{\textbf{Gemma}}
        \label{fig:appendix-dp_csimb}
    \end{subfigure}
    \begin{subfigure}{.48\linewidth}
        \includegraphics[scale=0.2,height=3cm]{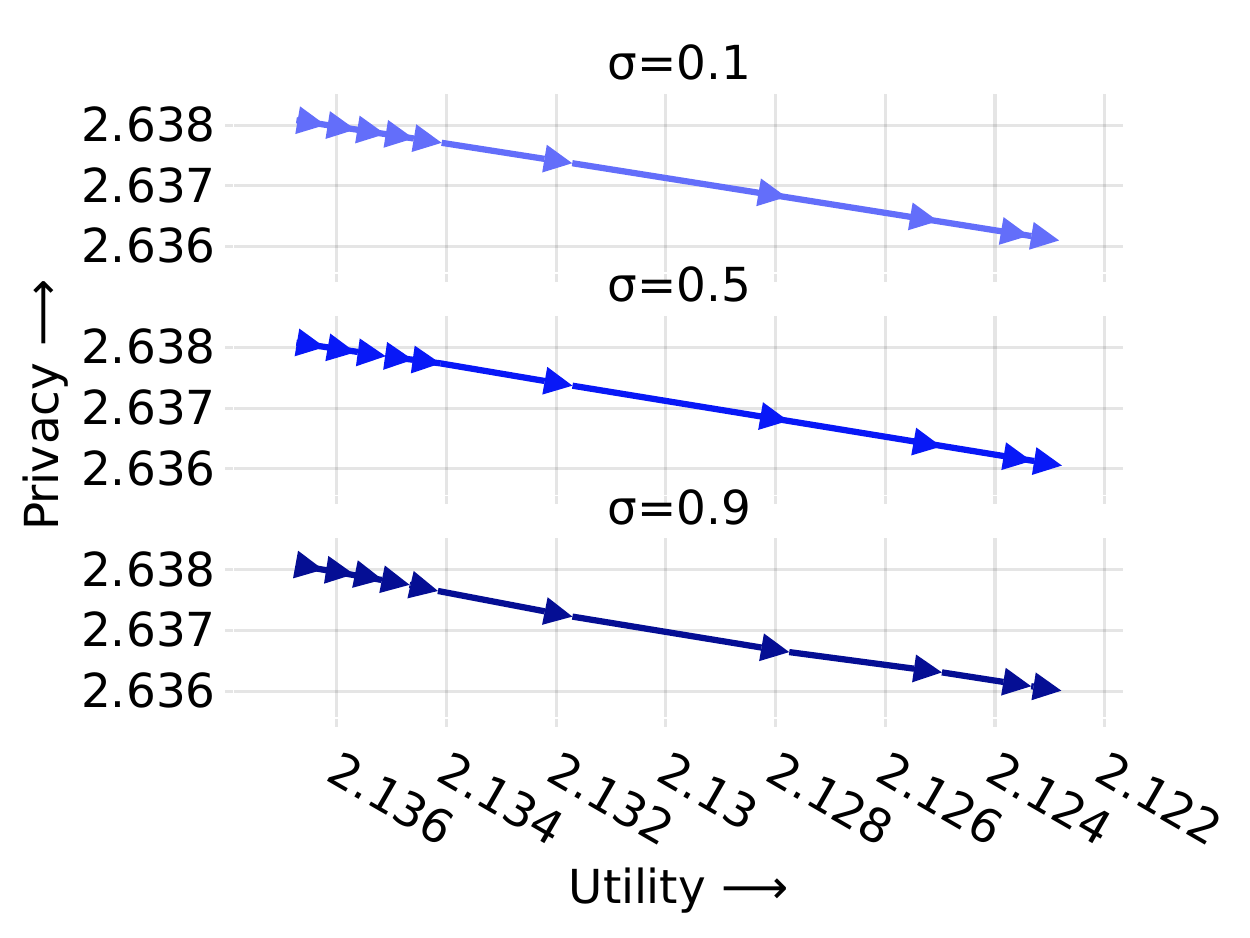}
        \caption{\textbf{Llama2}}
        \label{fig:appendix-dp_csimc}
    \end{subfigure}
    \caption{\textbf{DP} on \emph{CustomerSim} dataset using Presidio tool for annotating privacy-sensitive information.}
    \label{fig:appendix-dp-csim}
\end{figure}

\begin{figure}[!h]
    \centering
    \begin{subfigure}{0.48\linewidth}
    \centering
    \includegraphics[scale=0.2,height=3cm]{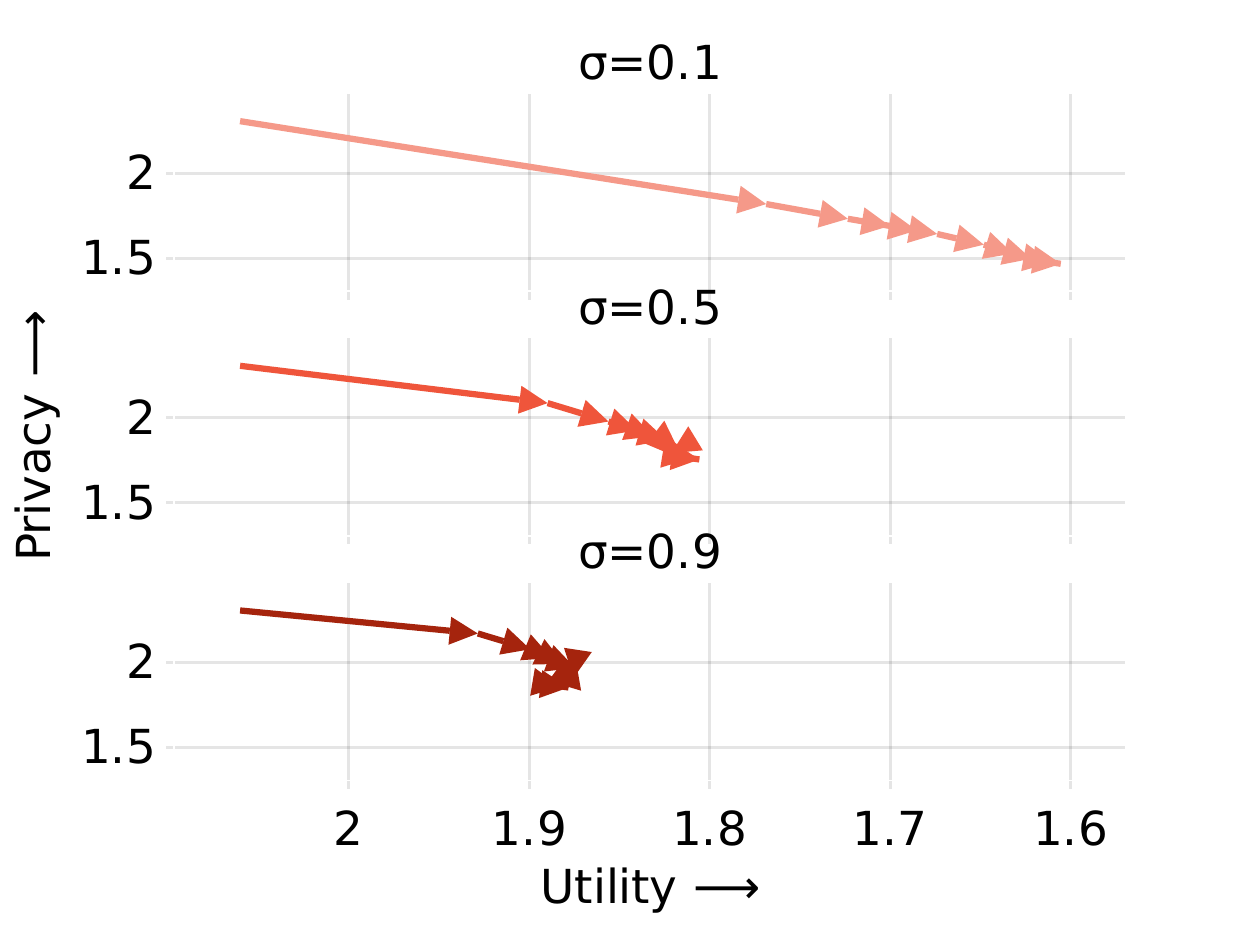}
    \caption{\textbf{Pythia}}
    \label{fig:appendix-dp_piia}
    \end{subfigure}
    \begin{subfigure}{0.48\linewidth}
    \centering
    \includegraphics[scale=0.2,height=3cm]{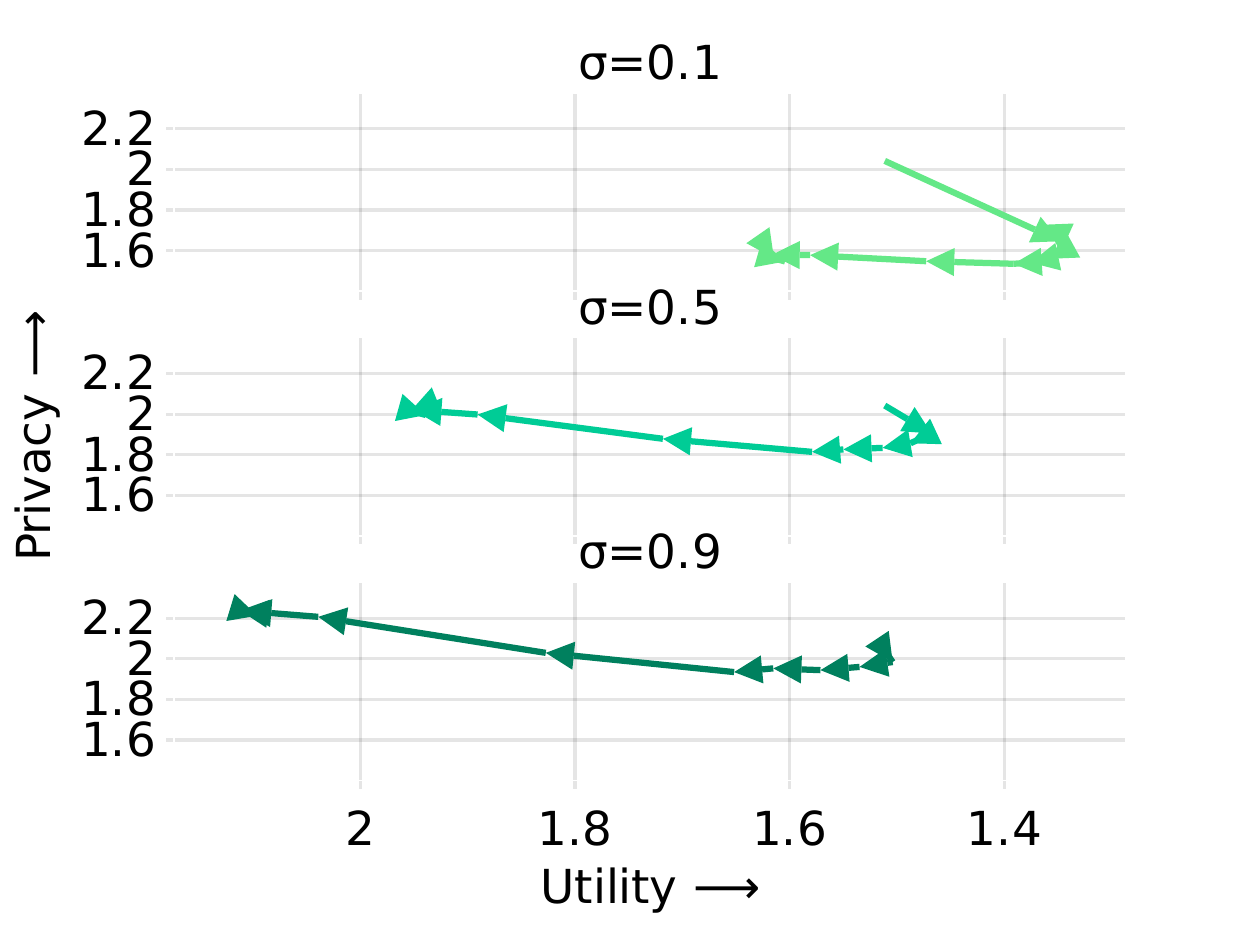}
    \caption{\textbf{Gemma}}
    \label{fig:appendix-dp_piib}
    \end{subfigure}
    \begin{subfigure}{0.48\linewidth}
    \centering
    \includegraphics[scale=0.2,height=3cm]{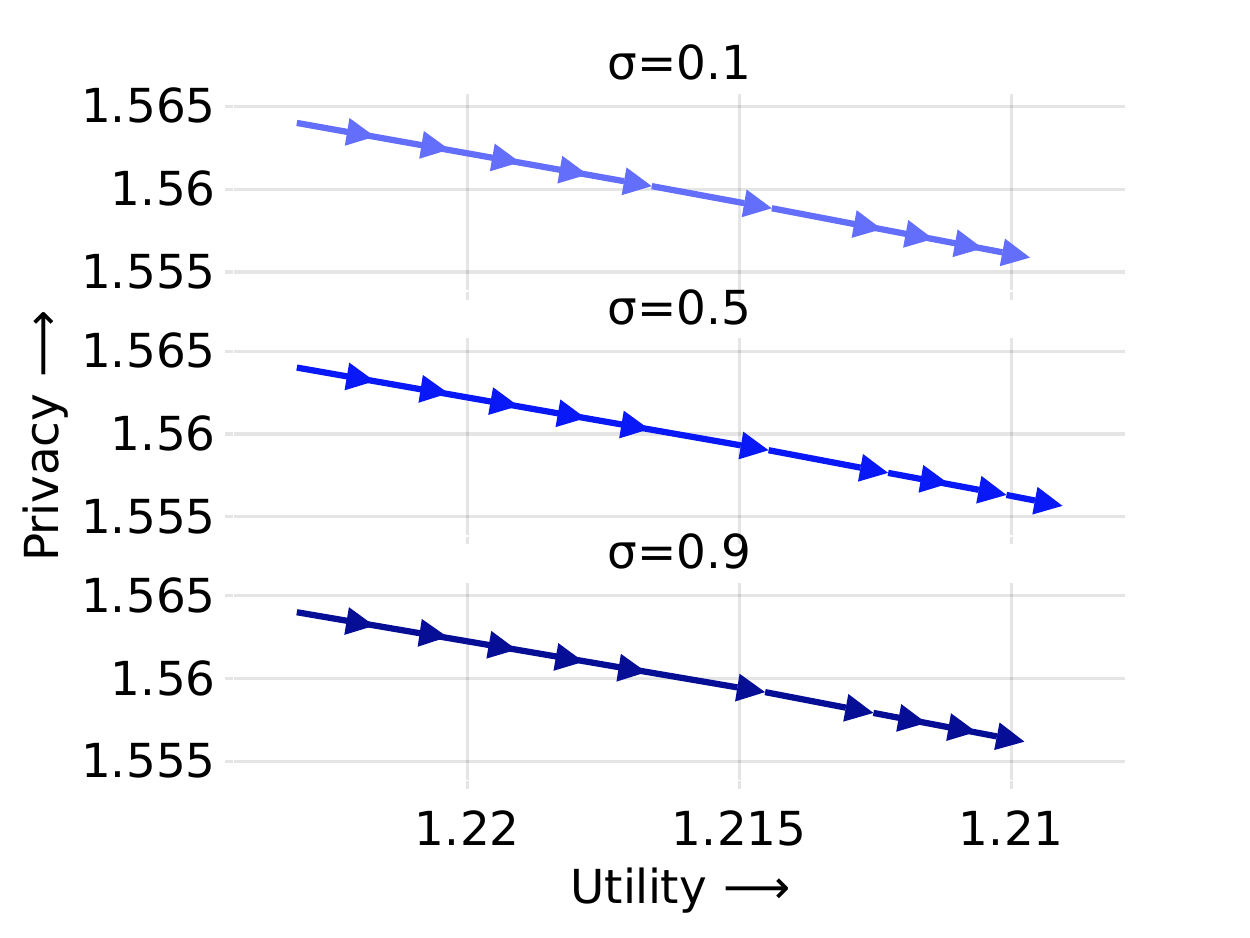}
    \caption{\textbf{Llama2}}
    \label{fig:appendix-dp_piic}
    \end{subfigure}
    \caption{\textbf{DP} on \emph{SynBio} dataset using Presidio tool for annotating privacy-sensitive information.}
    \label{fig:appendix-dp-pii}
\end{figure}

\textbf{DP-SGD} : A similar pattern as in Figure~\ref{fig:combined_dp} using GPT-4 annotations, has also emerged in Figure \ref{fig:appendix-dp-csim} using Presidio annotations. We observe that in the CustomerSim dataset, DP maintains privacy effectively with minimal degradation. However, when it comes to utility, we notice that lower noise values lead to better utility, but also result in a decrease in privacy. 
The same holds true for the SynBio dataset in Figure \ref{fig:appendix-dp-pii}, with the only exception being Gemma experiencing a decline in utility after a few iterations.

\begin{figure}[h!]
    \centering
    \begin{subfigure}{0.48\linewidth}
    \centering
    \includegraphics[scale=0.5,width=\linewidth]{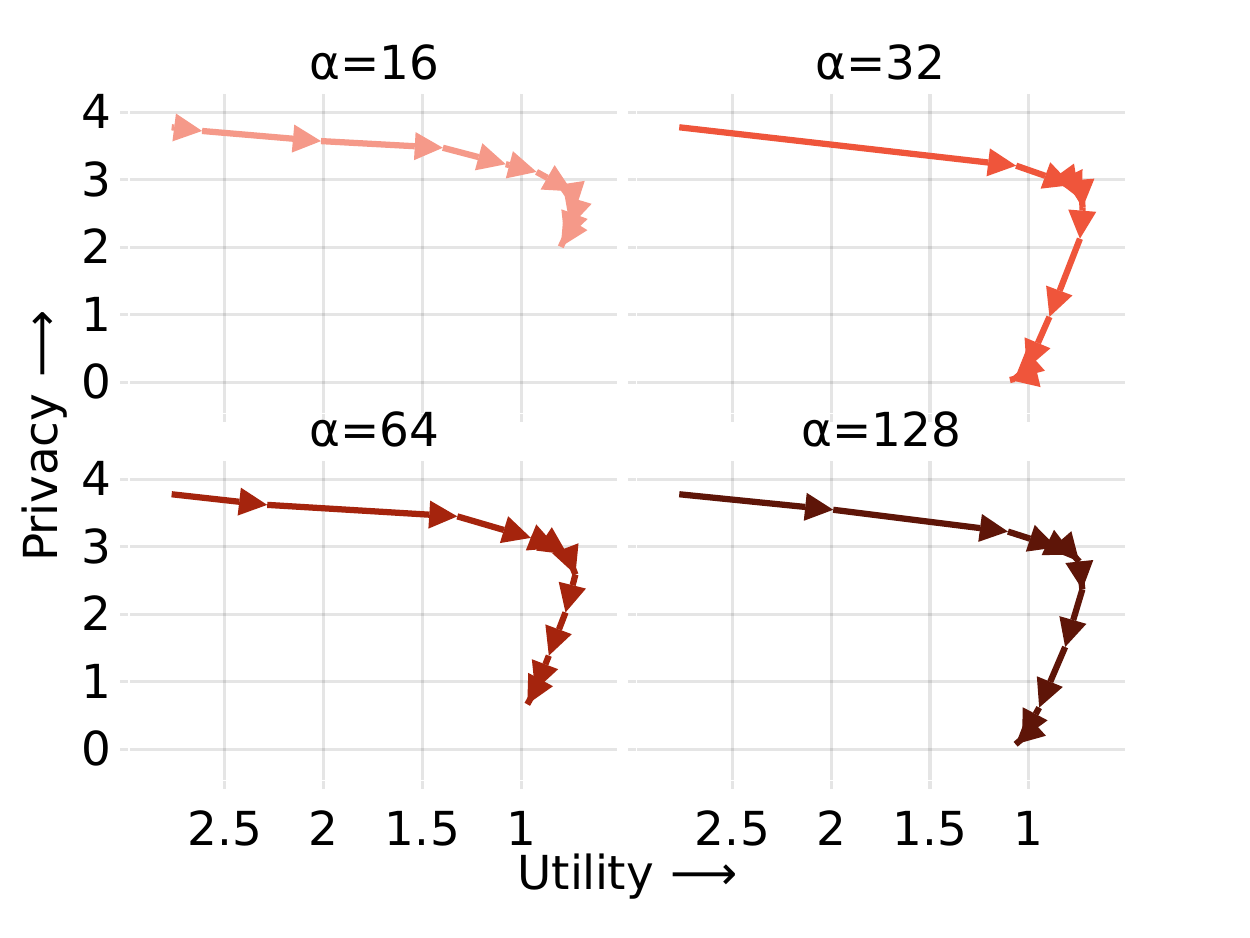}
    \caption{\textbf{Pythia}}
    \label{fig:appendix-lora16_csima}
    \end{subfigure}
    \begin{subfigure}{0.48\linewidth}
    \centering
    \includegraphics[scale=0.5,width=\linewidth]{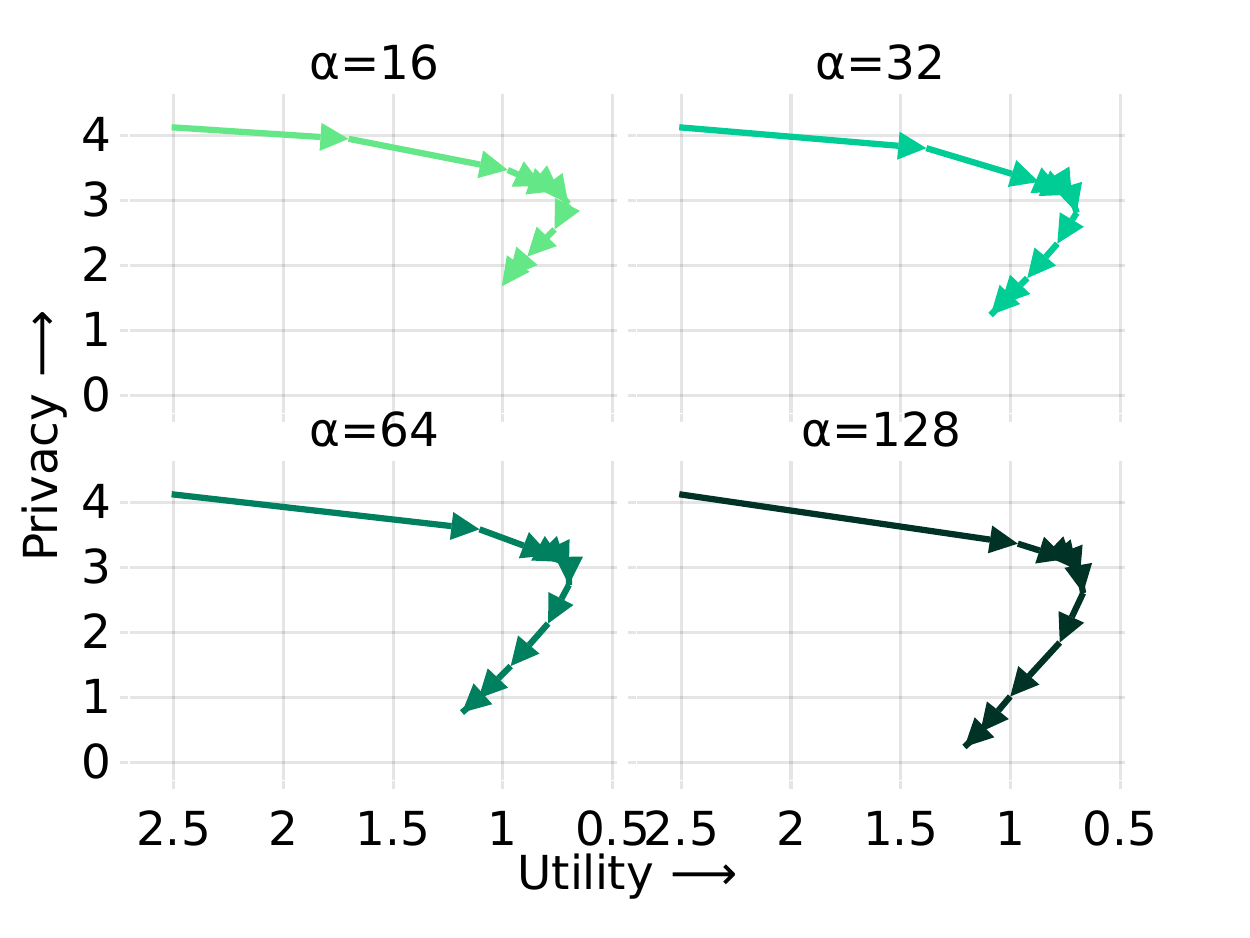}
    \caption{\textbf{Gemma}}
    \label{fig:appendix-lora16_csimb}
    \end{subfigure}
    \begin{subfigure}{0.48\linewidth}
    \centering
    \includegraphics[scale=0.5,width=\linewidth]{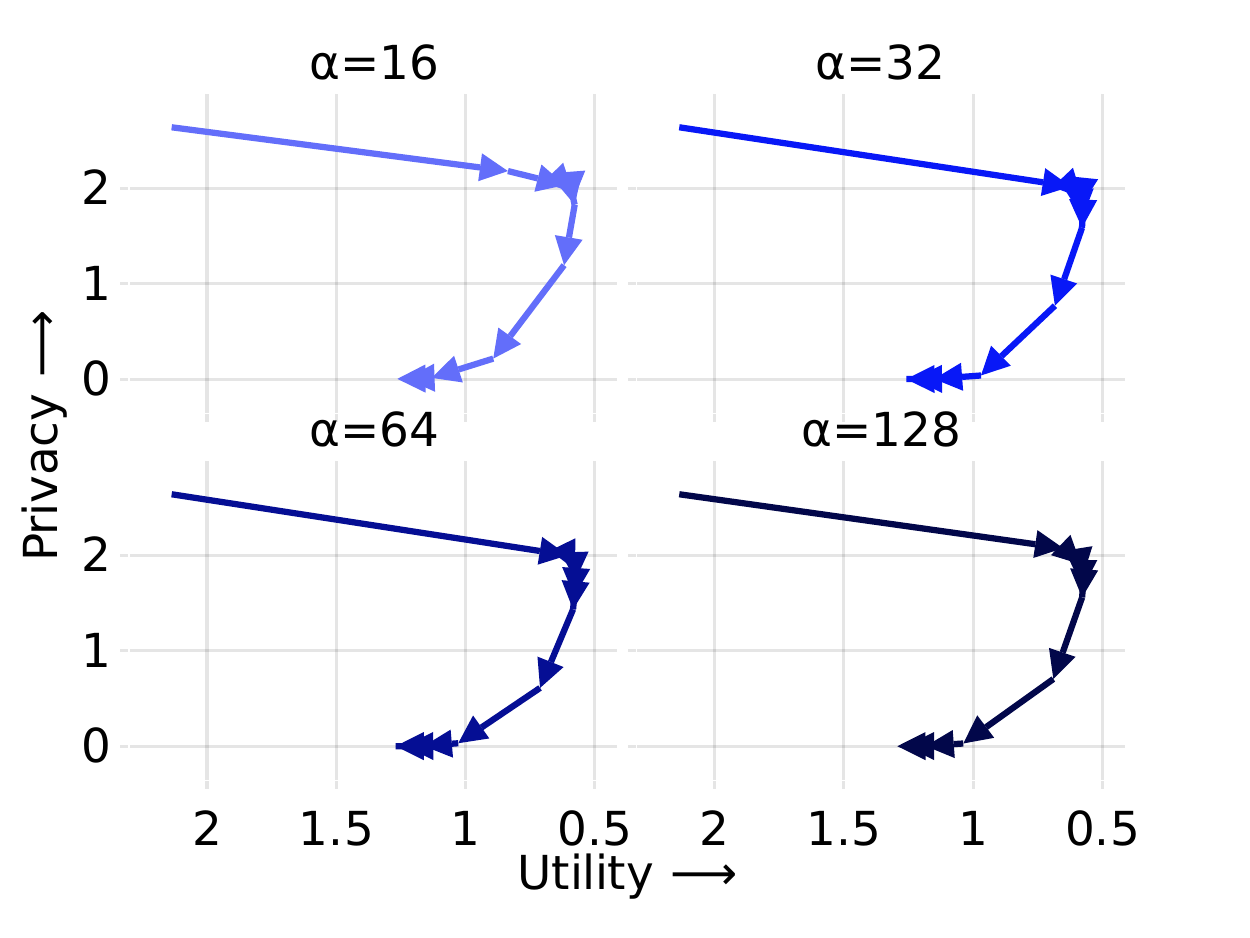}
    \caption{\textbf{Llama2}}
    \label{fig:appendix-lora16_csimc}
    \end{subfigure}
    \caption{\textbf{Lora with rank 16} on \emph{CustomerSim} dataset using Presidio tool for annotating privacy-sensitive information.}
    \label{fig:appendix-lora16-csim}
\end{figure}

\begin{figure}[h!]
    \centering
    \begin{subfigure}{0.48\linewidth}
    \centering
    \includegraphics[scale=0.5,width=\linewidth]{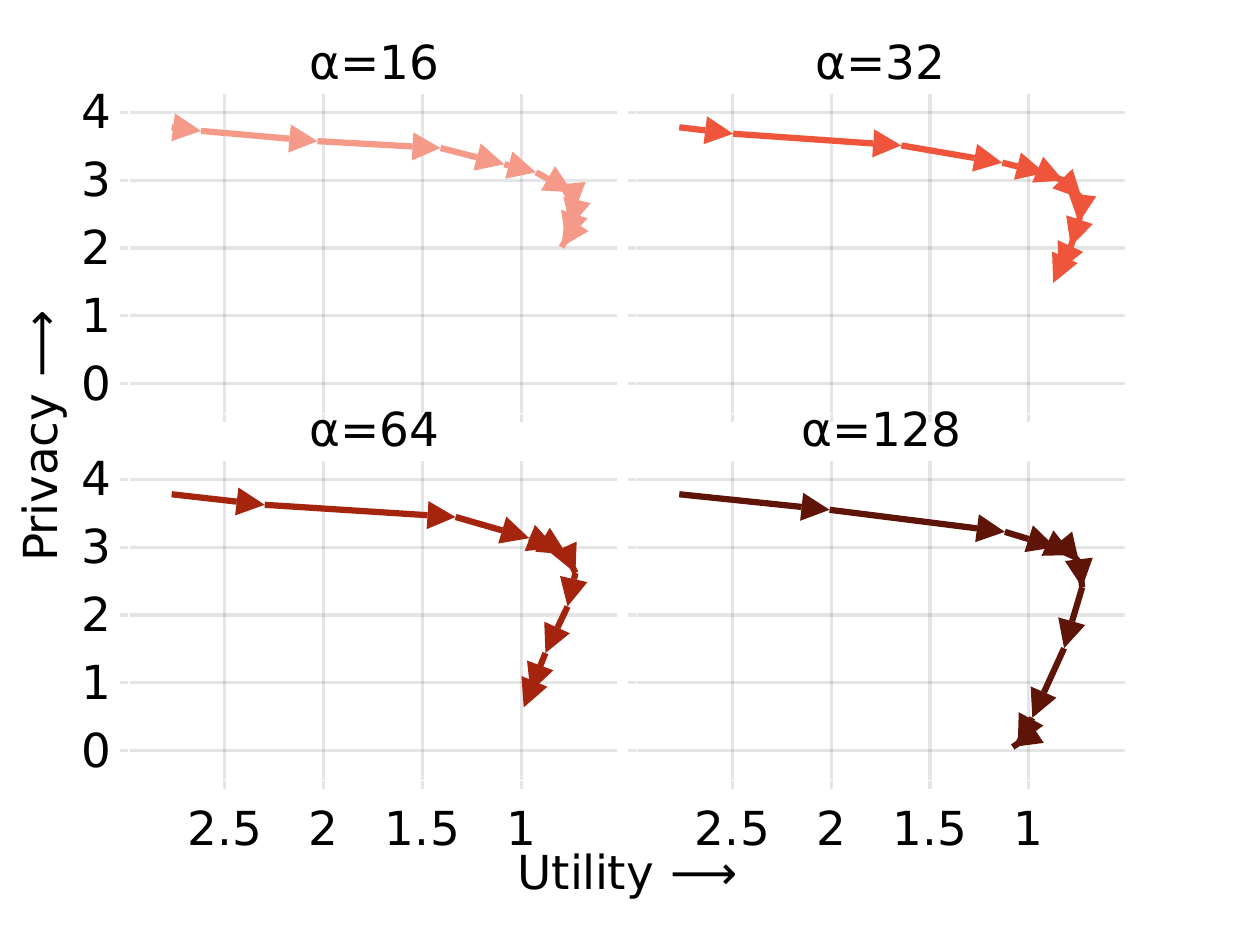}
    \caption{\textbf{Pythia}}
    \label{fig:appendix-lora32_csima}
    \end{subfigure}
    \begin{subfigure}{0.48\linewidth}
    \centering
    \includegraphics[scale=0.5,width=\linewidth]{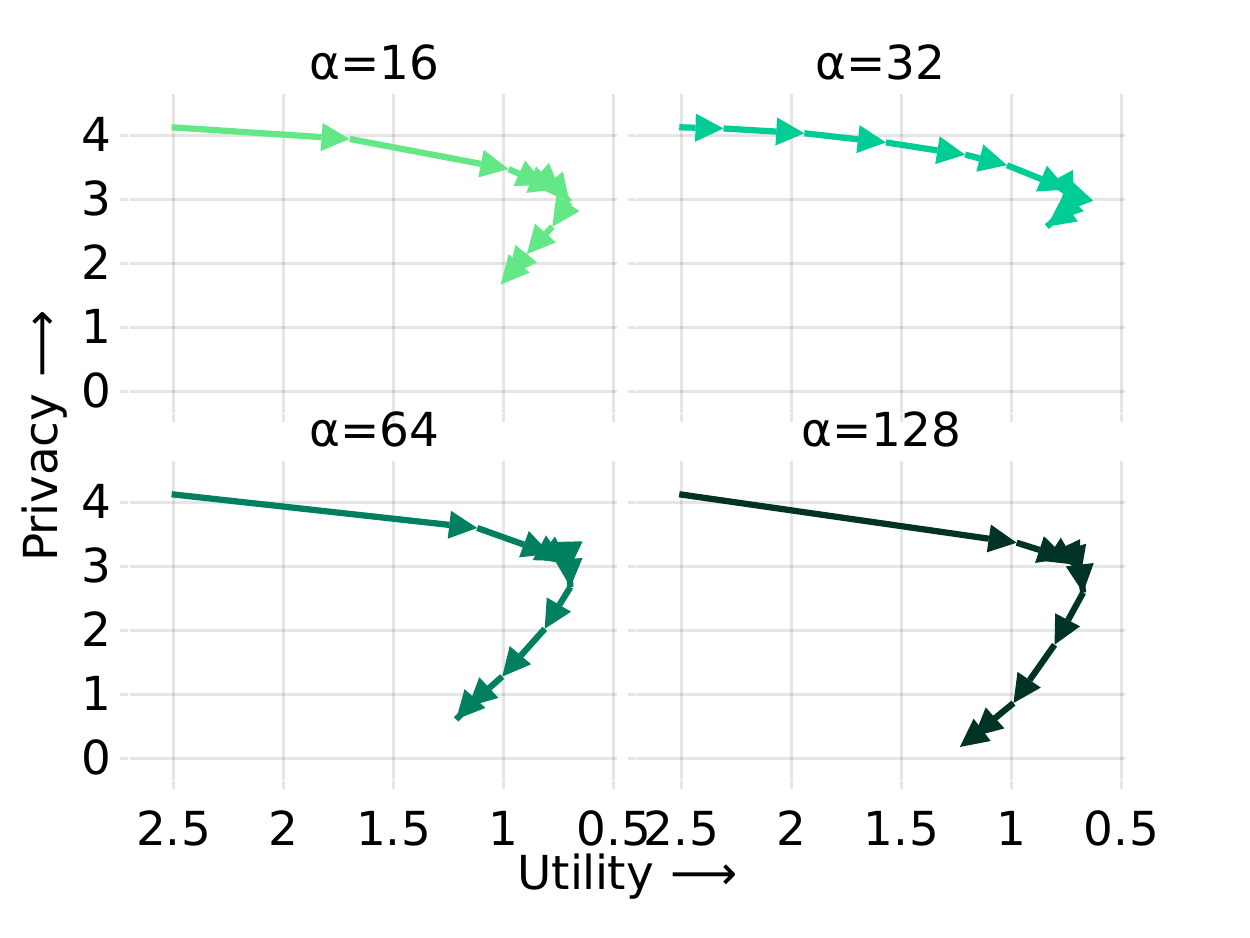}
    \caption{\textbf{Gemma}}
    \label{fig:appendix-lora32_csimb}
    \end{subfigure}
    \begin{subfigure}{0.48\linewidth}
    \centering
    \includegraphics[scale=0.5,width=\linewidth]{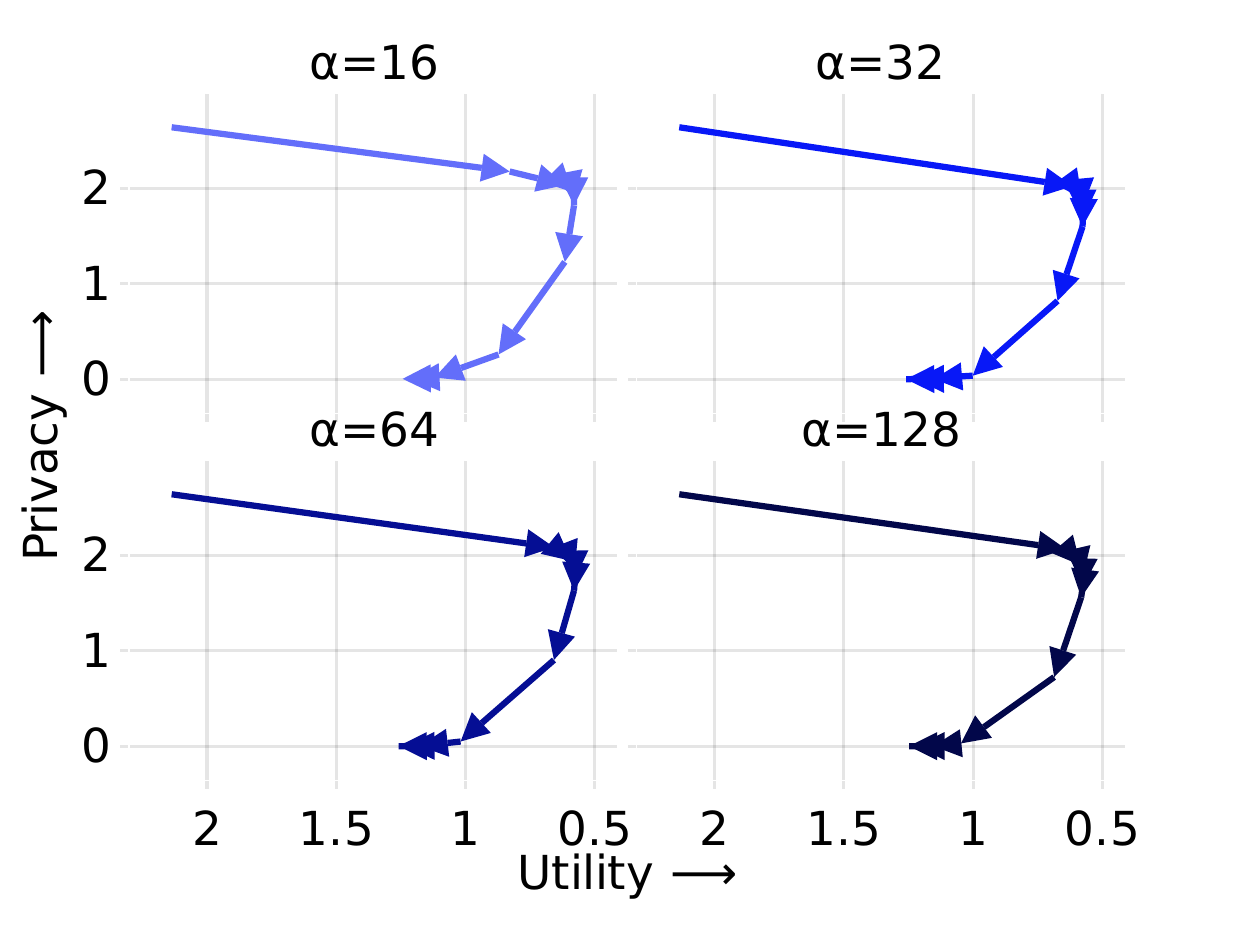}
    \caption{\textbf{Llama2}}
    \label{fig:appendix-lora32_csimc}
    \end{subfigure}
    \caption{\textbf{Lora with rank 32} on \emph{CustomerSim} dataset using Presidio tool for annotating privacy-sensitive information.}
    \label{fig:appendix-lora32-csim}
\end{figure}

\begin{figure}[h!]
    \centering
    \begin{subfigure}{0.48\linewidth}
    \centering
    \includegraphics[scale=0.5,width=\linewidth]{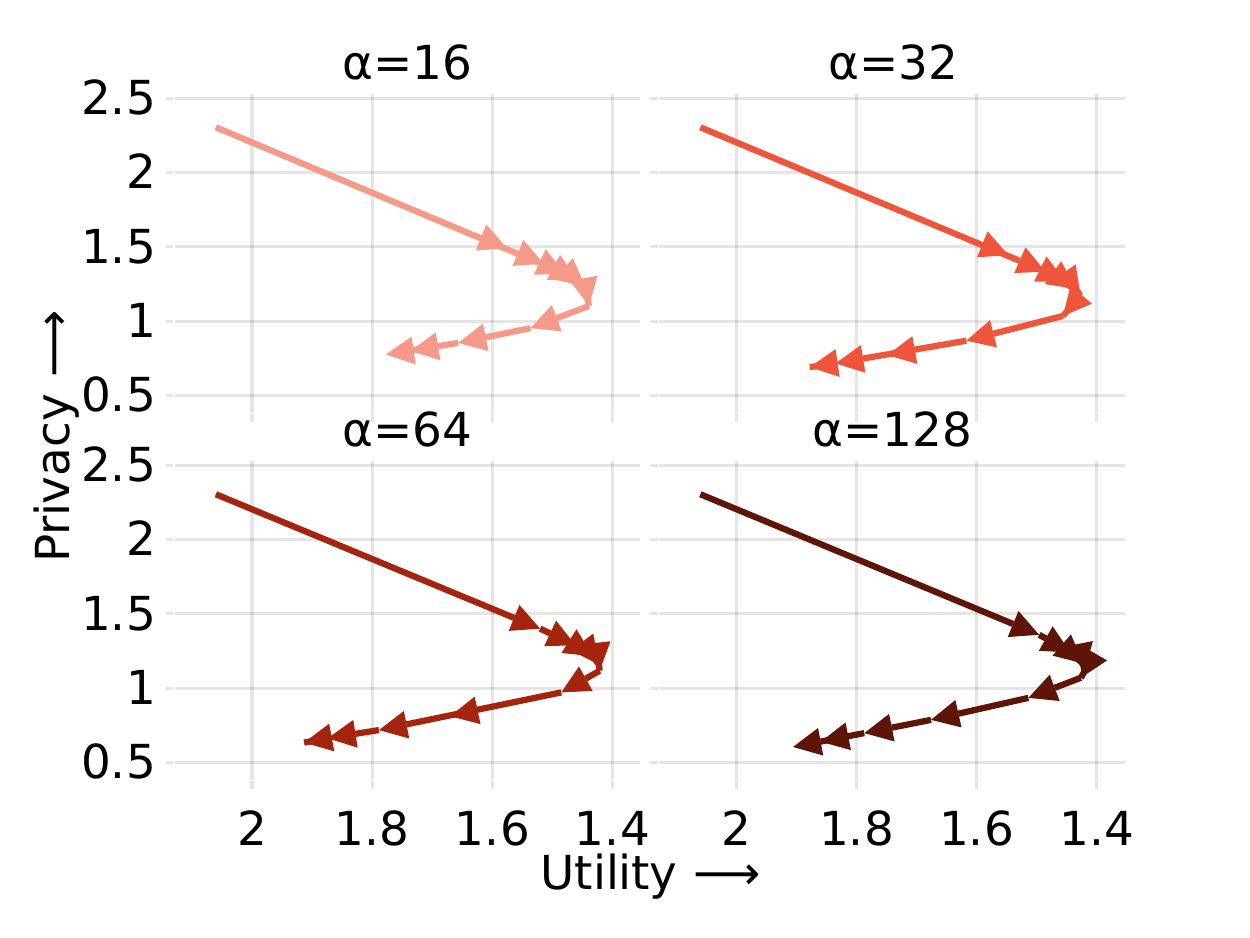}
    \caption{\textbf{Pythia}}
    \label{fig:appendix-lora16_piia}
    \end{subfigure}
    \begin{subfigure}{0.48\linewidth}
    \centering
    \includegraphics[scale=0.5,width=\linewidth]{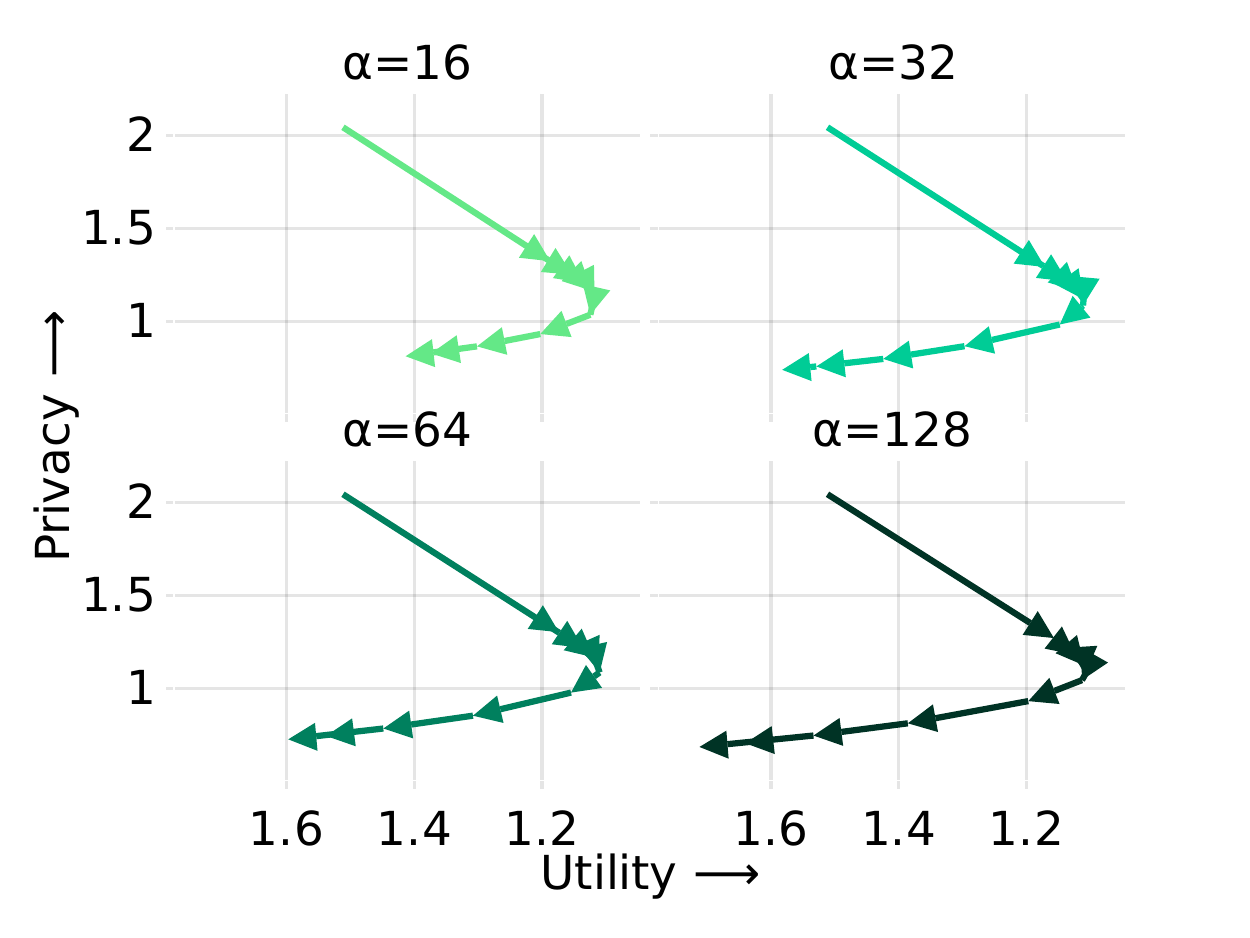}
    \caption{\textbf{Gemma}}
    \label{fig:appendix-lora16_piib}
    \end{subfigure}
    \begin{subfigure}{0.48\linewidth}
    \centering
    \includegraphics[scale=0.5,width=\linewidth]{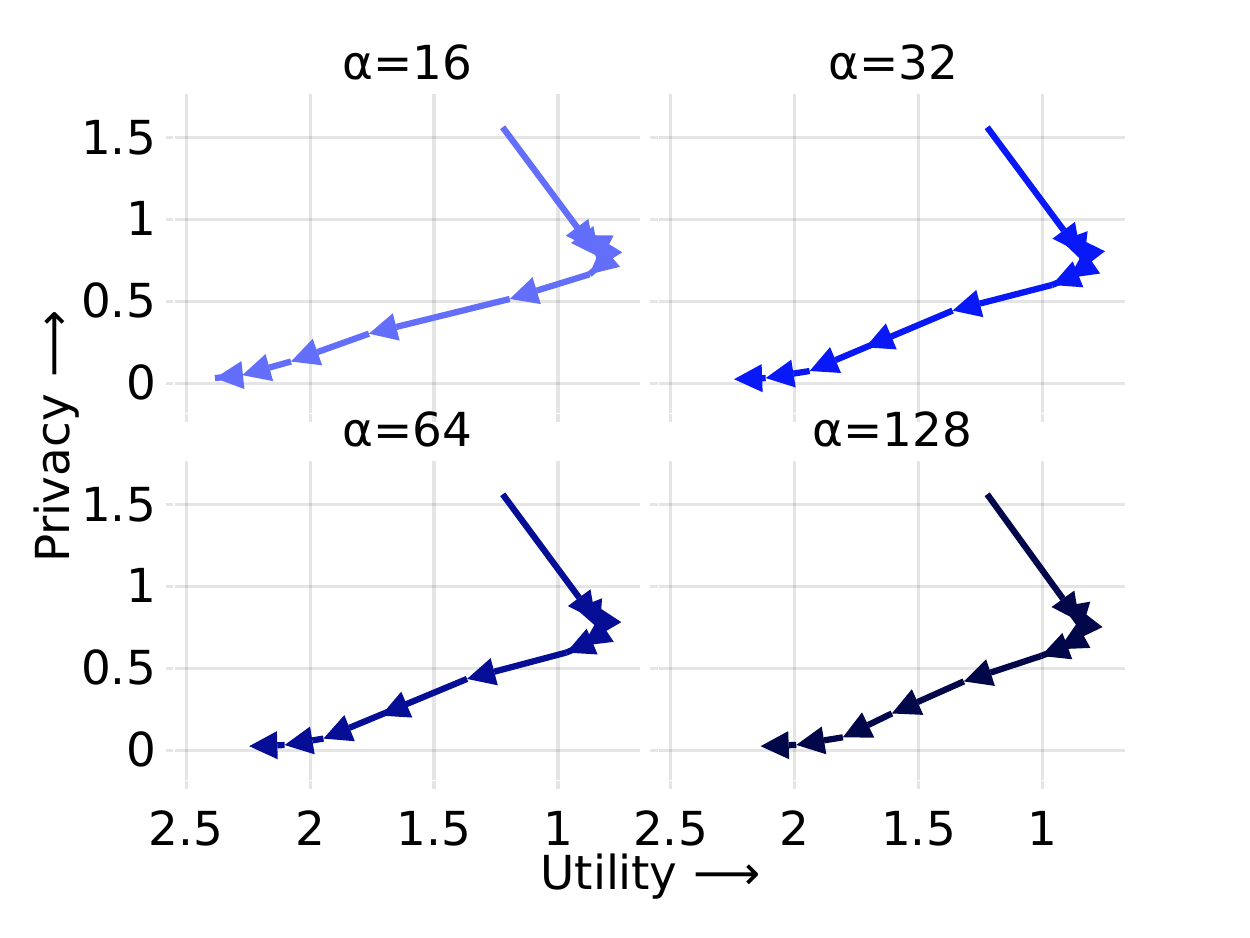}
    \caption{\textbf{Llama2}}
    \label{fig:appendix-lora16_piic}
    \end{subfigure}
    \caption{\textbf{Lora with rank 16} on \emph{SynBio} dataset using Presidio tool for annotating privacy-sensitive information.}
    \label{fig:appendix-lora16-pii}
\end{figure}

\begin{figure}[h!]
    \centering
    \begin{subfigure}{0.48\linewidth}
    \centering
    \includegraphics[scale=0.5,width=\linewidth]{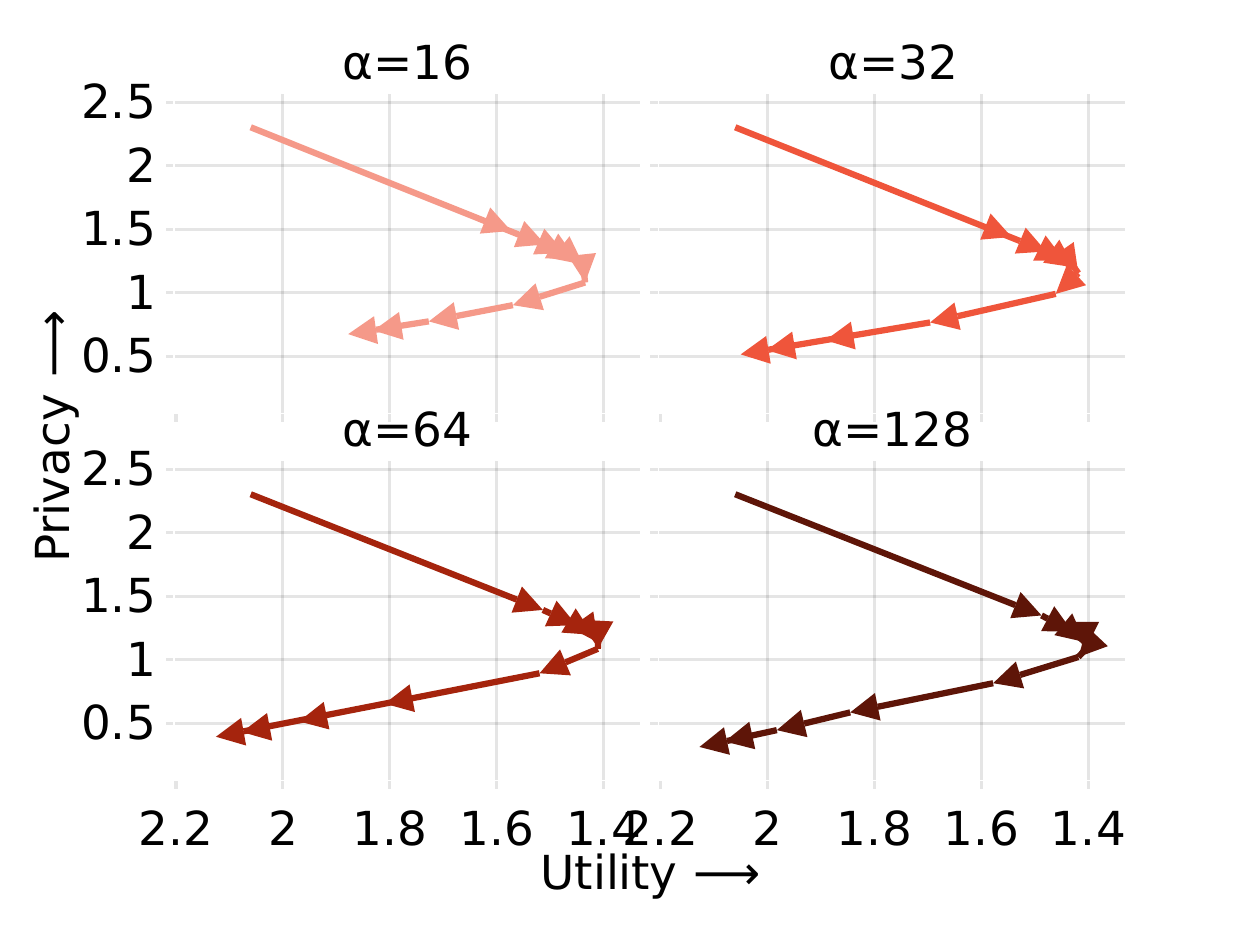}
    \caption{\textbf{Pythia}}
    \label{fig:appendix-lora32_piia}
    \end{subfigure}
    \begin{subfigure}{0.48\linewidth}
    \centering
    \includegraphics[scale=0.5,width=\linewidth]{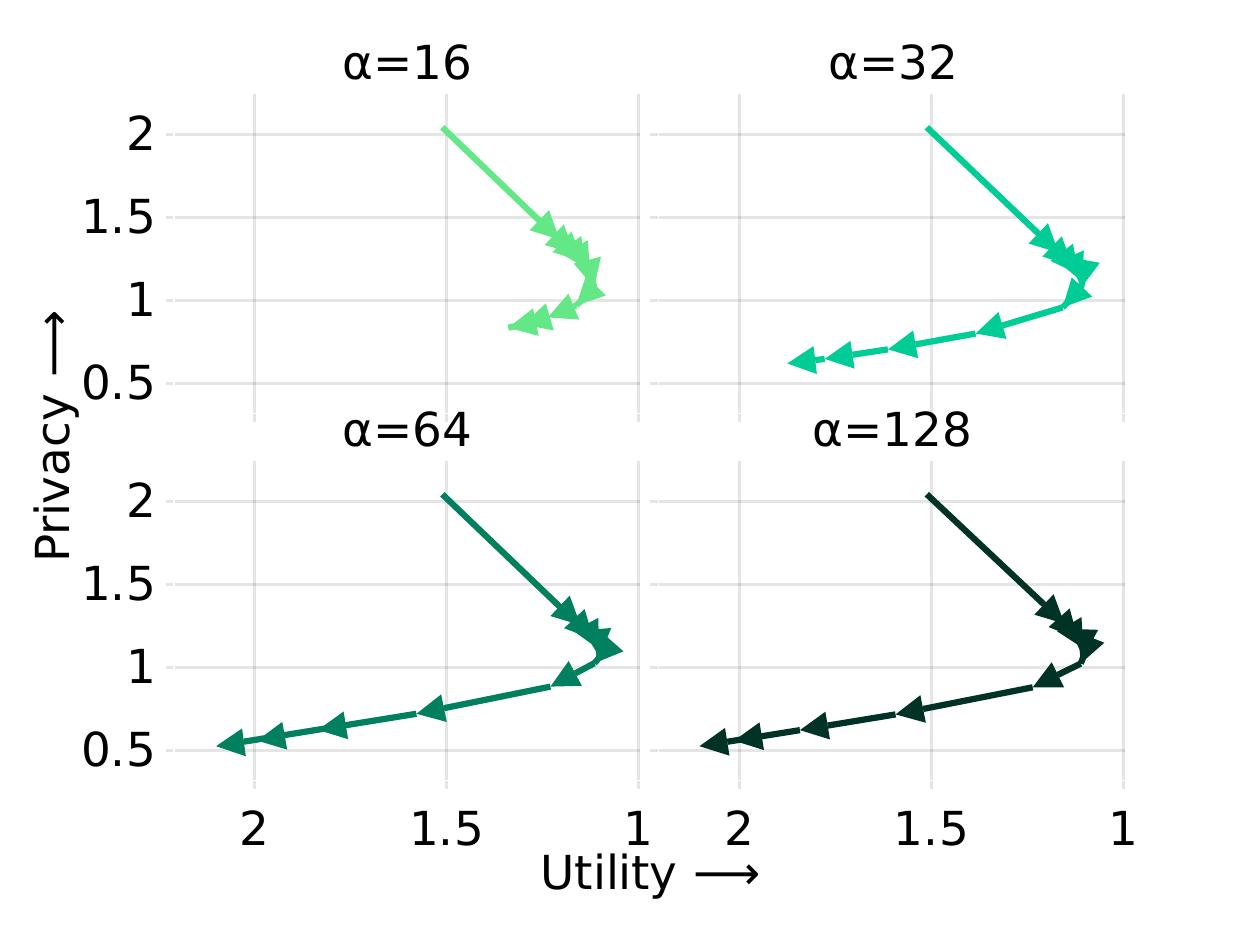}
    \caption{\textbf{Gemma}}
    \label{fig:appendix-lora32_piib}
    \end{subfigure}
    \begin{subfigure}{0.48\linewidth}
    \centering
    \includegraphics[scale=0.5,width=\linewidth]{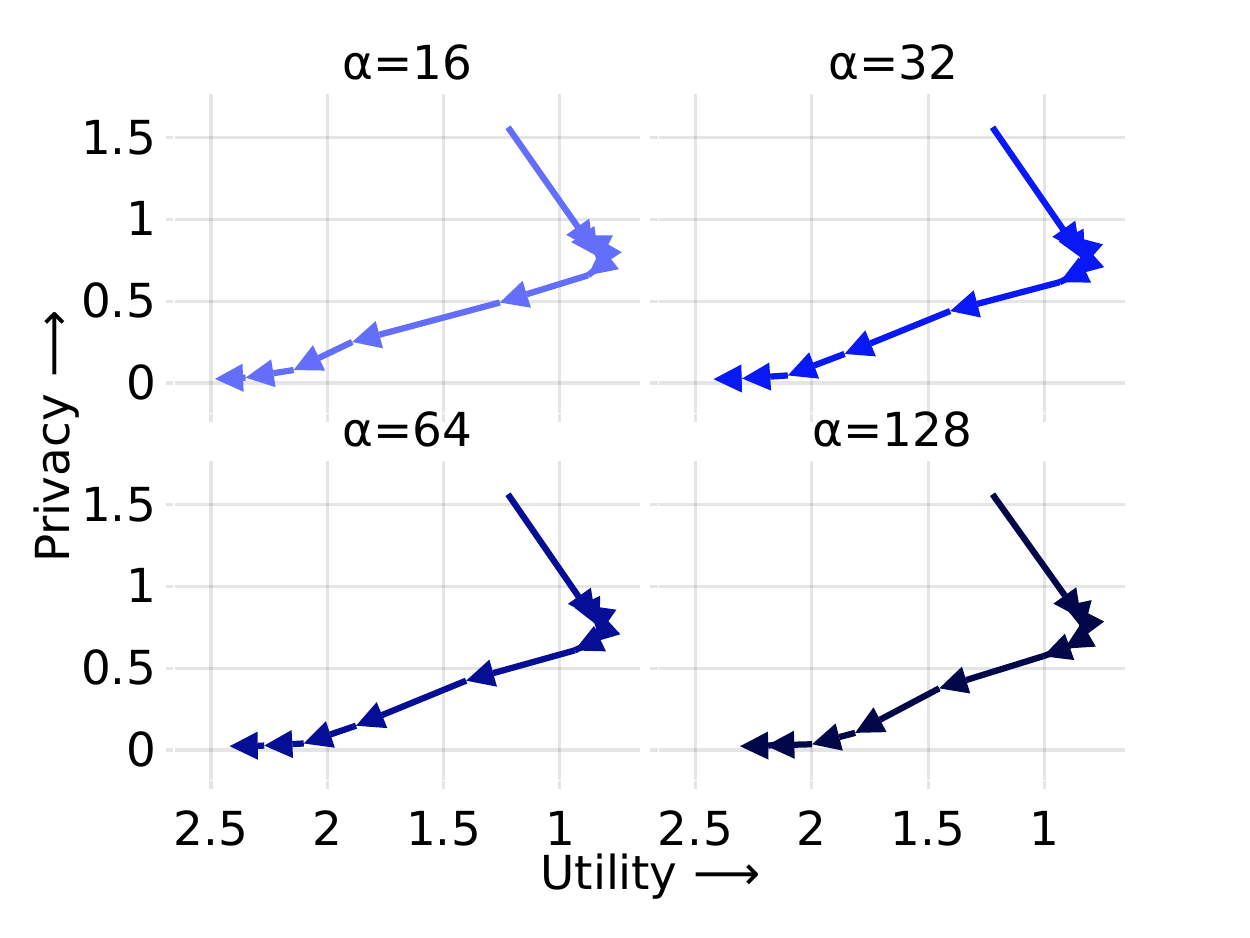}
    \caption{\textbf{Llama2}}
    \label{fig:appendix-lora32_piic}
    \end{subfigure}
    \caption{\textbf{Lora with rank 32} on \emph{SynBio} dataset using Presidio tool for annotating privacy-sensitive information.}
    \label{fig:appendix-lora32-pii}
\end{figure}

\textbf{LoRA }: Comparing the results from Figures \ref{fig:appendix-lora16-csim} and \ref{fig:appendix-lora32-csim} using Presidio annotations with Figure~\ref{fig:combined_lora}, which used GPT-4 annotations, we can observe the same trend. This suggests that the findings are consistent across different annotation methods. Similarly, when we analyze the trade-off for the SynBio dataset in Figures ~\ref{fig:appendix-lora16-pii} and \ref{fig:appendix-lora32-pii}, we see similar observations as we observed in Figure \ref{fig:combined_lora_synbio} using GPT-4 annotations.

\begin{figure}[h!]
    \centering
    \begin{subfigure}{0.48\linewidth}
    \centering
    \includegraphics[scale=0.5,width=\linewidth]{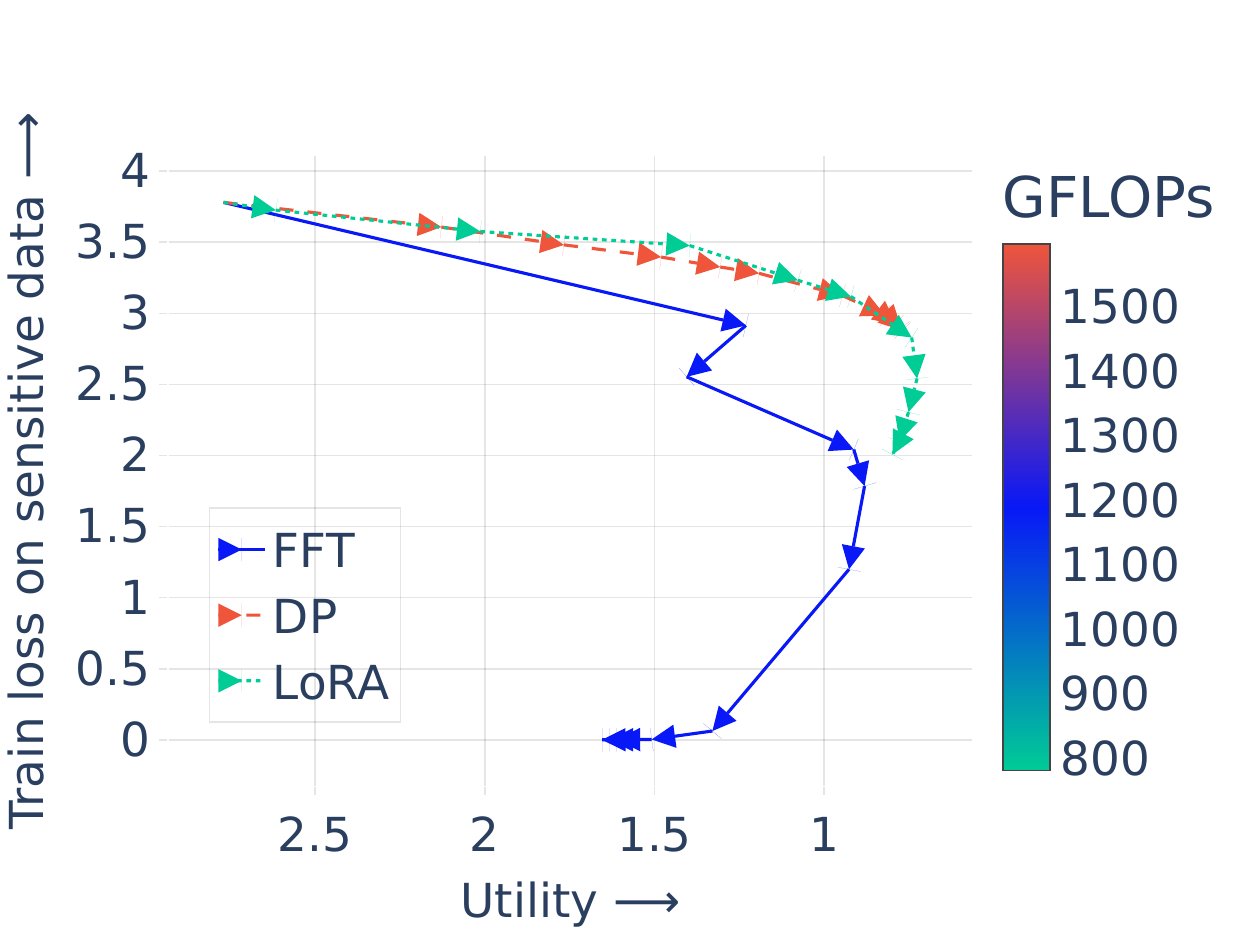}
    \caption{\textbf{Pythia}}
    \label{fig:appendix-fdl_csim1}
    \end{subfigure}
    \begin{subfigure}{0.48\linewidth}
    \centering
    \includegraphics[scale=0.5,width=\linewidth]{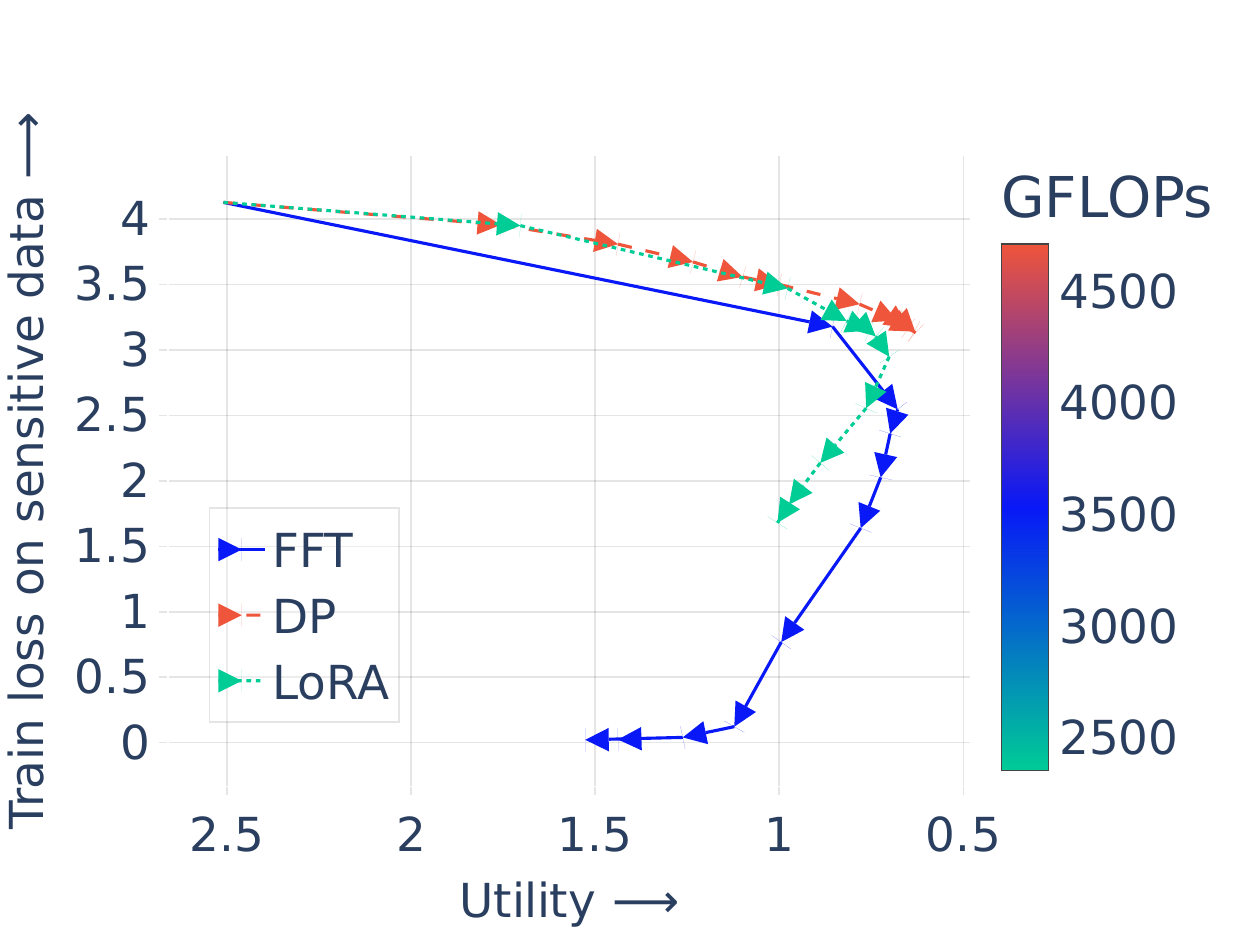}
    \caption{\textbf{Gemma}}
    \label{fig:appendix-fdl_csim2}
    \end{subfigure}
    \begin{subfigure}{0.48\linewidth}
    \centering
    \includegraphics[scale=0.5,width=\linewidth]{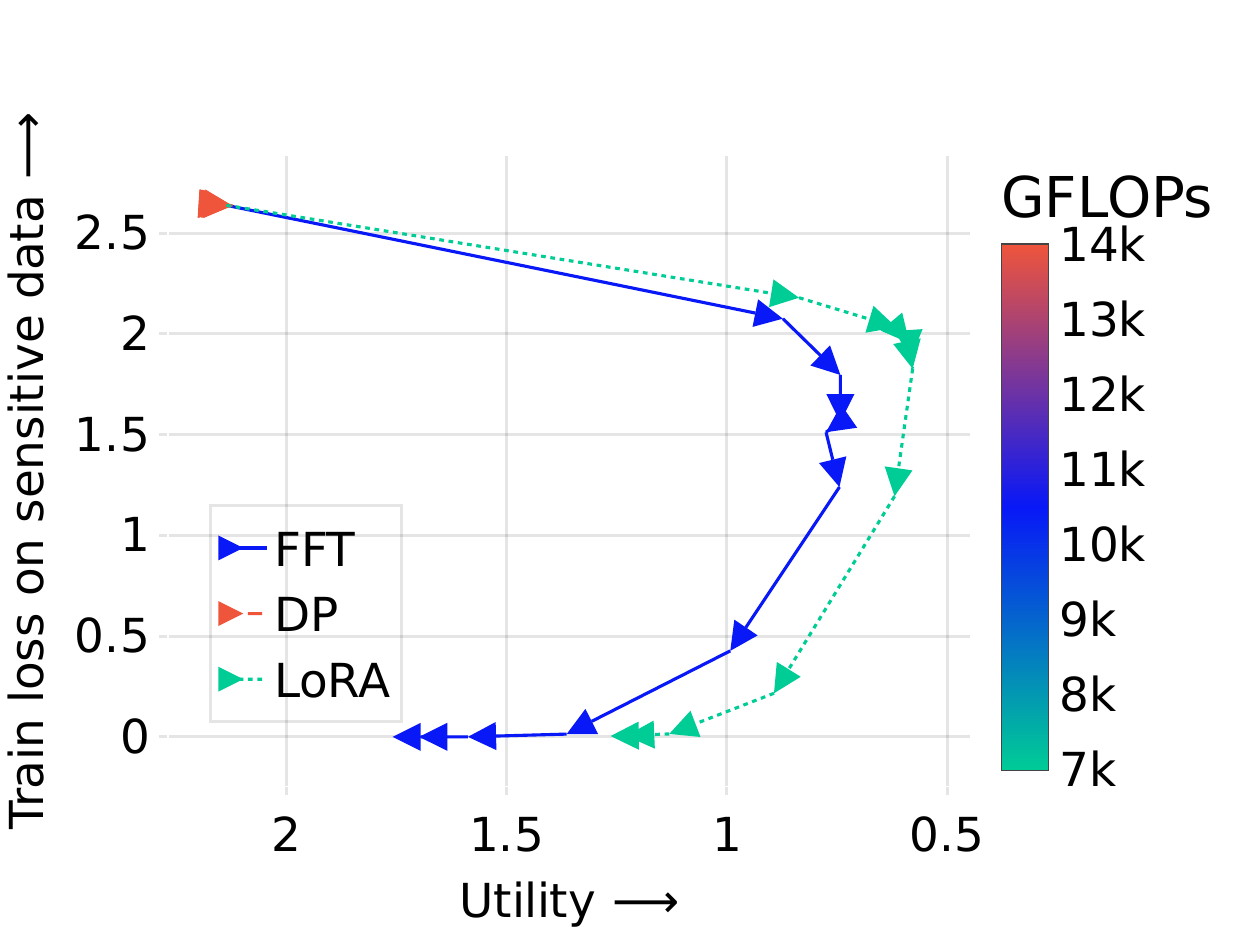}
    \caption{\textbf{Llama2}}
    \label{fig:appendix-fdl_csim3}
    \end{subfigure}
    \caption{\textbf{Full fine-tuning, DP and LoRA} on \emph{CustomerSim} with Presidio annotations for privacy-sensitive information.}
    \label{fig:appendix-fdl_csim}
\end{figure}
\begin{figure}[h!]
    \centering
    \begin{subfigure}{0.48\linewidth}
    \centering
    \includegraphics[scale=0.5,width=\linewidth]{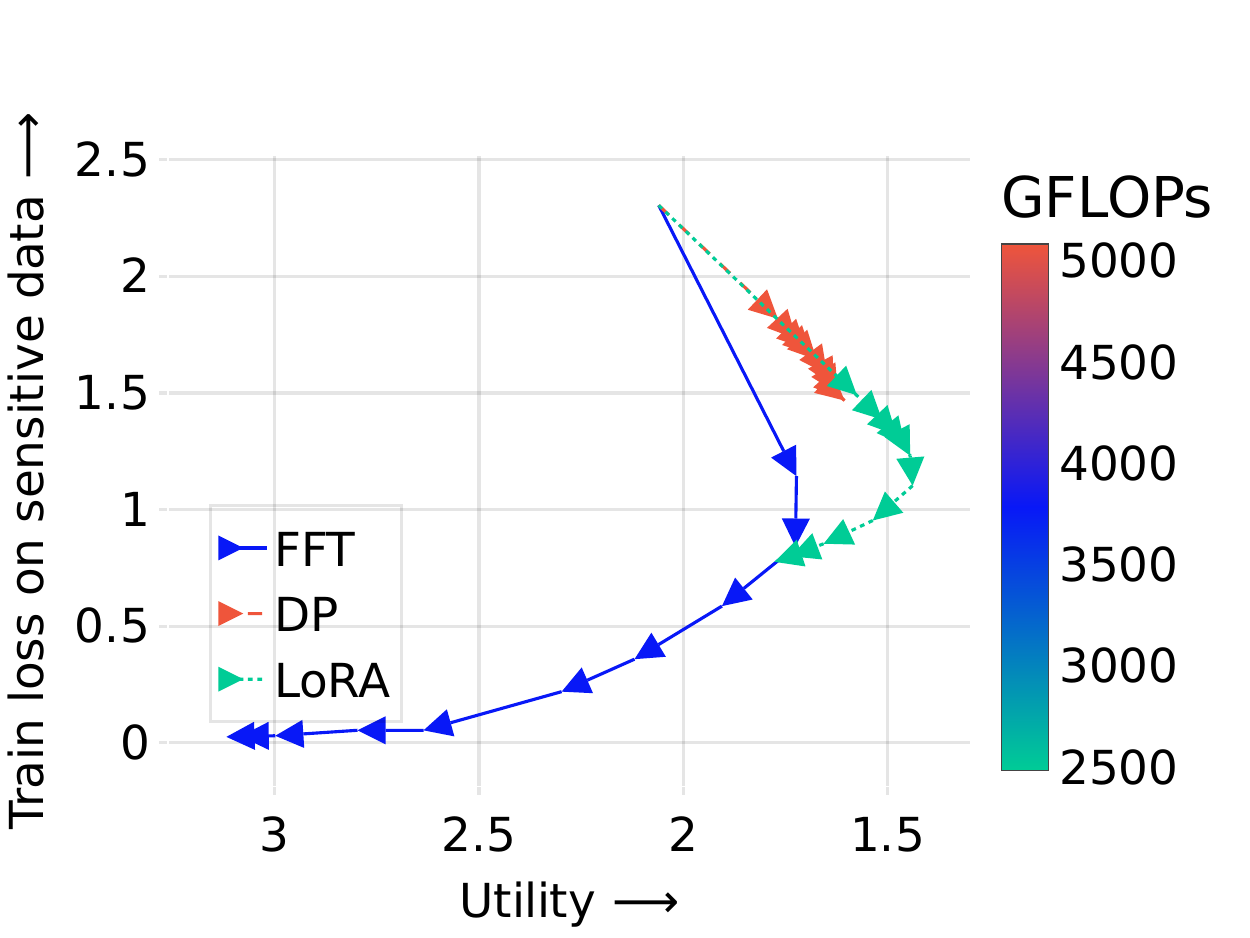}
    \caption{\textbf{Pythia}}
    \label{fig:appendix-fdl_pii1}
    \end{subfigure}
    \begin{subfigure}{0.48\linewidth}
    \centering
    \includegraphics[scale=0.5,width=\linewidth]{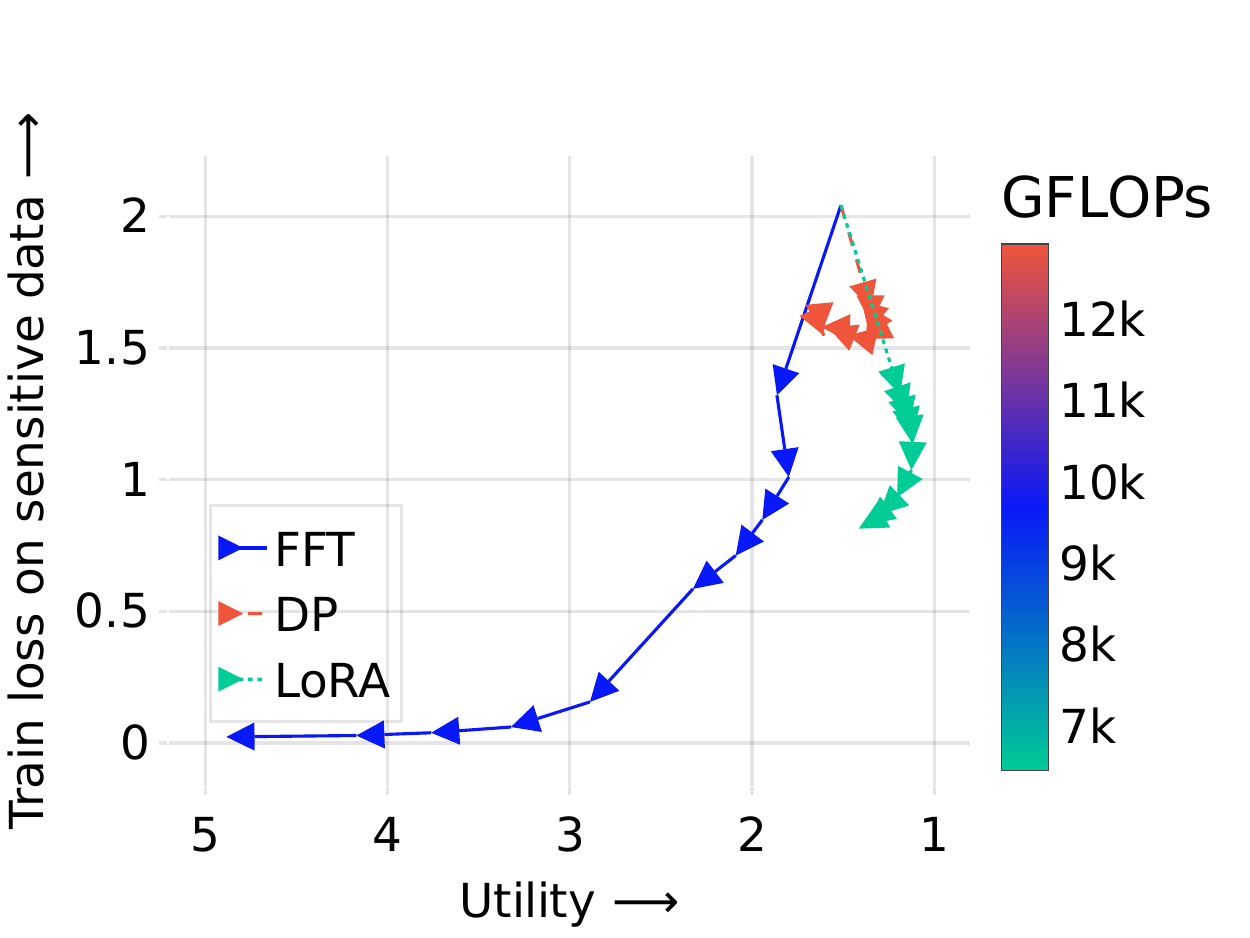}
    \caption{\textbf{Gemma}}
    \label{fig:appendix-fdl_pii2}
    \end{subfigure}
    \begin{subfigure}{0.48\linewidth}
    \centering
    \includegraphics[scale=0.5,width=\linewidth]{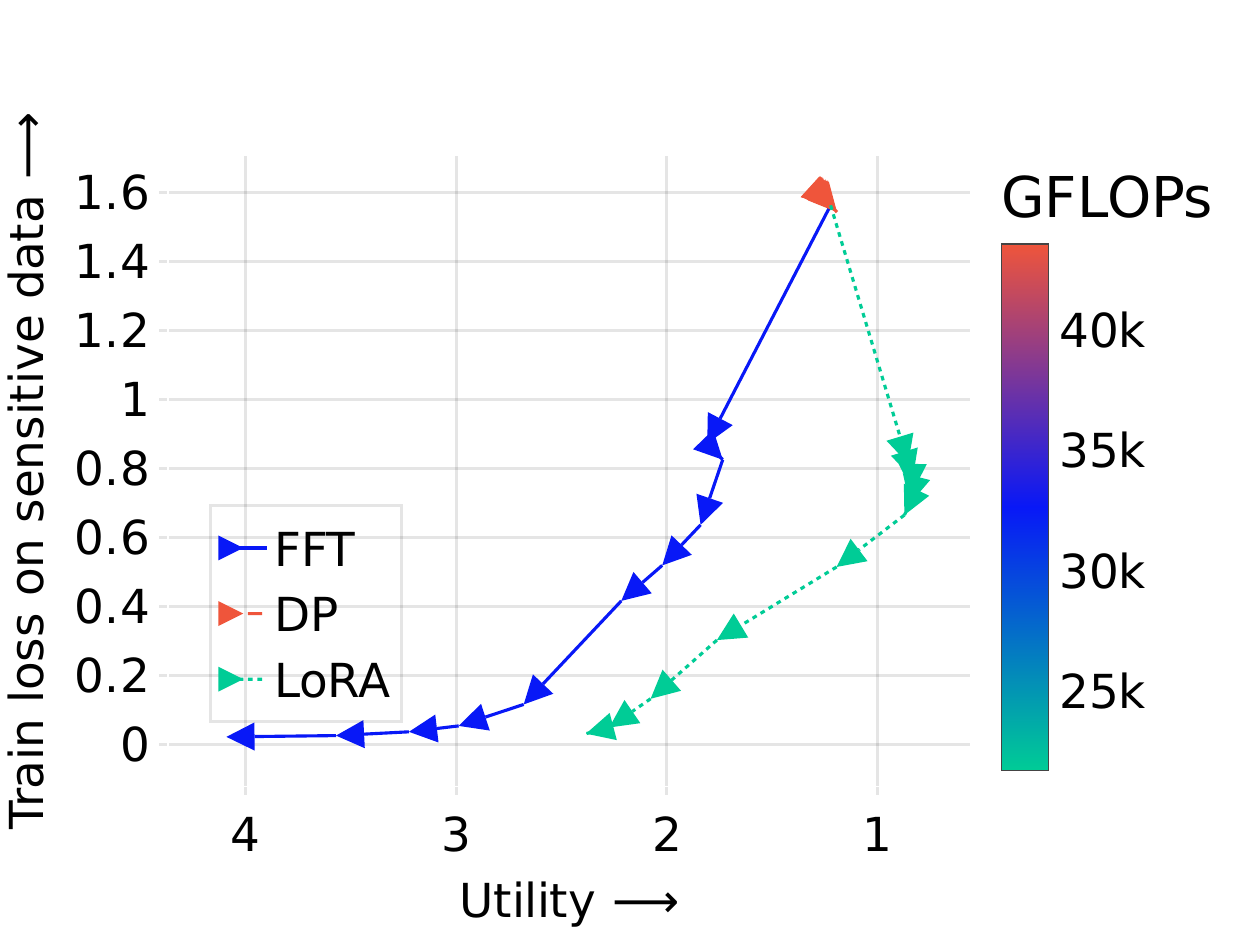}
    \caption{\textbf{Llama2}}
    \label{fig:appendix-fdl_pii3}
    \end{subfigure}
    \caption{\textbf{Full fine-tuning, DP and LoRA} on \emph{SynBio} with Presidio annotations for privacy-sensitive information.}
    \label{fig:appendix-fdl_pii}
\end{figure}
\begin{figure}[h!]
    \centering
    \begin{subfigure}{0.48\linewidth}
    \centering
    \includegraphics[width=\linewidth]{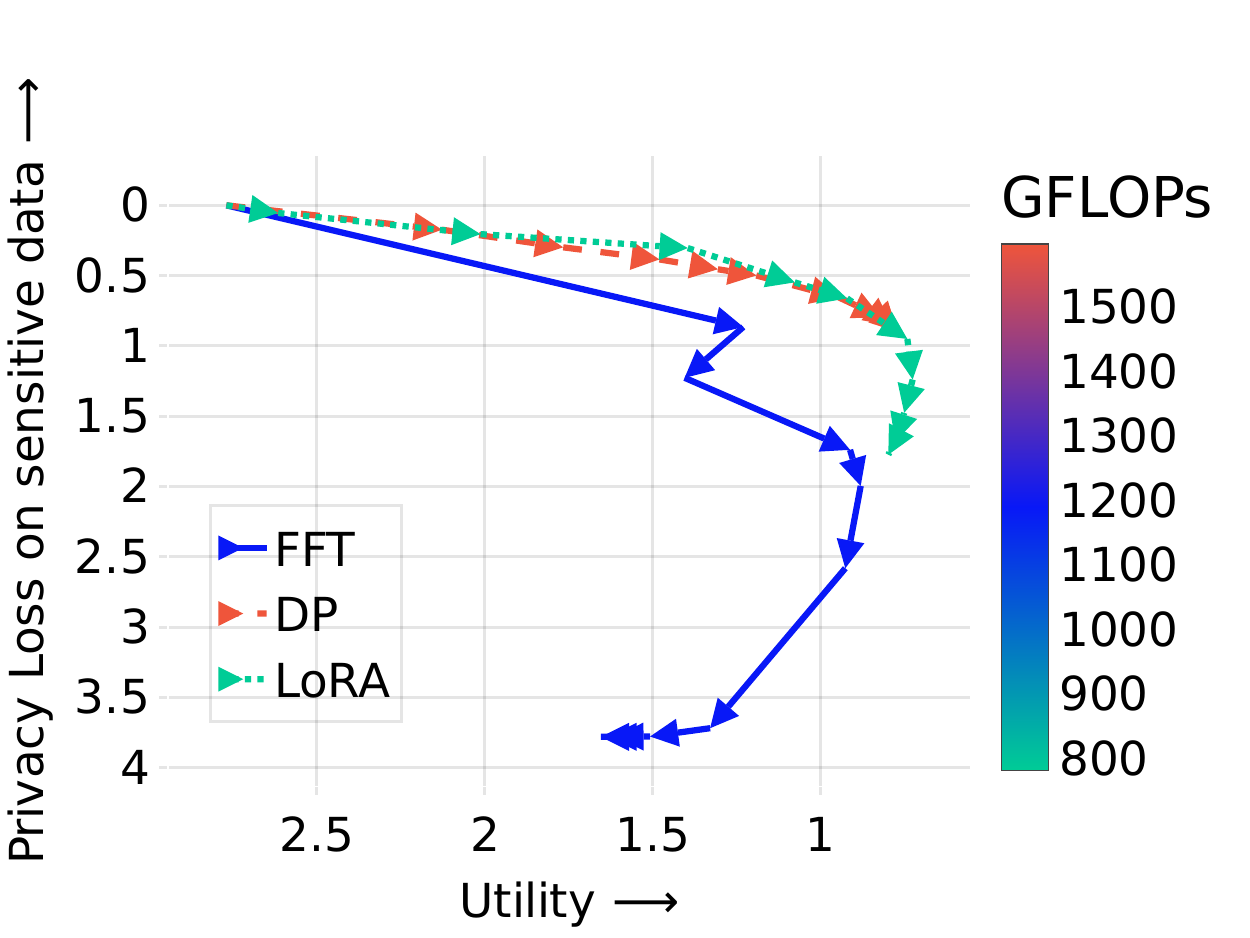}
    \caption{\textbf{Pythia}}
    \label{fig:appendix-fdl_pl_csim1}
    \end{subfigure}
    \begin{subfigure}{0.48\linewidth}
    \centering
    \includegraphics[width=\linewidth]{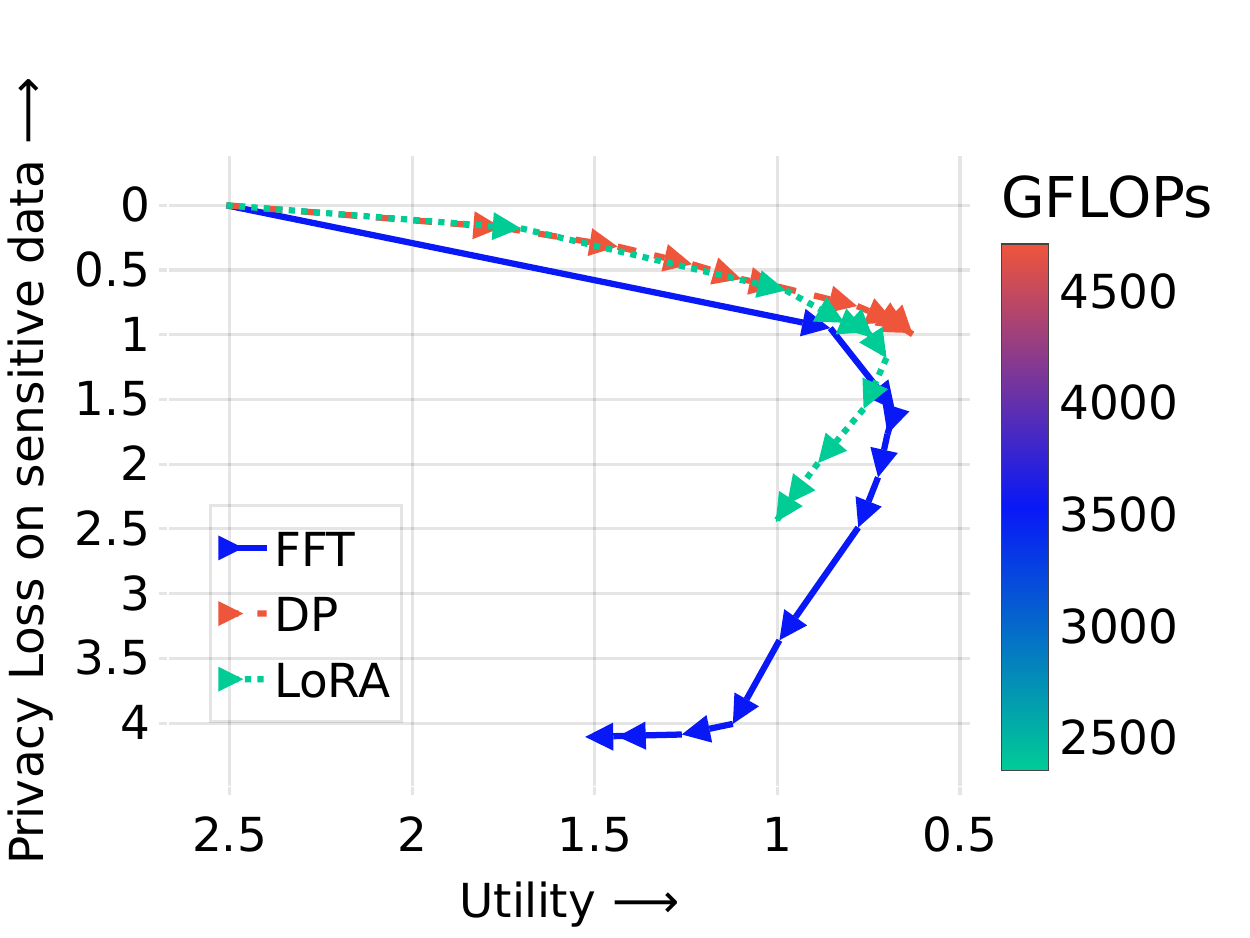}
    \caption{\textbf{Gemma}}
    \label{fig:appendix-fdl_pl_csim2}
    \end{subfigure}
    \begin{subfigure}{0.48\linewidth}
    \centering
    \includegraphics[width=\linewidth]{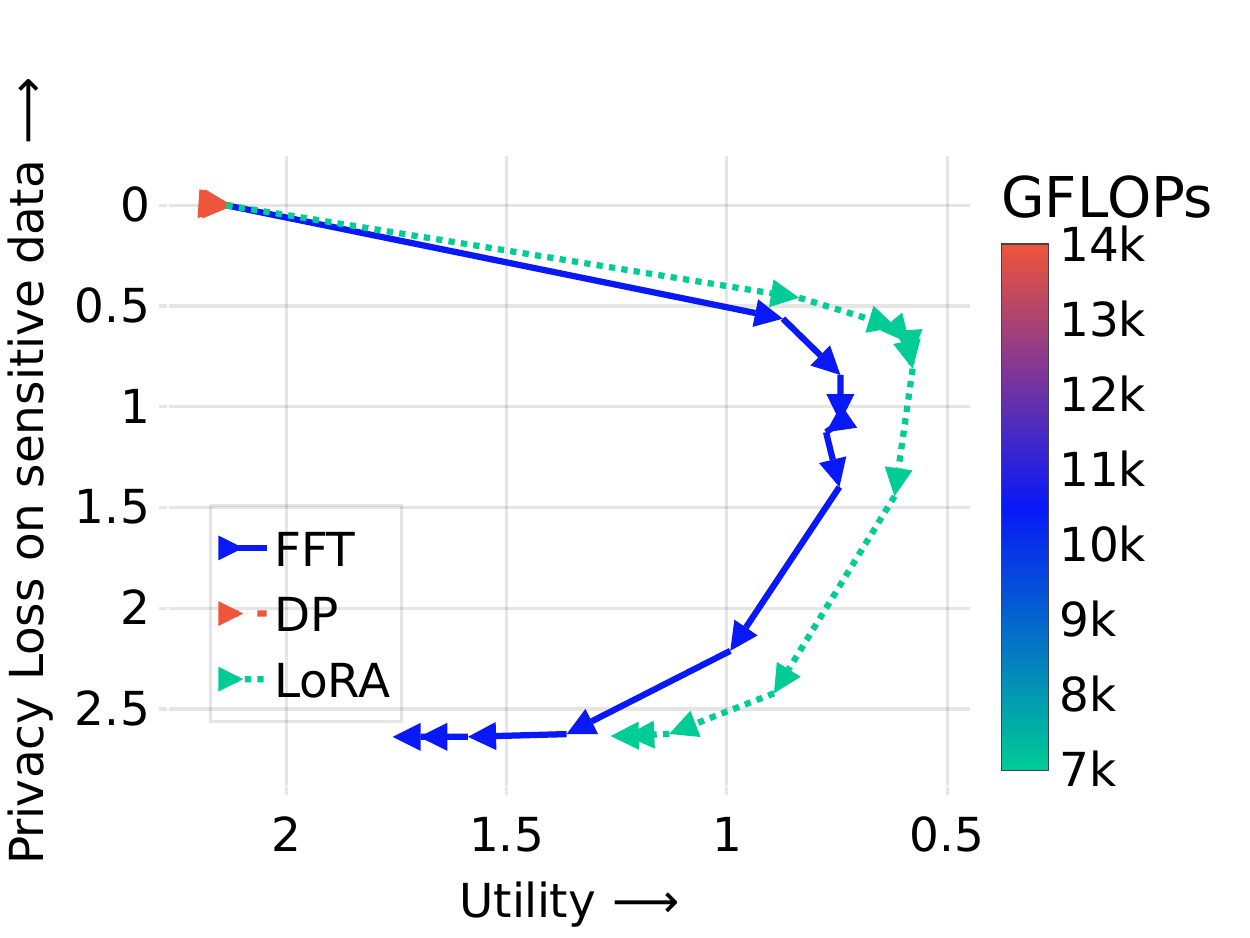}
    \caption{\textbf{Llama2}}
    \label{fig:appendix-fdl_pl_csim3}
    \end{subfigure}
    \begin{subfigure}{0.48\linewidth}
    \centering
    \includegraphics[width=\linewidth]{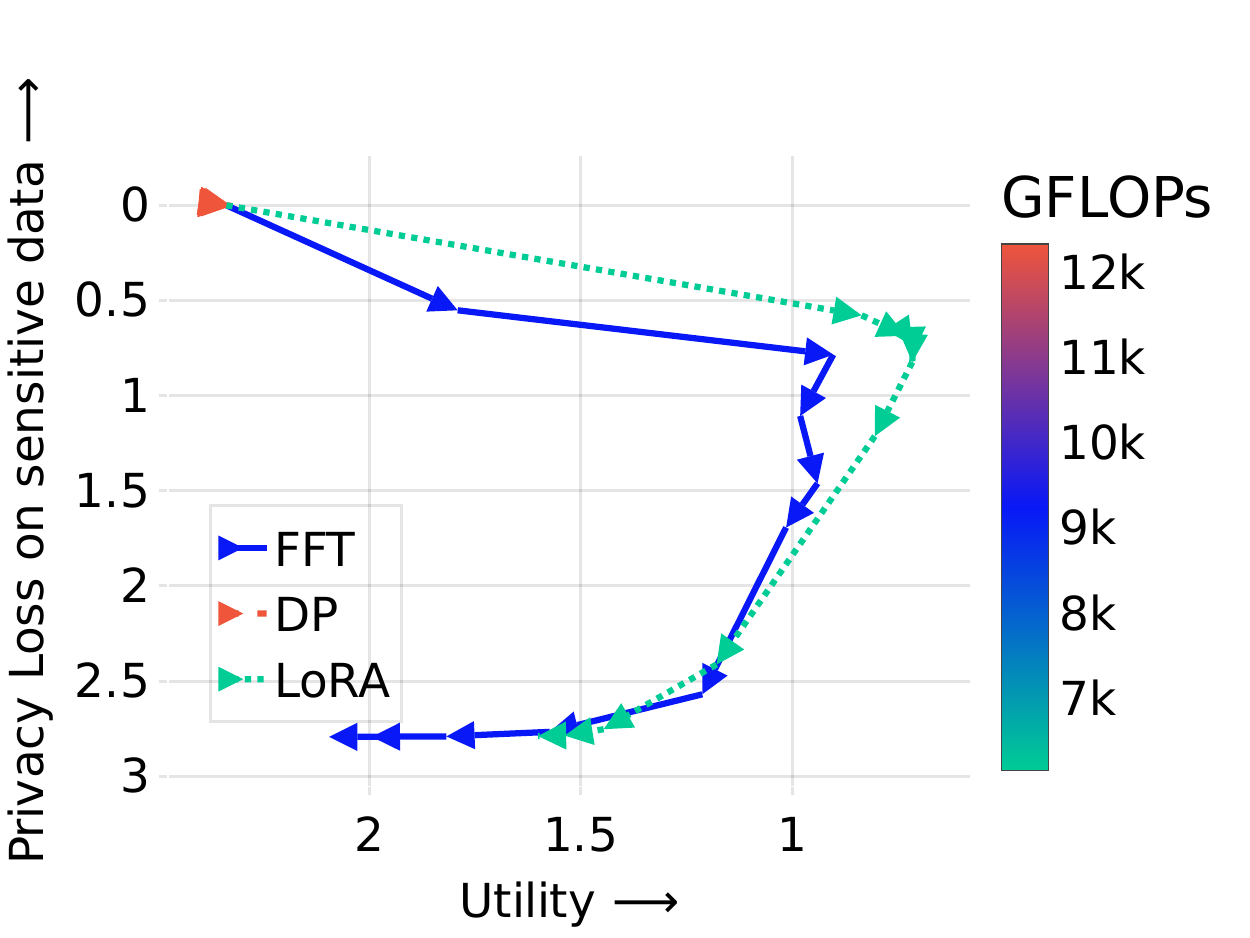}
    \caption{\textbf{Qwen2.5}}
    \label{fig:appendix-fdl_pl_csim4}
    \end{subfigure}
    \caption{\textbf{Full fine-tuning, DP and LoRA} on \emph{CustomerSim} with Presidio annotations for privacy loss}
    \label{fig:appendix-fdl_pl_csim}
\end{figure}
\begin{figure}[h!]
    \centering
    \begin{subfigure}{0.48\linewidth}
    \centering
    \includegraphics[width=\linewidth]{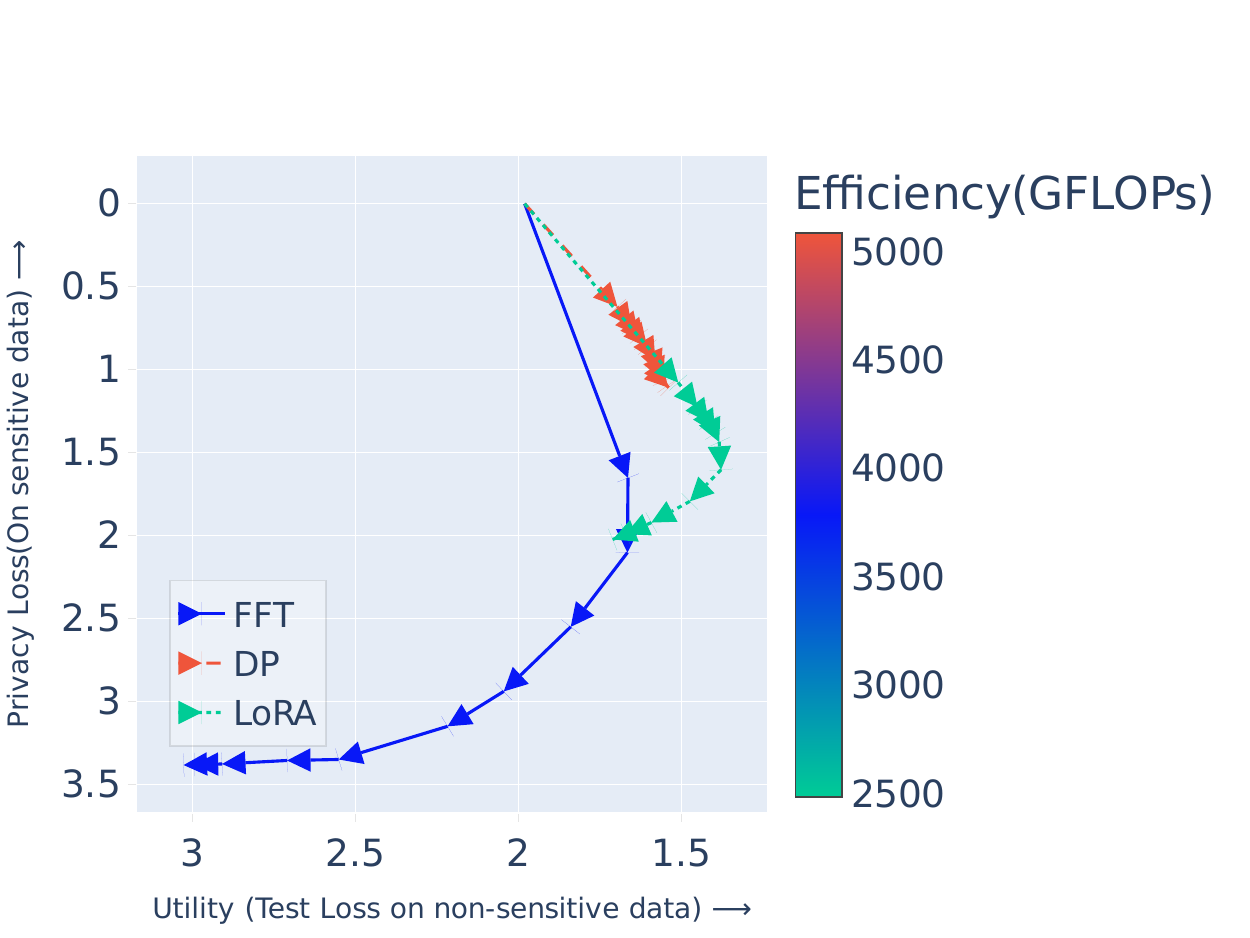}
    \caption{\textbf{Pythia}}
    \label{fig:appendix-fdl_pl_pii1}
    \end{subfigure}
    \begin{subfigure}{0.48\linewidth}
    \centering
    \includegraphics[width=\linewidth]{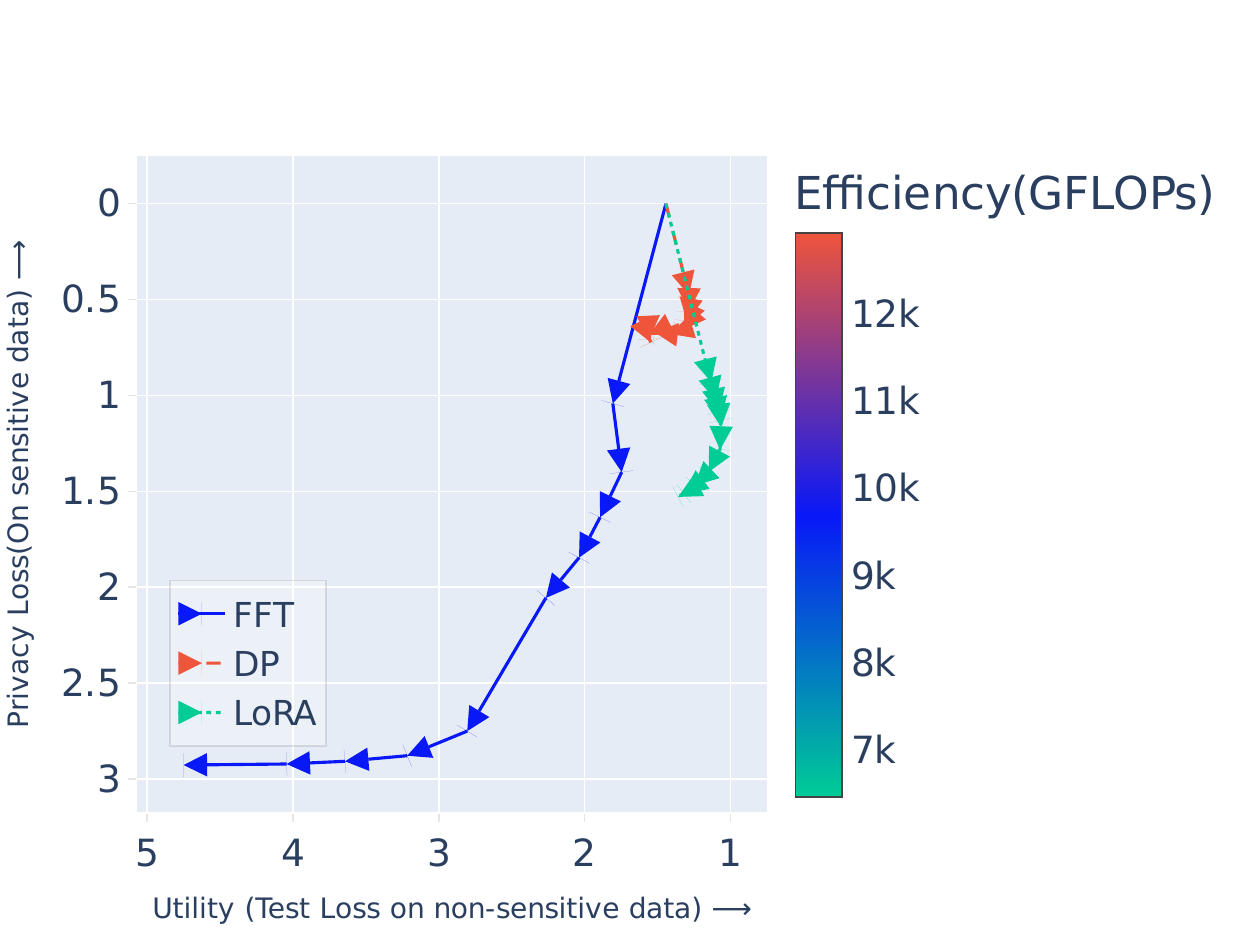}
    \caption{\textbf{Gemma}}
    \label{fig:appendix-fdl_pl_pii2}
    \end{subfigure}
    \begin{subfigure}{0.48\linewidth}
    \centering
    \includegraphics[width=\linewidth]{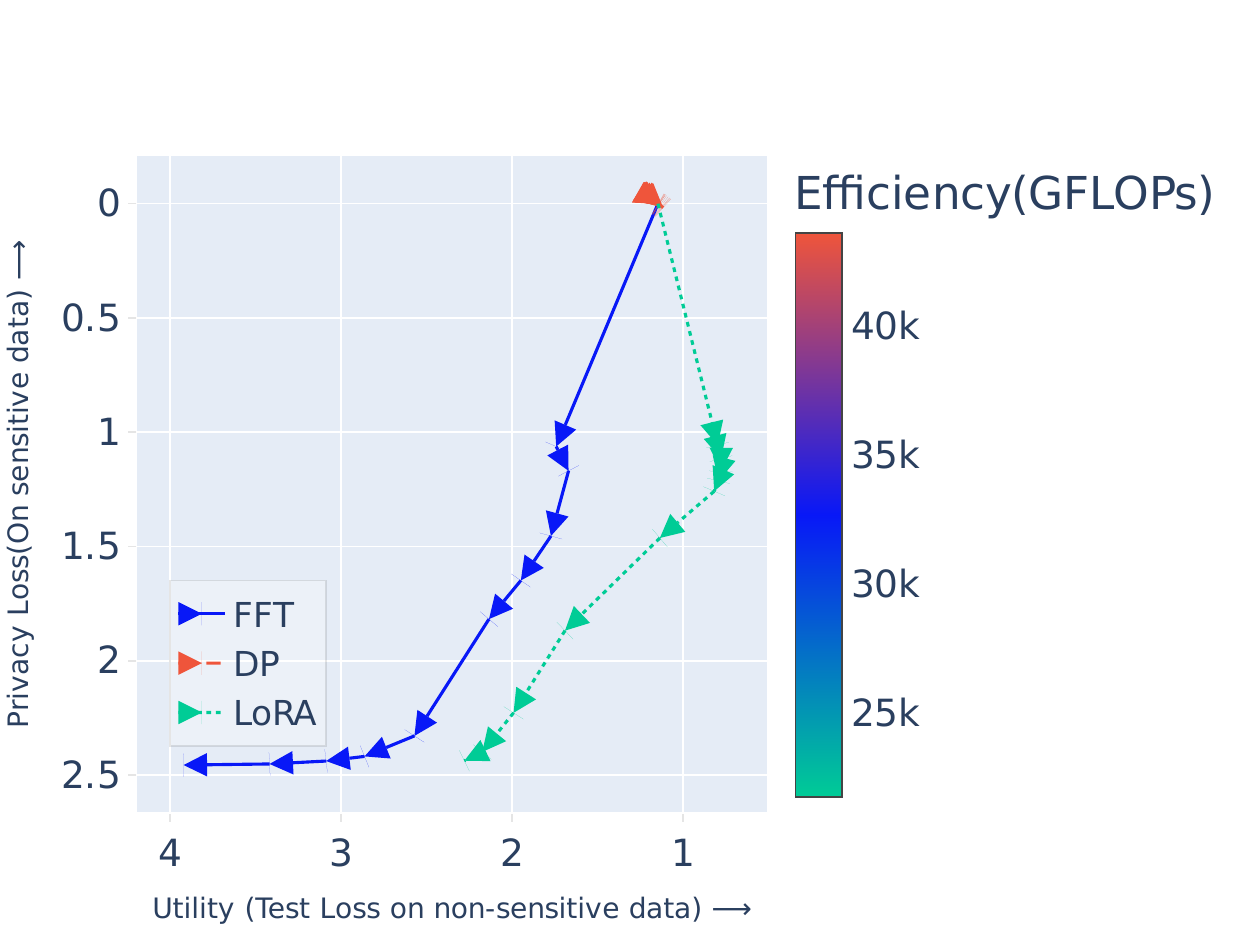}
    \caption{\textbf{Llama2}}
    \label{fig:appendix-fdl_pl_pii3}
    \end{subfigure}
    \begin{subfigure}{0.48\linewidth}
    \centering
    \includegraphics[width=\linewidth]{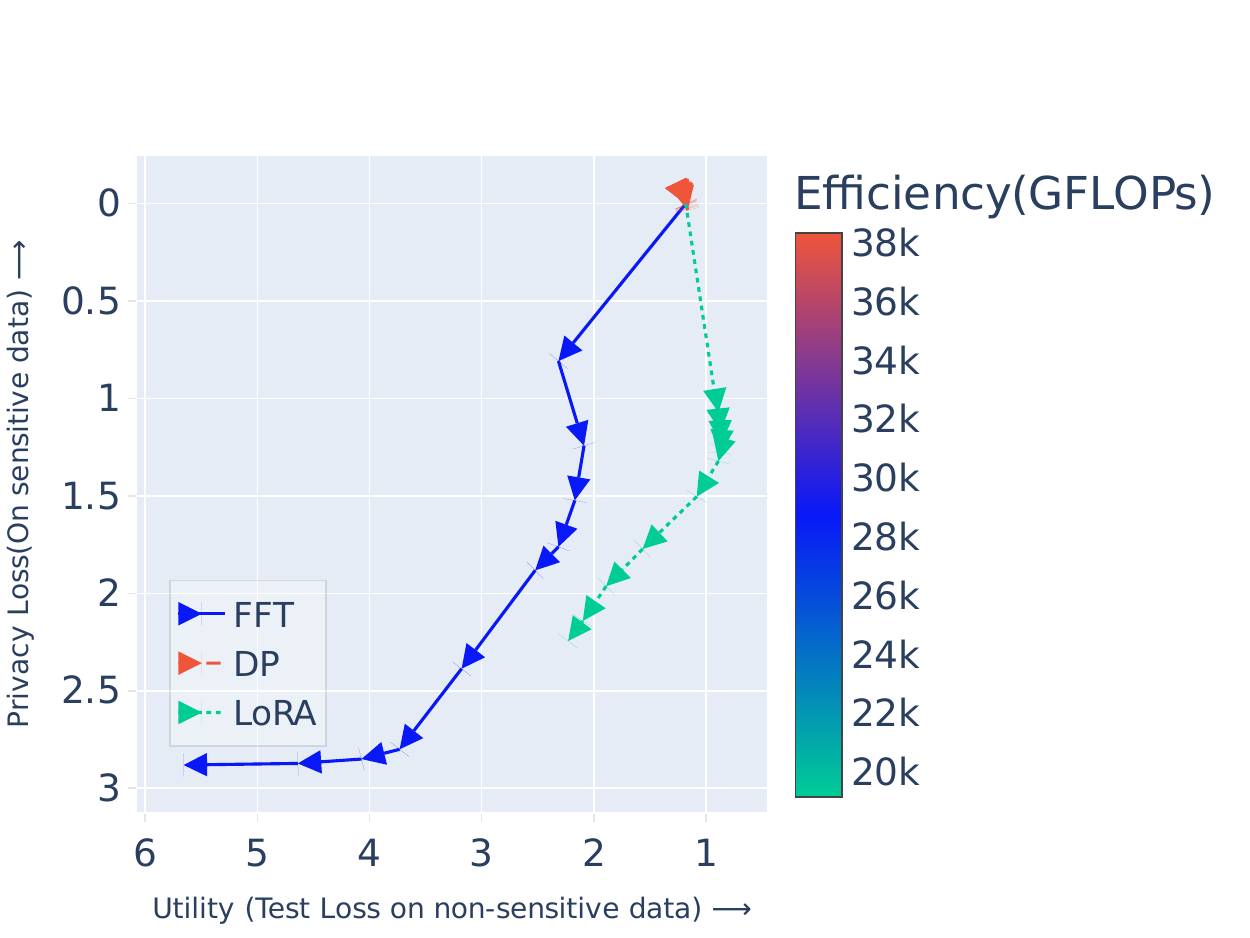}
    \caption{\textbf{Qwen2.5}}
    \label{fig:appendix-fdl_pl_pii4}
    \end{subfigure}
    \caption{\textbf{Full fine-tuning, DP and LoRA} on \emph{SynBio} with GPT4 annotations for privacy loss}
    \label{fig:appendix-fdl_pl_pii-gpt4}
\end{figure}
\begin{figure}[h!]
    \centering
    \begin{subfigure}{0.48\linewidth}
    \centering
    \includegraphics[width=\linewidth]{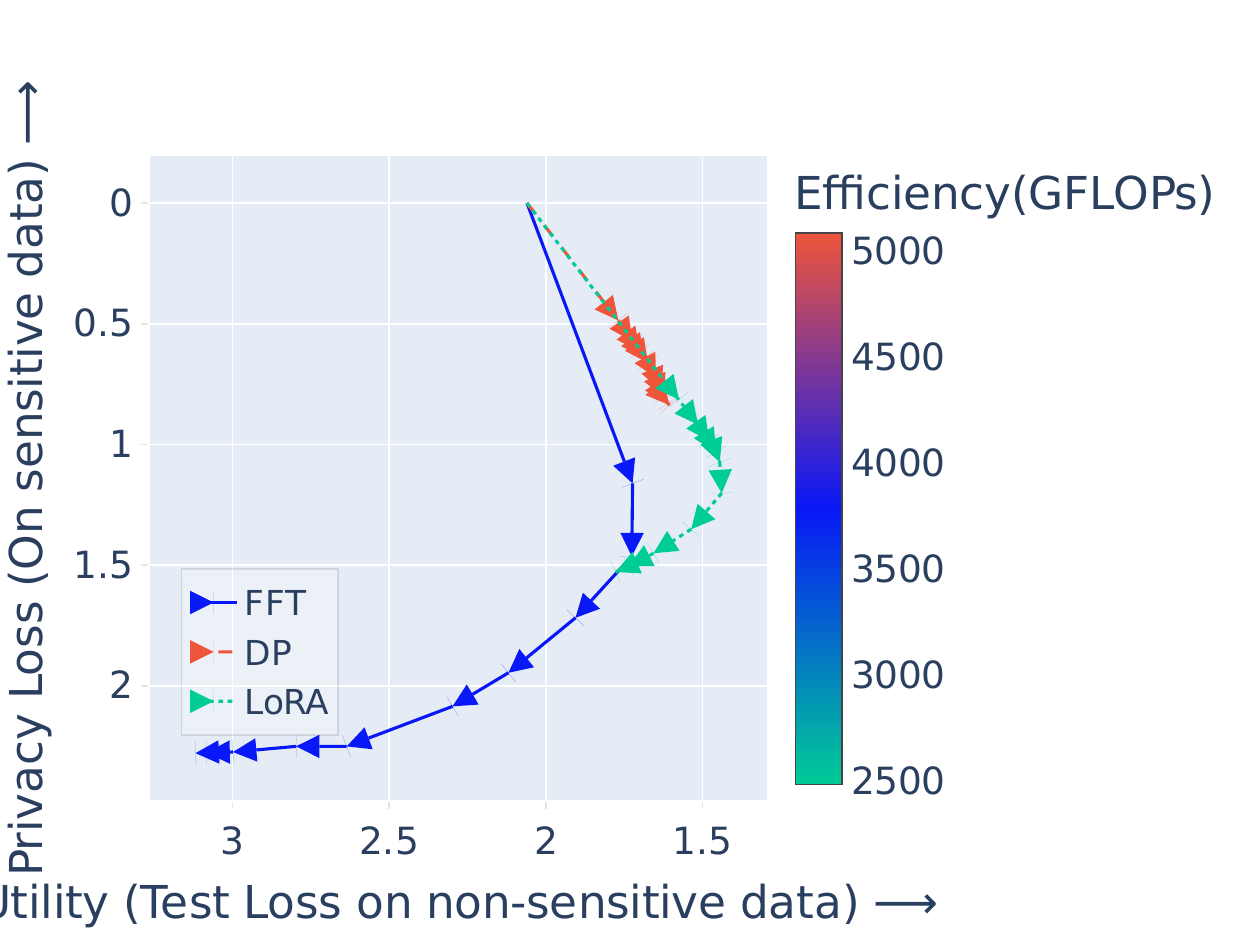}
    \caption{\textbf{Pythia}}
    \label{fig:appendix-fdl_pl_pii1}
    \end{subfigure}
    \begin{subfigure}{0.48\linewidth}
    \centering
    \includegraphics[width=\linewidth]{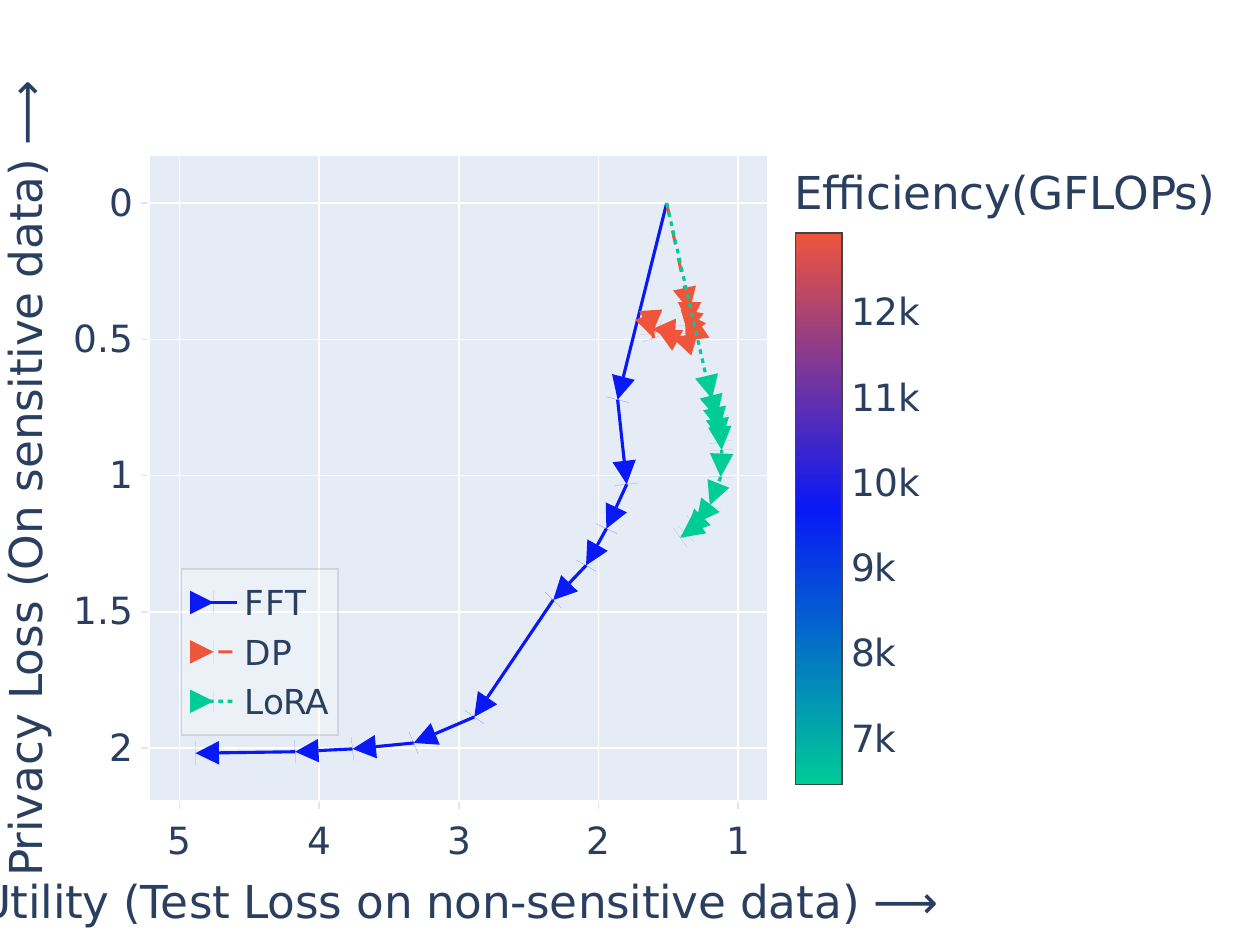}
    \caption{\textbf{Gemma}}
    \label{fig:appendix-fdl_pl_pii2}
    \end{subfigure}
    \begin{subfigure}{0.48\linewidth}
    \centering
    \includegraphics[width=\linewidth]{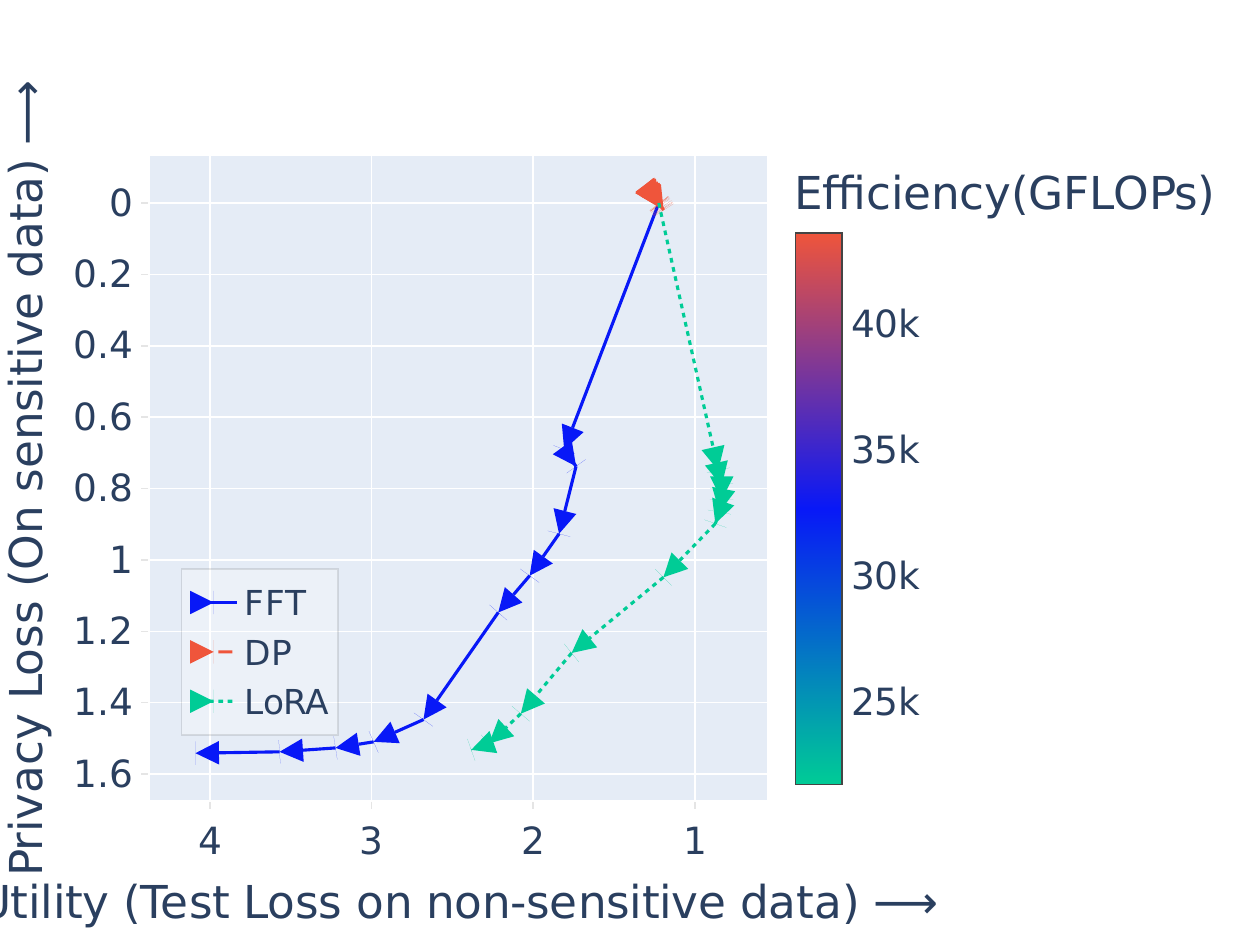}
    \caption{\textbf{Llama2}}
    \label{fig:appendix-fdl_pl_pii3}
    \end{subfigure}
    \begin{subfigure}{0.48\linewidth}
    \centering
    \includegraphics[width=\linewidth]{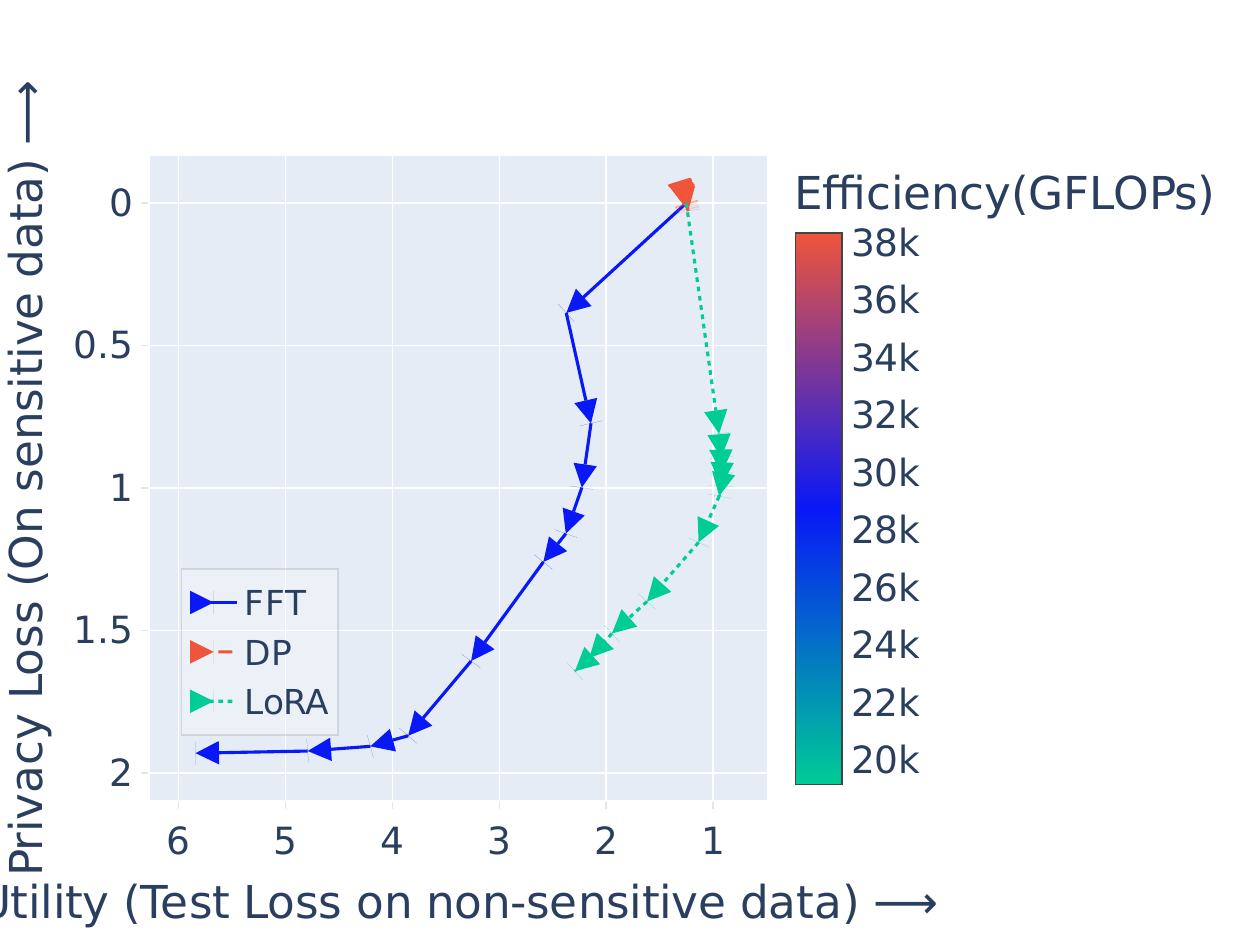}
    \caption{\textbf{Qwen2.5}}
    \label{fig:appendix-fdl_pl_pii4}
    \end{subfigure}
    \caption{\textbf{Full fine-tuning, DP and LoRA} on \emph{SynBio} with Presidio annotations for privacy loss}
    \label{fig:appendix-fdl_pl_pii}
\end{figure}
\begin{figure}[h!]
    \centering
    \begin{subfigure}{0.48\linewidth}
    \centering
    \includegraphics[width=\linewidth]{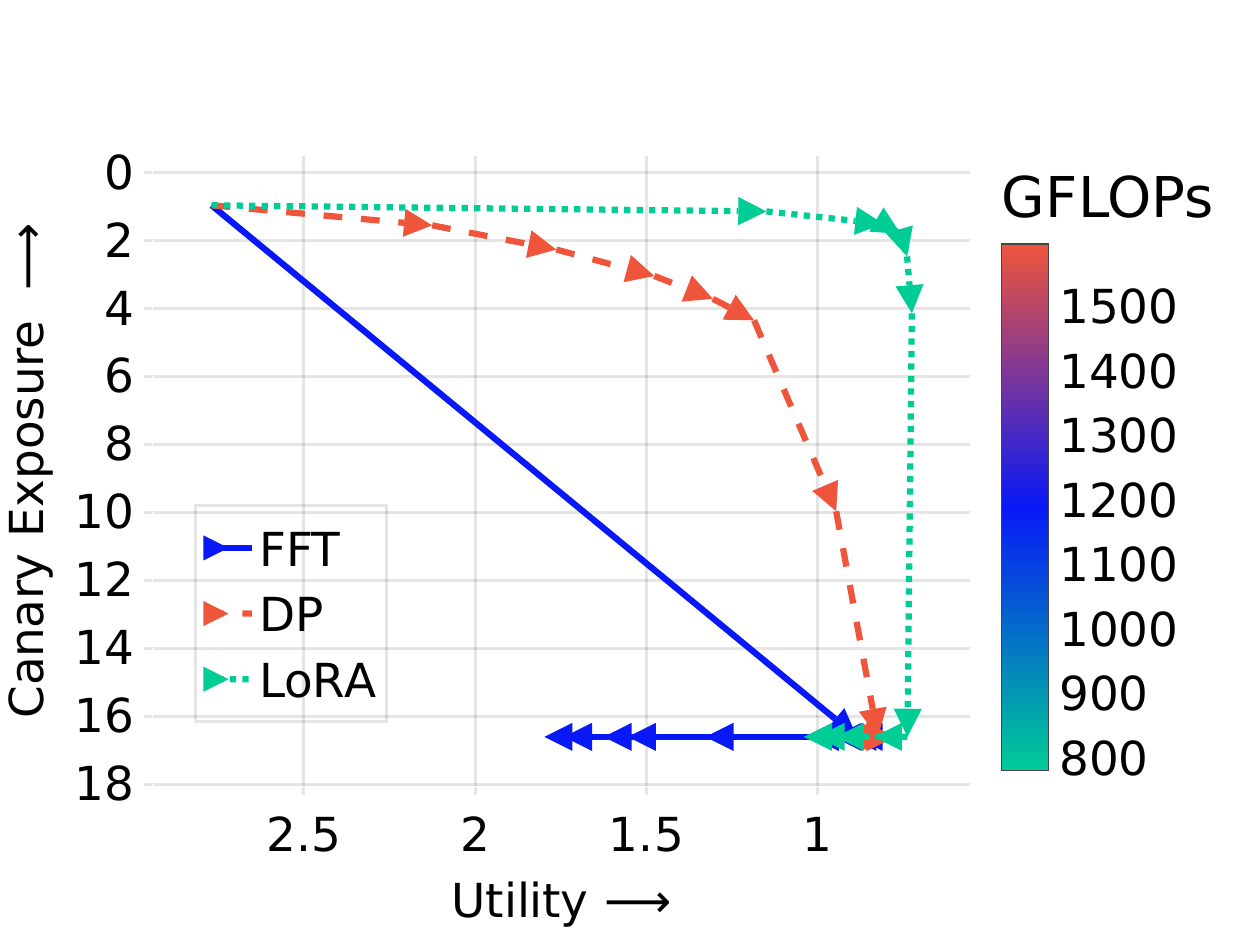}
    \caption{\textbf{Pythia}}
    \label{fig:appendix-fdl_exp_csim1}
    \end{subfigure}
    \begin{subfigure}{0.48\linewidth}
    \centering
    \includegraphics[width=\linewidth]{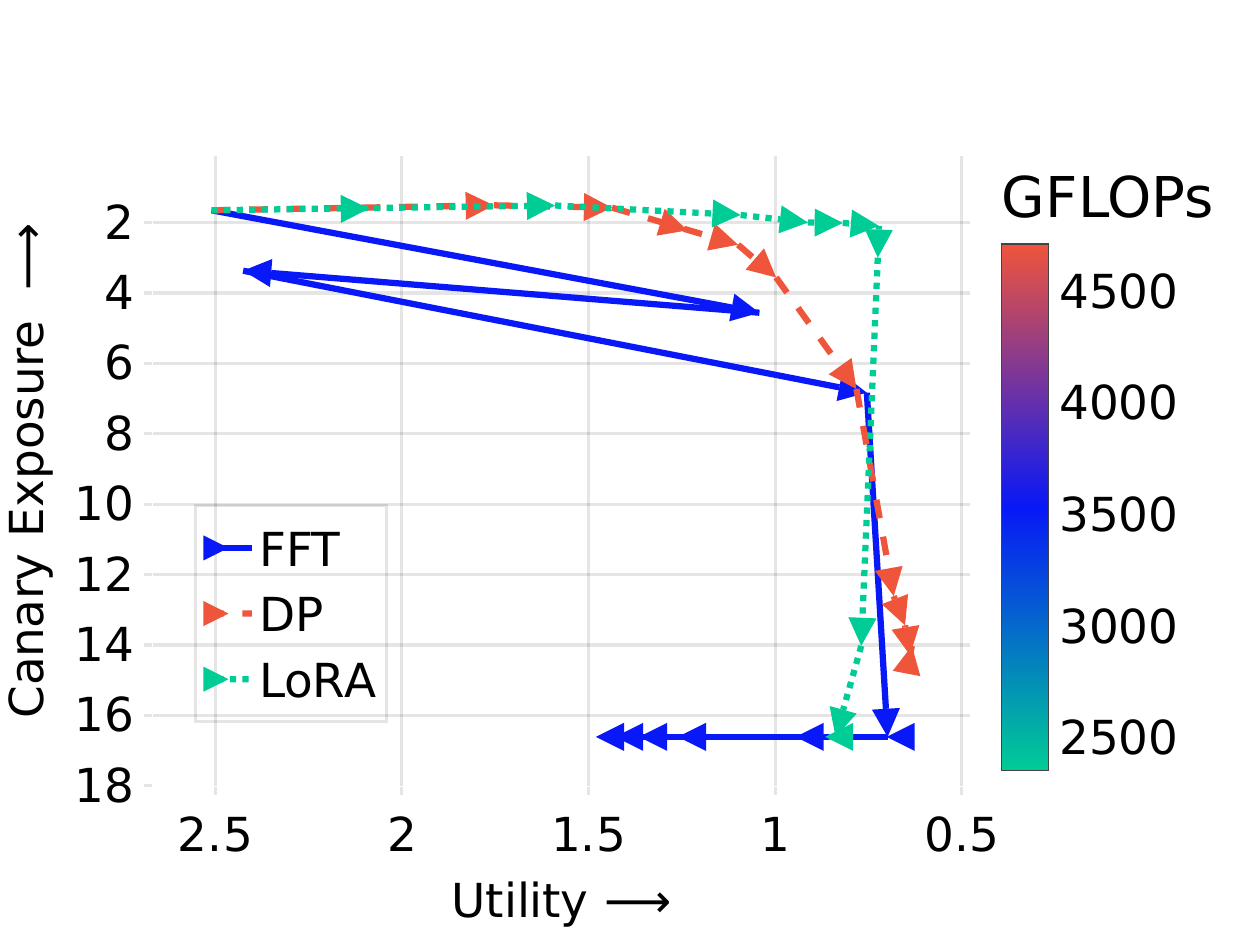}
    \caption{\textbf{Gemma}}
    \label{fig:appendix-fdl_exp_csim2}
    \end{subfigure}
    \begin{subfigure}{0.48\linewidth}
    \centering
    \includegraphics[width=\linewidth]{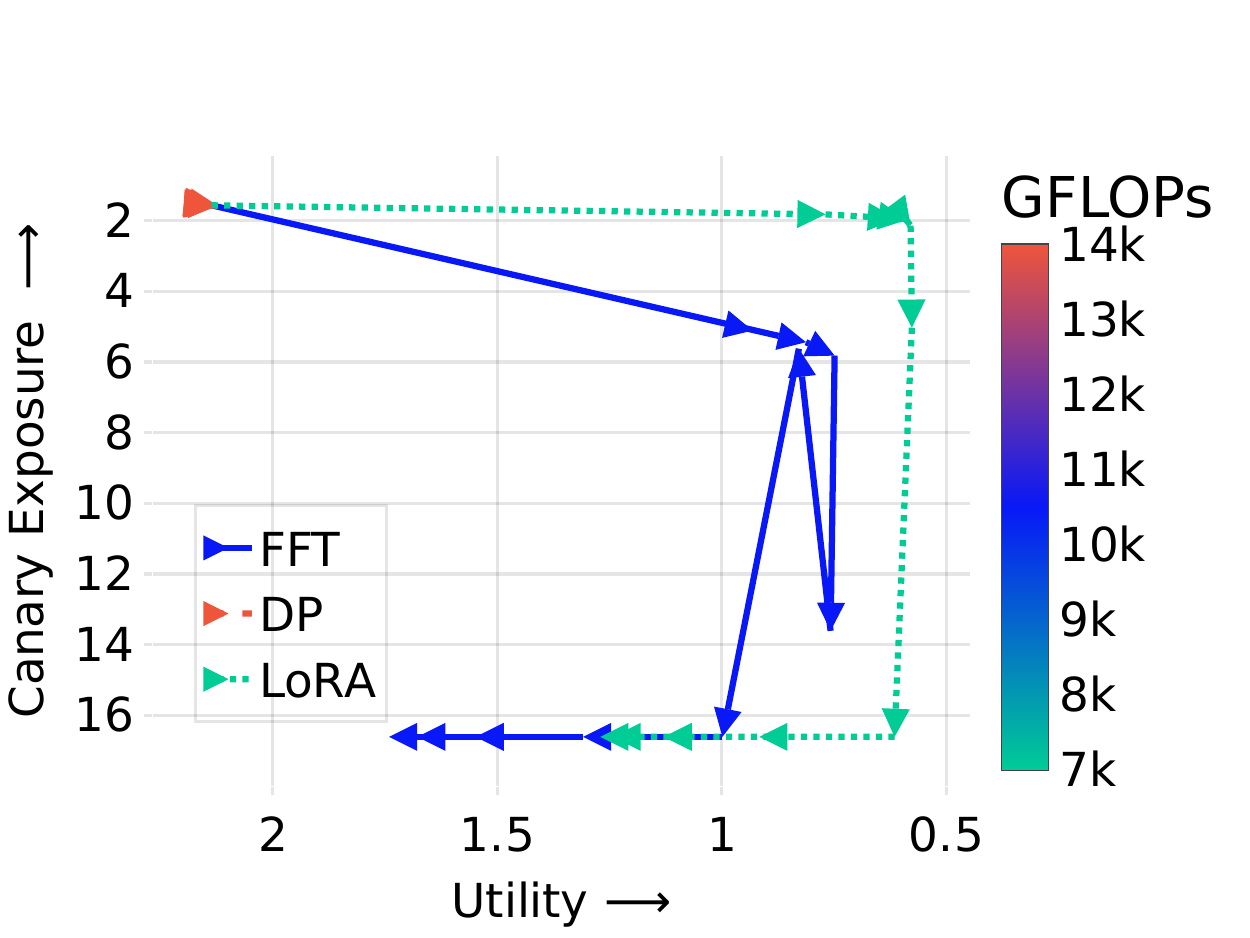}
    \caption{\textbf{Llama2}}
    \label{fig:appendix-fdl_exp_csim3}
    \end{subfigure}
    \begin{subfigure}{0.48\linewidth}
    \centering
    \includegraphics[width=\linewidth]{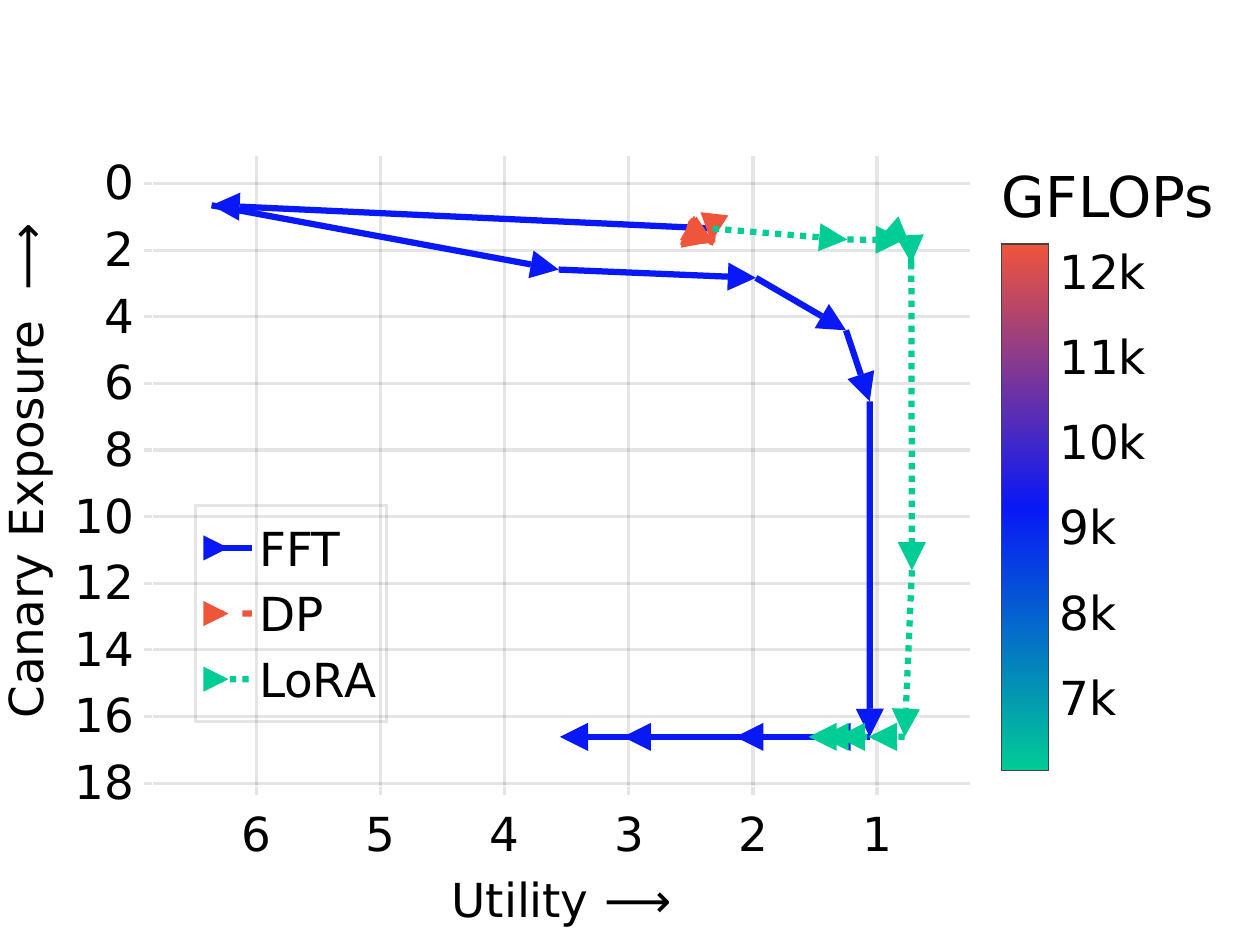}
    \caption{\textbf{Qwen2.5}}
    \label{fig:appendix-fdl_exp_csim4}
    \end{subfigure}
    \caption{\textbf{Full fine-tuning, DP and LoRA} on \emph{CustomerSim} with Presidio annotations for canary exposure.}
    \label{fig:appendix-fdl_exp_csim}
\end{figure}

We analyze the trade-offs between privacy, utility, and efficiency for different privacy measures over SynBio with both GPT4 and Presidio annotations. Figures~\ref{fig:appendix-fdl_csim} , \ref{fig:appendix-fdl_pii} for CustomerSim and SynBio using Presidio annotations and our privacy measure reveals the same as observed in Figure~\ref{fig:combined_fdl} with GPT-4 annotations. We also show the same for the other privacy measures -- (a) privacy loss in Figures \ref{fig:appendix-fdl_pl_csim}, \ref{fig:appendix-fdl_pl_pii-gpt4}, and \ref{fig:appendix-fdl_pl_pii} , and (b) exposure in Figure \ref{fig:appendix-fdl_exp_csim}.

\begin{figure}[!t]
    \centering
        \begin{subfigure}{.49\linewidth}
       \includegraphics[width=\linewidth]{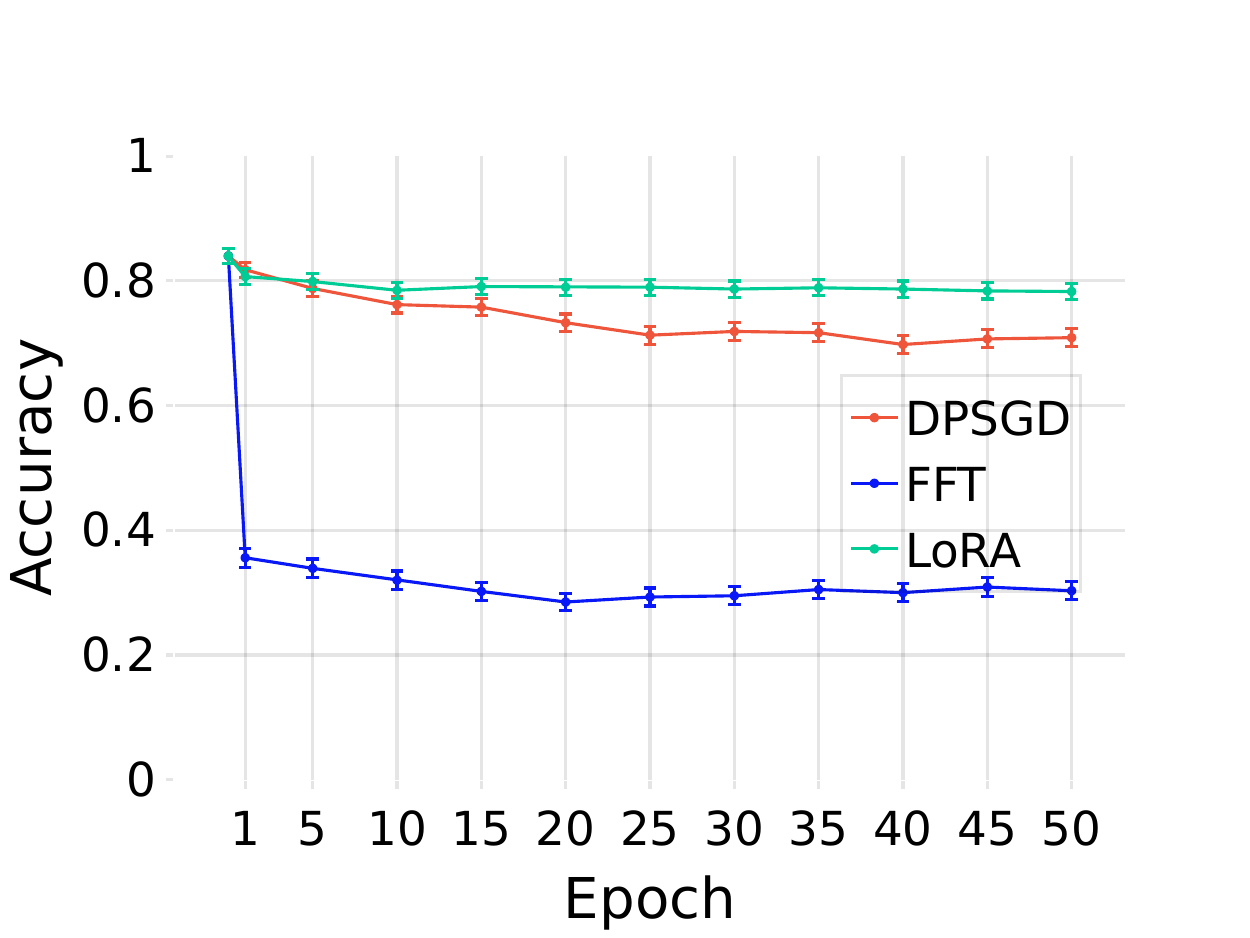}
        \caption{Pythia : SCIQ}
        \label{fig:sciq_all_pyt}
    \end{subfigure}  
     \begin{subfigure}{.49\linewidth}
       \includegraphics[width=\linewidth]{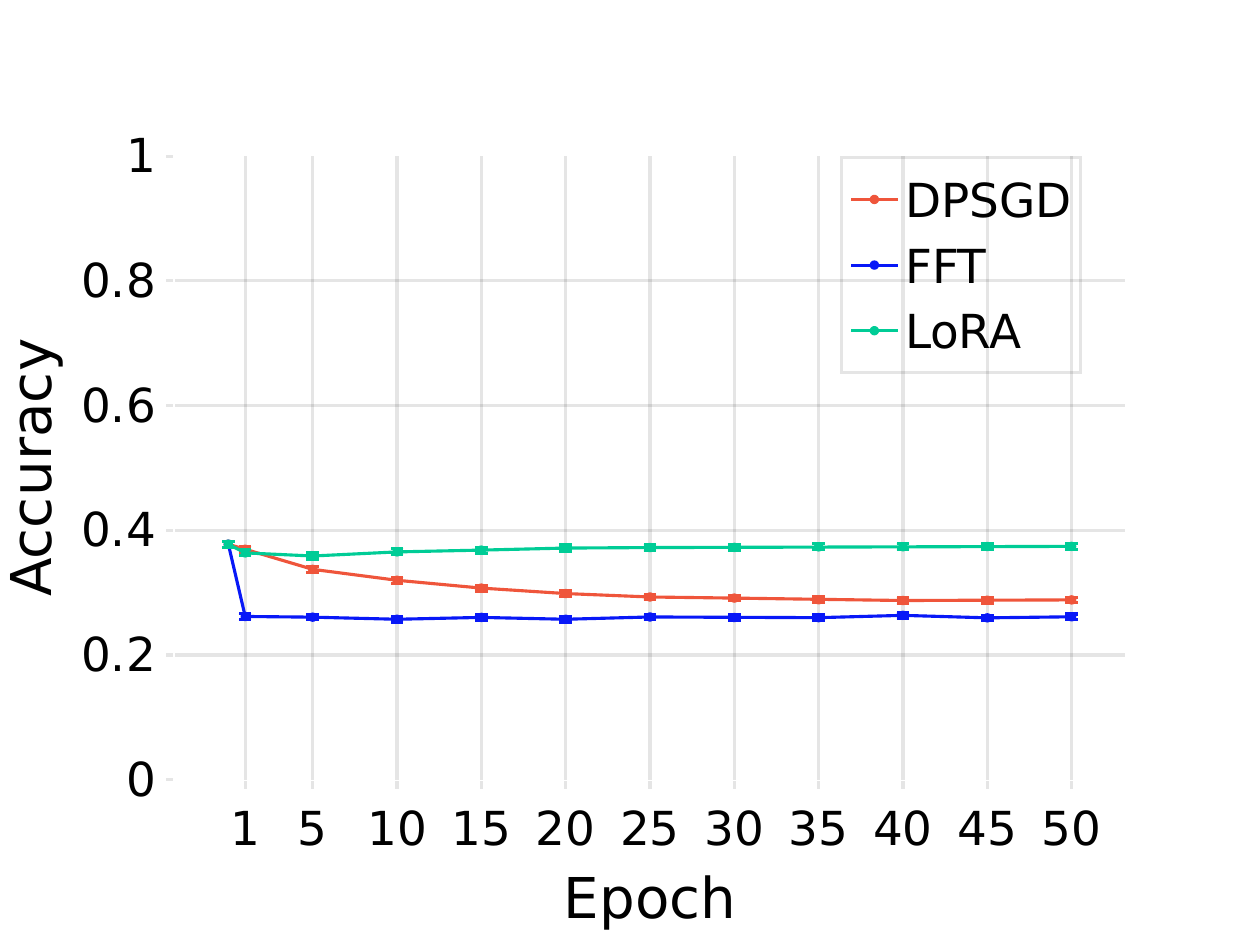}
        \caption{Pythia : HellaSwag}
        \label{fig:hs_all_pyt}
    \end{subfigure}

    \begin{subfigure}{.49\linewidth}
       \includegraphics[width=\linewidth]{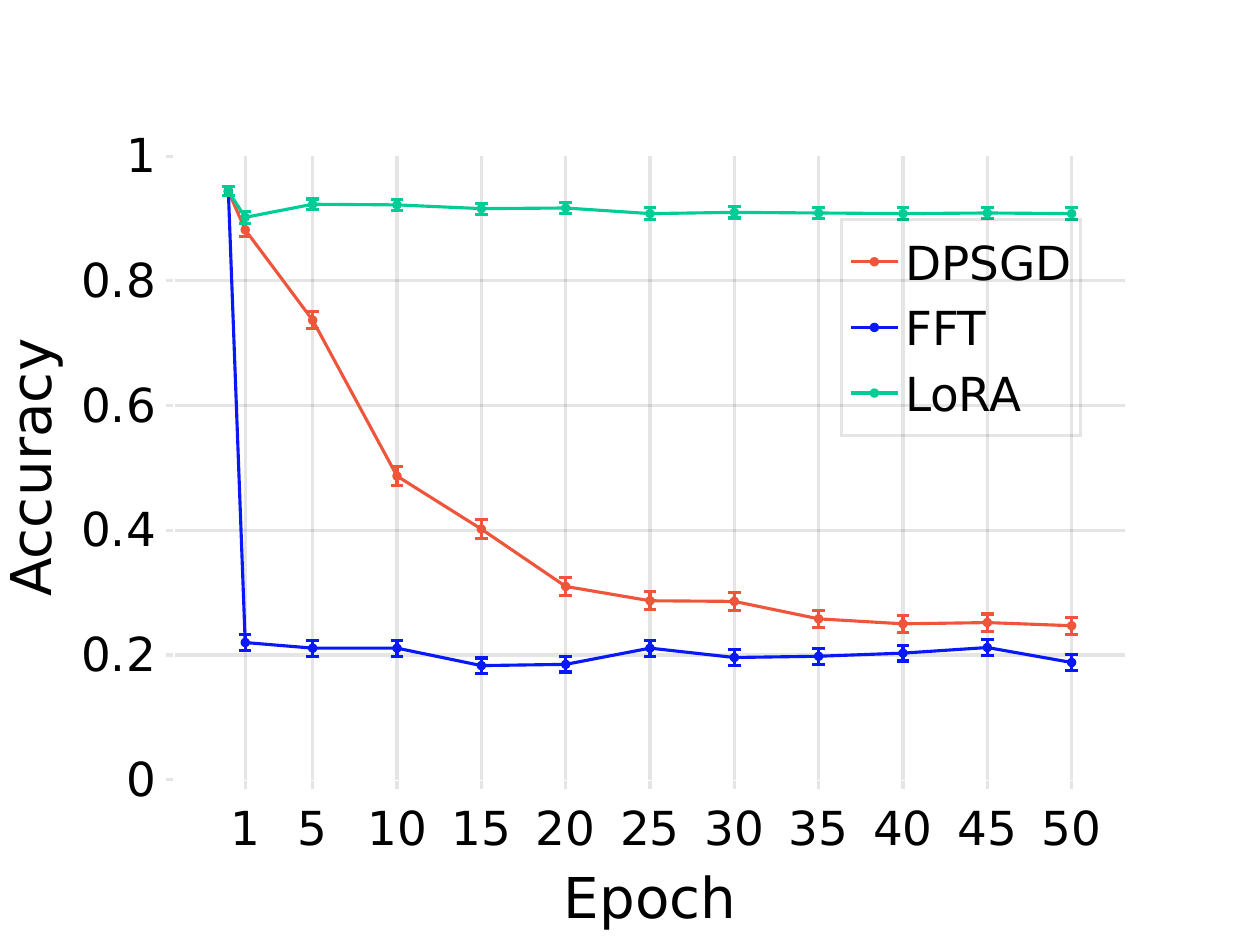}
        \caption{Gemma : SCIQ}
        \label{fig:sciq_all_pyt}
    \end{subfigure}  
     \begin{subfigure}{.49\linewidth}
       \includegraphics[width=\linewidth]{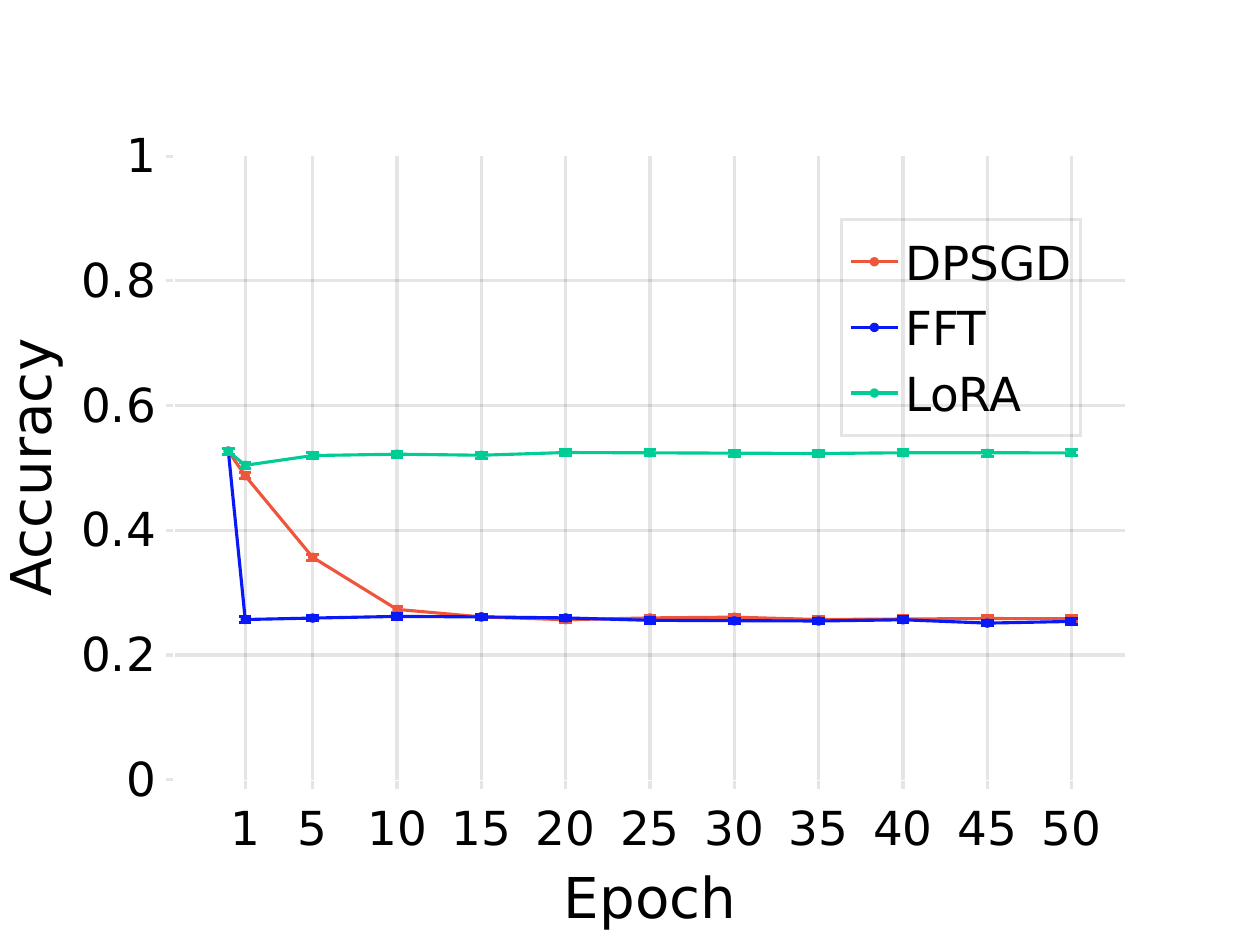}
        \caption{Gemma : HellaSwag}
        \label{fig:hs_all_pyt}
    \end{subfigure}
    
    \begin{subfigure}{.49\linewidth}
       \includegraphics[width=\linewidth]{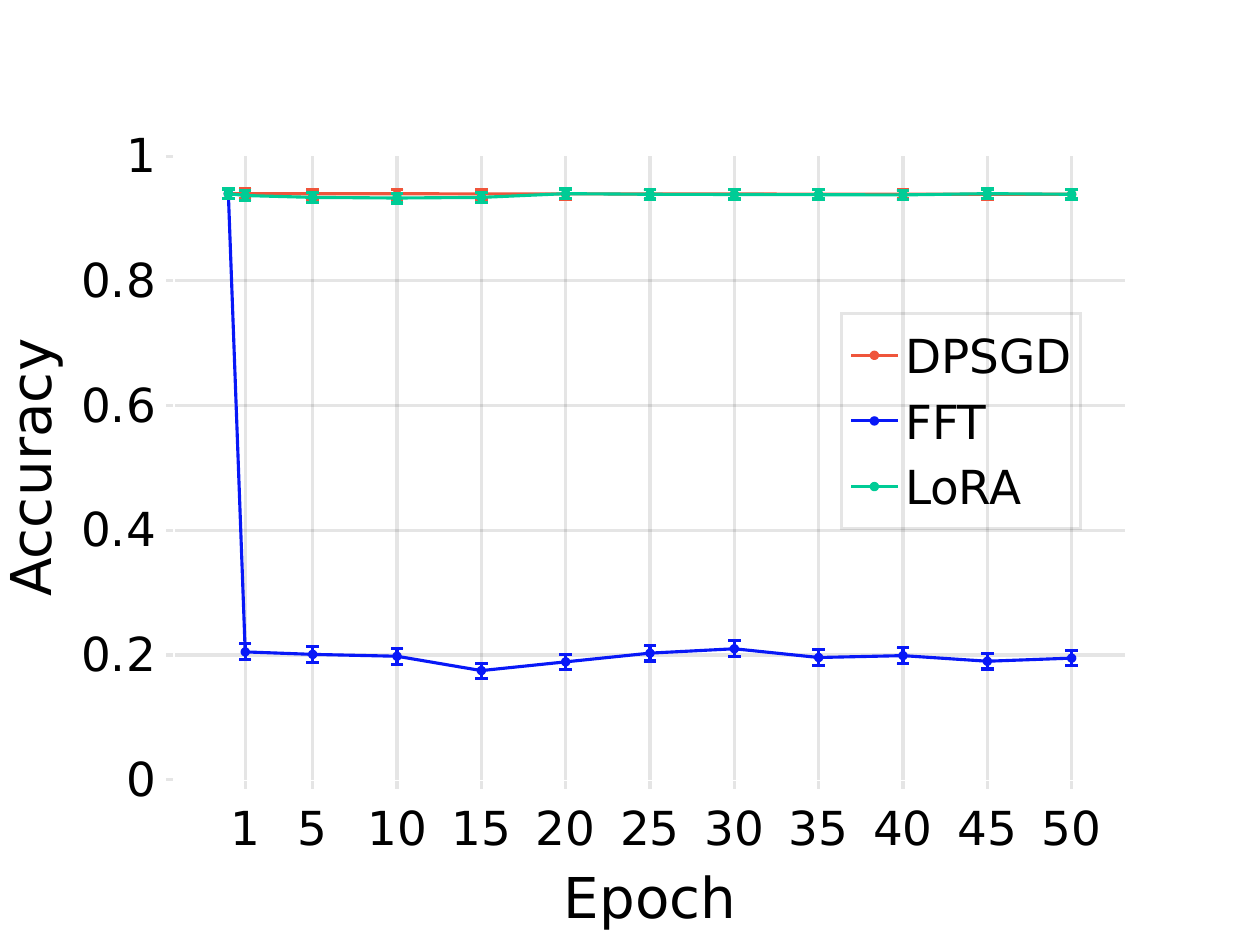}
        \caption{Llama2 : SCIQ}
        \label{fig:sciq_all_llama}
    \end{subfigure}  
     \begin{subfigure}{.49\linewidth}
       \includegraphics[width=\linewidth]{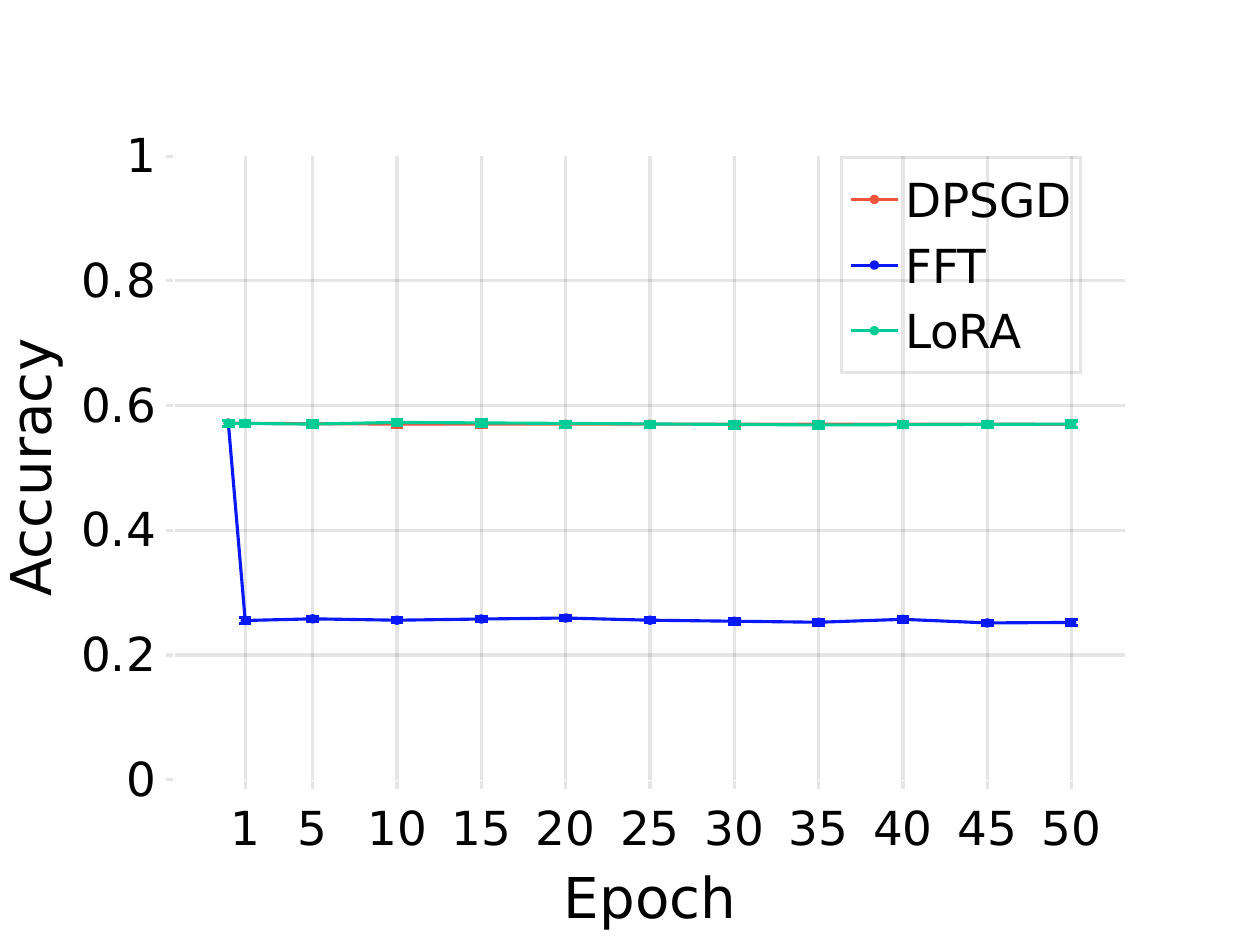}
        \caption{Llama 2 : HellaSwag}
        \label{fig:hs_all_llama}
    \end{subfigure}
    \caption{
    LoRA demonstrates knowledge retention on benchmark datasets throughout the training regime on CustomerSim for other models as well, while FFT and DP show fragility with sharp and gradual performance declines, respectively.
    }
    \label{fig:app-bench-all}
\end{figure}

The overall takeaway holds consistent -- LoRA is the most efficient method, followed by FFT while DP is the least efficient, requiring the highest number of FLOPs. For privacy, DP and LoRA perform at par with FFT leaking the most amount of information. In terms of utility, LoRA and FFT perform better while DP deteriorates on large scale models.  Additionally, LoRA also maintains benchmark performance across other models as shown in Figure \ref{fig:app-bench-all}.

\vspace{-10pt}

\vspace{4mm}

\section{Comparison of fine-tuning methods}
\label{appendix:comparison}

\begin{table}[H]
\centering
\scriptsize
\caption{Comparison of all fine-tuning methods across privacy, utility, efficiency, and benchmark performance. LoRA is seen to outperform FFT and DP over all dimensions.}
\label{tab:overall}
\begin{tabular}{|c|c|c|c|c|}
\hline
 & \textbf{Utility} & \textbf{Privacy} & \textbf{Efficiency} & \textbf{Benchmark} \\
\hline
\textbf{FFT} & \textcolor{gemmagreen}{Good} & \textcolor{red}{Poor} & \textcolor{yellow}{Moderate} & \textcolor{red}{Poor} \\
\hline
\textbf{DP} & \textcolor{yellow}{Moderate} & \textcolor{gemmagreen}{Good} & \textcolor{red}{Poor} & \textcolor{red}{Poor} \\
\hline
\rowcolor{green!20}
\textbf{LoRA} & \textcolor{gemmagreen}{Good} & \textcolor{gemmagreen}{Good} & \textcolor{gemmagreen}{Good} & \textcolor{gemmagreen}{Good} \\
\hline
\end{tabular}
\end{table}

\end{document}